\PassOptionsToPackage{table,x11names}{xcolor}
\documentclass{article}
\usepackage{iclr2027_conference,times}
\usepackage{amsmath,amssymb,amsfonts,mathtools}
\usepackage{iftex}
\ifPDFTeX
\else
  \ifXeTeX
    \AtBeginDocument{}
  \fi
\fi
\usepackage[T1]{fontenc}
\usepackage{hyperref,url,booktabs,multirow,longtable,array,graphicx,microtype}
\usepackage{xurl}
\usepackage{xcolor,placeins,wrapfig,needspace}
\usepackage{xspace}
\usepackage[capitalize,noabbrev]{cleveref}
\crefname{section}{\S}{\S\S}
\Crefname{section}{\S}{\S\S}
\crefname{appendix}{\S}{\S\S}
\Crefname{appendix}{\S}{\S\S}
\crefname{figure}{Fig.}{Figs.}
\Crefname{figure}{Fig.}{Figs.}
\crefname{table}{Table}{Tables}
\Crefname{table}{Table}{Tables}
\crefname{equation}{Eq.}{Eqs.}
\Crefname{equation}{Eq.}{Eqs.}
\newcommand{\missingcell}{\textcolor{gray}{--}}
\newcommand{\infeasiblecell}{\textcolor{gray}{NF}}
\newcommand{\statcell}[3]{\ensuremath{#1\if\relax\detokenize{#2}\relax\else_{\scriptscriptstyle\pm #2}\fi^{#3}}}
\newcommand{\beststatcell}[3]{\statcell{\mathbf{#1}}{#2}{#3}}
\newcommand{\secondstatcell}[3]{\statcell{\underline{#1}}{#2}{#3}}

\newsavebox{\tablefitbox}
\newcommand{\fittable}[1]{%
  \sbox{\tablefitbox}{#1}%
  \ifdim\wd\tablefitbox>\linewidth
    \resizebox{\linewidth}{!}{\usebox{\tablefitbox}}%
  \else
    \usebox{\tablefitbox}%
  \fi}
\newcommand{\systemname}{BELT\xspace}
\newcommand{\beltb}{\textsc{BELT-B-R}}
\newcommand{\belth}{\textsc{BELT-H-R}}

\newcommand{\about}{\ensuremath{\mathord{\sim}}}

\newcommand{\pnow}{p^{\mathrm{now}}}
\newcommand{\pend}{p^{\mathrm{end}}}
\newcommand{\plater}{p^{\mathrm{later}}}
\newcommand{\psame}{p^{\mathrm{same}}}
\newcommand{\E}{\mathbb{E}}
\newcommand{\R}{\mathbb{R}}
\newcommand{\ecoparagraph}[1]{\par\noindent\textbf{#1}\;}
\title{Same Beliefs, Different Policies: Learning Reusable Representations for Early Stopping in LLM Reasoning}
\author{Jingwei Ni\textsuperscript{1}\quad Dmitry Abulkhanov\textsuperscript{2}\quad Vladislav Smirnov\textsuperscript{2}\quad Markus Leippold\textsuperscript{3}\\
\textbf{Elliott Ash\textsuperscript{1}\quad Timothy Baldwin\textsuperscript{2,4}\quad Artem Shelmanov\textsuperscript{2}\quad Mrinmaya Sachan\textsuperscript{1}}\\
\mdseries\textsuperscript{1}ETH Z\"urich\quad \textsuperscript{2}MBZUAI\quad \textsuperscript{3}University of Z\"urich\quad \textsuperscript{4}The University of Melbourne}
\iclrfinalcopy
\begin{document}
\maketitle
\lhead{Preprint}
\begin{abstract}
Reasoning LLMs often continue reasoning after reaching a correct answer, wasting computation and sometimes replacing it with an incorrect one.
Deciding when to stop is a metareasoning problem: the model must weigh what further reasoning could achieve against its cost, given what it can already answer.
Direct utility training learns a stopping representation at a chosen token price; supporting new prices can require additional controller training.
We introduce \systemname{} (\underline{BEL}ief-based \underline{T}ermination), which separates outcome prediction from stopping control. It learns beliefs about current and future answers from a frozen LLM's hidden states, then fits a linear stopping policy on the frozen features.
The prediction targets are independent of token price, so one probe supports different accuracy--cost trade-offs through CPU policy refitting or threshold adjustment.
Averaged over five out-of-distribution (OOD) math benchmarks, \systemname{} reduces generated tokens by 41\% on Qwen3-1.7B~\citep{qwen3} and 51\% on OLMo-3-7B-Think~\citep{olmo2025olmo3} without loss of accuracy. The same math-trained BELT transfers to coding and agentic tasks without training on code or agentic data. In all three domains, its accuracy-cost trade-offs are competitive with or better than those of end-to-end controllers. On OLMo3, its probe has $21.2\times$ fewer parameters than one checkpoint of the strongest baseline, OS-Pruner, and needs $\about55\times$ fewer training GPU-hours than OS-Pruner's five price-specific checkpoints.
\unskip
\begingroup\renewcommand{\thefootnote}{\fnsymbol{footnote}}\footnote{Code and checkpoints: \url{https://github.com/EdisonNi-hku/BELT}}\addtocounter{footnote}{-1}\endgroup
\end{abstract}
\section{Introduction}

Reasoning large language models (LLMs) often overthink: they keep reasoning after the answer has stopped improving \citep{pmlr-v267-chen25bx,NEURIPS2025_3e22bea3}.
Yet deciding when to stop requires more than knowing how likely the current answer is to be correct.
A model may be uncertain about its current answer yet close to finding a better one.
The same uncertainty may also occur when further reasoning is unlikely to help \citep{muennighoff2025s1simpletesttimescaling,ni-etal-2026-reasoning}.
Even a correct answer can be overturned by continued reasoning \citep{wang2026rom}.
These situations call for different stopping decisions.
Stopping is therefore a metareasoning problem. The model must assess what further computation is likely to achieve and whether that improvement is worth its cost \citep{russell1991metareasoning,desabbata2025metareasoning}.

We separate outcome prediction from decision-making: first predict the outcomes of answering now or continuing, then learn when to act on those predictions.
The decision depends on the trade-off between accuracy and computation \citep{wang2024tokeneconomies}.
For a frozen reasoner with a fixed decoding rule, changing the token price changes how outcomes are valued, but not the outcome prediction targets at a given reasoning prefix.
The predictive representation can therefore be reused across prices instead of being learned anew for each stopping objective.

We study whether learning to predict reasoning outcomes yields effective, reusable representations for stopping control.
We introduce \systemname{} (\underline{BEL}ief-based \underline{T}ermination), which trains a compact probe (the BELT probe) \citep{uhead,ni-etal-2026-reprobe} on a frozen LLM's hidden states.
The probe learns a control representation by jointly predicting four outcomes.
Two reference outcomes are correctness when answering now (\emph{Now}, \citealp{zhang2025reasoning,yang2025deer,hosseini2026codestop}) and at natural completion (\emph{End}, \citealp{snell2024scaling}).
Continuing also preserves the option to stop before completion.
We therefore also predict whether a correct answer will be available at a later boundary (\emph{Later}, \citealp{wang2026rom}), even if the final answer is wrong.
Finally, stopping may preserve the natural-completion answer even when its correctness is uncertain.
We predict this agreement (\emph{Same}, \citealp{wu2025thought}).
We call these predicted outcome probabilities \emph{beliefs}.

\begin{figure}[t]\centering
    \includegraphics[width=\linewidth]{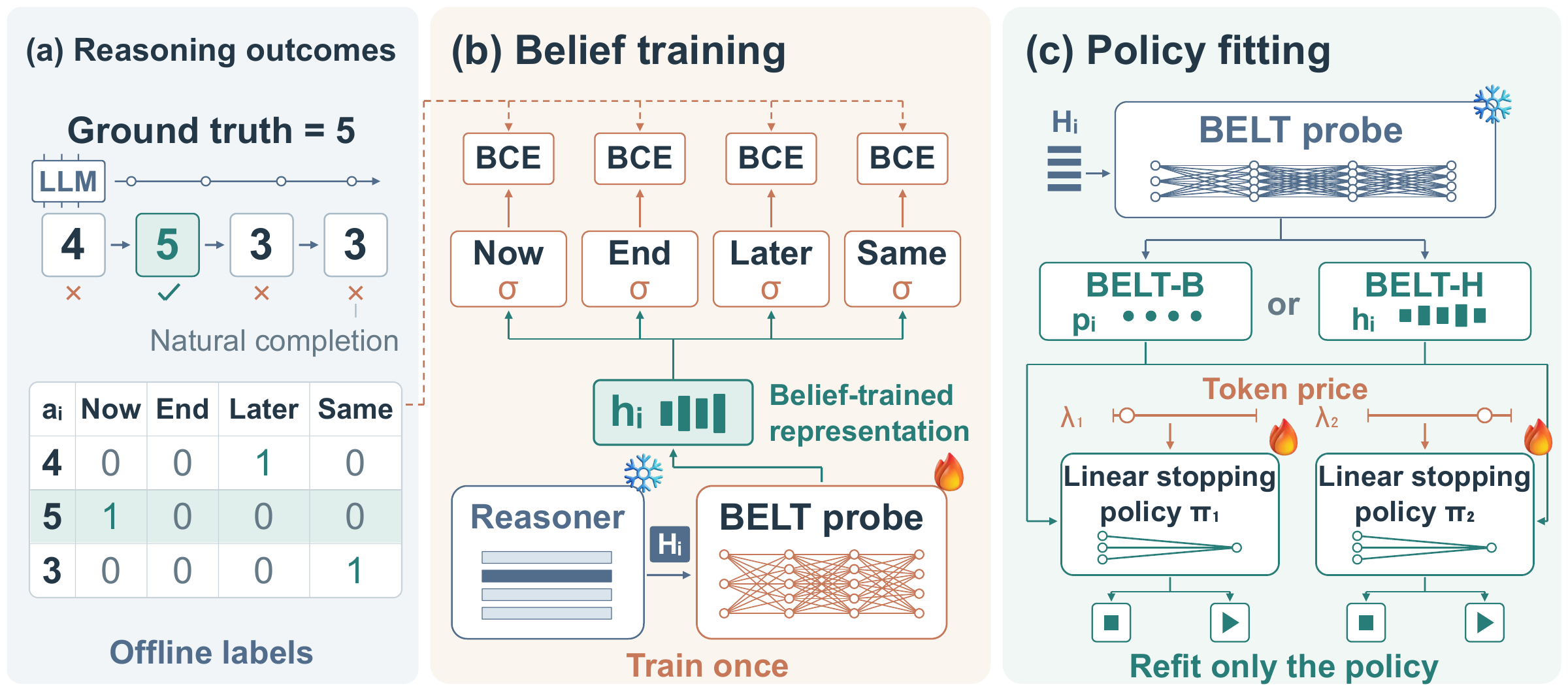}
    \caption{\textbf{\systemname overview.} (a) Label offline traces to supervise outcome beliefs. (b) Jointly train the BELT probe with four objectives of outcome prediction. (c) Fit stopping policies on frozen features for token prices $\lambda$; alternatively, fit once and vary its threshold. Both reuse one representation. B/H in panel (c) identify the feature interfaces.}
    \label{fig:method}
\end{figure}

During belief training, \systemname{} learns these predictions from offline outcome labels, independently of the token price (\cref{fig:method}).
During policy fitting, it freezes the probe and fits a linear stopping policy to the sequential stopping objective \citep{ehab2026ospruner}.
The policy uses either the explicit beliefs (\beltb{}) or the probe's learned representation (\belth{}).
At each boundary, it decides whether to stop and answer or continue and reassess at the next boundary.
The same probe supports trade-off control by refitting this small policy or shifting the threshold on one fitted direction.
Our evaluation tests the stopping performance and adaptation cost of this separation. Our contributions:

\textbf{(1) \systemname{} learns reusable representations for stopping control.}
It combines a jointly supervised probe with linear policies controlled by price-specific refitting or fixed-direction thresholds.
Averaged over five OOD math benchmarks, \systemname{} saves 41\% of generated tokens on Qwen3 and 51\% on OLMo3 without loss of accuracy.
BELT also beats the strongest baselines with significant parameter and computation advantages (\cref{sec:main-results}).
Trained only on mathematics, the same probes transfer to LiveCodeBench coding on both backbones and to the ALFWorld and ScienceWorld agent environments, without code or agent training (\cref{sec:main-results,sec:analysis-agentic}).
Reusing one probe across five token prices also reduces OLMo3 controller training by approximately $55\times$ in GPU-hours (\cref{sec:cost}).

\textbf{(2) Analyses show how outcome prediction supports stopping control.}
Belief-trained features give higher OOD stopping utility than raw-state controls and than features taken from an end-to-end stopping controller, at all five prices (\cref{sec:analysis-representation}).
Target and input ablations show that the \emph{Later} and \emph{Same} beliefs each improve stopping beyond \emph{Now} and \emph{End}
(\cref{sec:analysis-information}).
Policy diagnostics show that a new token price is accommodated by reweighting the same beliefs (\cref{sec:analysis-adaptation}).

\section{Background and Problem Formulation}
\label{sec:background}

\ecoparagraph{Problem formulation.}
We study the metareasoning decision of whether to continue an LLM's reasoning or stop and request an answer. For a trajectory $\tau$, let $i=1,\ldots,T-1$ index candidate stopping boundaries and let $T$ denote natural completion without early stopping. At boundary $i$, the controller observes $H_i$, the sequence of LLM hidden states at a selected layer up to that boundary. Stopping yields answer $a_i$, correctness $y_i\in\{0,1\}$ and total token cost $C_i$. Continuing advances to the next boundary. Reaching $T$ yields the final answer $a_T$, correctness $y_T$ and total cost $C_T$.

A \emph{stopping policy} with parameters $\xi$ assigns the conditional probability $\pi_i=\Pr(\text{stop at }i\mid\text{reached }i,H_i;\xi)$. Its reach probability $S_i$ and stopping probabilities $\rho_i$ are
\begin{equation}
S_i=\prod_{j<i}(1-\pi_j),\qquad
\rho_i=\begin{cases}S_i\pi_i,&i<T,\\S_T,&i=T,\end{cases}
\label{eq:stopping-probabilities}
\end{equation}
with $S_1=1$. Thus trajectories that continue past every stopping boundary complete at $T$, and $\sum_{i=1}^T\rho_i=1$. For token price $\lambda\geq0$, the \emph{utility} of stopping at $i$ is $u_i=y_i-\lambda C_i$. The stopping problem maximizes expected utility over trajectories and stopping decisions:
\begin{equation}
\max_\xi\mathcal U_\lambda(\xi),\qquad
\mathcal U_\lambda(\xi)=\E_{\tau\sim\mathcal D}\!\left[\sum_{i=1}^{T}\rho_i(\xi)u_i\right],
\label{eq:stopping-background}
\end{equation}
where $\mathcal D$ is the trajectory distribution induced by the questions, frozen reasoner and decoding rule. The token price $\lambda$ values correctness against computation, without imposing a hard token budget. For a fixed reasoner and decoding rule, changing $\lambda$ changes these values but not the outcome prediction targets at a given prefix.

\ecoparagraph{Direct utility training.}
Full rollouts and forced answers supply cached correctness labels and token costs for offline optimization of \cref{eq:stopping-background}. Jointly training the representation and stopping head on this utility ties representation learning to the token price $\lambda$ and the current policy \citep{ehab2026ospruner}. Adapting the representation to a new price can require another controller training run. Boundaries that the policy rarely reaches, or where its decisions saturate, provide weak learning signals even when outcome labels are available (\cref{sec:checks-direct}). BELT learns outcome predictions independently of the price and policy, then fits stopping decisions on the frozen features by optimizing utility.

\ecoparagraph{Outcome prediction with probes.}
A \emph{probe} predicts output properties from a frozen LLM's internal states. A \emph{belief} is an outcome probability conditioned on the observed prefix $H_i$. Following UHeads and ReProbe \citep{uhead,ni-etal-2026-reprobe}, BELT uses a causal encoder $h_i=E_\phi(H_i)$ and sigmoid readouts $\mathbf p_i=\sigma(W_{\rm bel}h_i+b_{\rm bel})$. \Cref{sec:probe-settings,sec:reader-efficiency} specify the architecture and its adaptations.

\section{BELT: Belief Training and Policy Fitting}
\label{sec:method}

BELT trains a price-independent outcome probe, then freezes it and fits linear stopping policies (\cref{fig:method}). It offers two ways to control the accuracy--cost trade-off: refit the policy for each token price, or vary the threshold on one fitted policy. Both reuse the same belief-trained features.

\ecoparagraph{Outcome beliefs.}\label{sec:beliefs}
Four outcomes describe the current answer and the prospects of continued reasoning, with binary labels at $i<T$:
\begin{equation}
\begin{aligned}
Y_i^{\rm now}&=y_i, & Y_i^{\rm end}&=y_T,\\
Y_i^{\rm later}&=\mathbf 1\{\exists j\in\{i+1,\ldots,T\}:y_j=1\},
& Y_i^{\rm same}&=\mathbf 1\{a_i=a_T\}.
\end{aligned}
\label{eq:belief-targets}
\end{equation}
Here $\mathbf 1\{\cdot\}$ is the indicator function. \emph{Now} and \emph{End} measure current and natural-completion correctness. \emph{Later} includes any subsequent correct answer, even if the final answer is wrong. \emph{Same} measures parsed-answer agreement with natural completion, regardless of correctness or agreement at intervening boundaries. Grading and comparing forced answers from completed rollouts supplies these labels (\cref{fig:method}(a)).

For each event $k\in\mathcal K=\{\mathrm{now},\mathrm{end},\mathrm{later},\mathrm{same}\}$, the probe estimates its probability from the observed prefix: $p_i^k\approx\Pr(Y_i^k=1\mid H_i)$.

The four targets summarize answer correctness and agreement relevant to the utility in \cref{eq:stopping-background}. Stopping at $i$ has expected utility $p_i^{\rm now}-\lambda\,\E[C_i\mid H_i]$. Continuing leaves natural completion as an option, with expected utility $p_i^{\rm end}-\lambda\,\E[C_T\mid H_i]$. The belief $p_i^{\rm later}$ indicates whether a later boundary could offer a correct answer; $p_i^{\rm same}$ predicts agreement with the natural-completion answer. Together, they distinguish a possible future correction (low \emph{Now}, high \emph{Later} or \emph{End}), likely final-answer agreement (high \emph{Same}), and a correct answer that may be lost (high \emph{Now}, low \emph{Same}). The value of continuing depends on future stopping decisions and the token price. Predicting the four trajectory outcomes keeps the probe reusable across prices (\cref{sec:cost}). Stage 2 learns the price-dependent trade-off from recorded correctness and token costs, with observed prefix length supplied to the Qwen3 BELT-B-R policy (\cref{sec:interfaces}). \Cref{sec:analysis-information} tests the contribution of \emph{Later} and \emph{Same}. Stage 1 learns these outcome predictions; Stage 2 uses them for stopping control.

\ecoparagraph{Stage 1: Belief training.}\label{sec:belief_training}
The BELT probe first projects each token's hidden state to width 768 with an MLP. Learned marker embeddings distinguish block-end tokens, and two causal Transformer layers encode the observed prefix (\cref{fig:method}(b)). We read the current block-end representation $h_i\in\R^{768}$ and apply four linear heads followed by sigmoids to obtain $\mathbf p_i=(\pnow_i,\pend_i,\plater_i,\psame_i)^\top$. With the reasoner frozen, we jointly train the shared encoder and readouts $\Theta=(\phi,W_{\rm bel},b_{\rm bel})$ on labelled prefixes using equally weighted binary cross-entropy:
\begin{equation}
\Theta^\star=\arg\min_\Theta\mathcal L_{\rm bel},\qquad
\mathcal L_{\rm bel}=\frac{1}{4}\sum_{k\in\mathcal K}
\E_{\tau\sim\mathcal D_{\rm bel}}\!\left[
\underset{i\in\mathcal I_k(\tau)}{\operatorname{mean}}
\operatorname{BCE}(p_i^k,Y_i^k)\right].
\label{eq:belief-training}
\end{equation}
Here $\mathcal D_{\rm bel}$ is the empirical training-trajectory distribution and $\mathcal I_k(\tau)$ contains the labelled positions for event $k$, unweighted by policy reach probabilities. The loss and validation checkpoint selection are independent of $\lambda$; the selected probe $\Theta^\star$ is frozen for Stage 2.

\ecoparagraph{Stage 2: Policy fitting.}\label{sec:interfaces}
We apply the frozen probe to a trajectory bank with recorded correctness $y_i$ and cost $C_i$. Two feature interfaces supply the policy inputs (\cref{fig:method}(c)): B uses the four beliefs and observed prefix length; H uses their shared representation. The OLMo3 B interface omits prefix length (\cref{sec:olmo-replication}). Both use a linear logistic stopping policy:
\begin{equation}
x_i^{\rm B}=\operatorname{Std}([\operatorname{logit}(\mathbf p_i);d_i]),\qquad
x_i^{\rm H}=\operatorname{Std}(h_i),\qquad
\pi_i=\sigma(w^\top x_i+b).
\label{eq:policy}
\end{equation}
Here $[\,;\,]$ concatenates features, $d_i$ is the observed prefix-token count divided by 12,000, and $\operatorname{Std}$ uses means and standard deviations from the fitting bank. The elementwise transform $\operatorname{logit}(p)=\log[p/(1-p)]$ uses clipped probabilities. %
The Qwen3 price-refit variant samples a stop decision from $\operatorname{Bernoulli}(\pi_i)$ at each boundary; otherwise, reasoning continues until the next boundary or natural completion.

For $\xi=(w,b)$ and the empirical bank distribution $\mathcal D_{\rm bank}$ in \cref{eq:stopping-background}, we fit
\begin{equation}
(w_\lambda,b_\lambda)=\arg\max_{w,b}J_\lambda(w,b),\qquad
J_\lambda(w,b)=\mathcal U_\lambda(w,b)-\kappa\lVert w\rVert_2^2.
\label{eq:utility}
\end{equation}
The empirical utility averages $\sum_i\rho_i(y_i-\lambda C_i)$ over banked trajectories, including natural completion, with $\rho_i$ from \cref{eq:stopping-probabilities}. It marginalizes stopping decisions exactly, conditional on the recorded trajectories and their fixed features, labels and costs. Development utility selects $\kappa\geq0$ and the optimization iterate (\cref{sec:probe-settings}).

\ecoparagraph{Two variants for controlling trade-offs.}\label{sec:tradeoff-variants}
\textbf{Refit (BELT-B-R/H-R)} solves \cref{eq:utility} at each price, allowing the direction, intercept and sigmoid sharpness to change. \textbf{Threshold (BELT-B-T/H-T)} solves it once at a reference price $\lambda_0$, then varies $t$ in the hard stopping rule
\begin{equation}
\pi_i^{(t)}=\mathbf1\{w_{\lambda_0}^{\top}x_i+b_{\lambda_0}\geq t\}.
\label{eq:threshold-policy}
\end{equation}
We call these variants BELT-R (refit) and BELT-T (threshold); B/H specify the feature interface. Both variants share one belief encoder and the same supervision. \Cref{sec:analysis-single-direction} compares the two families; threshold grids and the OLMo3 implementation appear in \cref{sec:setup,sec:olmo-replication}.

\section{Experimental Setup}
\label{sec:setup}
We evaluate stopping performance across models and tasks (\cref{sec:main-results}). We then compare representations under common policy fitting and examine the effects of supervision and price adaptation (\cref{sec:analysis}).
\ecoparagraph{Training data.}
We train the BELT probe on mathematics questions from DAPO-Math-17k \citep{yu2025dapo}, NuminaMath-1.5 \citep{numina_math_datasets} and DeepScaleR-Preview \citep{deepscaler2025}. Qwen3 uses 3,918 training questions and up to eight trajectories per question; OLMo3 uses up to four. A separate 320-question bank selects the probe checkpoint; disjoint 160-question halves then fit linear policies and select their hyperparameters (\cref{eq:policy,eq:utility}). \Cref{sec:dataset-recipe,sec:training-recipe,sec:olmo-replication} give the dataset composition and filtering rules. Main BELT probes are randomly initialized (three seeds per backbone). We force and grade an answer every four reasoning blocks for belief supervision (\cref{sec:belief_training}; \cref{fig:method}(a)).

\ecoparagraph{Evaluation data.}
ID sets contain held-out DAPO-Math, NuminaMath and DeepScaleR questions (DAPO, Numina and DSR). OOD sets are MATH-500 (M500) \citep{ICLR2024_aca97732}, OlympiadBench (Olymp.) \citep{he-etal-2024-olympiadbench}, Gaokao2023-Math-En (GK) \citep{liao-etal-2024-mario}, AIME 2022--2025 \citep{maa_aime}, and AMC 2022--2023 \citep{maa_amc}. We test BELT and OS-Pruner (OSP) variants \citep{ehab2026ospruner} on LiveCodeBench (LCB) \citep{ICLR2025_94074dd5} (Pre: before August 2024; Post: August 2024 onward), transferring to coding without code-side training (\cref{sec:code-transfer}). We also apply the math-trained controllers to the per-turn thinking of an OLMo3 agent in ALFWorld and ScienceWorld text games (\cref{sec:alfworld-protocol,sec:sciworld-protocol}).

\ecoparagraph{Metrics.} Following \citet{wu2025thought,ehab2026ospruner}, we compare test-frontier accuracy at token savings of 70\%, 60\%, 50\%, and 40\% relative to natural completion, and maximum saving at accuracy tolerances of 0, 1, and 2 percentage points. For each method and training run, distribution-level micro accuracy and total token cost define the frontier; interpolation between adjacent nondominated policies matches these constraints. Main macro readings evaluate those same selected policies or mixtures. We complement this comparison with split-half sensitivity checks and test utility, $\mathrm{Accuracy}-\lambda\,\mathrm{TokenCost}$, of policies selected by development utility at each $\lambda$ (dev-selected utility). Full protocols appear in \cref{sec:metric-protocol,sec:testcurve-reading,sec:checks-reading}; the agent tasks are read on task-specific budget ranges where their trade-offs lie (\cref{sec:alfworld-protocol,sec:sciworld-protocol}).

\ecoparagraph{Natural-completion grading.} If a mathematics trajectory closes its thinking without a parseable answer (16\% of the closed Qwen3 ID report chains and 10.4\% of the OOD chains), we make one seeded \texttt{\textbackslash boxed\{} request of at most 32 tokens from the completed prefix, grade the returned answer as the natural-completion answer, and charge its tokens to natural-completion cost. This recovers an answer for most such trajectories. LiveCodeBench chains always contain a program and need no request. \Cref{sec:natural-elicitation} gives the procedure and per-bank statistics.

\ecoparagraph{LLMs and baseline selection.}
We evaluate Qwen3-1.7B (Qwen3, reasoning mode) and OLMo-3-7B-Think (OLMo3), reading layers 16 and 18. Fixed caps, CoDE-Stop \citep{hosseini2026codestop} and DEER \citep{yang2025deer} test training-free stopping; Thought Calibration (TC) \citep{wu2025thought} and ROM \citep{wang2026rom} provide trained correctness/stability controls. We include OSP \citep{ehab2026ospruner} with five price-specific checkpoints on Qwen3 and OLMo3, and OSP-1 with one checkpoint and 13 score-quantile thresholds. OSP-1 is the primary comparator under a single-encoder constraint: like BELT-R/T (\cref{sec:tradeoff-variants}), it reuses one learned hidden-state encoder across operating points. Full OSP provides a five-encoder performance reference with greater training and storage requirements (\cref{sec:cost}). On Qwen3, refit has 17 stochastic policies; threshold fits at $\lambda_0=10^{-5}$ and uses 13 source-validation thresholds, as does OSP-1. On OLMo3, $\lambda_0=2\times10^{-5}$; refit retains its registered 221-member family, including per-direction thresholds (\cref{sec:olmo-replication}). Candidate counts differ across recipes; further OSP/BELT threshold expansion is reported in \cref{sec:threshold-expansion}. Probe Params excludes the reasoner; M denotes millions.

\section{Main Results}
\label{sec:main-results}

\begin{table}[t]
\caption{Token saving (\%) at Qwen3 accuracy tolerances (macro averages). Cells show mean with training-seed SD, or question-bootstrap SE ($\ddagger$); bold/underline mark best/second-best means. Probe Params excludes the frozen reasoner (M: millions). R/T denote refit/threshold control. OSP/OSP-1 extensions appear in Appendix \cref{sec:threshold-expansion}.}\label{tab:allbaselines}
\begin{center}
\begingroup\setlength{\tabcolsep}{2.5pt}
\begin{tabular}{@{}l|c|cc|cc|cc@{}}\toprule
\multirow{2}{*}{\textbf{Method}} & \multirow{2}{*}{\shortstack{\textbf{Probe}\\\textbf{Params}}} & \multicolumn{2}{c|}{\textbf{Acc. tolerance: 0 pp}} & \multicolumn{2}{c|}{\textbf{Acc. tolerance: 1 pp}} & \multicolumn{2}{c}{\textbf{Acc. tolerance: 2 pp}} \\
 &  & \textbf{ID} & \textbf{OOD} & \textbf{ID} & \textbf{OOD} & \textbf{ID} & \textbf{OOD} \\
\midrule
CoDE-Stop$^{\ddagger}$ & 0 & \cellcolor{Green3!18}\statcell{16.61}{0.50}{}& \cellcolor{Green3!24}\statcell{16.94}{0.75}{}& \cellcolor{Green3!2}\statcell{21.94}{0.51}{}& \cellcolor{Green3!16}\statcell{26.02}{0.78}{}& \cellcolor{Green3!1}\statcell{24.85}{0.51}{}& \cellcolor{Green3!15}\statcell{30.15}{0.79}{} \\
DEER$^{\ddagger}$ & 0 & \cellcolor{Green3!13}\statcell{12.77}{0.59}{}& \cellcolor{Green3!14}\statcell{10.36}{0.84}{}& \cellcolor{Green3!0}\statcell{20.96}{0.69}{}& \cellcolor{Green3!3}\statcell{20.07}{1.01}{}& \cellcolor{Green3!0}\statcell{24.22}{0.71}{}& \cellcolor{Green3!0}\statcell{23.19}{1.05}{} \\
\midrule
TC & 514 & \cellcolor{Green3!26}\statcell{22.15}{0.66}{}& \cellcolor{Green3!28}\statcell{19.97}{0.15}{}& \cellcolor{Green3!14}\statcell{28.12}{0.43}{}& \cellcolor{Green3!15}\statcell{25.46}{0.27}{}& \cellcolor{Green3!16}\statcell{32.24}{0.55}{}& \cellcolor{Green3!16}\statcell{30.45}{1.50}{} \\
ROM & 5M & \cellcolor{Green3!0}\statcell{3.50}{4.95}{}& \cellcolor{Green3!0}\statcell{1.22}{1.72}{}& \cellcolor{Green3!0}\statcell{21.01}{1.85}{}& \cellcolor{Green3!0}\statcell{18.76}{1.17}{}& \cellcolor{Green3!5}\statcell{26.57}{0.47}{}& \cellcolor{Green3!2}\statcell{24.00}{0.83}{} \\
OSP & 101M $\times 5$ & \cellcolor{Green3!58}\secondstatcell{44.77}{0.75}{}& \cellcolor{Green3!37}\statcell{25.92}{22.46}{}& \cellcolor{Green3!58}\secondstatcell{49.72}{1.09}{}& \cellcolor{Green3!58}\statcell{45.19}{1.69}{}& \cellcolor{Green3!60}\secondstatcell{54.15}{0.90}{}& \cellcolor{Green3!60}\beststatcell{50.68}{2.04}{} \\
OSP-1 & 101M & \cellcolor{Green3!50}\statcell{39.19}{5.46}{}& \cellcolor{Green3!12}\statcell{9.07}{15.71}{}& \cellcolor{Green3!52}\statcell{46.59}{2.01}{}& \cellcolor{Green3!40}\statcell{36.89}{0.18}{}& \cellcolor{Green3!54}\statcell{51.50}{1.49}{}& \cellcolor{Green3!46}\statcell{44.37}{0.81}{} \\
\midrule
\textbf{BELT-B-T} & 16M & \cellcolor{Green3!30}\statcell{25.21}{1.08}{}& \cellcolor{Green3!36}\statcell{25.22}{1.01}{}& \cellcolor{Green3!17}\statcell{29.57}{1.50}{}& \cellcolor{Green3!27}\statcell{30.87}{0.92}{}& \cellcolor{Green3!19}\statcell{33.54}{1.74}{}& \cellcolor{Green3!28}\statcell{36.01}{0.76}{} \\
\textbf{BELT-H-T} & 16M & \cellcolor{Green3!60}\beststatcell{46.52}{1.51}{}& \cellcolor{Green3!60}\beststatcell{40.77}{1.46}{}& \cellcolor{Green3!60}\beststatcell{50.64}{0.98}{}& \cellcolor{Green3!60}\beststatcell{45.98}{0.60}{}& \cellcolor{Green3!60}\beststatcell{54.29}{0.79}{}& \cellcolor{Green3!58}\statcell{49.82}{0.63}{} \\
\textbf{BELT-B-R} & 16M & \cellcolor{Green3!57}\statcell{44.45}{3.00}{}& \cellcolor{Green3!54}\statcell{37.14}{1.95}{}& \cellcolor{Green3!56}\statcell{48.74}{1.90}{}& \cellcolor{Green3!55}\statcell{43.76}{1.77}{}& \cellcolor{Green3!57}\statcell{52.58}{0.60}{}& \cellcolor{Green3!56}\statcell{48.74}{0.62}{} \\
\textbf{BELT-H-R} & 16M & \cellcolor{Green3!56}\statcell{43.76}{0.98}{}& \cellcolor{Green3!57}\secondstatcell{39.11}{0.63}{}& \cellcolor{Green3!56}\statcell{48.58}{0.44}{}& \cellcolor{Green3!58}\secondstatcell{45.25}{1.12}{}& \cellcolor{Green3!56}\statcell{52.28}{0.25}{}& \cellcolor{Green3!59}\secondstatcell{50.26}{1.09}{} \\
\bottomrule\end{tabular}\endgroup
\end{center}
\end{table}

\begin{table}[t]
\caption{Accuracy (\%) at Qwen3 token-saving budgets (macro averages); conventions follow \cref{tab:allbaselines}. Natural$^\dagger$ is the natural-completion reference; NF denotes an infeasible budget in at least one seed. OSP/OSP-1 threshold extensions and per-dataset results appear in Appendix \cref{sec:threshold-expansion,sec:extended-results}.}\label{tab:macro}
\begin{center}
\begingroup\setlength{\tabcolsep}{2.5pt}
\resizebox{\linewidth}{!}{%
\begin{tabular}{@{}l|c|cc|cc|cc|cc@{}}\toprule
\multirow{2}{*}{\textbf{Method}} & \multirow{2}{*}{\shortstack{\textbf{Probe}\\\textbf{Params}}} & \multicolumn{2}{c|}{\textbf{Saving: 70\%}} & \multicolumn{2}{c|}{\textbf{Saving: 60\%}} & \multicolumn{2}{c|}{\textbf{Saving: 50\%}} & \multicolumn{2}{c}{\textbf{Saving: 40\%}} \\
 &  & \textbf{ID} & \textbf{OOD} & \textbf{ID} & \textbf{OOD} & \textbf{ID} & \textbf{OOD} & \textbf{ID} & \textbf{OOD} \\
\midrule
Natural$^{\dagger\ddagger}$ & -- & \statcell{53.38}{1.1}{}& \statcell{66.66}{1.4}{}& \statcell{53.38}{1.1}{}& \statcell{66.66}{1.4}{}& \statcell{53.38}{1.1}{}& \statcell{66.66}{1.4}{}& \statcell{53.38}{1.1}{}& \statcell{66.66}{1.4}{} \\
Fixed cap$^{\ddagger}$ & 0 & \cellcolor{Green3!24}\statcell{31.66}{1.0}{}& \cellcolor{Green3!34}\statcell{44.05}{1.3}{}& \cellcolor{Green3!24}\statcell{37.08}{1.0}{}& \cellcolor{Green3!35}\statcell{50.37}{1.3}{}& \cellcolor{Green3!24}\statcell{42.32}{1.0}{}& \cellcolor{Green3!34}\statcell{55.38}{1.4}{}& \cellcolor{Green3!16}\statcell{46.29}{1.0}{}& \cellcolor{Green3!25}\statcell{58.89}{1.4}{} \\
CoDE-Stop$^{\ddagger}$ & 0 & \cellcolor{Green3!29}\statcell{33.29}{1.0}{}& \cellcolor{Green3!38}\statcell{46.26}{1.3}{}& \cellcolor{Green3!31}\statcell{39.46}{1.0}{}& \cellcolor{Green3!41}\statcell{53.30}{1.4}{}& \cellcolor{Green3!32}\statcell{44.71}{1.0}{}& \cellcolor{Green3!39}\statcell{57.12}{1.4}{}& \cellcolor{Green3!22}\statcell{47.32}{1.0}{}& \cellcolor{Green3!33}\statcell{60.77}{1.4}{} \\
DEER$^{\ddagger}$ & 0 & \cellcolor{Green3!0}\statcell{22.33}{0.8}{}& \cellcolor{Green3!0}\statcell{24.98}{1.1}{}& \cellcolor{Green3!0}\statcell{28.76}{0.8}{}& \cellcolor{Green3!0}\statcell{32.75}{1.1}{}& \cellcolor{Green3!0}\statcell{35.73}{0.9}{}& \cellcolor{Green3!0}\statcell{42.18}{1.2}{}& \cellcolor{Green3!0}\statcell{43.29}{1.0}{}& \cellcolor{Green3!0}\statcell{52.97}{1.3}{} \\
\midrule
TC & 514 & \cellcolor{Green3!23}\statcell{31.08}{0.5}{}& \cellcolor{Green3!32}\statcell{42.86}{0.4}{}& \cellcolor{Green3!31}\statcell{39.43}{0.4}{}& \cellcolor{Green3!38}\statcell{51.53}{0.7}{}& \cellcolor{Green3!33}\statcell{44.92}{0.1}{}& \cellcolor{Green3!40}\statcell{57.42}{0.5}{}& \cellcolor{Green3!31}\statcell{48.96}{0.3}{}& \cellcolor{Green3!38}\statcell{61.83}{0.1}{} \\
ROM & 5M & \cellcolor{Green3!18}\statcell{29.16}{1.5}{}& \cellcolor{Green3!26}\statcell{39.78}{2.9}{}& \cellcolor{Green3!19}\statcell{35.40}{1.3}{}& \cellcolor{Green3!28}\statcell{46.87}{2.1}{}& \cellcolor{Green3!19}\statcell{40.93}{1.0}{}& \cellcolor{Green3!28}\statcell{53.02}{1.4}{}& \cellcolor{Green3!15}\statcell{46.00}{0.7}{}& \cellcolor{Green3!23}\statcell{58.27}{0.7}{} \\
OSP & 101M $\times 5$ & \cellcolor{Green3!59}\secondstatcell{44.83}{0.1}{}& \cellcolor{Green3!60}\secondstatcell{58.91}{0.65}{}& \cellcolor{Green3!60}\statcell{49.24}{0.3}{}& \cellcolor{Green3!60}\beststatcell{62.58}{0.2}{}& \cellcolor{Green3!60}\secondstatcell{52.44}{0.3}{}& \cellcolor{Green3!60}\beststatcell{65.25}{0.3}{}& \cellcolor{Green3!60}\secondstatcell{54.41}{0.2}{}& \cellcolor{Green3!59}\secondstatcell{66.76}{0.1}{} \\
OSP-1 & 101M & \cellcolor{Green3!58}\statcell{44.73}{0.1}{}& \cellcolor{Green3!60}\beststatcell{58.93}{0.3}{}& \cellcolor{Green3!58}\statcell{48.72}{0.1}{}& \cellcolor{Green3!59}\statcell{62.10}{0.3}{}& \cellcolor{Green3!56}\statcell{51.33}{0.3}{}& \cellcolor{Green3!58}\statcell{64.30}{0.0}{}& \cellcolor{Green3!53}\statcell{53.20}{0.4}{}& \cellcolor{Green3!54}\statcell{65.73}{0.1}{} \\
\midrule
\textbf{BELT-B-T} & 16M & \cellcolor{Green3!39}\statcell{37.46}{2.1}{}& \cellcolor{Green3!44}\statcell{50.01}{2.6}{}& \cellcolor{Green3!42}\statcell{43.11}{1.4}{}& \cellcolor{Green3!47}\statcell{56.20}{1.2}{}& \cellcolor{Green3!41}\statcell{47.14}{0.8}{}& \cellcolor{Green3!46}\statcell{60.04}{0.7}{}& \cellcolor{Green3!37}\statcell{50.09}{0.5}{}& \cellcolor{Green3!42}\statcell{62.80}{0.4}{} \\
\textbf{BELT-H-T} & 16M & \cellcolor{Green3!56}\statcell{43.81}{0.3}{}& \cellcolor{Green3!54}\statcell{55.50}{0.7}{}& \cellcolor{Green3!60}\beststatcell{49.34}{0.4}{}& \cellcolor{Green3!58}\statcell{61.75}{0.3}{}& \cellcolor{Green3!60}\beststatcell{52.49}{0.2}{}& \cellcolor{Green3!60}\secondstatcell{65.12}{0.0}{}& \cellcolor{Green3!60}\beststatcell{54.42}{0.2}{}& \cellcolor{Green3!60}\beststatcell{67.09}{0.2}{} \\
\textbf{BELT-B-R} & 16M & \cellcolor{Green3!58}\statcell{44.65}{0.1}{}& \cellcolor{Green3!59}\statcell{58.42}{0.2}{}& \cellcolor{Green3!60}\secondstatcell{49.32}{0.1}{}& \cellcolor{Green3!60}\secondstatcell{62.53}{0.3}{}& \cellcolor{Green3!58}\statcell{51.90}{0.2}{}& \cellcolor{Green3!59}\statcell{64.93}{0.1}{}& \cellcolor{Green3!58}\statcell{53.99}{0.4}{}& \cellcolor{Green3!58}\statcell{66.66}{0.4}{} \\
\textbf{BELT-H-R} & 16M & \cellcolor{Green3!60}\beststatcell{45.39}{0.1}{}& \cellcolor{Green3!59}\statcell{58.47}{0.6}{}& \cellcolor{Green3!60}\statcell{49.27}{0.1}{}& \cellcolor{Green3!60}\statcell{62.42}{0.4}{}& \cellcolor{Green3!59}\statcell{52.12}{0.1}{}& \cellcolor{Green3!59}\statcell{64.90}{0.2}{}& \cellcolor{Green3!58}\statcell{54.05}{0.2}{}& \cellcolor{Green3!58}\statcell{66.73}{0.1}{} \\
\bottomrule\end{tabular}}

\endgroup
\end{center}
\end{table}

\begin{figure}[t]
\centering
\includegraphics[width=0.95\textwidth]{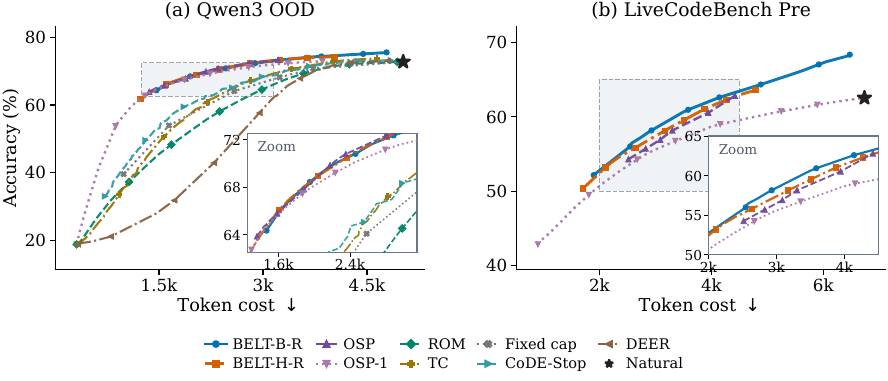}
\caption{Micro accuracy--token cost frontiers for Qwen3 OOD and LiveCodeBench Pre. Stars mark natural completion. Additional curves: datasets (\cref{fig:tradeoff-datasets}), Qwen3 ID and code Post (\cref{fig:tradeoff-full,fig:code-post}).}
\label{fig:tradeoff-main}
\end{figure}

\ecoparagraph{BELT matches full OSP within seed variation and exceeds OSP-1 on Qwen3.}
At a one-point accuracy tolerance, BELT saves 50.64\% of tokens on Qwen3 ID and 45.98\% on OOD, against OSP-1's 46.59\% and 36.89\%. These readings are within seed variation of full OSP's 49.72\% and 45.19\% (\cref{tab:allbaselines,tab:reb-belt-vs-baselines}). At natural accuracy BELT saves 40.77\% on OOD versus OSP's 25.92\%, which has high seed variation. Paired question-bootstrap intervals for the pre-registered comparisons are in \cref{tab:reb-bootstrap}. At the 40\% and 50\% saving budgets, BELT's accuracy is within 0.32~pp of full OSP's on ID and OOD and above OSP-1's at both budgets (\cref{tab:macro}); BELT also exceeds OSP-1 in all ten ID/OOD dev-selected utility readings (\cref{tab:source-transfer}).

\begin{table}[t]
\caption{\textbf{Top:} OLMo3 OOD macro averages. \textbf{Bottom:} math-to-code transfer to LiveCodeBench Pre. Conventions follow \cref{tab:allbaselines,tab:macro}; trained methods use three runs. OSP/OSP-1 and BELT threshold extensions appear in Appendix \cref{sec:threshold-expansion}.}\label{tab:olmo-main}
\begin{center}
\begingroup\setlength{\tabcolsep}{2.5pt}
\resizebox{\linewidth}{!}{%
\begin{tabular}{@{}l|c|ccc|cccc@{}}\toprule
\multirow{2}{*}{\textbf{Method}} & \multirow{2}{*}{\shortstack{\textbf{Probe}\\\textbf{Params}}} & \multicolumn{3}{c|}{\textbf{Token saving (\%) at tolerance}} & \multicolumn{4}{c}{\textbf{Accuracy (\%) at saving}} \\
 &  & \textbf{0.00} & \textbf{0.01} & \textbf{0.02} & \textbf{70\%} & \textbf{60\%} & \textbf{50\%} & \textbf{40\%} \\
\midrule
Natural$^{\dagger\ddagger}$ & -- & \missingcell& \missingcell& \missingcell& \statcell{78.63}{}{}& \statcell{78.63}{}{}& \statcell{78.63}{}{}& \statcell{78.63}{}{} \\
Fixed cap$^{\ddagger}$ & 0 & \missingcell& \missingcell& \missingcell& \cellcolor{Green3!33}\statcell{53.88}{1.5}{}& \cellcolor{Green3!31}\statcell{60.26}{1.5}{}& \cellcolor{Green3!29}\statcell{65.82}{1.5}{}& \cellcolor{Green3!28}\statcell{70.67}{1.4}{} \\
CoDE-Stop$^{\ddagger}$ & 0 & \cellcolor{Green3!0}\statcell{0.00}{0.0}{}& \cellcolor{Green3!8}\statcell{8.64}{0.3}{}& \cellcolor{Green3!8}\statcell{17.27}{0.6}{}& \cellcolor{Green3!26}\statcell{49.19}{1.4}{}& \cellcolor{Green3!23}\statcell{55.71}{1.4}{}& \cellcolor{Green3!19}\statcell{61.34}{1.5}{}& \cellcolor{Green3!16}\statcell{66.23}{1.4}{} \\
DEER$^{\ddagger}$ & 0 & \cellcolor{Green3!0}\statcell{0.00}{0.0}{}& \cellcolor{Green3!0}\statcell{1.57}{0.3}{}& \cellcolor{Green3!0}\statcell{10.80}{0.7}{}& \cellcolor{Green3!5}\statcell{35.65}{1.2}{}& \cellcolor{Green3!2}\statcell{44.25}{1.1}{}& \cellcolor{Green3!0}\statcell{52.85}{1.2}{}& \cellcolor{Green3!0}\statcell{60.52}{1.3}{} \\
\midrule
TC & 514 & \cellcolor{Green3!28}\statcell{24.27}{0.1}{}& \cellcolor{Green3!31}\statcell{29.16}{0.0}{}& \cellcolor{Green3!28}\statcell{33.21}{0.4}{}& \cellcolor{Green3!34}\statcell{54.02}{0.2}{}& \cellcolor{Green3!36}\statcell{63.04}{0.2}{}& \cellcolor{Green3!35}\statcell{68.50}{0.1}{}& \cellcolor{Green3!39}\statcell{74.33}{0.1}{} \\
ROM & 8M & \cellcolor{Green3!34}\statcell{29.17}{3.8}{}& \cellcolor{Green3!38}\statcell{35.63}{1.2}{}& \cellcolor{Green3!34}\statcell{37.72}{0.9}{}& \cellcolor{Green3!0}\statcell{32.17}{1.2}{}& \cellcolor{Green3!0}\statcell{43.08}{0.4}{}& \cellcolor{Green3!15}\statcell{59.79}{0.5}{}& \cellcolor{Green3!40}\statcell{74.78}{0.6}{} \\
OSP & 405M $\times 5$ & \cellcolor{Green3!53}\statcell{45.92}{0.6}{}& \cellcolor{Green3!54}\statcell{49.86}{0.3}{}& \cellcolor{Green3!54}\statcell{53.73}{0.5}{}& \cellcolor{Green3!58}\secondstatcell{69.80}{0.3}{}& \cellcolor{Green3!58}\statcell{75.09}{0.4}{}& \cellcolor{Green3!56}\statcell{78.26}{0.1}{}& \cellcolor{Green3!56}\statcell{80.66}{0.1}{} \\
OSP-1 & 405M & \cellcolor{Green3!46}\statcell{39.83}{1.4}{}& \cellcolor{Green3!51}\statcell{47.54}{1.9}{}& \cellcolor{Green3!52}\statcell{52.45}{1.7}{}& \cellcolor{Green3!58}\statcell{69.74}{0.6}{}& \cellcolor{Green3!58}\statcell{75.32}{0.8}{}& \cellcolor{Green3!56}\statcell{78.02}{0.4}{}& \cellcolor{Green3!53}\statcell{79.36}{0.1}{} \\
\midrule
\textbf{BELT-B-T} & 19M & \cellcolor{Green3!58}\statcell{50.14}{2.1}{}& \cellcolor{Green3!58}\statcell{53.09}{1.9}{}& \cellcolor{Green3!56}\statcell{55.26}{1.7}{}& \cellcolor{Green3!51}\statcell{65.16}{2.2}{}& \cellcolor{Green3!58}\statcell{75.19}{1.1}{}& \cellcolor{Green3!59}\secondstatcell{79.60}{1.0}{}& \cellcolor{Green3!59}\secondstatcell{81.51}{0.8}{} \\
\textbf{BELT-H-T} & 19M & \cellcolor{Green3!60}\secondstatcell{51.22}{3.0}{}& \cellcolor{Green3!59}\statcell{54.09}{3.0}{}& \cellcolor{Green3!57}\statcell{56.51}{3.2}{}& \cellcolor{Green3!52}\statcell{65.74}{3.8}{}& \cellcolor{Green3!58}\statcell{75.14}{2.1}{}& \cellcolor{Green3!58}\statcell{79.04}{0.9}{}& \cellcolor{Green3!57}\statcell{81.03}{0.6}{} \\
\textbf{BELT-B-R} & 19M & \cellcolor{Green3!59}\statcell{50.61}{1.5}{}& \cellcolor{Green3!59}\secondstatcell{54.30}{2.9}{}& \cellcolor{Green3!60}\secondstatcell{58.15}{0.9}{}& \cellcolor{Green3!56}\statcell{68.68}{1.1}{}& \cellcolor{Green3!60}\beststatcell{76.48}{0.6}{}& \cellcolor{Green3!60}\beststatcell{80.05}{0.7}{}& \cellcolor{Green3!60}\beststatcell{82.04}{0.3}{} \\
\textbf{BELT-H-R} & 19M & \cellcolor{Green3!60}\beststatcell{51.59}{2.3}{}& \cellcolor{Green3!60}\beststatcell{55.22}{2.4}{}& \cellcolor{Green3!60}\beststatcell{58.52}{2.0}{}& \cellcolor{Green3!60}\beststatcell{71.11}{0.4}{}& \cellcolor{Green3!60}\secondstatcell{76.30}{0.5}{}& \cellcolor{Green3!58}\statcell{79.19}{0.4}{}& \cellcolor{Green3!57}\statcell{81.05}{0.6}{} \\
\bottomrule\end{tabular}}
\endgroup
\end{center}

\label{tab:code-main}\vspace{2pt}
\begin{center}
\begingroup\setlength{\tabcolsep}{2pt}
\resizebox{\linewidth}{!}{%
\begin{tabular}{@{}lc|ccc|cccc@{}}\toprule
\multirow{2}{*}{\textbf{Method}} & \multirow{2}{*}{\shortstack{\textbf{Probe}\\\textbf{Params}}} & \multicolumn{3}{c|}{\textbf{Token saving (\%) at tolerance}} & \multicolumn{4}{c}{\textbf{Accuracy (\%) at saving}} \\
 &  & \textbf{0.00} & \textbf{0.01} & \textbf{0.02} & \textbf{70\%} & \textbf{60\%} & \textbf{50\%} & \textbf{40\%} \\
\midrule
Natural$^{\dagger}$ & -- & \missingcell& \missingcell& \missingcell& \statcell{62.52}{}{}& \statcell{62.52}{}{}& \statcell{62.52}{}{}& \statcell{62.52}{}{} \\
\midrule
OSP & 101M $\times 5$ & \cellcolor{Green3!49}\statcell{34.76}{3.37}{}& \cellcolor{Green3!43}\statcell{38.25}{2.80}{}& \cellcolor{Green3!39}\statcell{41.74}{2.78}{}& \infeasiblecell& \cellcolor{Green3!13}\statcell{55.10}{0.74}{}& \cellcolor{Green3!24}\statcell{58.39}{0.81}{}& \cellcolor{Green3!31}\statcell{60.99}{0.70}{} \\
OSP-1 & 101M & \cellcolor{Green3!0}\statcell{0.03}{0.06}{}& \cellcolor{Green3!0}\statcell{15.25}{7.45}{}& \cellcolor{Green3!0}\statcell{23.71}{7.40}{}& \cellcolor{Green3!0}\statcell{50.81}{0.34}{}& \cellcolor{Green3!0}\statcell{54.33}{0.35}{}& \cellcolor{Green3!0}\statcell{56.76}{0.24}{}& \cellcolor{Green3!0}\statcell{58.74}{0.90}{} \\
\midrule
\textbf{BELT-B-T} & 16M & \cellcolor{Green3!60}\beststatcell{42.96}{0.36}{}& \cellcolor{Green3!60}\beststatcell{47.65}{1.07}{}& \cellcolor{Green3!60}\beststatcell{51.40}{1.14}{}& \cellcolor{Green3!60}\beststatcell{54.76}{0.32}{}& \cellcolor{Green3!60}\beststatcell{57.84}{0.39}{}& \cellcolor{Green3!60}\beststatcell{60.89}{0.29}{}& \cellcolor{Green3!60}\beststatcell{63.09}{0.06}{} \\
\textbf{BELT-H-T} & 16M & \cellcolor{Green3!47}\statcell{33.40}{6.35}{}& \cellcolor{Green3!43}\statcell{38.25}{6.81}{}& \cellcolor{Green3!40}\statcell{42.22}{6.02}{}& \cellcolor{Green3!21}\statcell{52.22}{0.57}{}& \cellcolor{Green3!25}\statcell{55.81}{0.84}{}& \cellcolor{Green3!28}\statcell{58.67}{1.23}{}& \cellcolor{Green3!32}\statcell{61.08}{1.36}{} \\
\textbf{BELT-B-R} & 16M & \cellcolor{Green3!53}\secondstatcell{37.84}{4.79}{}& \cellcolor{Green3!51}\secondstatcell{42.87}{3.28}{}& \cellcolor{Green3!53}\secondstatcell{48.18}{1.92}{}& \cellcolor{Green3!32}\secondstatcell{52.89}{0.75}{}& \cellcolor{Green3!45}\secondstatcell{56.99}{1.07}{}& \cellcolor{Green3!48}\secondstatcell{60.06}{0.52}{}& \cellcolor{Green3!49}\secondstatcell{62.29}{1.08}{} \\
\textbf{BELT-H-R} & 16M & \cellcolor{Green3!49}\statcell{35.37}{2.51}{}& \cellcolor{Green3!46}\statcell{40.15}{2.32}{}& \cellcolor{Green3!44}\statcell{44.19}{1.56}{}& \cellcolor{Green3!27}\statcell{52.59}{0.15}{}& \cellcolor{Green3!30}\statcell{56.06}{0.48}{}& \cellcolor{Green3!32}\statcell{58.96}{0.31}{}& \cellcolor{Green3!39}\statcell{61.57}{0.48}{} \\
\bottomrule\end{tabular}}
\endgroup
\end{center}
\end{table}

\ecoparagraph{BELT improves on full OSP and OSP-1 on OLMo3.}
We train a new probe on OLMo3. BELT saves 54.09\% at a one-point tolerance against full OSP's 49.86\% and OSP-1's 47.54\%, and 51.22\% at natural accuracy against 45.92\% and 39.83\%. It has the higher mean at six of the seven macro points against full OSP and at five against OSP-1 (\cref{tab:olmo-main} upper). It also exceeds OSP-1 in dev-selected utility at all five prices and has a higher mean than full OSP at all five (\cref{tab:source-transfer}). The probe has $21.2\times$ fewer parameters than one OSP checkpoint and $106\times$ fewer than full OSP's five; replication scope appears in \cref{sec:checks-olmo}. Transferred to LiveCodeBench without code-side training, the same OLMo3 probe with one fitted direction and price thresholds saves 29.91\% and 43.78\% of tokens at a one-point tolerance on problems before and from August 2024, against OSP-1's 21.94\% and 35.35\% (\cref{sec:olmo-code}).

\ecoparagraph{Math-trained BELT probes transfer to coding.}
We transfer all probes and policies without code-side training. On Qwen3 LiveCodeBench Pre, BELT matches full OSP's one-point saving (38.25\% versus 38.25\%) with one encoder instead of five, saves 23.00~pp more than OSP-1 (15.25\%), and has the higher mean at the three jointly feasible budgets (\cref{tab:code-main}). LiveCodeBench Post results are reported in \cref{sec:code-post}; additional readings appear in \cref{sec:source-transfer}.

\ecoparagraph{One belief direction with price thresholds suffices.}
On the H interface, one fitted direction with price thresholds (the configuration recommended below) saves more tokens than OSP-1 at all 21 tolerance readings across the seven math and coding settings. Price-specific refitting adds further gains on math (\cref{sec:analysis-single-direction}). Selecting an operating point on one test half and evaluating it on the other shifts BELT's natural-accuracy readings by at most 0.8~pp on Qwen3 and OLMo3 mathematics and 1.1~pp on OLMo3 code; one Qwen3 code cell shifts by 2.5~pp. The corresponding OSP and OSP-1 shifts on Qwen3 OOD are 5.3 and 6.9~pp (\cref{sec:cross-fitted}).

Development selection at a one-point tolerance yields a BELT variant that stays within tolerance on all three seeds in each Qwen3 math and LiveCodeBench setting. On Qwen3 math, H-T saves 41--44\%; full OSP stays within tolerance on every ID seed and two of three OOD seeds (\cref{sec:cross-fitted,sec:dev-selected}). The main tables retain native OSP/OSP-1; \cref{sec:threshold-expansion} adds thresholds to OSP and BELT refit policies and adds the native stochastic policy to OSP-1's thresholds (OSP-1+T).

\ecoparagraph{Recommended configuration.} We recommend H-T: one fitted direction with price thresholds. It requires one policy fit, matches or exceeds full OSP in 40 of 48 feasible readings (34 within seed variation, 6 above), and exceeds OSP-1 in 35 of 49 (\cref{tab:reb-single-variant-wtl}). Price-specific refitting with threshold expansion (R+T) is useful on the B interface for math: it raises saving or accuracy by at least 1~pp on every seed at all seven ID and all seven OOD readings. On the H interface, R+T improves 3 of 35 cells by that margin (\cref{tab:reb-refit-threshold}), supporting the single-direction default.

\section{Analysis}
\label{sec:analysis}
We compare representations under common fitting, ablate supervision and inputs, test reuse across prices, and transfer the probe to two text-game agent tasks. Utility ($\mathrm{Acc}-\lambda\,\mathrm{Cost}$) scores the trade-off at a specified token price and complements the constrained readings of \cref{sec:main-results}.

\subsection{Belief-trained representations for stopping}
\label{sec:analysis-representation}
\begin{table}[t]
\footnotesize
\begin{minipage}[t]{0.555\textwidth}\centering
\caption{Math OOD micro mean utility. Means: five fitting starts (rows 1--4), three training runs (OSP-1, BELT-L, BELT). SDs in \cref{sec:utility-controls,tab:representation-policy-full}.}
\label{tab:representation-policy}
\begingroup\setlength{\tabcolsep}{2.5pt}
\begin{tabular}{@{}lccccc@{}}\toprule
\textbf{Features} / $\lambda$ ($\times10^{-6}$) & \textbf{2} & \textbf{5} & \textbf{10} & \textbf{20} & \textbf{50} \\\midrule
Pretrained ReProbe & \cellcolor{Green3!4}0.719& \cellcolor{Green3!24}0.713& \cellcolor{Green3!30}0.691& \cellcolor{Green3!33}0.650& \cellcolor{Green3!23}0.550 \\
Raw state (layer 16) & \cellcolor{Green3!16}0.724& \cellcolor{Green3!19}0.711& \cellcolor{Green3!24}0.688& \cellcolor{Green3!29}0.647& \cellcolor{Green3!23}0.550 \\
PCA (768-d) & \cellcolor{Green3!16}0.724& \cellcolor{Green3!17}0.710& \cellcolor{Green3!24}0.688& \cellcolor{Green3!26}0.645& \cellcolor{Green3!22}0.549 \\
Mean-pooled states & \cellcolor{Green3!0}0.717& \cellcolor{Green3!0}0.702& \cellcolor{Green3!0}0.677& \cellcolor{Green3!0}0.627& \cellcolor{Green3!0}0.527 \\
OSP-1 features & \cellcolor{Green3!24}0.728& \cellcolor{Green3!30}0.716& \cellcolor{Green3!41}0.696& \cellcolor{Green3!46}\underline{0.659}& \cellcolor{Green3!57}\underline{0.583} \\
BELT-L & \cellcolor{Green3!60}\textbf{0.744}& \cellcolor{Green3!60}\textbf{0.730}& \cellcolor{Green3!60}\textbf{0.705}& \cellcolor{Green3!44}0.658& \cellcolor{Green3!29}0.556 \\
BELT probe (BELT-H-R) & \cellcolor{Green3!40}\underline{0.735}& \cellcolor{Green3!45}\underline{0.723}& \cellcolor{Green3!54}\underline{0.702}& \cellcolor{Green3!60}\textbf{0.669}& \cellcolor{Green3!60}\textbf{0.586} \\
\bottomrule\end{tabular}\endgroup
\end{minipage}\hfill
\begin{minipage}[t]{0.425\textwidth}\centering
\caption{Sequence BELT-probe and OSP-checkpoint parameter counts, inference FLOPs per token, and archived RTX 4090 reference-schedule training estimates on Qwen3.}
\label{tab:cost-summary}
\begingroup\setlength{\tabcolsep}{3pt}
\begin{tabular}{@{}lrr@{}}\toprule
\textbf{Quantity} & \textbf{BELT} & \textbf{OSP} \\\midrule
Parameters (M) & 15.96 & 100.68 \\
FLOPs/token (M) & 85.5 & 347.6 \\\midrule
One probe (GPU-h) & $\sim$10.0 & $\sim$50.9 \\
Policies for 5 $\lambda$ (GPU-h) & $\sim$10.0 & $\sim$254.5 \\\bottomrule
\end{tabular}
\endgroup
\end{minipage}
\end{table}

\emph{Does belief training yield better features for stopping?} We freeze each feature set and fit the same linear stopping policies (\cref{tab:representation-policy}). On OOD math, BELT features give higher utility than raw hidden states, PCA, pooled states and a pretrained ReProbe probe without belief training at all five prices. They also exceed the features of the end-to-end OSP-1 controller under matched fitting, on OOD math and on LiveCodeBench Pre (\cref{sec:matched-transfer}). Even a linear probe trained on the four beliefs (BELT-L) beats every fixed-feature control at all five prices, pointing to the prediction targets as a main source of the gain; the sequence probe covers accuracy constraints more broadly (\cref{sec:linear-probe-frontiers}). Belief features also save 9--17~pp more tokens at a one-point tolerance than a linear policy on \emph{Now} and prefix length (\cref{sec:probe-controls}), and initializing the probe from ReProbe adds a further 3.2~pp on OOD (\cref{sec:init-push-further}).

\WFclear %

\subsection{Transfer to agentic tasks}
\label{sec:analysis-agentic}
\begin{table}[t]
\caption{OLMo3 agent per-turn thinking control with controllers transferred from mathematics. ALFWorld: token saving (\%) at success tolerance and success (\%) at saving budgets. ScienceWorld: saving at natural score and score (0--100) at saving budgets. Conventions follow \cref{tab:allbaselines,tab:macro}; subscripts are game-bootstrap SDs. Full grids appear in Appendix \cref{sec:alfworld-results,sec:sciworld-results}.}\label{tab:agentic}
\begin{center}
\begingroup\setlength{\tabcolsep}{2pt}
\resizebox{\linewidth}{!}{%
\begin{tabular}{@{}l|c|ccc|cc|c|cccc@{}}\toprule
\multirow{3}{*}{\textbf{Method}} & \multirow{3}{*}{\shortstack{\textbf{Probe}\\\textbf{Params}}} & \multicolumn{5}{c|}{\textbf{ALFWorld (success)}} & \multicolumn{5}{c}{\textbf{ScienceWorld (score)}} \\
 & & \multicolumn{3}{c|}{\textbf{Saving (\%) at tolerance}} & \multicolumn{2}{c|}{\textbf{Success at saving}} & \textbf{Saving (\%)} & \multicolumn{4}{c}{\textbf{Score at saving}} \\
 & & \textbf{0 pp} & \textbf{2 pp} & \textbf{5 pp} & \textbf{20\%} & \textbf{40\%} & \textbf{at 0 pt} & \textbf{80\%} & \textbf{70\%} & \textbf{60\%} & \textbf{50\%} \\
\midrule
Natural & -- & \missingcell& \missingcell& \missingcell& \statcell{33.6}{4.2}{}& \statcell{33.6}{4.2}{}& \missingcell& \statcell{21.4}{3.7}{}& \statcell{21.4}{3.7}{}& \statcell{21.4}{3.7}{}& \statcell{21.4}{3.7}{} \\
OSP-1 & 405M & \cellcolor{Green3!0}\statcell{0.0}{18.4}{}& \cellcolor{Green3!0}\statcell{11.5}{17.3}{}& \cellcolor{Green3!33}\secondstatcell{39.5}{15.8}{}& \cellcolor{Green3!12}\statcell{30.1}{3.6}{}& \cellcolor{Green3!23}\statcell{28.5}{3.6}{}& \cellcolor{Green3!0}\secondstatcell{84.9}{5.8}{}& \cellcolor{Green3!20}\statcell{25.2}{3.2}{}& \cellcolor{Green3!16}\statcell{25.7}{3.1}{}& \cellcolor{Green3!22}\secondstatcell{26.9}{3.2}{}& \cellcolor{Green3!42}\statcell{28.0}{3.4}{} \\
\midrule
\textbf{BELT-B-T} & 19M & \cellcolor{Green3!37}\secondstatcell{12.4}{13.9}{}& \cellcolor{Green3!45}\secondstatcell{25.4}{13.0}{}& \cellcolor{Green3!26}\statcell{38.0}{11.8}{}& \cellcolor{Green3!45}\secondstatcell{32.5}{3.1}{}& \cellcolor{Green3!47}\secondstatcell{29.2}{3.0}{}& \cellcolor{Green3!60}\beststatcell{85.0}{8.1}{}& \cellcolor{Green3!40}\secondstatcell{25.4}{3.1}{}& \cellcolor{Green3!27}\statcell{25.9}{3.0}{}& \cellcolor{Green3!0}\statcell{26.5}{2.9}{}& \cellcolor{Green3!7}\statcell{27.0}{2.9}{} \\
\textbf{BELT-H-T} & 19M & \cellcolor{Green3!0}\statcell{0.0}{12.1}{}& \cellcolor{Green3!5}\statcell{13.2}{13.3}{}& \cellcolor{Green3!0}\statcell{32.5}{13.8}{}& \cellcolor{Green3!0}\statcell{29.2}{3.1}{}& \cellcolor{Green3!0}\statcell{27.8}{3.0}{}& \cellcolor{Green3!60}\beststatcell{85.0}{7.4}{}& \cellcolor{Green3!60}\beststatcell{25.6}{3.1}{}& \cellcolor{Green3!49}\secondstatcell{26.3}{3.0}{}& \cellcolor{Green3!5}\statcell{26.6}{3.0}{}& \cellcolor{Green3!0}\statcell{26.8}{3.0}{} \\
\textbf{BELT-B-R} & 19M & \cellcolor{Green3!0}\statcell{0.0}{17.5}{}& \cellcolor{Green3!6}\statcell{13.4}{16.3}{}& \cellcolor{Green3!8}\statcell{34.2}{15.5}{}& \cellcolor{Green3!19}\statcell{30.6}{3.6}{}& \cellcolor{Green3!0}\statcell{27.8}{3.6}{}& \cellcolor{Green3!0}\secondstatcell{84.9}{6.6}{}& \cellcolor{Green3!0}\statcell{25.0}{3.1}{}& \cellcolor{Green3!0}\statcell{25.4}{3.0}{}& \cellcolor{Green3!11}\statcell{26.7}{3.0}{}& \cellcolor{Green3!49}\secondstatcell{28.2}{3.1}{} \\
\textbf{BELT-H-R} & 19M & \cellcolor{Green3!60}\beststatcell{20.0}{19.4}{}& \cellcolor{Green3!60}\beststatcell{30.2}{16.5}{}& \cellcolor{Green3!60}\beststatcell{45.4}{14.7}{}& \cellcolor{Green3!60}\beststatcell{33.6}{3.8}{}& \cellcolor{Green3!60}\beststatcell{29.6}{3.6}{}& \cellcolor{Green3!0}\secondstatcell{84.9}{6.6}{}& \cellcolor{Green3!40}\secondstatcell{25.4}{3.2}{}& \cellcolor{Green3!60}\beststatcell{26.5}{3.2}{}& \cellcolor{Green3!60}\beststatcell{27.6}{3.4}{}& \cellcolor{Green3!60}\beststatcell{28.5}{3.5}{} \\
\bottomrule\end{tabular}}
\endgroup
\end{center}
\end{table}

\emph{Does a math-trained probe control thinking in agent tasks?} We transfer the OLMo3 math probes, fitted policies and OSP-1 checkpoint unchanged to an agent that thinks before each action in ALFWorld \citep{shridhar2021alfworld} and ScienceWorld \citep{wang2022scienceworld}. Before each action, the controller decides whether to stop thinking and act. Operating points are calibrated on a disjoint split by projected token saving, without outcome labels (\cref{sec:alfworld-protocol,sec:sciworld-protocol}).

The two tasks need very different amounts of thinking. In ALFWorld, thinking raises success from 4.5\% to 33.6\%. There, BELT is comparable to OSP-1, and its refit H variant has the highest mean at every reported reading. In ScienceWorld, most thinking can be removed without lowering the score, and every BELT family saves 85\% of tokens at or above the natural score (\cref{tab:agentic}). One 19M-parameter math probe thus covers both regimes. Budget ranges, full grids and paired intervals appear in \cref{sec:alfworld-results,sec:sciworld-results}.

\subsection{Necessity of the \emph{Later} and \emph{Same} beliefs}
\label{sec:analysis-information}
\begin{table}[t]
\footnotesize\centering
\caption{Average math OOD utility for (a) prediction-target and (b) policy-input ablations (fixed four-belief probe). Columns give token prices $\lambda$ in $10^{-6}$. Means over three BELT probe training runs. SDs and accuracy--cost curves/readings appear in \cref{tab:belief-utility,fig:belief-ablation-frontiers,tab:belief-ablation-frontiers}.}
\label{tab:belief-ablation}
\begingroup\setlength{\tabcolsep}{3pt}
\begin{tabular*}{\textwidth}{@{\extracolsep{\fill}}lccccc@{\hspace{12pt}}ccccc@{}}\toprule
\multirow{2}{*}{\textbf{Belief set}} & \multicolumn{5}{c}{\textbf{(a) Prediction targets}} & \multicolumn{5}{c}{\textbf{(b) Policy inputs}} \\
\cmidrule(lr){2-6}\cmidrule(l){7-11}
 & \textbf{2} & \textbf{5} & \textbf{10} & \textbf{20} & \textbf{50} & \textbf{2} & \textbf{5} & \textbf{10} & \textbf{20} & \textbf{50} \\
\midrule
\emph{Now}, \emph{End} & \cellcolor{Green3!16}0.7451& \cellcolor{Green3!20}0.7307& \cellcolor{Green3!17}0.7067& \cellcolor{Green3!1}0.6586& \cellcolor{Green3!0}0.5820& \cellcolor{Green3!0}0.7447& \cellcolor{Green3!0}0.7302& \cellcolor{Green3!0}0.7061& \cellcolor{Green3!3}0.6595& \cellcolor{Green3!24}0.5831 \\
\emph{Now}, \emph{End}, \emph{Later} & \cellcolor{Green3!0}0.7448& \cellcolor{Green3!0}0.7304& \cellcolor{Green3!0}0.7065& \cellcolor{Green3!0}0.6585& \cellcolor{Green3!32}\underline{0.5853}& \cellcolor{Green3!13}0.7449& \cellcolor{Green3!13}0.7304& \cellcolor{Green3!11}0.7063& \cellcolor{Green3!0}0.6592& \cellcolor{Green3!0}0.5818 \\
\emph{Now}, \emph{End}, \emph{Same} & \cellcolor{Green3!60}\textbf{0.7459}& \cellcolor{Green3!60}\textbf{0.7313}& \cellcolor{Green3!26}\underline{0.7068}& \cellcolor{Green3!50}\underline{0.6648}& \cellcolor{Green3!60}\textbf{0.5882}& \cellcolor{Green3!60}\textbf{0.7456}& \cellcolor{Green3!60}\textbf{0.7311}& \cellcolor{Green3!49}\underline{0.7070}& \cellcolor{Green3!53}\underline{0.6652}& \cellcolor{Green3!36}\underline{0.5838} \\
All four & \cellcolor{Green3!33}\underline{0.7454}& \cellcolor{Green3!40}\underline{0.7310}& \cellcolor{Green3!60}\textbf{0.7072}& \cellcolor{Green3!60}\textbf{0.6660}& \cellcolor{Green3!30}0.5851& \cellcolor{Green3!47}\underline{0.7454}& \cellcolor{Green3!53}\underline{0.7310}& \cellcolor{Green3!60}\textbf{0.7072}& \cellcolor{Green3!60}\textbf{0.6660}& \cellcolor{Green3!60}\textbf{0.5851} \\
\bottomrule\end{tabular*}\endgroup
\end{table}

\emph{Are Later and Same needed beyond Now and End?} We remove them either as prediction targets during belief training (\cref{tab:belief-ablation}a) or as policy inputs with the trained probe fixed (\cref{tab:belief-ablation}b).

\ecoparagraph{Same raises utility at every price.}
Adding Same to Now/End raises utility at all five prices, both as a target and as an input. With the four-belief probe fixed, removing Same lowers utility at all five prices by 0.0005--0.0068. Each comparison gains 0.006--0.007 at $\lambda=2\times10^{-5}$. Same identifies prefixes whose current answer already matches the natural-completion answer, so stopping there gives up little; the fitted policy weights it positively (\cref{sec:analysis-adaptation}).

\ecoparagraph{Later improves saving under accuracy constraints.}
Later matters most when accuracy is constrained (\cref{tab:belief-ablation-frontiers}). As a target, adding it to Now/End raises token saving by 1.7--2.0~pp at all three tolerances; as an input, the full set has the best mean at six of seven constrained readings and the best utility at the three highest prices. Later credits continuing toward a correct answer that appears before completion, so the policy keeps reasoning where such an answer is still likely (\cref{sec:analysis-adaptation}).

\subsection{Interpreting the BELT-B-R logistic policy}
\label{sec:analysis-adaptation}
\begin{figure}[t]\centering
\includegraphics[width=\linewidth]{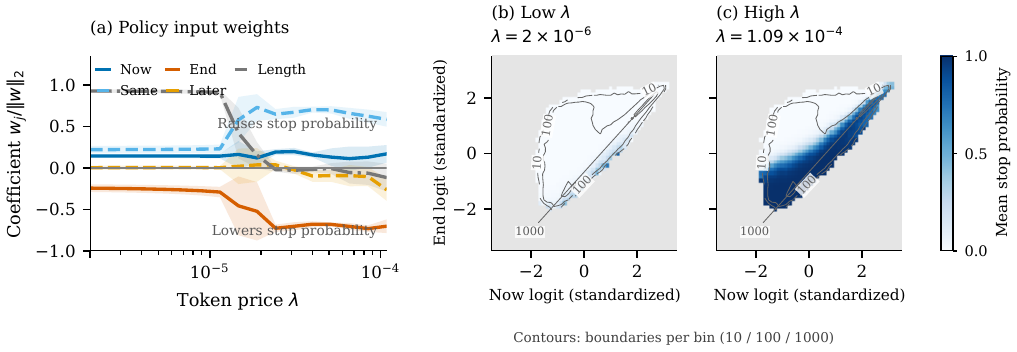}
\caption{How the \beltb{} policy changes with the token price $\lambda$. (a) Normalized input weights, shown as mean and sample SD over three independent BELT probe training runs; positive weights raise the stop probability. (b,c) Mean stop probability over the standardized \emph{Now}--\emph{End} plane at a low and a high price, from one representative training run. Gray contours show boundary counts; gray cells are empty.}
\label{fig:belief-roles}
\end{figure}
\beltb{} fits a logistic policy with six parameters, so its weights can be read directly (\cref{fig:belief-roles}a). At low prices, prefix length carries most of the weight, and the policy rarely stops anywhere in the \emph{Now}--\emph{End} plane (\cref{fig:belief-roles}b). As the price rises, weight moves from length to \emph{Same} (positive) and \emph{End} (negative); \emph{Later} turns negative at the highest prices, and \emph{Now} keeps a small positive weight throughout. The high-price policy therefore stops wherever \emph{End} is low relative to \emph{Now}, that is, where finishing the trace is not expected to improve the current answer, and continues where \emph{End} is high (\cref{fig:belief-roles}c). Refitting thus reuses the same four beliefs and reweights them by price. \Cref{sec:belief-details} gives the coefficients, the threshold decomposition and the heatmap construction.

\subsection{Refit and threshold cover different operating points}
\label{sec:analysis-single-direction}
A threshold moves a linear boundary's intercept; refitting can also rotate its normal and change stochastic sharpness, and the two families are nonnested (\cref{sec:single-direction-details}). On the H interface, the two families are within 1~pp at most readings, supporting one fitted direction with price thresholds. On the B interface, price-specific refitting raises math saving or accuracy by at least 1~pp on every seed at all seven ID and all seven OOD readings. Adding thresholds to the H-R policies raises OOD one-point saving from 45.25\% to 46.75\% (\cref{tab:threshold-expansion-qwen}).

\subsection{Why direct utility training collapses}
\label{sec:analysis-direct-training}
Training the same probe directly on the stopping utility, as OSP does, collapses: with random initialization, the source-validation-selected checkpoints either never stop (OOD utility 0.6269, equal to natural completion) or always stop (0.1814), whereas BELT selected on the same development data reaches 0.6689. The cause is that the utility gradient of a stop logit is scaled by the policy itself, through the factor $S_k\pi_k(1-\pi_k)$, where $\pi_k$ is the stop probability at boundary $k$ and $S_k$ the probability of reaching it. Once the policy collapses to always or never stopping, $\pi_k(1-\pi_k)$ approaches zero, and after early stops later boundaries are rarely reached; the gradient that should correct the policy vanishes with it, so training cannot leave the collapsed state. BELT training does not depend on the policy: each labelled boundary contributes the residual $\sigma(b_k)-y_k$ between the predicted belief and its label, which stays large whenever a prediction is wrong. \Cref{sec:checks-direct} gives the full gradient analysis, the recipe and the per-seed checkpoints.

\subsection{Training cost and reuse across token prices}
\label{sec:cost}
\ecoparagraph{New token prices reuse the BELT probe.}
Both variants train one probe and share its features. Refit performs $K$ CPU fits for $K$ prices; threshold fits once and varies its cutoff. Full OSP trains $K$ checkpoints. Each Qwen3 BELT-B-R/H-R policy adds 6/769 parameters; OLMo3 B-R omits length and uses five (\cref{sec:cost-adaptation}). Threshold expansion preserves this training reuse (\cref{sec:threshold-expansion}).
\ecoparagraph{A compact probe reduces per-checkpoint cost.}
The Qwen3 sequence probe has $6.3\times$ fewer parameters and approximately $4.1\times$ fewer inference FLOPs per token than one OSP checkpoint (\cref{tab:cost-summary}). Archived RTX 4090 training estimates and their scope appear in \cref{sec:cost-measured,sec:cost-analysis}. On OLMo3, the probe has $21.2\times$ fewer parameters than one OSP checkpoint; on GH200, its training needs approximately $11\times$ fewer GPU-hours than one OSP checkpoint and $55\times$ fewer than five (\cref{sec:cost-measured}).

\FloatBarrier

\section{Related Work}
\label{sec:related}
\ecoparagraph{Efficient reasoning and early stopping.}
Efficient reasoning methods train on concise or shortest-correct traces \citep{pmlr-v267-chen25bx,NEURIPS2025_3e22bea3} or control inference-time budgets \citep{muennighoff2025s1simpletesttimescaling,li2025steeringllmthinkingbudget}. Early-stopping methods use answer confidence \citep{yang2025deer,hosseini2026codestop}, convergence \citep{liu2025answerconvergencesignalearly}, or hidden-state overthinking detectors \citep{wang2026rom}. Thought Calibration \citep{wu2025thought} learns correctness or final-answer agreement scores and separately calibrates thresholds with statistical risk control. OS-Pruner \citep{ehab2026ospruner} jointly trains a representation and stopping head on sequential utility at a chosen token price. BELT reuses a belief-trained representation across prices through policy refitting or fixed-direction threshold control.

\ecoparagraph{Predictive representations for control.}
Hidden-state probes support hallucination detection \citep{uhead}, reasoning-step verification and search (ReProbe; \citealp{ni-etal-2026-reprobe}), and intermediate-answer correctness prediction \citep{zhang2025reasoning}. Cross-regressive decoding exploits latent information about future tokens in pretrained hidden states to propose multiple tokens in parallel and accelerate generation \citep{dualstream2026}. ZIP-RC predicts reward and remaining cost to allocate computation among samples \citep{ICLR2026_b22c5b6a}. AERA predicts recovery from cumulative response sets \citep{wang2026aeraadaptiveevidenceresidual}. BELT builds on ReProbe to control stopping within a single trajectory. \Cref{sec:analysis-representation,sec:analysis-information} compare representations and distinguish target-retraining from fixed-probe input ablations.

\section{Conclusion}
BELT separates learning what further reasoning could achieve from deciding whether its cost is worthwhile. One belief probe supports different token prices through a small CPU policy fit or a threshold shift. Its features exceed raw-state controls and matched OSP-1 features in OOD mean utility, and the Same belief improves both utility and constrained readings. At a one-point tolerance, BELT saves 45.98\% of Qwen3 OOD tokens, matching OSP's 45.19\% with one encoder instead of five. On OLMo3, it saves 4.23~pp more than full OSP and 6.55~pp more than OSP-1. On Qwen3 LiveCodeBench Pre, it matches full OSP's one-point saving without code-side training. The OLMo3 math probe also controls per-turn thinking in ALFWorld and ScienceWorld, saving 85\% of ScienceWorld tokens at or above natural score. We recommend one fitted direction with price thresholds: it saves more tokens than OSP-1 at all 21 tolerance readings across the seven math and coding settings. Reusing outcome features makes price adaptation a small policy update.
\subsection*{Acknowledgments}
This work was supported as part of the Swiss AI Initiative by a small grant from the Swiss National Supercomputing Centre (CSCS) under project ID a0142 on Alps, which provided the computing resources for our experiments.
\subsection*{Reproducibility statement}
\Cref{sec:implementation} specifies supervision, policy fitting, data partitions and metric conventions. Evaluation exports record question rosters, input hashes, frontier members and mixture weights. Mathematics and coding evaluations replay banked reasoning trajectories; agent evaluations replay a shared trajectory tree whose nodes are expanded on demand (\cref{sec:alfworld-protocol,sec:sciworld-protocol}). We release the code and the trained BELT and OSP checkpoints at \url{https://github.com/EdisonNi-hku/BELT}.
\subsection*{Ethics statement}
Our experiments use public reasoning and coding benchmarks and text-game environments. Reduced reasoning cost does not guarantee answer correctness, and matching an accuracy constraint on an evaluation set does not guarantee it on a new distribution. The method is not evaluated for safety-critical decisions.
\subsection*{AI use statement}
We used generative AI tools (LLM coding assistants) for implementing and debugging parts of the experiment and evaluation code, for data-analysis scripts and table generation, for producing source-linked numerical summaries and checking manuscript claims against the exported results, and for editing the manuscript for clarity. We did not use generative AI tools to generate synthetic data, to clean or reformat datasets, to design the method or the experiments, to propose or refine hypotheses, to derive the formulation in \cref{sec:background,sec:method} or any mathematical claim, or to interpret the results beyond producing summaries that the authors checked against the exported tables; translation, proofs and qualitative data analysis are not applicable to this work. All AI-assisted code and text were reviewed by the authors, who take full responsibility for the contents of this paper.
\bibliography{references}

@inproceedings{ehab2026ospruner,
  title = {{OS-Pruner}: Pruning Chains-of-Thought of Reasoning Models
  via Optimal Stopping},
  author = {Mohammed Ehab and Aymane El Gadarri and Vivek F. Farias
  and Adam Jozefiak and Ciamac C. Moallemi},
  booktitle = {Advances in Neural Information Processing Systems},
  year = {2026},
  eprint = {2607.11089},
  archiveprefix = {arXiv},
  primaryclass = {cs.AI},
  url = {https://openreview.net/forum?id=cB7emUeNVr},
}

@misc{wang2026rom,
  title = {{ROM}: Real-time Overthinking Mitigation via Streaming
  Detection and Intervention},
  author = {Xinyan Wang and Xiaogeng Liu and Ming Pei and Chaowei Xiao},
  year = {2026},
  eprint = {2603.22016},
  archiveprefix = {arXiv},
  primaryclass = {cs.LG},
  url = {https://arxiv.org/abs/2603.22016},
  journal = {arXiv.org},
  doi = {10.48550/arXiv.2603.22016},
}

@misc{hosseini2026codestop,
  title = {Early Stopping for Large Reasoning Models via Confidence Dynamics},
  author = {Parsa Hosseini and Sumit Nawathe and Mahdi Salmani and
  Meisam Razaviyayn and Soheil Feizi},
  year = {2026},
  eprint = {2604.04930},
  archiveprefix = {arXiv},
  primaryclass = {cs.CL},
  url = {https://arxiv.org/abs/2604.04930},
  journal = {arXiv.org},
  doi = {10.48550/arXiv.2604.04930},
}

@inproceedings{wu2025thought,
  title = {Thought calibration: Efficient and confident test-time scaling},
  author = {Wu, Menghua and Zhou, Cai and Bates, Stephen and
  Jaakkola, Tommi},
  editor = {Christodoulopoulos, Christos and Chakraborty, Tanmoy and
  Rose, Carolyn and Peng, Violet},
  booktitle = {Proceedings of the 2025 Conference on Empirical
  Methods in Natural Language Processing},
  month = nov,
  year = {2025},
  address = {Suzhou, China},
  publisher = {Association for Computational Linguistics},
  url = {https://aclanthology.org/2025.emnlp-main.722/},
  doi = {10.18653/v1/2025.emnlp-main.722},
  pages = {14291--14305},
}

@inproceedings{yang2025deer,
  title = {Dynamic Early Exit in Reasoning Models},
  author = {Yang, Chenxu and Si, Qingyi and Duan, Yongjie and Zhu,
    Zheliang and Zhu, Chenyu and Li, Qiaowei and Chen, Minghui and Lin,
  Zheng and Wang, Weiping},
  booktitle = {International Conference on Learning Representations},
  editor = {C. Vondrick and B. Hariharan and C. Raffel and L. Pinto
  and D. Yang and A. Faust},
  volume = {2026},
  pages = {88170--88210},
  url =
  {https://proceedings.iclr.cc/paper_files/paper/2026/file/8df90a1440ce782d1f5607b7a38f2531-Paper-Conference.pdf},
  year = {2026},
}

@misc{qwen3,
  title = {Qwen3 Technical Report},
  author = {{Qwen Team}},
  year = {2025},
  eprint = {2505.09388},
  archiveprefix = {arXiv},
  primaryclass = {cs.CL},
  url = {https://arxiv.org/abs/2505.09388},
}

@inproceedings{uhead,
  title = {A Head to Predict and a Head to Question: Pre-trained
    Uncertainty Quantification Heads for Hallucination Detection in
  {LLM} Outputs},
  author = {Shelmanov, Artem and Fadeeva, Ekaterina and Tsvigun, Akim
    and Tsvigun, Ivan and Xie, Zhuohan and Kiselev, Igor and Daheim,
    Nico and Zhang, Caiqi and Vazhentsev, Artem and Sachan, Mrinmaya
  and Nakov, Preslav and Baldwin, Timothy},
  editor = {Christodoulopoulos, Christos and Chakraborty, Tanmoy and
  Rose, Carolyn and Peng, Violet},
  booktitle = {Proceedings of the 2025 Conference on Empirical
  Methods in Natural Language Processing},
  month = nov,
  year = {2025},
  address = {Suzhou, China},
  publisher = {Association for Computational Linguistics},
  url = {https://aclanthology.org/2025.emnlp-main.1809/},
  doi = {10.18653/v1/2025.emnlp-main.1809},
  pages = {35712--35731},
}

@inproceedings{zhang2025reasoning,
  title = {Reasoning Models Know When They{\textquoteright}re Right:
  Probing Hidden States for Self-Verification},
  author = {Anqi Zhang and Yulin Chen and Jane Pan and Chen Zhao and
  Aurojit Panda and Jinyang Li and He He},
  booktitle = {Second Conference on Language Modeling},
  year = {2025},
  url = {https://openreview.net/forum?id=O6I0Av7683},
}

@inproceedings{snell2024scaling,
  title = {Scaling {LLM} Test-Time Compute Optimally Can be More
  Effective than Scaling Parameters for Reasoning},
  author = {Snell, Charlie and Lee, Jaehoon and Xu, Kelvin and Kumar, Aviral},
  booktitle = {International Conference on Learning Representations},
  editor = {Y. Yue and A. Garg and N. Peng and F. Sha and R. Yu},
  volume = {2025},
  pages = {10131--10165},
  url =
  {https://proceedings.iclr.cc/paper_files/paper/2025/file/1b623663fd9b874366f3ce019fdfdd44-Paper-Conference.pdf},
  year = {2025},
  journal = {International Conference on Learning Representations},
}

@inproceedings{dualstream2026,
  title = {Dual-Stream Decoding for Accelerated Large Language Models},
  author = {Abulkhanov, Dmitry and Strizhakov, Daniil and Panov, Maxim},
  booktitle = {Third Conference on Language Modeling},
  year = {2026},
  url = {https://openreview.net/forum?id=QCfy43en5P},
}

@inproceedings{yu2025dapo,
  title = {{DAPO}: An Open-Source {LLM} Reinforcement Learning System at Scale},
  author = {Yu, Qiying and Zhang, Zheng and Zhu, Ruofei and Yuan,
    Yufeng and Zuo, Xiaochen and Yue, Yu and Dai, Weinan and Fan,
    Tiantian and Liu, Gaohong and Liu, Juncai and Liu, LingJun and
  Liu, Xin and others},
  booktitle = {Advances in Neural Information Processing Systems},
  editor = {D. Belgrave and C. Zhang and H. Lin and R. Pascanu and P.
  Koniusz and M. Ghassemi and N. Chen},
  volume = {38},
  pages = {113222--113244},
  publisher = {Curran Associates, Inc.},
  doi = {10.52202/085713-3775},
  url = {https://proceedings.neurips.cc/paper_files/paper/2025/file/a4277440d50f1f15d2cb4c14f7e0c0d2-Paper-Conference.pdf},
  year = {2025},
}

@inproceedings{wang2024tokeneconomies,
  title = {Reasoning in Token Economies: Budget-Aware Evaluation of
  {LLM} Reasoning Strategies},
  author = {Wang, Junlin  and Jain, Siddhartha  and Zhang, Dejiao
  and Ray, Baishakhi  and Kumar, Varun  and Athiwaratkun, Ben},
  editor = {Al-Onaizan, Yaser  and Bansal, Mohit  and Chen, Yun-Nung},
  booktitle = {Proceedings of the 2024 Conference on Empirical
  Methods in Natural Language Processing},
  year = {2024},
  journal = {Proceedings of the 2024 Conference on Empirical Methods
  in Natural Language Processing},
  pages = {19916-19939},
  doi = {10.18653/v1/2024.emnlp-main.1112},
  publisher = {Association for Computational Linguistics},
}

@inproceedings{ni-etal-2026-reprobe,
  title = {{R}e{P}robe: Efficient Test-Time Scaling of Multi-Step
  Reasoning by Probing Internal States of Large Language Models},
  author = {Ni, Jingwei and Fadeeva, Ekaterina and Wu, Tianyi and
    Akhtar, Mubashara and Zhang, Jiaheng and Ash, Elliott and Leippold,
    Markus and Baldwin, Timothy and Ng, See-Kiong and Shelmanov, Artem
  and others},
  editor = {Liakata, Maria  and Moreira, Viviane P.  and Zhang,
  Jiajun  and Jurgens, David},
  booktitle = {Proceedings of the 64th Annual Meeting of the
  Association for Computational Linguistics (Volume 1: Long Papers)},
  year = {2026},
  journal = {Proceedings of the 64th Annual Meeting of the
  Association for Computational Linguistics (Volume 1: Long Papers)},
  pages = {11667-11689},
  doi = {10.18653/v1/2026.acl-long.536},
  publisher = {Association for Computational Linguistics},
}

@misc{wang2026aeraadaptiveevidenceresidual,
  title = {{AERA}: Adaptive Evidence Residual Allocation for Efficient
  Test-Time Reasoning},
  author = {Ziming Wang and Ivor Tsang and Hangwei Qian},
  year = {2026},
  eprint = {2608.27964},
  archiveprefix = {arXiv},
  primaryclass = {cs.AI},
  url = {https://arxiv.org/abs/2608.27964},
}

@inproceedings{muennighoff2025s1simpletesttimescaling,
  title = {s1: Simple test-time scaling},
  author = {Muennighoff, Niklas and Yang, Zitong and Shi, Weijia and
    Li, Xiang Lisa and Fei-Fei, Li and Hajishirzi, Hannaneh and
    Zettlemoyer, Luke and Liang, Percy and Cand{\`e}s, Emmanuel and
  Hashimoto, Tatsunori},
  editor = {Christodoulopoulos, Christos and Chakraborty, Tanmoy and
  Rose, Carolyn and Peng, Violet},
  booktitle = {Proceedings of the 2025 Conference on Empirical
  Methods in Natural Language Processing},
  month = nov,
  year = {2025},
  address = {Suzhou, China},
  publisher = {Association for Computational Linguistics},
  url = {https://aclanthology.org/2025.emnlp-main.1025/},
  doi = {10.18653/v1/2025.emnlp-main.1025},
  pages = {20275--20321},
}

@inproceedings{li2025steeringllmthinkingbudget,
  title = {Steering {LLM} Thinking with Budget Guidance},
  author = {Li, Junyan  and Zhao, Wenshuo  and Zhang, Yang  and Gan, Chuang},
  editor = {Liakata, Maria  and Moreira, Viviane P.  and Zhang,
  Jiajun  and Jurgens, David},
  booktitle = {Findings of the Association for Computational
  Linguistics: ACL 2026},
  year = {2026},
  journal = {Findings of the Association for Computational
  Linguistics: ACL 2026},
  pages = {37437-37451},
  doi = {10.18653/v1/2026.findings-acl.1866},
  publisher = {Association for Computational Linguistics},
}

@inproceedings{liu2025answerconvergencesignalearly,
  title = {Answer Convergence as a Signal for Early Stopping in Reasoning},
  author = {Liu, Xin and Wang, Lu},
  editor = {Christodoulopoulos, Christos and Chakraborty, Tanmoy and
  Rose, Carolyn and Peng, Violet},
  booktitle = {Proceedings of the 2025 Conference on Empirical
  Methods in Natural Language Processing},
  month = nov,
  year = {2025},
  address = {Suzhou, China},
  publisher = {Association for Computational Linguistics},
  url = {https://aclanthology.org/2025.emnlp-main.904/},
  doi = {10.18653/v1/2025.emnlp-main.904},
  pages = {17896--17907},
}

@misc{olmo2025olmo3,
  title = {Olmo 3},
  author = {{Team Olmo} and Ettinger, Allyson and Bertsch, Amanda and
    Kuehl, Bailey and Graham, David and Heineman, David and Groeneveld,
    Dirk and Brahman, Faeze and Timbers, Finbarr and Ivison, Hamish and
  Morrison, Jacob and others},
  year = {2025},
  eprint = {2512.13961},
  archiveprefix = {arXiv},
  primaryclass = {cs.CL},
  url = {https://arxiv.org/abs/2512.13961},
}

@inproceedings{ni-etal-2026-reasoning,
  title = {Can Reasoning Help Large Language Models Capture Human
  Annotator Disagreement?},
  author = {Ni, Jingwei  and Fan, Yu  and Zouhar, Vil{\'e}m  and
    Rooein, Donya  and Hoyle, Alexander Miserlis  and Sachan, Mrinmaya
  and Leippold, Markus  and Hovy, Dirk  and Ash, Elliott},
  editor = {Demberg, Vera  and Inui, Kentaro  and Marquez, Llu{\'i}s},
  booktitle = {Proceedings of the 19th Conference of the European Chapter
  of the Association for Computational Linguistics (Volume 1: Long Papers)},
  year = {2026},
  journal = {Proceedings of the 19th Conference of the European Chapter
  of the Association for Computational Linguistics (Volume 1: Long Papers)},
  pages = {36-54},
  doi = {10.18653/v1/2026.eacl-long.3},
  publisher = {Association for Computational Linguistics},
}

@inproceedings{pmlr-v267-chen25bx,
  title = {Do {NOT} Think That Much for 2+3=? {O}n the Overthinking of
  Long Reasoning Models},
  author = {Chen, Xingyu and Xu, Jiahao and Liang, Tian and He,
    Zhiwei and Pang, Jianhui and Yu, Dian and Song, Linfeng and Liu,
    Qiuzhi and Zhou, Mengfei and Zhang, Zhuosheng and Wang, Rui and Tu,
  Zhaopeng and Mi, Haitao and Yu, Dong},
  booktitle = {Proceedings of the 42nd International Conference on
  Machine Learning},
  pages = {9487--9499},
  year = {2025},
  editor = {Singh, Aarti and Fazel, Maryam and Hsu, Daniel and
    Lacoste-Julien, Simon and Berkenkamp, Felix and Maharaj, Tegan and
  Wagstaff, Kiri and Zhu, Jerry},
  volume = {267},
  series = {Proceedings of Machine Learning Research},
  publisher = {PMLR},
  url = {https://proceedings.mlr.press/v267/chen25bx.html},
}

@inproceedings{NEURIPS2025_3e22bea3,
  title = {Towards Thinking-Optimal Scaling of Test-Time Compute for
  {LLM} Reasoning},
  author = {Wenkai Yang and Shuming Ma and Yankai Lin and Furu Wei},
  booktitle = {The Thirty-ninth Annual Conference on Neural
  Information Processing Systems},
  year = {2025},
  url = {https://openreview.net/forum?id=6ICFqmixlS},
  journal = {Advances in Neural Information Processing Systems 38},
  pages = {48704-48730},
  doi = {10.52202/085713-1452},
  publisher = {Neural Information Processing Systems Foundation, Inc.
  (NeurIPS)},
}

@inproceedings{ICLR2026_b22c5b6a,
  title = {Zero-Overhead Introspection for Adaptive Test-Time Compute},
  author = {Rohin Manvi and Joey Hong and Tim Seyde and Maxime
  Labonne and Mathias Lechner and Sergey Levine},
  booktitle = {The Fourteenth International Conference on Learning
  Representations},
  year = {2026},
  url = {https://openreview.net/forum?id=GqZYGOYuF2},
}

@misc{numina_math_datasets,
  author = {Jia LI and Edward Beeching and Lewis Tunstall and Ben
    Lipkin and Roman Soletskyi and Shengyi Costa Huang and Kashif Rasul
    and Longhui Yu and Albert Jiang and Ziju Shen and Zihan Qin and Bin
  Dong and Li Zhou and Yann Fleureau and Guillaume Lample and Stanislas Polu},
  title = {{NuminaMath}},
  year = {2024},
  publisher = {Numina},
  journal = {Hugging Face repository},
  url = {https://huggingface.co/datasets/AI-MO/NuminaMath-1.5},
}

@misc{deepscaler2025,
  title = {{DeepScaleR}: Surpassing {O1-Preview} with a {1.5B} Model
  by Scaling {RL}},
  author = {Michael Luo and Sijun Tan and Justin Wong and Xiaoxiang
    Shi and William Y. Tang and Manan Roongta and Colin Cai and
    Jeffrey Luo and Tianjun Zhang and Li Erran Li and Raluca Ada Popa
  and Ion Stoica},
  year = {2025},
  howpublished = {Notion Blog},
  url = {https://pretty-radio-b75.notion.site/DeepScaleR-Surpassing-O1-Preview-with-a-1-5B-Model-by-Scaling-RL-19681902c1468005bed8ca303013a4e2},
}

@inproceedings{ICLR2024_aca97732,
  author = {Lightman, Hunter and Kosaraju, Vineet and Burda, Yuri and
    Edwards, Harrison and Baker, Bowen and Lee, Teddy and Leike, Jan and
  Schulman, John and Sutskever, Ilya and Cobbe, Karl},
  booktitle = {International Conference on Learning Representations},
  editor = {B. Kim and Y. Yue and S. Chaudhuri and K. Fragkiadaki and
  M. Khan and Y. Sun},
  pages = {39578--39601},
  title = {Let's Verify Step by Step},
  url = {https://proceedings.iclr.cc/paper_files/paper/2024/file/aca97732e30bcf1303bc22ac3924fd16-Paper-Conference.pdf},
  volume = {2024},
  year = {2024},
}

@inproceedings{he-etal-2024-olympiadbench,
  title = {{O}lympiad{B}ench: A Challenging Benchmark for Promoting
  {AGI} with Olympiad-Level Bilingual Multimodal Scientific Problems},
  author = {He, Chaoqun and Luo, Renjie and Bai, Yuzhuo and Hu,
    Shengding and Thai, Zhen and Shen, Junhao and Hu, Jinyi and Han, Xu
  and Huang, Yujie and Zhang, Yuxiang and others},
  editor = {Ku, Lun-Wei  and Martins, Andre  and Srikumar, Vivek},
  booktitle = {Proceedings of the 62nd Annual Meeting of the
  Association for Computational Linguistics (Volume 1: Long Papers)},
  year = {2024},
  journal = {Proceedings of the 62nd Annual Meeting of the
  Association for Computational Linguistics (Volume 1: Long Papers)},
  pages = {3828-3850},
  doi = {10.18653/v1/2024.acl-long.211},
  publisher = {Association for Computational Linguistics},
}

@inproceedings{liao-etal-2024-mario,
  title = {{MARIO}: {MA}th Reasoning with code Interpreter Output - A
  Reproducible Pipeline},
  author = {Liao, Minpeng and Li, Chengxi and Luo, Wei and Jing, Wu
  and Fan, Kai},
  editor = {Ku, Lun-Wei  and Martins, Andre  and Srikumar, Vivek},
  booktitle = {Findings of the Association for Computational
  Linguistics: ACL 2024},
  year = {2024},
  journal = {Findings of the Association for Computational
  Linguistics ACL 2024},
  pages = {905-924},
  doi = {10.18653/v1/2024.findings-acl.53},
  publisher = {Association for Computational Linguistics},
}

@inproceedings{ICLR2025_94074dd5,
  author = {Jain, Naman and Han, King and Gu, Alex and Li, Wen-Ding
    and Yan, Fanjia and Zhang, Tianjun and Wang, Sida I. and
  Solar-Lezama, Armando and Sen, Koushik and Stoica, Ion},
  booktitle = {International Conference on Learning Representations},
  editor = {Y. Yue and A. Garg and N. Peng and F. Sha and R. Yu},
  pages = {58791--58831},
  title = {{LiveCodeBench}: Holistic and Contamination Free
  Evaluation of Large Language Models for Code},
  url =
  {https://proceedings.iclr.cc/paper_files/paper/2025/file/94074dd5a072d28ff75a76dabed43767-Paper-Conference.pdf},
  volume = {2025},
  year = {2025},
}

@misc{maa_aime,
  author = {{MAA}},
  year = {2025},
  title = {{American Invitational Mathematics Examination} ({AIME}), 2022--2025},
  howpublished = {Official competition website},
  url = {https://maa.org/maa-invitational-competitions/},
  note = {Accessed September 22, 2026},
}

@misc{maa_amc,
  author = {{MAA}},
  year = {2023},
  title = {{American Mathematics Competitions} ({AMC}), 2022--2023},
  howpublished = {Official competition website},
  url = {https://maa.org/student-programs/amc/},
  note = {Accessed September 22, 2026},
}

@article{russell1991metareasoning,
  title = {Principles of metareasoning},
  journal = {Artificial Intelligence},
  volume = {49},
  number = {1--3},
  pages = {361-395},
  year = {1991},
  issn = {0004-3702},
  doi = {10.1016/0004-3702(91)90015-C},
  url = {https://www.sciencedirect.com/science/article/pii/000437029190015C},
  author = {Stuart Russell and Eric Wefald},
}

@misc{desabbata2025metareasoning,
  title = {Rational Metareasoning for Large Language Models},
  author = {De Sabbata, C. Nicolò and Theodore R. Sumers and Badr
  AlKhamissi and Antoine Bosselut and Thomas L. Griffiths},
  year = {2025},
  eprint = {2410.05563},
  archiveprefix = {arXiv},
  primaryclass = {cs.CL},
  url = {https://arxiv.org/abs/2410.05563},
}

@inproceedings{shridhar2021alfworld,
  title = {{ALFWorld}: Aligning Text and Embodied Environments for
  Interactive Learning},
  author = {Shridhar, Mohit and Yuan, Xingdi and C{\^o}t{\'e},
  Marc-Alexandre and Bisk, Yonatan and Trischler, Adam and Hausknecht, Matthew},
  booktitle = {International Conference on Learning Representations},
  year = {2021},
}

@inproceedings{wang2022scienceworld,
  title = {{ScienceWorld}: Is your Agent Smarter than a 5th Grader?},
  author = {Wang, Ruoyao and Jansen, Peter Alexander and Côté,
  Marc-Alexandre and Ammanabrolu, Prithviraj},
  booktitle = {Proceedings of the 2022 Conference on Empirical
  Methods in Natural Language Processing},
  pages = {11279--11298},
  year = {2022},
  journal = {Proceedings of the 2022 Conference on Empirical Methods
  in Natural Language Processing},
  doi = {10.18653/v1/2022.emnlp-main.775},
  publisher = {Association for Computational Linguistics},
}
\bibliographystyle{iclr2027_conference}
\clearpage\appendix
\raggedbottom %
\crefalias{section}{appendix}
\crefalias{subsection}{appendix}
\crefalias{subsubsection}{appendix}
\section{Experimental Protocol and Reproducibility}
\label{sec:implementation}
\subsection{Evaluation grading}
\label{sec:dataset-recipe}
\ecoparagraph{Admission and grading.}
Following DAPO-Math's integer-answer transformation \citep[Section~3.5]{yu2025dapo} and the integer-answer TheoremQA evaluation of \citet[Appendix~D.3]{wang2024tokeneconomies}, we use text-only integer-answer subsets to reduce answer-format ambiguity for intermediate and final answers. Questions are filtered, not rewritten; scores describe the admitted populations in \cref{tab:dataset-recipe}.

The parser strips whitespace, dollar delimiters, outer \texttt{\textbackslash boxed\{\}} wrappers, and trailing periods, then accepts only an optional minus sign followed by digits, so fractions, symbolic expressions, decimals, and comma-separated numerals fail. Source adapters convert integral numeric fields to integer strings. We exclude empty questions, questions over 2,000 characters, and those containing \texttt{[asy]}, \texttt{\textbackslash includegraphics}, or \texttt{<img}, gates that do not exhaustively detect images.

Methods in a comparison share the roster, reasoner trajectories, strict grader, token accounting, and operating-point constraints. Admission depends on reference format, not method performance; missing method coverage does not change the roster in \cref{tab:dataset-recipe}.

\begin{table}[!ht]

\caption{ID/OOD test questions after filtering and deduplication. DAPO, Numina, DSR, M500, Olymp., and GK denote DAPO-Math, NuminaMath, DeepScaleR, MATH-500, OlympiadBench, and Gaokao2023-Math-En.}
\label{tab:dataset-recipe}
\begin{center}
\setlength{\tabcolsep}{5pt}
\begin{tabular}{@{}lccc@{}}
\toprule
Dataset & Distribution & Questions & Composition \\
\midrule
DAPO & ID & 740 & Held-out DAPO source questions \\
Numina & ID & 473 & Held-out NuminaMath-hard source questions \\
DSR & ID & 481 & Held-out DeepScaleR source questions \\
M500 & OOD & 295 & Retained MATH-500 questions \\
Olymp. & OOD & 366 & English text-only mathematics, single answer \\
GK & OOD & 191 & Gaokao 2023 mathematics, English \\
AIME & OOD & 98 & 2022--25, nine known duplicates removed \\
AMC & OOD & 75 & 35 from 2022 + 40 from 2023 \\
\bottomrule
\end{tabular}
\end{center}
\end{table}

\ecoparagraph{Sources and deduplication.}
The ID bank draws from DAPO-Math-17k (prompt instructions stripped), NuminaMath-1.5 (valid problems tagged \texttt{olympiads}, \texttt{amc\_aime}, or \texttt{aops\_forum}), and DeepScaleR-Preview. After prior-use exclusions, normalized-text deduplication, and exact/13-word overlap screening against recorded external references, two disjoint batches each admit 1,500 DAPO-Math, 1,000 NuminaMath, and 1,000 DeepScaleR questions. Salted question-key hashing splits them 60\%/15\%/25\% into fitting, selection, and reporting; the reporting portions form the ID set.

MATH-500 keeps $500-189-16=295$ questions after integer and figure filtering, and Gaokao $385-194=191$ after integer parsing. OlympiadBench keeps 366 of 581 single-answer candidates after integer, length, and figure gates; its adapter separately excludes 93 multi-answer entries. AIME deduplication by contest identity keeps the original 2024 record without combining alternative trajectories. Missing identities for some AIME-2025 and AMC-2023 records prevent exhaustive semantic deduplication. Partial-coverage methods are unavailable outside aligned pools.

\ecoparagraph{Baseline implementations.}
We adapt the released implementations of TC, ROM, CoDE-Stop, and DEER to the shared protocol. OSP had no public code at the time of reproduction, so we implemented it from its published architecture and expected-utility objective \citep{ehab2026ospruner}; \cref{sec:checks-osp} checks the reproduction's performance.

\ecoparagraph{OS-Pruner implementation and training.}
OSP trains copies of the reasoner's final two decoder blocks and a sigmoid stopping head on their final-normalized boundary states, with the reasoner frozen. It optimizes exact expected utility on banked trajectories, with strict correctness labels and total generated-token cost including the forced answer. AdamW uses learning rate $10^{-5}$, weight decay $0.01$, $(\beta_1,\beta_2)=(0.9,0.999)$, $\epsilon=10^{-8}$, gradient clipping at norm 1, and three trajectories per step, in bfloat16 with float32 trainable weights; the head starts at stopping probability $0.1$. The training price is $\lambda_e=\lambda\,10^{1-\min(e,4)/4}$ for zero-based epoch $e$; validation utility at the final price selects checkpoints from $e\geq4$, with patience two. Qwen3 allows eight epochs and 16,384-token sequences, with training prices $\{2,5,10,20,40\}\times10^{-6}$ ($10^{-5}$ for OSP-1). OLMo3 OSP-1 uses $2\times10^{-5}$, seven epochs, and 20,480-token sequences. Both use training seeds 0, 1, and 2 and the 320-question validation bank. At the report price $5\times10^{-5}$, above the highest training price, development utility selects the $5\times10^{-6}$ checkpoint for all three seeds, so the full-OSP column there reports a low-price checkpoint; the $4\times10^{-5}$ checkpoint saves 1.1--1.4k more tokens per question but loses 11--12~pp accuracy.

\ecoparagraph{Qwen3 policy families.}
Full OSP supplies five checkpoints per utility-training run and OSP-1 13 threshold policies. Per probe training run, the price-refit variants BELT-B-R/H-R each supply 17 linear stopping policies, and the threshold variants BELT-B-T/H-T apply 13 source-validation logit-quantile thresholds to one policy fitted at $\lambda=10^{-5}$. BELT and OSP use three independent training runs. ROM uses two, with its layer-16 R1 configuration; TC uses two, with six configurations (correctness and stability targets at three smoothing weights). Fixed caps are deterministic.

CoDE-Stop and DEER decode a trial answer with their own prompt at each stopping boundary and read its confidence; total cost includes trial-answer tokens. Their Qwen3 grids of raw, logistic-calibrated, and isotonic-calibrated readouts, with calibrators fitted on the shared validation bank, have 225 and 54 configurations; the split-half diagnostic retains rosters of the same sizes (\cref{tab:cross-reading}). Family sizes and training effort differ; \cref{sec:testcurve-reading} specifies how they are evaluated.

\subsection{Natural-completion answer elicitation}
\label{sec:natural-elicitation}
A mathematics trajectory can close its thinking without a parseable final answer. For such a trajectory, unless it hit the reasoning-length cap, we elicit a boxed answer once from the completed prefix with the generator's backbone-specific prompt, sampler, trajectory-identity seed, and 32-token answer budget, and charge its tokens to natural-completion cost. In the two ID report banks and the OOD report bank, this returns a parseable answer for 504 of 1,032, 527 of 1,085 and 625 of 706 such Qwen3 chains. The elicited answer sets terminal correctness and agreement labels; if still unparseable, it counts as incorrect and agreement targets are masked. These outcomes enter \emph{End}, \emph{Later}, and \emph{Same} supervision and every policy's natural-completion option. LiveCodeBench is graded by unit tests and no closed code chain lacks a program, so code needs no elicitation. Per-bank statistics are in \cref{tab:natural-elicitation}.
{\scriptsize\setlength{\tabcolsep}{2.5pt}
\begin{longtable}{@{}lllrrrrr@{}}
\caption{Natural completion per bank and dataset with answer elicitation. Unparseable = the thinking text contains no parseable answer (math: no integer; code: no program). Recovered = the seeded \texttt{\textbackslash boxed\{} request returned a parseable / correct answer. Accuracy is over all rows and uses the recovered answers. Code banks are graded by unit tests and need no elicitation.}\label{tab:natural-elicitation}\\
\toprule Backbone & Bank & Dataset & Rows & Unpars.\ \% & Recovered & Still unpars.\ \% & Natural acc.\ \% \\\midrule\endfirsthead
\toprule Backbone & Bank & Dataset & Rows & Unpars.\ \% & Recovered & Still unpars.\ \% & Natural acc.\ \% \\\midrule\endhead
\bottomrule\endfoot
Qwen3-1.7B & math\_union\_fit & ALL & 42,408 & 18.0 & 2,918 / 2,097 & 9.1 & 53.57 \\
 &  & dapo & 18,048 & 10.7 & 1,169 / 797 & 1.8 & 56.42 \\
 &  & deepscaler & 12,288 & 11.8 & 837 / 661 & 3.5 & 61.25 \\
 &  & numina\_hard & 12,072 & 35.0 & 912 / 639 & 25.3 & 41.48 \\
\midrule
Qwen3-1.7B & math\_union\_select & ALL & 10,648 & 15.9 & 581 / 418 & 8.7 & 54.86 \\
 &  & dapo & 4,656 & 8.7 & 249 / 158 & 1.3 & 57.50 \\
 &  & deepscaler & 2,912 & 9.5 & 167 / 153 & 2.4 & 61.57 \\
 &  & numina\_hard & 3,080 & 31.9 & 165 / 107 & 25.1 & 44.51 \\
\midrule
Qwen3-1.7B & math\_t1\_report & ALL & 8,408 & 16.0 & 504 / 337 & 8.2 & 53.78 \\
 &  & dapo & 3,608 & 9.8 & 200 / 137 & 1.9 & 55.21 \\
 &  & deepscaler & 2,464 & 11.9 & 186 / 132 & 3.1 & 64.25 \\
 &  & numina\_hard & 2,336 & 30.0 & 118 / 68 & 23.4 & 40.54 \\
\midrule
Qwen3-1.7B & math\_t2\_report & ALL & 8,528 & 16.1 & 527 / 378 & 8.3 & 54.70 \\
 &  & dapo & 3,608 & 9.2 & 211 / 133 & 1.2 & 58.12 \\
 &  & deepscaler & 2,488 & 10.7 & 145 / 101 & 3.9 & 58.56 \\
 &  & numina\_hard & 2,432 & 31.3 & 171 / 144 & 22.6 & 45.68 \\
\midrule
Qwen3-1.7B & math\_ood\_report & ALL & 8,400 & 10.4 & 625 / 577 & 1.2 & 72.61 \\
 &  & aime24 & 464 & 0.9 & 0 / 0 & 0.9 & 34.91 \\
 &  & aime25 & 480 & 0.5 & 1 / 0 & 0.0 & 28.33 \\
 &  & amc23 & 640 & 1.6 & 6 / 6 & 0.4 & 71.25 \\
 &  & gaokao23en & 1,528 & 22.5 & 279 / 251 & 2.4 & 80.96 \\
 &  & math500 & 2,360 & 14.7 & 327 / 311 & 0.2 & 92.33 \\
 &  & olympiadbench & 2,928 & 2.3 & 12 / 9 & 1.7 & 65.88 \\
\midrule
Qwen3-1.7B & ood2\_math\_report & ALL & 1,328 & 0.4 & 1 / 0 & 0.3 & 41.72 \\
 &  & aime\_val & 768 & 0.6 & 0 / 0 & 0.6 & 30.47 \\
 &  & amc\_val & 560 & 0.3 & 1 / 0 & 0.0 & 57.14 \\
\midrule
Qwen3-1.7B & ood2\_code\_report & ALL & 4,392 & 0.0 & n/a & 0.0 & 62.52 \\
 &  & lcb\_rest & 4,392 & 0.0 & n/a & 0.0 & 62.52 \\
\midrule
Qwen3-1.7B & lcb\_post\_f2048 & ALL & 3,328 & 0.0 & n/a & 0.0 & 28.67 \\
 &  & lcb\_post & 3,328 & 0.0 & n/a & 0.0 & 28.67 \\
\midrule
OLMo-3-7B-Think & olmo3\_fit & ALL & 16,964 & 9.5 & 112 / 66 & 8.7 & 66.36 \\
 &  & dapo & 7,196 & 1.6 & 9 / 5 & 1.5 & 72.92 \\
 &  & deepscaler & 4,908 & 4.7 & 32 / 10 & 3.9 & 70.74 \\
 &  & numina\_hard & 4,860 & 26.1 & 71 / 51 & 24.3 & 52.22 \\
\midrule
OLMo-3-7B-Think & olmo3\_report & ALL & 3,804 & 1.2 & 13 / 11 & 0.8 & 83.65 \\
 &  & aime24 & 116 & 0.0 & 0 / 0 & 0.0 & 58.62 \\
 &  & aime25 & 120 & 0.0 & 0 / 0 & 0.0 & 41.67 \\
 &  & amc23 & 160 & 0.0 & 0 / 0 & 0.0 & 88.12 \\
 &  & gaokao23en & 764 & 4.1 & 12 / 10 & 2.4 & 88.35 \\
 &  & math500 & 1,180 & 0.3 & 1 / 1 & 0.2 & 96.19 \\
 &  & olympiadbench & 1,464 & 0.7 & 0 / 0 & 0.7 & 76.02 \\
\midrule
OLMo-3-7B-Think & olmo3\_ext74 & ALL & 296 & 0.0 & n/a & 0.0 & 62.16 \\
 &  & aime\_val & 156 & 0.0 & n/a & 0.0 & 48.08 \\
 &  & amc\_val & 140 & 0.0 & n/a & 0.0 & 77.86 \\
\end{longtable}}

\subsection{Generation and answer elicitation}
\label{sec:generation-settings}
We pin model weights and tokenizers to \texttt{Qwen/Qwen3-1.7B}, revision \nolinkurl{70d244cc86ccca08cf5af4e1e306ecf908b1ad5e}, and \texttt{allenai/Olmo-3-7B-Think}, revision \nolinkurl{d97e442d7cc678210054dbcc9b440894d62c89a4}. Both use the native chat template with an assistant generation prompt; Qwen3 sets \texttt{enable\_thinking=True}, and OLMo3 uses its template's default system turn and opening thinking tag. Reasoning and forced-answer decoding use temperature $0.6$, top-$p=0.95$, no min-$p$ filtering, and repetition penalty 1; top-$k$ is 20 for Qwen3 and disabled for OLMo3.

The user message is \texttt{Answer the following question. \{instruction\}\textbackslash n<Question>: \{question\}}, where \texttt{\textbackslash n} denotes a newline. For mathematics, \texttt{\{instruction\}} is \texttt{Reply with the answer only. Answer enclosed with \textbackslash boxed\{\}.} For code, it is \texttt{Reply with the complete Python 3 program only, inside one ```python code block.} Code questions retain their function signature or starter code, or, for standard-input tasks, append \texttt{Read the input from standard input and write the answer to standard output.}

At a forced-answer boundary, we append \verb|</think>\n\n\boxed{| (math) or \verb|</think>\n\n```python\n| (code) to the reasoning prefix and decode until \verb|}| or \verb|```|, keeping the closing delimiter. The opening delimiter counts as prompt; only newly generated answer tokens, at most 32 for math and 2,048 for code, enter the cost. The reasoning and replication budgets are specified in \cref{sec:training-recipe,sec:olmo-replication,sec:code-transfer}.

\subsection{Training trajectories and belief labels}
\label{sec:training-recipe}
The source bank has 4,241 questions from the three training datasets. Qwen3 rollouts allow 12,000 thinking tokens and 128 blank-line-delimited blocks, with a forced-answer query after every fourth block. Of the eight requested trajectories per question (16 for a historical subset), belief training keeps at most eight, with the lowest eligible replicate indices. After filtering, 3,918 questions (31,248 trajectories) serve probe training and 320 (2,554) probe validation. The full validation bank selects the BELT probe checkpoint; its disjoint 160-question halves serve policy fitting and development selection.

Belief training excludes trajectories missing reasoning-block boundaries or aligned forced-answer queries, but not incorrect or length-truncated ones; inconsistent query positions and duplicate boundary assignments raise validation errors. Evaluation retains trajectories without eligible queries and lets them complete naturally.

The labels follow \cref{eq:belief-targets} with natural-completion outcomes graded as in \cref{sec:natural-elicitation}. All belief probes, all Qwen3 OSP checkpoints and OLMo3 full OSP are trained on these labels, and policy fitting and development selection use the same outcomes. The OLMo3 OSP-1 checkpoints were trained on labels without answer elicitation, which differ in \emph{End} for 62 of 15,672 OLMo3 training chains (0.40\%); they are evaluated under the same grading as every other method. \emph{End} repeats natural-completion correctness at every block boundary; \emph{Later} is positive if a strictly later recorded forced answer or the natural-completion answer is correct. \emph{Now} and \emph{Same} are labelled at forced-answer positions, and an unparseable forced answer has \emph{Now} label zero. \emph{Same} is masked if the natural-completion answer is unparseable and otherwise indicates equality with it, with zero for an unparseable forced answer. Only available labels enter the belief loss.

\subsection{BELT probe and policy fitting}
\label{sec:probe-settings}
\ecoparagraph{Probe architecture.}
ReProbe's causal architecture encodes token-level states into a boundary representation and predicts outcome probabilities through a classification head:
\begin{equation}
h_i=E_\phi(H_i)
=\left[\operatorname{Transformer}^{\rm causal}_{\theta}
\!\left(\operatorname{MLP}_{\psi}(H_i)+M_i\right)\right]_{n_i},\qquad
\mathbf p_i=\sigma(W_{\rm bel}h_i+b_{\rm bel}),
\label{eq:reader}
\end{equation}
Here $\operatorname{MLP}_{\psi}$ (two linear layers with LayerNorm and GELU) projects each token state, $M_i$ adds learned embeddings distinguishing ordinary from block-end tokens, and $[\cdot]_{n_i}$ takes the causal Transformer's state at the current boundary, the last of the $n_i$ tokens in $H_i$. The parameters $\phi$ comprise the projection, marker embeddings and Transformer; $W_{\rm bel},b_{\rm bel}$ and the elementwise sigmoid $\sigma$ map $h_i$ to the beliefs $\mathbf p_i$. We adapt this architecture to predict the outcomes of answering now and of continuing.

\ecoparagraph{Instantiation and training.}
The Qwen3 BELT probe implements \cref{eq:reader} on 2,048-dimensional layer-16 states with a width-768 token-wise projection, block-end markers, and two causal Transformer layers with eight attention heads, read at marked block ends. It has 15,957,765 trainable parameters including belief heads. Inference uses only observed prefix states and, for BELT-B-R, prefix length; future answers, correctness, and final trajectory length are unavailable.

\Cref{tab:belief-hyperparameters} gives the belief training settings; each step averages trajectory losses, and evaluation mode disables the configured dropout of 0.1. Target weights are equal: Qwen3 sums the four losses ($4\mathcal L_{\rm bel}$) and OLMo3 averages them ($\mathcal L_{\rm bel}$). One minimum-validation-loss checkpoint serves all token prices; the three full-target Qwen3 runs stop after three epochs and select epoch 1 (epochs are one-based).

\begin{table}[t]

\caption{BELT probe training hyperparameters.}
\label{tab:belief-hyperparameters}
\begin{center}
\begin{tabular}{@{}lcc@{}}
\toprule
Setting & Qwen3 & OLMo3 \\
\midrule
Optimizer & \multicolumn{2}{c}{AdamW} \\
Encoder learning rate & \multicolumn{2}{c}{$5\times10^{-5}$} \\
Belief-head learning rate & \multicolumn{2}{c}{$10^{-3}$} \\
AdamW $(\beta_1,\beta_2)$; $\epsilon$ & \multicolumn{2}{c}{$(0.9,0.999)$; $10^{-8}$} \\
Weight decay (all parameter groups) & \multicolumn{2}{c}{$0.01$} \\
Gradient clipping (global norm) & \multicolumn{2}{c}{$1.0$} \\
Learning-rate schedule / warm-up & \multicolumn{2}{c}{Constant / none} \\
Trajectories per optimizer step & 8 & 1 \\
Additional gradient accumulation & \multicolumn{2}{c}{None} \\
Maximum epochs & 8 & 3 \\
Early-stopping patience (epochs) & 2 & 3 \\
Validation frequency & \multicolumn{2}{c}{Every epoch} \\
Checkpoint selection & \multicolumn{2}{c}{Minimum validation belief loss} \\
Coefficient per target loss & 1 & $1/4$ \\
Dropout during training & \multicolumn{2}{c}{Disabled (evaluation mode)} \\
Independent training runs & \multicolumn{2}{c}{3} \\
\bottomrule
\end{tabular}
\end{center}
\end{table}

For \cref{eq:policy}, belief probabilities are clipped to $[10^{-4},1-10^{-4}]$ before logit conversion. BELT-B-R adds prefix tokens divided by 12,000; BELT-H-R uses all 768 representation coordinates. Features are standardized on policy-fitting boundaries, with zero SD replaced by one. \Cref{tab:policy-hyperparameters} gives the Qwen3 fitting grid and optimizer; development utility selects regularization and iterate. Reported prices are $\{2,5,10,20,50\}\times10^{-6}$. BELT and OSP evaluation integrates stopping probabilities exactly over each banked trajectory, including natural completion.

\begin{table}[t]

\caption{Qwen3 linear stopping policy fitting hyperparameters.}
\label{tab:policy-hyperparameters}
\begin{center}
\begin{tabular}{@{}lp{0.62\textwidth}@{}}
\toprule
Setting & Value \\
\midrule
Optimizer / precision & L-BFGS / float64 \\
Learning rate / line search & 1.0 / strong Wolfe \\
History size & 20 \\
Maximum iterations & 500 \\
Gradient / change tolerance & $10^{-9}$ / $10^{-12}$ \\
Development evaluation & Every 25 iterations \\
Initial weights & $w=0$ \\
Initial intercepts & $b=\operatorname{logit}(p)$, $p\in\{0.1,0.01\}$ \\
Initial regularization grid & $\kappa\in\{0,10^{-4},10^{-3},10^{-2},10^{-1},1\}$ \\
Grid extension & If the largest $\kappa$ is selected, extend by $\times10$ up to $10^4$ \\
Penalized parameters & $w$ only; intercept unpenalized \\
Token-cost-weight grid & $2\times10^{-6}\cdot20^{k/12}$, $k=0,\ldots,16$ \\
Selection criterion & Maximum development utility \\
\bottomrule
\end{tabular}
\end{center}
\end{table}

\subsection{Efficient probe processing for long trajectories}
\label{sec:reader-efficiency}
We adapt ReProbe's causal step head\footnote{\url{https://github.com/ReProbe/ReProbe}, revision \texttt{fac0c16}.} to observed blank-line boundaries: one causal pass with scaled dot-product attention encodes the sequence with all boundaries marked and gathers their representations. Trajectory batches are right-padded, frozen reasoner features are kept in a bounded store, and exported BELT probe features are reused across token prices.

\subsection{Metrics, aggregation and uncertainty}
\label{sec:metric-protocol}
For training run $r$ and dataset $d$, averaging trajectories within questions, then questions, gives accuracy $a_{r,d}$ and total generated-token cost $c_{r,d}$; with natural cost $c^{\rm nat}_{d}$, token saving is $100(1-c_{r,d}/c^{\rm nat}_{d})$. Macro averages weight datasets equally; micro accuracy weights questions equally, and micro saving pools costs. Main tables are macro; \cref{tab:dataset-saving,tab:dataset-budget,tab:olmo-full} add micro, and Avg. columns pool ID and OOD. An aggregate missing a required dataset, question, or training run is unavailable.

Learned methods are scored per independent training run and reported as the mean (prose and main tables) with the sample SD (denominator $S-1$) as table subscripts. Single-run methods and deterministic controls have no training-run variation; varied policy-fitting starts are identified separately, and repeated evaluations are not independent runs. Runs are not paired across methods.

For Natural, Fixed cap, CoDE-Stop, and DEER, $\ddagger$ denotes a question-bootstrap standard error: 5,000 draws from a fixed pseudorandom sequence resample each dataset's questions independently at its size, keep trajectory-averaged outcomes, share identities between method and natural costs, and recompute dataset and aggregate metrics, including pooled-cost micro saving. Frontier members and mixture weights stay fixed, so these errors exclude frontier-selection uncertainty and variation across training runs.

\subsection{Operating points on accuracy--cost frontiers}
\label{sec:testcurve-reading}
\ecoparagraph{Test-frontier operating points.}
For each method, training run, and distribution, we score the complete exported family, plus natural completion as the never-stop policy, by micro accuracy and mean total generated-token cost. Nondominated members, in ascending cost, form the frontier; linear interpolation between adjacent points corresponds to randomizing between their policies with fixed probabilities.

At an accuracy tolerance (pp), we report the cheapest frontier point within that drop from natural accuracy; at a token-saving target, the accuracy at that fraction of natural cost, with costs below the cheapest member infeasible. The single-member comparison in \crefrange{tab:ceiling}{tab:ceiling-ood} instead takes the most accurate member within budget, without interpolation. Per-dataset and macro values evaluate the same distribution-level policy or mixture.

\ecoparagraph{Two utility protocols.}
Dev-selected utility selects policies on development data before test evaluation (\cref{sec:source-transfer,sec:utility-controls}). Test-family-max utility, used only for the direct-training diagnostic, takes the member or natural completion with the highest test utility at each price. Utility fixes the accuracy--cost trade-off without a matched constraint, whereas test-frontier operating points hold test accuracy or cost fixed; a constraint selected on source development data need not hold on the target distribution.

\ecoparagraph{Frontier plots.}
Curves average interpolated accuracy at common costs over the training runs' shared support, without extrapolation; per-dataset panels project the distribution-level mixtures. Uniform markers are selected mean-curve knots, including endpoints, not table operating points or necessarily single policies. Tables average constrained readings separately, so a mean cost at fixed accuracy need not lie on the plotted mean-accuracy curve. Validation supports checkpoint, calibration, and hyperparameter selection; test data determine frontier readings after the families are frozen.

\subsection{Compute resources}
Experiments use NVIDIA GH200, RTX 4090, and RTX PRO 6000 GPUs. Qwen3 generation on RTX 4090 uses vLLM 0.9.0.1, Transformers 4.52.4, and PyTorch \texttt{2.7.0+cu126}. The archived GH200 generation image contains vLLM build \nolinkurl{0.1.dev1+gf7e62e3d6.d20260427.cu132}, Transformers 4.57.6, and PyTorch \nolinkurl{2.11.0a0+a6c236b9fd.nv26.3.46836102}. A versioned roster fixes question identities, benchmark membership, and duplicate exclusions; primary Qwen3 exports record it with frontier members, mixture weights, grading, and cost conventions. \Cref{sec:cost-measured} specifies the scope of the training-cost estimates.

\subsection{OLMo3 cross-family replication}
\label{sec:olmo-replication}
OLMo3 uses its native thinking template, up to four trajectories per question, and a 16,384-token generation cap. A new BELT probe on layer-18 states, with the same two-layer, width-768 architecture and four belief targets, replicates the recipe without transferring Qwen3 weights. Of 4,241 source questions, 320 are held out for probe validation and split into 160-question policy-fitting and development sets. Policies are fitted with the OLMo3 fitting code, whose utility labels grade forced answers by exact match to the reference answer (loose grading) and use the natural-completion outcomes of \cref{sec:natural-elicitation}; test evaluation uses strict grading and total generated tokens.

All comparisons use the main 1,025-question OOD set, with natural accuracy 82.10\% (micro average; 78.63\% macro, \cref{tab:olmo-main}) and mean cost 6,868 tokens. Metrics are aggregated separately for each backbone. OLMo3 BELT-B-R omits prefix length, giving four standardized belief logits and five policy parameters including the intercept.

Per probe training run, BELT-B-R/H-R have 221 refit members, and BELT-B-T/H-T apply 13 thresholds to one policy fitted at $\lambda=2\times10^{-5}$. TC has 150 members across three smoothing settings, ROM 21 thresholds, full OSP five price-specific checkpoints, OSP-1 13, CoDE-Stop 75 configurations, DEER 18, and Fixed cap eight, each plus natural completion. BELT, TC, ROM, OSP and OSP-1 each use three independent training runs. Operating points follow \cref{sec:testcurve-reading}, with full results in \cref{tab:olmo-full}.

\subsection{Math to code domain transfer}
\label{sec:code-transfer}
LiveCodeBench provides 549 problems before August 2024 (Pre) and 416 from August 2024 onward (Post). Pre precedes the Qwen3 backbone's stated cutoff and may overlap its training period; both sets precede OLMo3's data cutoff. Qwen3 generates eight trajectories per question with a 12,000-token thinking cap and a forced-answer query every fourth reasoning block. Unit tests grade the forced and natural programs; a program not completed within 2,048 tokens counts as wrong, and cost includes the forced program.

Tests are the public and decoded private tests in \texttt{livecodebench/code\_generation\_lite}, revision \nolinkurl{0fe84c3912ea0c4d4a78037083943e8f0c4dd505}. We keep up to 15 tests per problem, longer serialized inputs first (ties by source order), excluding inputs over 1\,MiB and capping serialized tests at 4\,MiB in total; admission requires at least five retained tests. Each test runs in a fresh \texttt{python -I} subprocess with 10-second CPU and wall-clock limits, a 1\,GiB address-space limit, and an 8\,MiB file-size limit. Standard-output comparison normalizes whitespace, checks integer tokens exactly, and allows floating-point error $10^{-6}\max(1,|y|)$ for reference value $y$; function-call tests use the supplied entry point. A program is correct only if all retained tests pass; malformed programs, runtime errors, timeouts, and memory failures count as incorrect.

All controller and policy weights are trained on mathematics. Each pool receives 17 BELT-B-R/H-R policies and 13 BELT-B-T/H-T thresholds per probe training run, 13 OSP-1 thresholds and five full-OSP checkpoints per utility-training run, and natural completion. Frontier operating points are selected on code test outcomes; dev-selected utility keeps the mathematics-side selection (\cref{sec:source-transfer}). Each method has three independent training runs.

OLMo3 uses the same sets, tests, and grader, with four trajectories per question, its native thinking template, the 16,384-token thinking cap of its mathematics bank, and a forced program every fourth reasoning block under the same 2,048-token limit. Its BELT, OSP-1, and full-OSP families are those of \cref{sec:olmo-replication}, unchanged, plus natural completion (\cref{sec:olmo-code}).

\subsection{Per-turn thinking control in ALFWorld}
\label{sec:alfworld-protocol}
\ecoparagraph{Environment and prompt.} We use the ALFWorld text environment (\texttt{alfworld} 0.4.2, TextWorld 1.7.0, data release 2.1.1) with its own wrappers \citep{shridhar2021alfworld}: the 134 unseen-split games for evaluation and the 140 seen-split games only for threshold calibration. Each step is one user turn in OLMo3's native chat template with the task description, previous commands and observations, and current admissible commands, asking for one line \texttt{Action: <command>}; earlier thinking is never kept in context. Inadmissible commands are still sent and count as steps. An episode succeeds on task completion and fails after 50 actions. Sampling uses temperature 0.6 and top-$p$ 0.95 with one generation seed.

\ecoparagraph{Forced actions and cost.} Thinking is capped at 8,192 tokens per step and split into blank-line blocks as in the math data. A controller acts only before any thinking, at every fourth block boundary, and at the cap, where every method takes the forced action so the episode continues. Stopping appends \texttt{</think>\textbackslash n\textbackslash nAction:\textvisiblespace} and generates at most 32 tokens; natural completion uses the model's own closing tag and action line (at most 64 tokens). Cost is generated tokens per episode (thinking up to the stop plus action tokens; prompt tokens excluded).

\ecoparagraph{Shared trajectory tree.} All methods replay a shared tree of turns, expanded on demand and stored, whose nodes record the thinking, natural action and forced actions for one environment state. A stop whose forced action matches the natural action follows the natural child, so methods making the same decisions see the same observations and tokens, enabling game-paired comparisons (\cref{tab:alfworld-bootstrap}).

\ecoparagraph{Transferred controllers and calibration.} BELT uses the OLMo3 math probes (\cref{sec:olmo-replication}): BELT-H-T and BELT-B-T use the direction fitted at $\lambda=2\times10^{-5}$, and BELT-H-R and BELT-B-R choose among the 221 refit policies of one probe run. OSP-1 is the archived math checkpoint. No weight is trained or refit on ALFWorld. Each family has four members, calibrated on the disjoint seen split's natural trajectories without reading outcomes: for each target saving of 20, 35, 50 and 65\%, we keep the threshold (or refit policy) whose projected saving, when stopping at the first probe at or above the threshold, is closest. Realised evaluation savings lie within 6.1~pp of the targets (\cref{tab:alfworld-members}). The math-trained OSP-1 head assigns high stop probabilities from the first probe onward (median 0.9987 over the seen split's natural probes, 91.8\% at or above 0.97, calibrated thresholds between 0.9991 and 0.9996); calibration ranks them rather than reading them as absolute values, and its calibrated members reach their projected savings (\cref{tab:alfworld-members}). Operating points follow \cref{sec:testcurve-reading}, with success for accuracy. Agent tasks differ in how much per-turn thinking they need, so each is read on the range where its trade-off lies, with the same grid for every method within a task: ALFWorld at 20--50\% saving (\cref{sec:alfworld-results}) and ScienceWorld at 50--80\% (\cref{sec:sciworld-protocol}).

\ecoparagraph{Scale.} BELT-H-T and BELT-B-T use three probe training runs of four members each; BELT-H-R, BELT-B-R and OSP-1 use one run of four. Readings average over runs; the game bootstrap conditions on these runs and excludes run-to-run variation (\cref{sec:checks-alfworld}).

\subsection{Per-turn thinking control in ScienceWorld}
\label{sec:sciworld-protocol}
\ecoparagraph{Environment.} ScienceWorld \citep{wang2022scienceworld} (\texttt{scienceworld} 1.2.3) has 30 task types with train, development and test variations; we use the \emph{easy} simplification common to LLM-agent evaluations. We evaluate on the first three test variations of every task type (90 games) and calibrate only on the first three development variations (89 games; one task type has two). Each user turn gives the task, initial observation, previous commands and observations, action templates and visible objects, but not the valid commands. Rejected commands count as steps. An episode ends on task completion or failure, or after 50 actions, and scores the simulator's progress (0--100; 0 for a failed task). Forced actions, thinking cap and blocks, cost, the shared tree, sampling and the transferred controllers follow ALFWorld (\cref{sec:alfworld-protocol}), with a 24,576-token context limit. For deterministic replay, the simulator's Java virtual machine uses a constant identity hash.

\ecoparagraph{Calibration and readings.} Calibration applies the ALFWorld rule to the development natural trajectories at projected savings of 20, 35, 50, 65, 80, 90 and 95\%. The 95\% target exceeds the rule's 91.5\% calibration-split ceiling, a stop at the first probe of every turn, so every family's 95\% member is that first-probe stop; these members act identically in 87 of 90 test games. Realised test savings fall 3.2--20.0~pp short of the targets (mean 9.3; \cref{tab:sciworld-members}). Operating points follow \cref{sec:testcurve-reading} with mean score for accuracy. ScienceWorld needs far less per-turn thinking than ALFWorld: the score stays at or above natural up to 85\% saving, so budgets are read at 50--80\% and tolerances at 0, 2 and 5 points. BELT-H-T and BELT-B-T use three probe training runs, the other families one.

\section{Detailed Results by Dataset and Distribution}
\label{sec:extended-results}
The tables below follow the aggregation and uncertainty conventions in \cref{sec:metric-protocol}. Each dataset inherits its distribution-level operating point, so its accuracy or cost need not satisfy that constraint on its own. Avg. columns pool all ID/OOD questions for micro averages and weight datasets equally for macro averages. Natural and Fixed cap are omitted from token-saving tables.

\ecoparagraph{The single-encoder comparison.}
OSP-1 is BELT's primary comparator under a single-encoder constraint: both reuse one learned hidden-state encoder across operating points, whereas full OSP trains five price-specific encoders (\cref{sec:setup,sec:cost-adaptation}). Counting each BELT variant separately over the 49 readings of the seven settings (Qwen3 ID/OOD, LiveCodeBench Pre/Post, OLMo3 mathematics and both OLMo3 LiveCodeBench sets; \cref{tab:reb-single-variant-wtl,tab:allbaselines,tab:macro,tab:olmo-main,tab:code-main,tab:code-post,tab:code-post-accuracy,tab:olmo-code-pre,tab:olmo-code-post}), the recommended fixed-direction H configuration exceeds OSP-1 in 35, ties in 10 and trails in 4; the refit H configuration exceeds it in 38 and trails in none.

\ecoparagraph{Competitive trade-offs with full OSP.}
Against full OSP, 48 readings are jointly feasible; the OSP reading at 70\% saving on Qwen3 Pre is infeasible and excluded. The fixed-direction H configuration matches or exceeds full OSP in 40 readings (6 wins, 34 ties) and trails in 8; the refit B configuration wins 11, ties 30 and trails in 7. A tie means the difference of seed means is within the larger seed SD. On OLMo3 mathematics every BELT variant has more wins than losses against full OSP (the refit H configuration wins 6 and ties 1), and on OLMo3 code before August 2024 no BELT variant has a loss.

BELT uses one reusable approximately 16M-parameter probe rather than five approximately 101M-parameter OSP checkpoints (19M versus five 405M on OLMo3). Against one OSP checkpoint, it has $6.3\times$ fewer controller parameters and approximately $4.1\times$ fewer controller inference FLOPs per banked token (\cref{sec:cost,sec:cost-inference,tab:cost-summary}), and the archived reference schedules estimate approximately $5.1\times$ fewer training GPU-hours (10.0 versus 50.9, \cref{tab:cost-summary}; scope in \cref{sec:cost-measured}). Per-request OSP executes only one selected checkpoint.

\subsection{Per-dataset results}
\begin{table}[!htp]
\caption{Token saving given the accuracy tolerance, reported by dataset. Reporting convention follows \cref{tab:allbaselines}. Micro and Macro aggregate ID and OOD. DAPO, Numina, DSR, M500, Olymp., and GK denote DAPO-Math, NuminaMath, DeepScaleR, MATH-500, OlympiadBench, and Gaokao2023-Math-En, respectively. Panel: 0 pp tolerance, ID and pooled averages.}\label{tab:dataset-saving}
\begin{center}
\begingroup\setlength{\tabcolsep}{2.5pt}
\begin{tabular}{@{}lcccccc@{}}\toprule
\textbf{Method} & \shortstack{\textbf{Probe}\\\textbf{Params}} & \textbf{DAPO} & \textbf{Numina} & \textbf{DSR} & \textbf{Micro} & \textbf{Macro} \\\midrule
CoDE-Stop$^{\ddagger}$ & 0 & \cellcolor{Green3!21}\statcell{17.45}{0.63}{}& \cellcolor{Green3!12}\statcell{13.12}{0.94}{}& \cellcolor{Green3!23}\statcell{19.27}{1.00}{}& \cellcolor{Green3!19}\statcell{16.29}{0.37}{}& \cellcolor{Green3!22}\statcell{16.82}{0.50}{} \\
DEER$^{\ddagger}$ & 0 & \cellcolor{Green3!12}\statcell{11.08}{0.76}{}& \cellcolor{Green3!7}\statcell{9.22}{1.08}{}& \cellcolor{Green3!21}\statcell{18.01}{1.23}{}& \cellcolor{Green3!12}\statcell{11.41}{0.45}{}& \cellcolor{Green3!13}\statcell{11.26}{0.57}{} \\
TC & 514 & \cellcolor{Green3!24}\statcell{18.86}{0.59}{}& \cellcolor{Green3!24}\statcell{23.40}{1.03}{}& \cellcolor{Green3!31}\statcell{24.19}{0.37}{}& \cellcolor{Green3!26}\statcell{21.03}{0.54}{}& \cellcolor{Green3!27}\statcell{20.79}{0.35}{} \\
ROM & 5M & \cellcolor{Green3!0}\statcell{3.02}{4.27}{}& \cellcolor{Green3!0}\statcell{2.95}{4.17}{}& \cellcolor{Green3!0}\statcell{4.54}{6.41}{}& \cellcolor{Green3!0}\statcell{2.67}{3.78}{}& \cellcolor{Green3!0}\statcell{2.08}{2.93}{} \\
OSP & 101M $\times 5$ & \cellcolor{Green3!60}\secondstatcell{43.11}{1.50}{}& \cellcolor{Green3!56}\statcell{49.72}{1.33}{}& \cellcolor{Green3!58}\statcell{41.49}{0.59}{}& \cellcolor{Green3!52}\statcell{39.07}{7.13}{}& \cellcolor{Green3!45}\statcell{32.99}{13.86}{} \\
OSP-1 & 101M & \cellcolor{Green3!43}\statcell{31.80}{5.00}{}& \cellcolor{Green3!50}\statcell{44.94}{5.69}{}& \cellcolor{Green3!57}\statcell{40.82}{6.01}{}& \cellcolor{Green3!37}\statcell{28.64}{7.76}{}& \cellcolor{Green3!27}\statcell{20.36}{11.06}{} \\
\textbf{BELT-B-T} & 16M & \cellcolor{Green3!36}\statcell{27.25}{0.93}{}& \cellcolor{Green3!28}\statcell{26.59}{1.30}{}& \cellcolor{Green3!27}\statcell{21.80}{1.03}{}& \cellcolor{Green3!33}\statcell{25.82}{1.07}{}& \cellcolor{Green3!34}\statcell{25.22}{1.03}{} \\
\bottomrule\end{tabular}\endgroup
\end{center}
\end{table}

\begin{table}[!htp]
\caption{Token saving given the accuracy tolerance, reported by dataset. Reporting convention follows \cref{tab:allbaselines}. Micro and Macro aggregate ID and OOD. DAPO, Numina, DSR, M500, Olymp., and GK denote DAPO-Math, NuminaMath, DeepScaleR, MATH-500, OlympiadBench, and Gaokao2023-Math-En, respectively. Panel: 0 pp tolerance, OOD.}\label{tab:dataset-saving-0-1}
\begin{center}
\begingroup\setlength{\tabcolsep}{2.5pt}
\begin{tabular}{@{}lcccccc@{}}\toprule
\textbf{Method} & \shortstack{\textbf{Probe}\\\textbf{Params}} & \textbf{M500} & \textbf{Olymp.} & \textbf{GK} & \textbf{AIME} & \textbf{AMC} \\\midrule
CoDE-Stop$^{\ddagger}$ & 0 & \cellcolor{Green3!34}\statcell{24.11}{1.90}{}& \cellcolor{Green3!14}\statcell{11.48}{0.93}{}& \cellcolor{Green3!25}\statcell{16.02}{1.49}{}& \cellcolor{Green3!22}\statcell{15.22}{1.49}{}& \cellcolor{Green3!26}\statcell{17.88}{2.16}{} \\
DEER$^{\ddagger}$ & 0 & \cellcolor{Green3!33}\statcell{23.82}{2.23}{}& \cellcolor{Green3!9}\statcell{7.73}{1.11}{}& \cellcolor{Green3!14}\statcell{9.72}{1.66}{}& \cellcolor{Green3!4}\statcell{2.79}{1.47}{}& \cellcolor{Green3!11}\statcell{7.72}{2.49}{} \\
TC & 514 & \cellcolor{Green3!32}\statcell{22.97}{0.54}{}& \cellcolor{Green3!27}\statcell{21.86}{0.63}{}& \cellcolor{Green3!38}\statcell{23.16}{0.16}{}& \cellcolor{Green3!19}\statcell{13.15}{0.52}{}& \cellcolor{Green3!27}\statcell{18.73}{0.00}{} \\
ROM & 5M & \cellcolor{Green3!0}\statcell{2.61}{3.69}{}& \cellcolor{Green3!0}\statcell{1.03}{1.46}{}& \cellcolor{Green3!0}\statcell{1.53}{2.15}{}& \cellcolor{Green3!0}\statcell{0.27}{0.38}{}& \cellcolor{Green3!0}\statcell{0.66}{0.93}{} \\
OSP & 101M $\times 5$ & \cellcolor{Green3!35}\statcell{24.89}{21.69}{}& \cellcolor{Green3!39}\statcell{31.06}{26.92}{}& \cellcolor{Green3!37}\statcell{22.58}{19.55}{}& \cellcolor{Green3!36}\statcell{25.15}{21.98}{}& \cellcolor{Green3!38}\statcell{25.90}{22.51}{} \\
OSP-1 & 101M & \cellcolor{Green3!13}\statcell{10.89}{18.87}{}& \cellcolor{Green3!13}\statcell{10.89}{18.85}{}& \cellcolor{Green3!14}\statcell{9.57}{16.57}{}& \cellcolor{Green3!8}\statcell{5.93}{10.27}{}& \cellcolor{Green3!11}\statcell{8.08}{13.99}{} \\
\textbf{BELT-B-T} & 16M & \cellcolor{Green3!17}\statcell{13.38}{0.39}{}& \cellcolor{Green3!34}\statcell{27.23}{1.46}{}& \cellcolor{Green3!32}\statcell{20.02}{0.85}{}& \cellcolor{Green3!52}\statcell{36.30}{1.28}{}& \cellcolor{Green3!43}\statcell{29.18}{1.08}{} \\
\bottomrule\end{tabular}\endgroup
\end{center}
\end{table}

\begin{table}[!htp]
\caption{Token saving given the accuracy tolerance, reported by dataset. Reporting convention follows \cref{tab:allbaselines}. Micro and Macro aggregate ID and OOD. DAPO, Numina, DSR, M500, Olymp., and GK denote DAPO-Math, NuminaMath, DeepScaleR, MATH-500, OlympiadBench, and Gaokao2023-Math-En, respectively. Panel: 1 pp tolerance, ID and pooled averages.}\label{tab:dataset-saving-1-0}
\begin{center}
\begingroup\setlength{\tabcolsep}{2.5pt}
\begin{tabular}{@{}lcccccc@{}}\toprule
\textbf{Method} & \shortstack{\textbf{Probe}\\\textbf{Params}} & \textbf{DAPO} & \textbf{Numina} & \textbf{DSR} & \textbf{Micro} & \textbf{Macro} \\\midrule
CoDE-Stop$^{\ddagger}$ & 0 & \cellcolor{Green3!12}\statcell{23.83}{0.65}{}& \cellcolor{Green3!0}\statcell{18.08}{0.99}{}& \cellcolor{Green3!0}\statcell{23.90}{1.00}{}& \cellcolor{Green3!7}\statcell{22.95}{0.39}{}& \cellcolor{Green3!10}\statcell{24.49}{0.52}{} \\
DEER$^{\ddagger}$ & 0 & \cellcolor{Green3!0}\statcell{18.05}{0.90}{}& \cellcolor{Green3!1}\statcell{18.77}{1.35}{}& \cellcolor{Green3!6}\statcell{26.07}{1.32}{}& \cellcolor{Green3!1}\statcell{20.05}{0.53}{}& \cellcolor{Green3!2}\statcell{20.40}{0.68}{} \\
TC & 514 & \cellcolor{Green3!13}\statcell{24.42}{0.26}{}& \cellcolor{Green3!18}\statcell{29.91}{0.21}{}& \cellcolor{Green3!16}\statcell{30.03}{0.82}{}& \cellcolor{Green3!15}\statcell{26.82}{0.41}{}& \cellcolor{Green3!15}\statcell{26.46}{0.33}{} \\
ROM & 5M & \cellcolor{Green3!0}\statcell{18.14}{1.51}{}& \cellcolor{Green3!2}\statcell{19.49}{1.56}{}& \cellcolor{Green3!4}\statcell{25.38}{2.48}{}& \cellcolor{Green3!0}\statcell{19.55}{1.44}{}& \cellcolor{Green3!0}\statcell{19.60}{1.43}{} \\
OSP & 101M $\times 5$ & \cellcolor{Green3!60}\beststatcell{47.74}{1.36}{}& \cellcolor{Green3!57}\statcell{55.44}{2.37}{}& \cellcolor{Green3!56}\statcell{45.99}{0.49}{}& \cellcolor{Green3!59}\secondstatcell{48.86}{1.45}{}& \cellcolor{Green3!58}\secondstatcell{46.89}{1.46}{} \\
OSP-1 & 101M & \cellcolor{Green3!42}\statcell{38.60}{2.47}{}& \cellcolor{Green3!54}\statcell{53.64}{1.62}{}& \cellcolor{Green3!60}\beststatcell{47.54}{2.56}{}& \cellcolor{Green3!47}\statcell{43.11}{1.68}{}& \cellcolor{Green3!45}\statcell{40.53}{0.85}{} \\
\textbf{BELT-B-T} & 16M & \cellcolor{Green3!28}\statcell{31.77}{1.30}{}& \cellcolor{Green3!20}\statcell{31.08}{1.77}{}& \cellcolor{Green3!5}\statcell{25.85}{1.43}{}& \cellcolor{Green3!22}\statcell{30.66}{1.33}{}& \cellcolor{Green3!23}\statcell{30.38}{1.13}{} \\
\bottomrule\end{tabular}\endgroup
\end{center}
\end{table}

\begin{table}[!htp]
\caption{Token saving given the accuracy tolerance, reported by dataset. Reporting convention follows \cref{tab:allbaselines}. Micro and Macro aggregate ID and OOD. DAPO, Numina, DSR, M500, Olymp., and GK denote DAPO-Math, NuminaMath, DeepScaleR, MATH-500, OlympiadBench, and Gaokao2023-Math-En, respectively. Panel: 1 pp tolerance, OOD.}\label{tab:dataset-saving-1-1}
\begin{center}
\begingroup\setlength{\tabcolsep}{2.5pt}
\begin{tabular}{@{}lcccccc@{}}\toprule
\textbf{Method} & \shortstack{\textbf{Probe}\\\textbf{Params}} & \textbf{M500} & \textbf{Olymp.} & \textbf{GK} & \textbf{AIME} & \textbf{AMC} \\\midrule
CoDE-Stop$^{\ddagger}$ & 0 & \cellcolor{Green3!34}\statcell{32.42}{1.86}{}& \cellcolor{Green3!5}\statcell{20.20}{1.08}{}& \cellcolor{Green3!8}\statcell{25.10}{1.69}{}& \cellcolor{Green3!26}\statcell{25.05}{1.67}{}& \cellcolor{Green3!23}\statcell{27.31}{2.20}{} \\
DEER$^{\ddagger}$ & 0 & \cellcolor{Green3!40}\statcell{35.18}{2.41}{}& \cellcolor{Green3!5}\statcell{19.77}{1.42}{}& \cellcolor{Green3!2}\statcell{23.07}{2.35}{}& \cellcolor{Green3!0}\statcell{6.44}{1.68}{}& \cellcolor{Green3!0}\statcell{15.88}{3.04}{} \\
TC & 514 & \cellcolor{Green3!25}\statcell{28.62}{0.74}{}& \cellcolor{Green3!18}\statcell{27.84}{0.86}{}& \cellcolor{Green3!20}\statcell{28.91}{0.42}{}& \cellcolor{Green3!16}\statcell{17.87}{1.01}{}& \cellcolor{Green3!16}\statcell{24.06}{0.17}{} \\
ROM & 5M & \cellcolor{Green3!28}\statcell{29.77}{2.00}{}& \cellcolor{Green3!0}\statcell{16.95}{0.13}{}& \cellcolor{Green3!0}\statcell{22.57}{2.01}{}& \cellcolor{Green3!2}\statcell{8.15}{0.45}{}& \cellcolor{Green3!1}\statcell{16.38}{2.16}{} \\
OSP & 101M $\times 5$ & \cellcolor{Green3!50}\statcell{40.10}{2.31}{}& \cellcolor{Green3!60}\beststatcell{53.82}{2.52}{}& \cellcolor{Green3!51}\statcell{38.43}{0.72}{}& \cellcolor{Green3!57}\secondstatcell{47.94}{6.31}{}& \cellcolor{Green3!59}\secondstatcell{45.68}{2.72}{} \\
OSP-1 & 101M & \cellcolor{Green3!46}\statcell{37.98}{3.82}{}& \cellcolor{Green3!46}\statcell{45.51}{3.07}{}& \cellcolor{Green3!35}\statcell{33.41}{2.66}{}& \cellcolor{Green3!38}\statcell{34.37}{5.21}{}& \cellcolor{Green3!34}\statcell{33.20}{1.01}{} \\
\textbf{BELT-B-T} & 16M & \cellcolor{Green3!0}\statcell{17.01}{0.41}{}& \cellcolor{Green3!27}\statcell{33.31}{1.43}{}& \cellcolor{Green3!8}\statcell{25.09}{0.78}{}& \cellcolor{Green3!51}\statcell{43.47}{1.09}{}& \cellcolor{Green3!39}\statcell{35.47}{0.89}{} \\
\bottomrule\end{tabular}\endgroup
\end{center}
\end{table}

\begin{table}[!htp]
\caption{Token saving given the accuracy tolerance, reported by dataset. Reporting convention follows \cref{tab:allbaselines}. Micro and Macro aggregate ID and OOD. DAPO, Numina, DSR, M500, Olymp., and GK denote DAPO-Math, NuminaMath, DeepScaleR, MATH-500, OlympiadBench, and Gaokao2023-Math-En, respectively. Panel: 2 pp tolerance, ID and pooled averages.}\label{tab:dataset-saving-2-0}
\begin{center}
\begingroup\setlength{\tabcolsep}{2.5pt}
\begin{tabular}{@{}lcccccc@{}}\toprule
\textbf{Method} & \shortstack{\textbf{Probe}\\\textbf{Params}} & \textbf{DAPO} & \textbf{Numina} & \textbf{DSR} & \textbf{Micro} & \textbf{Macro} \\\midrule
CoDE-Stop$^{\ddagger}$ & 0 & \cellcolor{Green3!13}\statcell{27.23}{0.65}{}& \cellcolor{Green3!0}\statcell{21.35}{0.99}{}& \cellcolor{Green3!0}\statcell{25.96}{0.96}{}& \cellcolor{Green3!6}\statcell{26.34}{0.39}{}& \cellcolor{Green3!10}\statcell{28.16}{0.53}{} \\
DEER$^{\ddagger}$ & 0 & \cellcolor{Green3!0}\statcell{20.76}{0.95}{}& \cellcolor{Green3!2}\statcell{22.66}{1.39}{}& \cellcolor{Green3!8}\statcell{29.26}{1.34}{}& \cellcolor{Green3!0}\statcell{23.27}{0.55}{}& \cellcolor{Green3!0}\statcell{23.58}{0.71}{} \\
TC & 514 & \cellcolor{Green3!15}\statcell{28.44}{0.30}{}& \cellcolor{Green3!19}\statcell{34.19}{0.61}{}& \cellcolor{Green3!19}\statcell{34.10}{0.73}{}& \cellcolor{Green3!16}\statcell{31.21}{0.87}{}& \cellcolor{Green3!16}\statcell{31.12}{1.14}{} \\
ROM & 5M & \cellcolor{Green3!4}\statcell{22.75}{0.22}{}& \cellcolor{Green3!7}\statcell{26.00}{0.15}{}& \cellcolor{Green3!12}\statcell{30.97}{1.05}{}& \cellcolor{Green3!3}\statcell{24.90}{0.41}{}& \cellcolor{Green3!3}\statcell{24.97}{0.69}{} \\
OSP & 101M $\times 5$ & \cellcolor{Green3!60}\beststatcell{51.72}{0.79}{}& \cellcolor{Green3!59}\secondstatcell{60.37}{2.55}{}& \cellcolor{Green3!56}\statcell{50.34}{0.54}{}& \cellcolor{Green3!60}\beststatcell{53.55}{1.26}{}& \cellcolor{Green3!60}\beststatcell{51.98}{1.57}{} \\
OSP-1 & 101M & \cellcolor{Green3!44}\statcell{43.47}{2.08}{}& \cellcolor{Green3!57}\statcell{59.10}{1.00}{}& \cellcolor{Green3!60}\beststatcell{51.92}{1.86}{}& \cellcolor{Green3!51}\statcell{48.86}{0.95}{}& \cellcolor{Green3!50}\statcell{47.04}{0.06}{} \\
\textbf{BELT-B-T} & 16M & \cellcolor{Green3!29}\statcell{35.80}{1.49}{}& \cellcolor{Green3!21}\statcell{35.16}{2.08}{}& \cellcolor{Green3!9}\statcell{29.65}{1.66}{}& \cellcolor{Green3!23}\statcell{35.05}{1.44}{}& \cellcolor{Green3!24}\statcell{35.09}{1.12}{} \\
\bottomrule\end{tabular}\endgroup
\end{center}
\end{table}

\begin{table}[!htp]
\caption{Token saving given the accuracy tolerance, reported by dataset. Reporting convention follows \cref{tab:allbaselines}. Micro and Macro aggregate ID and OOD. DAPO, Numina, DSR, M500, Olymp., and GK denote DAPO-Math, NuminaMath, DeepScaleR, MATH-500, OlympiadBench, and Gaokao2023-Math-En, respectively. Panel: 2 pp tolerance, OOD.}\label{tab:dataset-saving-2-1}
\begin{center}
\begingroup\setlength{\tabcolsep}{2.5pt}
\begin{tabular}{@{}lcccccc@{}}\toprule
\textbf{Method} & \shortstack{\textbf{Probe}\\\textbf{Params}} & \textbf{M500} & \textbf{Olymp.} & \textbf{GK} & \textbf{AIME} & \textbf{AMC} \\\midrule
CoDE-Stop$^{\ddagger}$ & 0 & \cellcolor{Green3!27}\statcell{32.77}{1.84}{}& \cellcolor{Green3!2}\statcell{24.20}{1.08}{}& \cellcolor{Green3!2}\statcell{28.20}{1.85}{}& \cellcolor{Green3!30}\statcell{32.83}{1.70}{}& \cellcolor{Green3!25}\statcell{32.77}{2.19}{} \\
DEER$^{\ddagger}$ & 0 & \cellcolor{Green3!40}\statcell{38.20}{2.42}{}& \cellcolor{Green3!2}\statcell{24.02}{1.48}{}& \cellcolor{Green3!0}\statcell{27.62}{2.48}{}& \cellcolor{Green3!0}\statcell{7.82}{1.78}{}& \cellcolor{Green3!0}\statcell{18.31}{3.15}{} \\
TC & 514 & \cellcolor{Green3!31}\statcell{34.42}{1.86}{}& \cellcolor{Green3!16}\statcell{32.85}{2.07}{}& \cellcolor{Green3!22}\statcell{33.85}{1.18}{}& \cellcolor{Green3!18}\statcell{22.52}{1.30}{}& \cellcolor{Green3!18}\statcell{28.60}{1.10}{} \\
ROM & 5M & \cellcolor{Green3!35}\statcell{35.94}{1.94}{}& \cellcolor{Green3!0}\statcell{22.89}{0.29}{}& \cellcolor{Green3!3}\statcell{28.40}{1.20}{}& \cellcolor{Green3!4}\statcell{11.03}{0.30}{}& \cellcolor{Green3!6}\statcell{21.76}{1.58}{} \\
OSP & 101M $\times 5$ & \cellcolor{Green3!50}\statcell{42.88}{2.86}{}& \cellcolor{Green3!60}\beststatcell{59.41}{2.67}{}& \cellcolor{Green3!56}\statcell{43.57}{0.51}{}& \cellcolor{Green3!57}\secondstatcell{54.92}{6.90}{}& \cellcolor{Green3!60}\beststatcell{52.60}{3.60}{} \\
OSP-1 & 101M & \cellcolor{Green3!51}\statcell{43.28}{3.30}{}& \cellcolor{Green3!50}\statcell{53.32}{1.57}{}& \cellcolor{Green3!43}\statcell{39.78}{2.68}{}& \cellcolor{Green3!44}\statcell{43.79}{5.48}{}& \cellcolor{Green3!41}\statcell{41.67}{1.98}{} \\
\textbf{BELT-B-T} & 16M & \cellcolor{Green3!0}\statcell{20.64}{0.32}{}& \cellcolor{Green3!26}\statcell{38.82}{1.33}{}& \cellcolor{Green3!8}\statcell{29.85}{0.62}{}& \cellcolor{Green3!51}\statcell{49.65}{0.86}{}& \cellcolor{Green3!40}\statcell{41.10}{0.67}{} \\
\bottomrule\end{tabular}\endgroup
\end{center}
\end{table}

\begin{table}[!htp]
\caption{Acc. controlling the token budget, reported by dataset. Reporting convention follows \cref{tab:macro}. Micro and Macro aggregate ID and OOD. DAPO, Numina, DSR, M500, Olymp., and GK denote DAPO-Math, NuminaMath, DeepScaleR, MATH-500, OlympiadBench, and Gaokao2023-Math-En, respectively. Panel: 70\% token saving, ID and pooled averages.}\label{tab:dataset-budget}
\begin{center}
\begingroup\setlength{\tabcolsep}{2.5pt}
\begin{tabular}{@{}lcccccc@{}}\toprule
\textbf{Method} & \shortstack{\textbf{Probe}\\\textbf{Params}} & \textbf{DAPO} & \textbf{Numina} & \textbf{DSR} & \textbf{Micro} & \textbf{Macro} \\\midrule
Natural$^{\dagger\ddagger}$ & -- & \statcell{56.59}{1.51}{}& \statcell{42.77}{2.02}{}& \statcell{60.77}{1.90}{}& \statcell{61.01}{0.77}{}& \statcell{61.68}{0.95}{} \\
Fixed cap$^{\ddagger}$ & 0 & \cellcolor{Green3!16}\statcell{26.63}{1.32}{}& \cellcolor{Green3!27}\statcell{29.48}{1.78}{}& \cellcolor{Green3!30}\statcell{38.86}{1.93}{}& \cellcolor{Green3!30}\statcell{38.43}{0.73}{}& \cellcolor{Green3!31}\statcell{39.40}{0.89}{} \\
CoDE-Stop$^{\ddagger}$ & 0 & \cellcolor{Green3!25}\statcell{30.82}{1.32}{}& \cellcolor{Green3!32}\statcell{30.81}{1.75}{}& \cellcolor{Green3!29}\statcell{38.23}{1.85}{}& \cellcolor{Green3!34}\statcell{40.52}{0.73}{}& \cellcolor{Green3!35}\statcell{41.40}{0.92}{} \\
DEER$^{\ddagger}$ & 0 & \cellcolor{Green3!0}\statcell{19.47}{0.96}{}& \cellcolor{Green3!0}\statcell{22.06}{1.46}{}& \cellcolor{Green3!0}\statcell{25.45}{1.50}{}& \cellcolor{Green3!0}\statcell{24.41}{0.60}{}& \cellcolor{Green3!0}\statcell{23.99}{0.71}{} \\
TC & 514 & \cellcolor{Green3!17}\statcell{27.10}{0.40}{}& \cellcolor{Green3!30}\statcell{30.27}{0.74}{}& \cellcolor{Green3!24}\statcell{35.88}{0.22}{}& \cellcolor{Green3!28}\statcell{37.52}{0.56}{}& \cellcolor{Green3!29}\statcell{38.44}{0.43}{} \\
ROM & 5M & \cellcolor{Green3!16}\statcell{26.67}{1.18}{}& \cellcolor{Green3!19}\statcell{27.34}{1.27}{}& \cellcolor{Green3!18}\statcell{33.48}{2.11}{}& \cellcolor{Green3!23}\statcell{35.12}{2.50}{}& \cellcolor{Green3!24}\statcell{35.80}{2.38}{} \\
OSP & 101M $\times 5$ & \cellcolor{Green3!60}\secondstatcell{46.73}{0.41}{}& \cellcolor{Green3!58}\statcell{37.92}{0.27}{}& \cellcolor{Green3!55}\statcell{49.83}{0.33}{}& \cellcolor{Green3!60}\secondstatcell{52.72}{0.24}{}& \cellcolor{Green3!60}\beststatcell{53.63}{0.46}{} \\
\bottomrule\end{tabular}\endgroup
\end{center}
\end{table}

\begin{table}[!htp]
\caption{Acc. controlling the token budget, reported by dataset. Reporting convention follows \cref{tab:macro}. Micro and Macro aggregate ID and OOD. DAPO, Numina, DSR, M500, Olymp., and GK denote DAPO-Math, NuminaMath, DeepScaleR, MATH-500, OlympiadBench, and Gaokao2023-Math-En, respectively. Panel: 70\% token saving, OOD.}\label{tab:dataset-budget-0-1}
\begin{center}
\begingroup\setlength{\tabcolsep}{2.5pt}
\begin{tabular}{@{}lcccccc@{}}\toprule
\textbf{Method} & \shortstack{\textbf{Probe}\\\textbf{Params}} & \textbf{M500} & \textbf{Olymp.} & \textbf{GK} & \textbf{AIME} & \textbf{AMC} \\\midrule
Natural$^{\dagger\ddagger}$ & -- & \statcell{92.33}{1.19}{}& \statcell{65.88}{2.20}{}& \statcell{80.96}{2.52}{}& \statcell{29.46}{3.71}{}& \statcell{64.67}{4.51}{} \\
Fixed cap$^{\ddagger}$ & 0 & \cellcolor{Green3!44}\statcell{74.67}{1.96}{}& \cellcolor{Green3!30}\statcell{40.66}{2.07}{}& \cellcolor{Green3!42}\statcell{62.42}{3.02}{}& \cellcolor{Green3!4}\statcell{6.17}{1.93}{}& \cellcolor{Green3!27}\statcell{36.32}{4.61}{} \\
CoDE-Stop$^{\ddagger}$ & 0 & \cellcolor{Green3!45}\statcell{74.88}{1.84}{}& \cellcolor{Green3!40}\statcell{46.20}{2.15}{}& \cellcolor{Green3!37}\statcell{59.70}{3.02}{}& \cellcolor{Green3!14}\statcell{8.98}{2.43}{}& \cellcolor{Green3!36}\statcell{41.57}{4.71}{} \\
DEER$^{\ddagger}$ & 0 & \cellcolor{Green3!0}\statcell{39.47}{2.14}{}& \cellcolor{Green3!0}\statcell{23.29}{1.61}{}& \cellcolor{Green3!0}\statcell{37.50}{2.80}{}& \cellcolor{Green3!0}\statcell{5.17}{1.22}{}& \cellcolor{Green3!0}\statcell{19.47}{3.35}{} \\
TC & 514 & \cellcolor{Green3!33}\statcell{65.96}{1.62}{}& \cellcolor{Green3!36}\statcell{44.26}{0.67}{}& \cellcolor{Green3!34}\statcell{57.74}{0.61}{}& \cellcolor{Green3!17}\statcell{9.70}{0.29}{}& \cellcolor{Green3!28}\statcell{36.63}{0.56}{} \\
ROM & 5M & \cellcolor{Green3!31}\statcell{64.40}{8.13}{}& \cellcolor{Green3!27}\statcell{38.80}{3.18}{}& \cellcolor{Green3!26}\statcell{52.79}{4.77}{}& \cellcolor{Green3!15}\statcell{9.16}{2.16}{}& \cellcolor{Green3!23}\statcell{33.76}{0.55}{} \\
OSP & 101M $\times 5$ & \cellcolor{Green3!60}\statcell{86.90}{0.44}{}& \cellcolor{Green3!59}\secondstatcell{57.45}{0.40}{}& \cellcolor{Green3!59}\secondstatcell{72.79}{0.44}{}& \cellcolor{Green3!58}\secondstatcell{20.82}{0.94}{}& \cellcolor{Green3!60}\secondstatcell{56.60}{1.51}{} \\
\bottomrule\end{tabular}\endgroup
\end{center}
\end{table}

\begin{table}[!htp]
\caption{Acc. controlling the token budget, reported by dataset. Reporting convention follows \cref{tab:macro}. Micro and Macro aggregate ID and OOD. DAPO, Numina, DSR, M500, Olymp., and GK denote DAPO-Math, NuminaMath, DeepScaleR, MATH-500, OlympiadBench, and Gaokao2023-Math-En, respectively. Panel: 60\% token saving, ID and pooled averages.}\label{tab:dataset-budget-1-0}
\begin{center}
\begingroup\setlength{\tabcolsep}{2.5pt}
\begin{tabular}{@{}lcccccc@{}}\toprule
\textbf{Method} & \shortstack{\textbf{Probe}\\\textbf{Params}} & \textbf{DAPO} & \textbf{Numina} & \textbf{DSR} & \textbf{Micro} & \textbf{Macro} \\\midrule
Natural$^{\dagger\ddagger}$ & -- & \statcell{56.59}{1.51}{}& \statcell{42.77}{2.02}{}& \statcell{60.77}{1.90}{}& \statcell{61.01}{0.77}{}& \statcell{61.68}{0.95}{} \\
Fixed cap$^{\ddagger}$ & 0 & \cellcolor{Green3!14}\statcell{33.82}{1.34}{}& \cellcolor{Green3!28}\statcell{33.45}{1.82}{}& \cellcolor{Green3!31}\statcell{43.98}{1.89}{}& \cellcolor{Green3!30}\statcell{44.61}{0.73}{}& \cellcolor{Green3!32}\statcell{45.39}{0.92}{} \\
CoDE-Stop$^{\ddagger}$ & 0 & \cellcolor{Green3!29}\statcell{39.77}{1.38}{}& \cellcolor{Green3!33}\statcell{34.55}{1.80}{}& \cellcolor{Green3!31}\statcell{44.05}{1.81}{}& \cellcolor{Green3!37}\statcell{47.54}{0.73}{}& \cellcolor{Green3!38}\statcell{48.11}{0.94}{} \\
DEER$^{\ddagger}$ & 0 & \cellcolor{Green3!0}\statcell{28.37}{1.15}{}& \cellcolor{Green3!0}\statcell{26.06}{1.57}{}& \cellcolor{Green3!0}\statcell{31.85}{1.64}{}& \cellcolor{Green3!0}\statcell{31.98}{0.65}{}& \cellcolor{Green3!0}\statcell{31.25}{0.75}{} \\
TC & 514 & \cellcolor{Green3!25}\statcell{38.09}{0.41}{}& \cellcolor{Green3!39}\statcell{36.26}{0.38}{}& \cellcolor{Green3!31}\statcell{43.96}{0.39}{}& \cellcolor{Green3!35}\statcell{46.45}{0.46}{}& \cellcolor{Green3!36}\statcell{46.99}{0.56}{} \\
ROM & 5M & \cellcolor{Green3!16}\statcell{34.53}{1.17}{}& \cellcolor{Green3!23}\statcell{32.15}{1.20}{}& \cellcolor{Green3!19}\statcell{39.51}{1.59}{}& \cellcolor{Green3!24}\statcell{42.01}{1.85}{}& \cellcolor{Green3!26}\statcell{42.57}{1.84}{} \\
OSP & 101M $\times 5$ & \cellcolor{Green3!59}\statcell{51.51}{0.36}{}& \cellcolor{Green3!60}\secondstatcell{41.67}{0.32}{}& \cellcolor{Green3!57}\statcell{54.55}{0.93}{}& \cellcolor{Green3!60}\secondstatcell{56.81}{0.02}{}& \cellcolor{Green3!60}\beststatcell{57.58}{0.08}{} \\
\bottomrule\end{tabular}\endgroup
\end{center}
\end{table}

\begin{table}[!htp]
\caption{Acc. controlling the token budget, reported by dataset. Reporting convention follows \cref{tab:macro}. Micro and Macro aggregate ID and OOD. DAPO, Numina, DSR, M500, Olymp., and GK denote DAPO-Math, NuminaMath, DeepScaleR, MATH-500, OlympiadBench, and Gaokao2023-Math-En, respectively. Panel: 60\% token saving, OOD.}\label{tab:dataset-budget-1-1}
\begin{center}
\begingroup\setlength{\tabcolsep}{2.5pt}
\begin{tabular}{@{}lcccccc@{}}\toprule
\textbf{Method} & \shortstack{\textbf{Probe}\\\textbf{Params}} & \textbf{M500} & \textbf{Olymp.} & \textbf{GK} & \textbf{AIME} & \textbf{AMC} \\\midrule
Natural$^{\dagger\ddagger}$ & -- & \statcell{92.33}{1.19}{}& \statcell{65.88}{2.20}{}& \statcell{80.96}{2.52}{}& \statcell{29.46}{3.71}{}& \statcell{64.67}{4.51}{} \\
Fixed cap$^{\ddagger}$ & 0 & \cellcolor{Green3!47}\statcell{81.55}{1.74}{}& \cellcolor{Green3!33}\statcell{49.46}{2.13}{}& \cellcolor{Green3!44}\statcell{67.78}{2.96}{}& \cellcolor{Green3!0}\statcell{9.03}{2.31}{}& \cellcolor{Green3!30}\statcell{44.04}{4.82}{} \\
CoDE-Stop$^{\ddagger}$ & 0 & \cellcolor{Green3!50}\statcell{83.48}{1.63}{}& \cellcolor{Green3!43}\statcell{54.29}{2.17}{}& \cellcolor{Green3!45}\statcell{68.09}{2.91}{}& \cellcolor{Green3!9}\statcell{11.39}{2.66}{}& \cellcolor{Green3!39}\statcell{49.26}{5.03}{} \\
DEER$^{\ddagger}$ & 0 & \cellcolor{Green3!0}\statcell{49.57}{2.04}{}& \cellcolor{Green3!0}\statcell{33.61}{1.78}{}& \cellcolor{Green3!0}\statcell{44.28}{2.66}{}& \cellcolor{Green3!0}\statcell{9.01}{1.69}{}& \cellcolor{Green3!0}\statcell{27.29}{3.61}{} \\
TC & 514 & \cellcolor{Green3!40}\statcell{76.47}{0.77}{}& \cellcolor{Green3!42}\statcell{53.90}{0.30}{}& \cellcolor{Green3!40}\statcell{65.67}{0.65}{}& \cellcolor{Green3!25}\statcell{15.81}{0.22}{}& \cellcolor{Green3!33}\statcell{45.79}{1.33}{} \\
ROM & 5M & \cellcolor{Green3!34}\statcell{72.61}{5.90}{}& \cellcolor{Green3!28}\statcell{47.21}{1.37}{}& \cellcolor{Green3!28}\statcell{58.95}{3.05}{}& \cellcolor{Green3!18}\statcell{13.94}{0.64}{}& \cellcolor{Green3!26}\statcell{41.64}{1.07}{} \\
OSP & 101M $\times 5$ & \cellcolor{Green3!60}\secondstatcell{89.93}{0.59}{}& \cellcolor{Green3!57}\statcell{60.93}{0.13}{}& \cellcolor{Green3!60}\beststatcell{76.23}{0.75}{}& \cellcolor{Green3!58}\statcell{24.96}{0.75}{}& \cellcolor{Green3!60}\beststatcell{60.88}{0.36}{} \\
\bottomrule\end{tabular}\endgroup
\end{center}
\end{table}

\begin{table}[!htp]
\caption{Acc. controlling the token budget, reported by dataset. Reporting convention follows \cref{tab:macro}. Micro and Macro aggregate ID and OOD. DAPO, Numina, DSR, M500, Olymp., and GK denote DAPO-Math, NuminaMath, DeepScaleR, MATH-500, OlympiadBench, and Gaokao2023-Math-En, respectively. Panel: 50\% token saving, ID and pooled averages.}\label{tab:dataset-budget-2-0}
\begin{center}
\begingroup\setlength{\tabcolsep}{2.5pt}
\begin{tabular}{@{}lcccccc@{}}\toprule
\textbf{Method} & \shortstack{\textbf{Probe}\\\textbf{Params}} & \textbf{DAPO} & \textbf{Numina} & \textbf{DSR} & \textbf{Micro} & \textbf{Macro} \\\midrule
Natural$^{\dagger\ddagger}$ & -- & \statcell{56.59}{1.51}{}& \statcell{42.77}{2.02}{}& \statcell{60.77}{1.90}{}& \statcell{61.01}{0.77}{}& \statcell{61.68}{0.95}{} \\
Fixed cap$^{\ddagger}$ & 0 & \cellcolor{Green3!11}\statcell{40.77}{1.46}{}& \cellcolor{Green3!27}\statcell{37.28}{1.96}{}& \cellcolor{Green3!31}\statcell{48.92}{1.94}{}& \cellcolor{Green3!30}\statcell{50.09}{0.77}{}& \cellcolor{Green3!31}\statcell{50.48}{0.96}{} \\
CoDE-Stop$^{\ddagger}$ & 0 & \cellcolor{Green3!29}\statcell{46.08}{1.45}{}& \cellcolor{Green3!32}\statcell{38.33}{1.89}{}& \cellcolor{Green3!34}\statcell{49.73}{1.82}{}& \cellcolor{Green3!37}\statcell{52.49}{0.75}{}& \cellcolor{Green3!37}\statcell{52.47}{0.96}{} \\
DEER$^{\ddagger}$ & 0 & \cellcolor{Green3!0}\statcell{37.45}{1.27}{}& \cellcolor{Green3!0}\statcell{31.61}{1.68}{}& \cellcolor{Green3!0}\statcell{38.14}{1.71}{}& \cellcolor{Green3!0}\statcell{40.45}{0.69}{}& \cellcolor{Green3!0}\statcell{39.76}{0.81}{} \\
TC & 514 & \cellcolor{Green3!26}\statcell{45.21}{0.47}{}& \cellcolor{Green3!39}\statcell{39.69}{0.13}{}& \cellcolor{Green3!34}\statcell{49.87}{0.19}{}& \cellcolor{Green3!37}\statcell{52.31}{0.26}{}& \cellcolor{Green3!38}\statcell{52.73}{0.37}{} \\
ROM & 5M & \cellcolor{Green3!14}\statcell{41.45}{0.66}{}& \cellcolor{Green3!22}\statcell{36.17}{1.51}{}& \cellcolor{Green3!20}\statcell{45.16}{0.82}{}& \cellcolor{Green3!23}\statcell{48.02}{1.11}{}& \cellcolor{Green3!25}\statcell{48.48}{1.22}{} \\
OSP & 101M $\times 5$ & \cellcolor{Green3!58}\statcell{54.42}{0.23}{}& \cellcolor{Green3!60}\beststatcell{44.08}{0.17}{}& \cellcolor{Green3!60}\beststatcell{58.82}{0.53}{}& \cellcolor{Green3!60}\secondstatcell{59.77}{0.31}{}& \cellcolor{Green3!60}\beststatcell{60.44}{0.28}{} \\
\bottomrule\end{tabular}\endgroup
\end{center}
\end{table}

\begin{table}[!htp]
\caption{Acc. controlling the token budget, reported by dataset. Reporting convention follows \cref{tab:macro}. Micro and Macro aggregate ID and OOD. DAPO, Numina, DSR, M500, Olymp., and GK denote DAPO-Math, NuminaMath, DeepScaleR, MATH-500, OlympiadBench, and Gaokao2023-Math-En, respectively. Panel: 50\% token saving, OOD.}\label{tab:dataset-budget-2-1}
\begin{center}
\begingroup\setlength{\tabcolsep}{2.5pt}
\begin{tabular}{@{}lcccccc@{}}\toprule
\textbf{Method} & \shortstack{\textbf{Probe}\\\textbf{Params}} & \textbf{M500} & \textbf{Olymp.} & \textbf{GK} & \textbf{AIME} & \textbf{AMC} \\\midrule
Natural$^{\dagger\ddagger}$ & -- & \statcell{92.33}{1.19}{}& \statcell{65.88}{2.20}{}& \statcell{80.96}{2.52}{}& \statcell{29.46}{3.71}{}& \statcell{64.67}{4.51}{} \\
Fixed cap$^{\ddagger}$ & 0 & \cellcolor{Green3!49}\statcell{86.26}{1.60}{}& \cellcolor{Green3!36}\statcell{56.90}{2.22}{}& \cellcolor{Green3!42}\statcell{71.45}{2.94}{}& \cellcolor{Green3!0}\statcell{11.89}{2.72}{}& \cellcolor{Green3!31}\statcell{50.40}{5.05}{} \\
CoDE-Stop$^{\ddagger}$ & 0 & \cellcolor{Green3!51}\statcell{87.52}{1.49}{}& \cellcolor{Green3!42}\statcell{59.05}{2.22}{}& \cellcolor{Green3!45}\statcell{72.54}{2.91}{}& \cellcolor{Green3!5}\statcell{13.40}{2.81}{}& \cellcolor{Green3!36}\statcell{53.09}{5.11}{} \\
DEER$^{\ddagger}$ & 0 & \cellcolor{Green3!0}\statcell{61.92}{1.94}{}& \cellcolor{Green3!0}\statcell{44.26}{1.98}{}& \cellcolor{Green3!0}\statcell{54.78}{2.62}{}& \cellcolor{Green3!6}\statcell{13.57}{2.16}{}& \cellcolor{Green3!0}\statcell{36.35}{3.97}{} \\
TC & 514 & \cellcolor{Green3!43}\statcell{83.51}{0.55}{}& \cellcolor{Green3!45}\statcell{59.86}{0.20}{}& \cellcolor{Green3!40}\statcell{70.71}{0.00}{}& \cellcolor{Green3!35}\statcell{21.52}{0.39}{}& \cellcolor{Green3!33}\statcell{51.48}{1.45}{} \\
ROM & 5M & \cellcolor{Green3!34}\statcell{79.07}{3.33}{}& \cellcolor{Green3!29}\statcell{54.42}{0.06}{}& \cellcolor{Green3!26}\statcell{64.93}{1.15}{}& \cellcolor{Green3!25}\statcell{18.69}{0.88}{}& \cellcolor{Green3!25}\statcell{47.98}{1.34}{} \\
OSP & 101M $\times 5$ & \cellcolor{Green3!60}\beststatcell{91.83}{0.84}{}& \cellcolor{Green3!57}\statcell{64.29}{0.22}{}& \cellcolor{Green3!60}\beststatcell{78.39}{0.60}{}& \cellcolor{Green3!57}\statcell{27.74}{0.65}{}& \cellcolor{Green3!60}\beststatcell{63.98}{0.55}{} \\
\bottomrule\end{tabular}\endgroup
\end{center}
\end{table}

\begin{table}[!htp]
\caption{Acc. controlling the token budget, reported by dataset. Reporting convention follows \cref{tab:macro}. Micro and Macro aggregate ID and OOD. DAPO, Numina, DSR, M500, Olymp., and GK denote DAPO-Math, NuminaMath, DeepScaleR, MATH-500, OlympiadBench, and Gaokao2023-Math-En, respectively. Panel: 40\% token saving, ID and pooled averages.}\label{tab:dataset-budget-3-0}
\begin{center}
\begingroup\setlength{\tabcolsep}{2.5pt}
\begin{tabular}{@{}lcccccc@{}}\toprule
\textbf{Method} & \shortstack{\textbf{Probe}\\\textbf{Params}} & \textbf{DAPO} & \textbf{Numina} & \textbf{DSR} & \textbf{Micro} & \textbf{Macro} \\\midrule
Natural$^{\dagger\ddagger}$ & -- & \statcell{56.59}{1.51}{}& \statcell{42.77}{2.02}{}& \statcell{60.77}{1.90}{}& \statcell{61.01}{0.77}{}& \statcell{61.68}{0.95}{} \\
Fixed cap$^{\ddagger}$ & 0 & \cellcolor{Green3!0}\statcell{46.04}{1.49}{}& \cellcolor{Green3!20}\statcell{40.10}{1.95}{}& \cellcolor{Green3!27}\statcell{52.72}{1.90}{}& \cellcolor{Green3!22}\statcell{54.00}{0.76}{}& \cellcolor{Green3!22}\statcell{54.16}{0.96}{} \\
CoDE-Stop$^{\ddagger}$ & 0 & \cellcolor{Green3!9}\statcell{47.82}{1.48}{}& \cellcolor{Green3!23}\statcell{40.56}{1.96}{}& \cellcolor{Green3!31}\statcell{53.58}{1.88}{}& \cellcolor{Green3!28}\statcell{55.33}{0.75}{}& \cellcolor{Green3!29}\statcell{55.73}{0.95}{} \\
DEER$^{\ddagger}$ & 0 & \cellcolor{Green3!2}\statcell{46.44}{1.40}{}& \cellcolor{Green3!0}\statcell{37.37}{1.83}{}& \cellcolor{Green3!0}\statcell{46.05}{1.78}{}& \cellcolor{Green3!0}\statcell{49.59}{0.74}{}& \cellcolor{Green3!0}\statcell{49.34}{0.88}{} \\
TC & 514 & \cellcolor{Green3!24}\statcell{50.64}{0.03}{}& \cellcolor{Green3!35}\statcell{42.16}{0.50}{}& \cellcolor{Green3!33}\statcell{54.09}{0.37}{}& \cellcolor{Green3!34}\statcell{56.49}{0.20}{}& \cellcolor{Green3!35}\statcell{57.01}{0.18}{} \\
ROM & 5M & \cellcolor{Green3!8}\statcell{47.50}{0.34}{}& \cellcolor{Green3!19}\statcell{39.96}{1.20}{}& \cellcolor{Green3!18}\statcell{50.56}{0.42}{}& \cellcolor{Green3!18}\statcell{53.24}{0.63}{}& \cellcolor{Green3!20}\statcell{53.67}{0.71}{} \\
OSP & 101M $\times 5$ & \cellcolor{Green3!57}\statcell{56.92}{0.21}{}& \cellcolor{Green3!60}\beststatcell{45.60}{0.31}{}& \cellcolor{Green3!60}\beststatcell{60.69}{0.44}{}& \cellcolor{Green3!59}\secondstatcell{61.61}{0.14}{}& \cellcolor{Green3!59}\secondstatcell{62.13}{0.12}{} \\
\bottomrule\end{tabular}\endgroup
\end{center}
\end{table}

\begin{table}[!htp]
\caption{Acc. controlling the token budget, reported by dataset. Reporting convention follows \cref{tab:macro}. Micro and Macro aggregate ID and OOD. DAPO, Numina, DSR, M500, Olymp., and GK denote DAPO-Math, NuminaMath, DeepScaleR, MATH-500, OlympiadBench, and Gaokao2023-Math-En, respectively. Panel: 40\% token saving, OOD.}\label{tab:dataset-budget-3-1}
\begin{center}
\begingroup\setlength{\tabcolsep}{2.5pt}
\begin{tabular}{@{}lcccccc@{}}\toprule
\textbf{Method} & \shortstack{\textbf{Probe}\\\textbf{Params}} & \textbf{M500} & \textbf{Olymp.} & \textbf{GK} & \textbf{AIME} & \textbf{AMC} \\\midrule
Natural$^{\dagger\ddagger}$ & -- & \statcell{92.33}{1.19}{}& \statcell{65.88}{2.20}{}& \statcell{80.96}{2.52}{}& \statcell{29.46}{3.71}{}& \statcell{64.67}{4.51}{} \\
Fixed cap$^{\ddagger}$ & 0 & \cellcolor{Green3!47}\statcell{89.00}{1.44}{}& \cellcolor{Green3!32}\statcell{61.33}{2.20}{}& \cellcolor{Green3!31}\statcell{73.75}{2.87}{}& \cellcolor{Green3!0}\statcell{15.48}{3.01}{}& \cellcolor{Green3!23}\statcell{54.88}{5.03}{} \\
CoDE-Stop$^{\ddagger}$ & 0 & \cellcolor{Green3!50}\statcell{89.94}{1.32}{}& \cellcolor{Green3!40}\statcell{63.10}{2.20}{}& \cellcolor{Green3!39}\statcell{75.40}{2.78}{}& \cellcolor{Green3!12}\statcell{18.41}{3.12}{}& \cellcolor{Green3!30}\statcell{57.02}{4.91}{} \\
DEER$^{\ddagger}$ & 0 & \cellcolor{Green3!0}\statcell{76.49}{1.76}{}& \cellcolor{Green3!0}\statcell{54.15}{2.11}{}& \cellcolor{Green3!0}\statcell{66.92}{2.65}{}& \cellcolor{Green3!16}\statcell{19.31}{2.73}{}& \cellcolor{Green3!0}\statcell{47.97}{4.27}{} \\
TC & 514 & \cellcolor{Green3!42}\statcell{87.69}{0.22}{}& \cellcolor{Green3!41}\statcell{63.28}{0.16}{}& \cellcolor{Green3!36}\statcell{74.82}{0.16}{}& \cellcolor{Green3!42}\statcell{25.87}{0.19}{}& \cellcolor{Green3!32}\statcell{57.51}{0.23}{} \\
ROM & 5M & \cellcolor{Green3!28}\statcell{84.11}{1.04}{}& \cellcolor{Green3!24}\statcell{59.55}{0.49}{}& \cellcolor{Green3!17}\statcell{70.58}{0.09}{}& \cellcolor{Green3!32}\statcell{23.35}{1.91}{}& \cellcolor{Green3!19}\statcell{53.75}{0.40}{} \\
OSP & 101M $\times 5$ & \cellcolor{Green3!60}\beststatcell{92.56}{0.36}{}& \cellcolor{Green3!55}\statcell{66.29}{0.54}{}& \cellcolor{Green3!59}\secondstatcell{79.83}{0.20}{}& \cellcolor{Green3!57}\statcell{29.58}{0.74}{}& \cellcolor{Green3!58}\secondstatcell{65.56}{0.60}{} \\
\bottomrule\end{tabular}\endgroup
\end{center}
\end{table}

\FloatBarrier

\subsection{OLMo3 micro- and macro-average results}
\Crefrange{tab:olmo-full}{tab:olmo-full-1-1} retain both aggregation schemes for the OLMo3 comparison.
\begin{table}[!htp]
\caption{Token saving and accuracy on OLMo3, reported as Micro and Macro averages. Reporting convention follows \cref{tab:olmo-main}. Panel: token saving, micro averages. OSP-1+T retains the native stochastic policy alongside its existing 13 source-validation thresholds. OSP+T adds 13 source-validation thresholds to each of the five OSP checkpoints and keeps their native policies.}\label{tab:olmo-full}
\begin{center}
\begingroup\setlength{\tabcolsep}{4pt}
\begin{tabular}{@{}lcccc@{}}\toprule
\textbf{Method} & \shortstack{\textbf{Probe}\\\textbf{Params}} & \textbf{0 pp} & \textbf{1 pp} & \textbf{2 pp} \\\midrule
CoDE-Stop$^{\ddagger}$ & 0 & \cellcolor{Green3!0}\statcell{0.00}{0.00}{}& \cellcolor{Green3!7}\statcell{8.43}{0.27}{}& \cellcolor{Green3!7}\statcell{16.86}{0.53}{} \\
DEER$^{\ddagger}$ & 0 & \cellcolor{Green3!0}\statcell{0.00}{0.00}{}& \cellcolor{Green3!0}\statcell{1.71}{0.29}{}& \cellcolor{Green3!0}\statcell{10.98}{0.64}{} \\
TC & 514 & \cellcolor{Green3!28}\statcell{24.23}{0.07}{}& \cellcolor{Green3!30}\statcell{29.32}{0.19}{}& \cellcolor{Green3!27}\statcell{32.88}{0.18}{} \\
ROM & 8M & \cellcolor{Green3!31}\statcell{27.58}{3.79}{}& \cellcolor{Green3!35}\statcell{34.15}{1.16}{}& \cellcolor{Green3!31}\statcell{36.34}{0.87}{} \\
OSP & 405M $\times 5$ & \cellcolor{Green3!54}\statcell{47.32}{0.53}{}& \cellcolor{Green3!54}\statcell{51.31}{0.29}{}& \cellcolor{Green3!54}\statcell{55.15}{0.47}{} \\
OSP+T & 405M $\times 5$ & \cellcolor{Green3!56}\statcell{49.58}{3.37}{}& \cellcolor{Green3!54}\statcell{51.69}{1.72}{}& \cellcolor{Green3!55}\statcell{56.09}{0.45}{} \\
OSP-1 & 405M & \cellcolor{Green3!46}\statcell{40.64}{1.38}{}& \cellcolor{Green3!51}\statcell{48.93}{2.17}{}& \cellcolor{Green3!53}\statcell{54.04}{1.99}{} \\
OSP-1+T &405M & \cellcolor{Green3!46}\statcell{40.64}{1.38}{}& \cellcolor{Green3!51}\statcell{48.93}{2.17}{}& \cellcolor{Green3!53}\statcell{54.04}{1.99}{} \\
\textbf{BELT-B-T} & 19M & \cellcolor{Green3!59}\statcell{52.16}{2.18}{}& \cellcolor{Green3!58}\statcell{55.15}{2.03}{}& \cellcolor{Green3!57}\statcell{57.35}{1.75}{} \\
\textbf{BELT-H-T} & 19M & \cellcolor{Green3!60}\statcell{52.46}{3.22}{}& \cellcolor{Green3!59}\statcell{55.52}{3.23}{}& \cellcolor{Green3!57}\statcell{58.05}{3.46}{} \\
\textbf{BELT-B-R} & 19M & \cellcolor{Green3!60}\secondstatcell{52.71}{1.47}{}& \cellcolor{Green3!60}\secondstatcell{56.42}{2.75}{}& \cellcolor{Green3!60}\secondstatcell{60.12}{0.84}{} \\
\textbf{BELT-H-R} & 19M & \cellcolor{Green3!60}\beststatcell{52.78}{2.64}{}& \cellcolor{Green3!60}\beststatcell{56.78}{2.32}{}& \cellcolor{Green3!60}\beststatcell{60.14}{1.70}{} \\
\bottomrule\end{tabular}\endgroup
\end{center}
\end{table}

\begin{table}[!htp]
\caption{Token saving and accuracy on OLMo3, reported as Micro and Macro averages. Reporting convention follows \cref{tab:olmo-main}. Panel: token saving, macro averages.}\label{tab:olmo-full-0-1}
\begin{center}
\begingroup\setlength{\tabcolsep}{4pt}
\begin{tabular}{@{}lcccc@{}}\toprule
\textbf{Method} & \shortstack{\textbf{Probe}\\\textbf{Params}} & \textbf{0 pp} & \textbf{1 pp} & \textbf{2 pp} \\\midrule
CoDE-Stop$^{\ddagger}$ & 0 & \cellcolor{Green3!0}\statcell{0.00}{0.00}{}& \cellcolor{Green3!8}\statcell{8.64}{0.31}{}& \cellcolor{Green3!8}\statcell{17.27}{0.63}{} \\
DEER$^{\ddagger}$ & 0 & \cellcolor{Green3!0}\statcell{0.00}{0.00}{}& \cellcolor{Green3!0}\statcell{1.57}{0.25}{}& \cellcolor{Green3!0}\statcell{10.80}{0.68}{} \\
TC & 514 & \cellcolor{Green3!28}\statcell{24.27}{0.07}{}& \cellcolor{Green3!31}\statcell{29.16}{0.04}{}& \cellcolor{Green3!28}\statcell{33.21}{0.41}{} \\
ROM & 8M & \cellcolor{Green3!34}\statcell{29.17}{3.83}{}& \cellcolor{Green3!38}\statcell{35.63}{1.17}{}& \cellcolor{Green3!34}\statcell{37.72}{0.94}{} \\
OSP & 405M $\times 5$ & \cellcolor{Green3!53}\statcell{45.92}{0.64}{}& \cellcolor{Green3!54}\statcell{49.86}{0.27}{}& \cellcolor{Green3!54}\statcell{53.73}{0.50}{} \\
OSP+T & 405M $\times 5$ & \cellcolor{Green3!56}\statcell{48.09}{3.49}{}& \cellcolor{Green3!55}\statcell{50.36}{1.68}{}& \cellcolor{Green3!55}\statcell{54.37}{0.53}{} \\
OSP-1 & 405M & \cellcolor{Green3!46}\statcell{39.83}{1.36}{}& \cellcolor{Green3!51}\statcell{47.54}{1.87}{}& \cellcolor{Green3!52}\statcell{52.45}{1.69}{} \\
OSP-1+T &405M & \cellcolor{Green3!46}\statcell{39.83}{1.36}{}& \cellcolor{Green3!51}\statcell{47.54}{1.87}{}& \cellcolor{Green3!52}\statcell{52.45}{1.69}{} \\
\textbf{BELT-B-T} & 19M & \cellcolor{Green3!58}\statcell{50.14}{2.14}{}& \cellcolor{Green3!58}\statcell{53.09}{1.94}{}& \cellcolor{Green3!56}\statcell{55.26}{1.72}{} \\
\textbf{BELT-H-T} & 19M & \cellcolor{Green3!60}\secondstatcell{51.22}{3.01}{}& \cellcolor{Green3!59}\statcell{54.09}{3.00}{}& \cellcolor{Green3!57}\statcell{56.51}{3.18}{} \\
\textbf{BELT-B-R} & 19M & \cellcolor{Green3!59}\statcell{50.61}{1.51}{}& \cellcolor{Green3!59}\secondstatcell{54.30}{2.95}{}& \cellcolor{Green3!60}\secondstatcell{58.15}{0.86}{} \\
\textbf{BELT-H-R} & 19M & \cellcolor{Green3!60}\beststatcell{51.59}{2.26}{}& \cellcolor{Green3!60}\beststatcell{55.22}{2.39}{}& \cellcolor{Green3!60}\beststatcell{58.52}{1.98}{} \\
\bottomrule\end{tabular}\endgroup
\end{center}
\end{table}

\begin{table}[!htp]
\caption{Token saving and accuracy on OLMo3, reported as Micro and Macro averages. Reporting convention follows \cref{tab:olmo-main}. Panel: accuracy, micro averages.}\label{tab:olmo-full-1-0}
\begin{center}
\begingroup\setlength{\tabcolsep}{4pt}
\begin{tabular}{@{}lccccc@{}}\toprule
\textbf{Method} & \shortstack{\textbf{Probe}\\\textbf{Params}} & \textbf{70\%} & \textbf{60\%} & \textbf{50\%} & \textbf{40\%} \\\midrule
Natural$^{\dagger\ddagger}$ & -- & \statcell{82.10}{}{}& \statcell{82.10}{}{}& \statcell{82.10}{}{}& \statcell{82.10}{}{} \\
Fixed cap$^{\ddagger}$ & 0 & \cellcolor{Green3!39}\statcell{61.60}{1.22}{}& \cellcolor{Green3!39}\statcell{68.00}{1.20}{}& \cellcolor{Green3!35}\statcell{72.69}{1.14}{}& \cellcolor{Green3!34}\statcell{76.63}{1.07}{} \\
CoDE-Stop$^{\ddagger}$ & 0 & \cellcolor{Green3!33}\statcell{56.94}{1.22}{}& \cellcolor{Green3!32}\statcell{63.68}{1.17}{}& \cellcolor{Green3!27}\statcell{69.34}{1.15}{}& \cellcolor{Green3!25}\statcell{73.50}{1.09}{} \\
DEER$^{\ddagger}$ & 0 & \cellcolor{Green3!9}\statcell{40.28}{1.05}{}& \cellcolor{Green3!7}\statcell{49.13}{1.00}{}& \cellcolor{Green3!0}\statcell{57.97}{1.05}{}& \cellcolor{Green3!0}\statcell{65.55}{1.04}{} \\
TC & 514 & \cellcolor{Green3!38}\statcell{60.67}{0.19}{}& \cellcolor{Green3!41}\statcell{69.24}{0.11}{}& \cellcolor{Green3!38}\statcell{74.12}{0.08}{}& \cellcolor{Green3!40}\statcell{78.36}{0.06}{} \\
ROM & 8M & \cellcolor{Green3!0}\statcell{34.13}{1.03}{}& \cellcolor{Green3!0}\statcell{45.03}{0.80}{}& \cellcolor{Green3!11}\statcell{62.60}{1.10}{}& \cellcolor{Green3!38}\statcell{77.90}{0.69}{} \\
OSP & 405M $\times 5$ & \cellcolor{Green3!57}\statcell{74.05}{0.17}{}& \cellcolor{Green3!57}\statcell{78.43}{0.34}{}& \cellcolor{Green3!56}\statcell{81.39}{0.12}{}& \cellcolor{Green3!56}\statcell{83.80}{0.14}{} \\
OSP+T & 405M $\times 5$ & \cellcolor{Green3!58}\secondstatcell{74.28}{0.31}{}& \cellcolor{Green3!57}\statcell{78.56}{0.21}{}& \cellcolor{Green3!57}\statcell{81.82}{0.89}{}& \cellcolor{Green3!58}\statcell{84.30}{0.25}{} \\
OSP-1 & 405M & \cellcolor{Green3!57}\statcell{74.06}{0.98}{}& \cellcolor{Green3!58}\statcell{78.66}{0.61}{}& \cellcolor{Green3!55}\statcell{80.91}{0.37}{}& \cellcolor{Green3!51}\statcell{82.12}{0.09}{} \\
OSP-1+T &405M & \cellcolor{Green3!58}\statcell{74.13}{1.00}{}& \cellcolor{Green3!58}\statcell{78.66}{0.61}{}& \cellcolor{Green3!55}\statcell{80.91}{0.37}{}& \cellcolor{Green3!51}\statcell{82.12}{0.09}{} \\
\textbf{BELT-B-T} & 19M & \cellcolor{Green3!50}\statcell{69.13}{2.75}{}& \cellcolor{Green3!57}\statcell{78.59}{1.00}{}& \cellcolor{Green3!59}\secondstatcell{82.77}{0.88}{}& \cellcolor{Green3!58}\statcell{84.46}{0.80}{} \\
\textbf{BELT-H-T} & 19M & \cellcolor{Green3!52}\statcell{70.23}{3.64}{}& \cellcolor{Green3!58}\statcell{78.97}{1.92}{}& \cellcolor{Green3!59}\statcell{82.65}{0.79}{}& \cellcolor{Green3!58}\statcell{84.30}{0.52}{} \\
\textbf{BELT-B-R} & 19M & \cellcolor{Green3!56}\statcell{73.05}{1.27}{}& \cellcolor{Green3!60}\secondstatcell{80.06}{0.44}{}& \cellcolor{Green3!60}\beststatcell{83.15}{0.48}{}& \cellcolor{Green3!60}\beststatcell{84.99}{0.34}{} \\
\textbf{BELT-H-R} & 19M & \cellcolor{Green3!60}\beststatcell{75.86}{0.19}{}& \cellcolor{Green3!60}\beststatcell{80.08}{0.71}{}& \cellcolor{Green3!59}\statcell{82.70}{0.39}{}& \cellcolor{Green3!58}\secondstatcell{84.47}{0.49}{} \\
\bottomrule\end{tabular}\endgroup
\end{center}
\end{table}

\begin{table}[!htp]
\caption{Token saving and accuracy on OLMo3, reported as Micro and Macro averages. Reporting convention follows \cref{tab:olmo-main}. Panel: accuracy, macro averages.}\label{tab:olmo-full-1-1}
\begin{center}
\begingroup\setlength{\tabcolsep}{4pt}
\begin{tabular}{@{}lccccc@{}}\toprule
\textbf{Method} & \shortstack{\textbf{Probe}\\\textbf{Params}} & \textbf{70\%} & \textbf{60\%} & \textbf{50\%} & \textbf{40\%} \\\midrule
Natural$^{\dagger\ddagger}$ & -- & \statcell{78.63}{}{}& \statcell{78.63}{}{}& \statcell{78.63}{}{}& \statcell{78.63}{}{} \\
Fixed cap$^{\ddagger}$ & 0 & \cellcolor{Green3!33}\statcell{53.88}{1.47}{}& \cellcolor{Green3!31}\statcell{60.26}{1.52}{}& \cellcolor{Green3!29}\statcell{65.82}{1.47}{}& \cellcolor{Green3!28}\statcell{70.67}{1.43}{} \\
CoDE-Stop$^{\ddagger}$ & 0 & \cellcolor{Green3!26}\statcell{49.19}{1.40}{}& \cellcolor{Green3!23}\statcell{55.71}{1.44}{}& \cellcolor{Green3!19}\statcell{61.34}{1.47}{}& \cellcolor{Green3!16}\statcell{66.23}{1.41}{} \\
DEER$^{\ddagger}$ & 0 & \cellcolor{Green3!5}\statcell{35.65}{1.15}{}& \cellcolor{Green3!2}\statcell{44.25}{1.13}{}& \cellcolor{Green3!0}\statcell{52.85}{1.22}{}& \cellcolor{Green3!0}\statcell{60.52}{1.25}{} \\
TC & 514 & \cellcolor{Green3!34}\statcell{54.02}{0.21}{}& \cellcolor{Green3!36}\statcell{63.04}{0.19}{}& \cellcolor{Green3!35}\statcell{68.50}{0.08}{}& \cellcolor{Green3!39}\statcell{74.33}{0.11}{} \\
ROM & 8M & \cellcolor{Green3!0}\statcell{32.17}{1.22}{}& \cellcolor{Green3!0}\statcell{43.08}{0.37}{}& \cellcolor{Green3!15}\statcell{59.79}{0.49}{}& \cellcolor{Green3!40}\statcell{74.78}{0.57}{} \\
OSP & 405M $\times 5$ & \cellcolor{Green3!58}\statcell{69.80}{0.27}{}& \cellcolor{Green3!58}\statcell{75.09}{0.45}{}& \cellcolor{Green3!56}\statcell{78.26}{0.10}{}& \cellcolor{Green3!56}\statcell{80.66}{0.14}{} \\
OSP+T & 405M $\times 5$ & \cellcolor{Green3!59}\secondstatcell{70.15}{0.25}{}& \cellcolor{Green3!58}\statcell{75.10}{0.59}{}& \cellcolor{Green3!57}\statcell{78.57}{0.85}{}& \cellcolor{Green3!58}\statcell{81.21}{0.25}{} \\
OSP-1 & 405M & \cellcolor{Green3!58}\statcell{69.74}{0.63}{}& \cellcolor{Green3!58}\statcell{75.32}{0.78}{}& \cellcolor{Green3!56}\statcell{78.02}{0.44}{}& \cellcolor{Green3!53}\statcell{79.36}{0.07}{} \\
OSP-1+T &405M & \cellcolor{Green3!58}\statcell{69.83}{0.68}{}& \cellcolor{Green3!58}\statcell{75.32}{0.78}{}& \cellcolor{Green3!56}\statcell{78.02}{0.44}{}& \cellcolor{Green3!53}\statcell{79.36}{0.07}{} \\
\textbf{BELT-B-T} & 19M & \cellcolor{Green3!51}\statcell{65.16}{2.24}{}& \cellcolor{Green3!58}\statcell{75.19}{1.11}{}& \cellcolor{Green3!59}\secondstatcell{79.60}{0.98}{}& \cellcolor{Green3!59}\secondstatcell{81.51}{0.80}{} \\
\textbf{BELT-H-T} & 19M & \cellcolor{Green3!52}\statcell{65.74}{3.82}{}& \cellcolor{Green3!58}\statcell{75.14}{2.09}{}& \cellcolor{Green3!58}\statcell{79.04}{0.92}{}& \cellcolor{Green3!57}\statcell{81.03}{0.63}{} \\
\textbf{BELT-B-R} & 19M & \cellcolor{Green3!56}\statcell{68.68}{1.11}{}& \cellcolor{Green3!60}\beststatcell{76.48}{0.60}{}& \cellcolor{Green3!60}\beststatcell{80.05}{0.75}{}& \cellcolor{Green3!60}\beststatcell{82.04}{0.31}{} \\
\textbf{BELT-H-R} & 19M & \cellcolor{Green3!60}\beststatcell{71.11}{0.39}{}& \cellcolor{Green3!60}\secondstatcell{76.30}{0.54}{}& \cellcolor{Green3!58}\statcell{79.19}{0.41}{}& \cellcolor{Green3!57}\statcell{81.05}{0.57}{} \\
\bottomrule\end{tabular}\endgroup
\end{center}
\end{table}

\FloatBarrier

\subsection{LiveCodeBench Post results}
\label{sec:code-post}
\begin{figure}[t]\centering
\includegraphics[width=.55\textwidth]{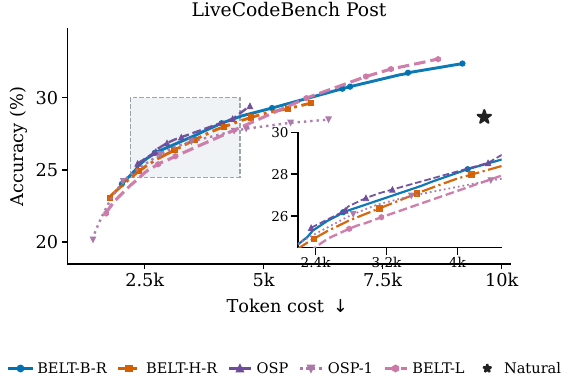}
\caption{LiveCodeBench Post native frontiers. Mean curves use common seed cost support; their endpoints are not per-method maximum accuracies. Other conventions follow the LiveCodeBench Pre panel of \cref{fig:tradeoff-main}.}
\label{fig:code-post}
\end{figure}

On LiveCodeBench Post (416 problems), BELT-B-R exceeds OSP-1 at all seven readings. Against full OSP, every BELT variant ties at most of the seven readings (\cref{tab:reb-single-variant-wtl}); at the four budgets the closest BELT variant is within 0.45~pp of OSP's accuracy, and OSP saves 1.07--1.46~pp more at the three accuracy tolerances (\cref{tab:code-post-accuracy,tab:code-post,fig:code-post}).
\ecoparagraph{Coverage of the higher-accuracy regime.}
Beyond the seven readings, \cref{fig:code-post} shows a higher-cost regime where BELT-B-R improves accuracy substantially above natural completion. Over the three native-family seed frontiers, its maximum accuracy is $32.68\pm0.15\%$, versus $30.05\pm0.51\%$ for full OSP, $28.81\pm0.24\%$ for OSP-1 and 28.67\% for natural completion. With per-seed test-frontier interpolation, every B-R seed reaches 31\% and 32\% accuracy (mean token savings 25.53\% and 12.75\%); no native full-OSP seed reaches either.

\begin{table}[!htp]
\caption{Token saving given the accuracy tolerance and accuracy given the token budget on LiveCodeBench Post (from August 2024). Reporting convention follows \cref{tab:code-main}. Panel: token saving. OSP-1+T retains the native stochastic policy alongside its existing 13 source-validation thresholds.}\label{tab:code-post}
\begin{center}
\begingroup\setlength{\tabcolsep}{4pt}
\begin{tabular}{@{}lcccc@{}}\toprule
\textbf{Method} & \shortstack{\textbf{Probe}\\\textbf{Params}} & \textbf{0 pp} & \textbf{1 pp} & \textbf{2 pp} \\\midrule
Natural$^{\dagger}$ & -- & \missingcell& \missingcell& \missingcell \\
OSP & 101M $\times 5$ & \cellcolor{Green3!60}\beststatcell{54.58}{2.30}{}& \cellcolor{Green3!60}\beststatcell{62.38}{2.01}{}& \cellcolor{Green3!60}\beststatcell{70.23}{0.20}{} \\
OSP-1 & 101M & \cellcolor{Green3!0}\statcell{14.07}{24.37}{}& \cellcolor{Green3!0}\statcell{53.92}{5.13}{}& \cellcolor{Green3!26}\statcell{66.75}{1.65}{} \\
OSP-1+T &101M & \cellcolor{Green3!0}\statcell{14.07}{24.37}{}& \cellcolor{Green3!0}\statcell{53.92}{5.13}{}& \cellcolor{Green3!27}\statcell{66.87}{1.53}{} \\
\textbf{BELT-B-R} & 16M & \cellcolor{Green3!58}\secondstatcell{53.51}{2.49}{}& \cellcolor{Green3!52}\secondstatcell{61.21}{1.07}{}& \cellcolor{Green3!46}\secondstatcell{68.77}{2.91}{} \\
\textbf{BELT-H-R} & 16M & \cellcolor{Green3!54}\statcell{50.48}{5.43}{}& \cellcolor{Green3!34}\statcell{58.76}{3.94}{}& \cellcolor{Green3!17}\statcell{65.80}{2.33}{} \\
\textbf{BELT-B-T} & 16M & \cellcolor{Green3!57}\statcell{52.77}{0.66}{}& \cellcolor{Green3!37}\statcell{59.08}{1.13}{}& \cellcolor{Green3!0}\statcell{64.04}{1.04}{} \\
\textbf{BELT-H-T} & 16M & \cellcolor{Green3!51}\statcell{48.30}{8.74}{}& \cellcolor{Green3!32}\statcell{58.47}{5.73}{}& \cellcolor{Green3!35}\statcell{67.62}{3.12}{} \\
\bottomrule\end{tabular}\endgroup
\end{center}
\end{table}

\begin{table}[!htp]
\caption{Token saving given the accuracy tolerance and accuracy given the token budget on LiveCodeBench Post (from August 2024). Reporting convention follows \cref{tab:code-main}. Panel: accuracy.}\label{tab:code-post-accuracy}
\begin{center}
\begingroup\setlength{\tabcolsep}{4pt}
\begin{tabular}{@{}lccccc@{}}\toprule
\textbf{Method} & \shortstack{\textbf{Probe}\\\textbf{Params}} & \textbf{70\%} & \textbf{60\%} & \textbf{50\%} & \textbf{40\%} \\\midrule
Natural$^{\dagger}$ & -- & \statcell{28.67}{}{}& \statcell{28.67}{}{}& \statcell{28.67}{}{}& \statcell{28.67}{}{} \\
OSP & 101M $\times 5$ & \cellcolor{Green3!60}\beststatcell{26.71}{0.04}{}& \cellcolor{Green3!60}\beststatcell{27.96}{0.29}{}& \cellcolor{Green3!60}\beststatcell{29.53}{0.54}{}& \cellcolor{Green3!52}\secondstatcell{30.05}{0.51}{} \\
OSP-1 & 101M & \cellcolor{Green3!37}\statcell{26.20}{0.14}{}& \cellcolor{Green3!0}\statcell{27.27}{0.21}{}& \cellcolor{Green3!0}\statcell{27.94}{0.35}{}& \cellcolor{Green3!0}\statcell{28.33}{0.41}{} \\
OSP-1+T &101M & \cellcolor{Green3!40}\statcell{26.27}{0.07}{}& \cellcolor{Green3!0}\statcell{27.27}{0.21}{}& \cellcolor{Green3!0}\statcell{27.94}{0.35}{}& \cellcolor{Green3!0}\statcell{28.33}{0.41}{} \\
\textbf{BELT-B-R} & 16M & \cellcolor{Green3!48}\secondstatcell{26.44}{0.41}{}& \cellcolor{Green3!50}\secondstatcell{27.85}{0.18}{}& \cellcolor{Green3!40}\statcell{28.99}{0.22}{}& \cellcolor{Green3!46}\statcell{29.84}{0.16}{} \\
\textbf{BELT-H-R} & 16M & \cellcolor{Green3!27}\statcell{25.98}{0.42}{}& \cellcolor{Green3!26}\statcell{27.57}{0.46}{}& \cellcolor{Green3!29}\statcell{28.72}{0.55}{}& \cellcolor{Green3!35}\statcell{29.47}{0.34}{} \\
\textbf{BELT-B-T} & 16M & \cellcolor{Green3!0}\statcell{25.39}{0.08}{}& \cellcolor{Green3!20}\statcell{27.50}{0.21}{}& \cellcolor{Green3!43}\secondstatcell{29.08}{0.04}{}& \cellcolor{Green3!60}\beststatcell{30.30}{0.09}{} \\
\textbf{BELT-H-T} & 16M & \cellcolor{Green3!40}\statcell{26.28}{0.51}{}& \cellcolor{Green3!28}\statcell{27.59}{0.53}{}& \cellcolor{Green3!26}\statcell{28.64}{0.76}{}& \cellcolor{Green3!34}\statcell{29.43}{0.76}{} \\
\bottomrule\end{tabular}\endgroup
\end{center}
\end{table}

\FloatBarrier

\subsection{OLMo3 math-to-code transfer}
\label{sec:olmo-code}
The OLMo3 probes and policies of \cref{sec:olmo-replication} transfer to the two LiveCodeBench sets without code-side training (protocol in \cref{sec:code-transfer}). The fixed-direction H configuration saves more tokens than OSP-1 at all three accuracy tolerances on both sets (\cref{tab:olmo-code-pre,tab:olmo-code-post}), and before August 2024 also has the higher mean at every budget. Against full OSP before August 2024, no BELT variant falls behind by more than the seed spread at any of the seven readings, and the refit H configuration exceeds it at five (\cref{tab:reb-single-variant-wtl}). From August 2024, full OSP has the higher mean at the one-point tolerance than every BELT variant; the refit H configuration has the higher mean at the two-point tolerance and at natural accuracy, and is within 0.10~pp of full OSP's accuracy or above it at all four budgets. Dev-selected utility is in \cref{tab:source-transfer}.
\begin{table}[!htp]
\caption{OLMo3 math-to-code transfer on LiveCodeBench problems before August 2024 (549 problems). Conventions follow \cref{tab:code-main,tab:olmo-main}; policies are trained on mathematics only and read on the code test frontier. Natural$^{\dagger}$ is the natural-completion reference.}\label{tab:olmo-code-pre}
\begin{center}
\begingroup\setlength{\tabcolsep}{2pt}
\resizebox{\linewidth}{!}{%
\begin{tabular}{@{}lc|ccc|cccc@{}}\toprule
\multirow{2}{*}{\textbf{Method}} & \multirow{2}{*}{\shortstack{\textbf{Probe}\\\textbf{Params}}} & \multicolumn{3}{c|}{\textbf{Token saving (\%) at tolerance}} & \multicolumn{4}{c}{\textbf{Accuracy (\%) at saving}} \\
 &  & \textbf{0.00} & \textbf{0.01} & \textbf{0.02} & \textbf{70\%} & \textbf{60\%} & \textbf{50\%} & \textbf{40\%} \\
\midrule
Natural$^{\dagger}$ & -- & \missingcell & \missingcell & \missingcell & \statcell{69.35}{}{} & \statcell{69.35}{}{} & \statcell{69.35}{}{} & \statcell{69.35}{}{} \\
\midrule
OSP & 405M $\times 5$ & \cellcolor{Green3!6}\statcell{9.10}{15.8}{} & \cellcolor{Green3!35}\statcell{27.58}{5.6}{} & \cellcolor{Green3!48}\statcell{34.01}{0.5}{} & \cellcolor{Green3!10}\statcell{52.83}{0.6}{} & \cellcolor{Green3!7}\statcell{57.79}{0.8}{} & \cellcolor{Green3!20}\statcell{62.04}{0.8}{} & \cellcolor{Green3!26}\statcell{65.50}{0.4}{} \\
OSP-1 & 405M & \cellcolor{Green3!0}\statcell{6.94}{12.0}{} & \cellcolor{Green3!0}\statcell{21.94}{7.3}{} & \cellcolor{Green3!0}\statcell{27.67}{6.0}{} & \cellcolor{Green3!0}\statcell{52.44}{0.5}{} & \cellcolor{Green3!0}\statcell{57.56}{0.5}{} & \cellcolor{Green3!0}\statcell{61.53}{0.6}{} & \cellcolor{Green3!0}\statcell{64.61}{0.8}{} \\
\midrule
\textbf{BELT-B-T} & 19M & \cellcolor{Green3!44}\statcell{21.84}{4.2}{} & \cellcolor{Green3!34}\statcell{27.33}{4.7}{} & \cellcolor{Green3!38}\statcell{32.69}{3.8}{} & \cellcolor{Green3!14}\statcell{53.01}{0.7}{} & \cellcolor{Green3!23}\statcell{58.32}{1.2}{} & \cellcolor{Green3!42}\statcell{62.62}{1.1}{} & \cellcolor{Green3!32}\statcell{65.69}{1.0}{} \\
\textbf{BELT-H-T} & 19M & \cellcolor{Green3!52}\secondstatcell{24.39}{5.2}{} & \cellcolor{Green3!50}\statcell{29.91}{4.9}{} & \cellcolor{Green3!49}\secondstatcell{34.10}{4.9}{} & \cellcolor{Green3!40}\secondstatcell{54.06}{1.1}{} & \cellcolor{Green3!50}\secondstatcell{59.20}{1.3}{} & \cellcolor{Green3!56}\secondstatcell{62.98}{1.2}{} & \cellcolor{Green3!40}\statcell{65.96}{1.2}{} \\
\textbf{BELT-B-R} & 19M & \cellcolor{Green3!45}\statcell{22.16}{3.2}{} & \cellcolor{Green3!52}\secondstatcell{30.37}{3.2}{} & \cellcolor{Green3!47}\statcell{33.89}{4.4}{} & \cellcolor{Green3!22}\statcell{53.33}{0.9}{} & \cellcolor{Green3!18}\statcell{58.16}{1.4}{} & \cellcolor{Green3!48}\statcell{62.76}{1.3}{} & \cellcolor{Green3!46}\secondstatcell{66.18}{0.9}{} \\
\textbf{BELT-H-R} & 19M & \cellcolor{Green3!60}\beststatcell{27.07}{2.1}{} & \cellcolor{Green3!60}\beststatcell{31.59}{3.2}{} & \cellcolor{Green3!60}\beststatcell{35.54}{2.6}{} & \cellcolor{Green3!60}\beststatcell{54.85}{1.4}{} & \cellcolor{Green3!60}\beststatcell{59.53}{1.3}{} & \cellcolor{Green3!60}\beststatcell{63.08}{0.9}{} & \cellcolor{Green3!60}\beststatcell{66.64}{0.5}{} \\
\bottomrule\end{tabular}}
\endgroup
\end{center}
\end{table}
\begin{table}[!htp]
\caption{OLMo3 math-to-code transfer on LiveCodeBench problems from August 2024 (416 problems). Conventions follow \cref{tab:code-main,tab:olmo-main}; policies are trained on mathematics only and read on the code test frontier. Natural$^{\dagger}$ is the natural-completion reference.}\label{tab:olmo-code-post}
\begin{center}
\begingroup\setlength{\tabcolsep}{2pt}
\resizebox{\linewidth}{!}{%
\begin{tabular}{@{}lc|ccc|cccc@{}}\toprule
\multirow{2}{*}{\textbf{Method}} & \multirow{2}{*}{\shortstack{\textbf{Probe}\\\textbf{Params}}} & \multicolumn{3}{c|}{\textbf{Token saving (\%) at tolerance}} & \multicolumn{4}{c}{\textbf{Accuracy (\%) at saving}} \\
 &  & \textbf{0.00} & \textbf{0.01} & \textbf{0.02} & \textbf{70\%} & \textbf{60\%} & \textbf{50\%} & \textbf{40\%} \\
\midrule
Natural$^{\dagger}$ & -- & \missingcell & \missingcell & \missingcell & \statcell{45.25}{}{} & \statcell{45.25}{}{} & \statcell{45.25}{}{} & \statcell{45.25}{}{} \\
\midrule
OSP & 405M $\times 5$ & \cellcolor{Green3!38}\statcell{29.59}{25.6}{} & \cellcolor{Green3!60}\beststatcell{48.17}{1.7}{} & \cellcolor{Green3!56}\secondstatcell{50.95}{1.5}{} & \cellcolor{Green3!60}\beststatcell{36.32}{0.4}{} & \cellcolor{Green3!59}\secondstatcell{40.55}{0.2}{} & \cellcolor{Green3!60}\beststatcell{43.63}{0.6}{} & \cellcolor{Green3!51}\secondstatcell{45.45}{0.9}{} \\
OSP-1 & 405M & \cellcolor{Green3!0}\statcell{9.11}{15.8}{} & \cellcolor{Green3!0}\statcell{35.35}{6.8}{} & \cellcolor{Green3!0}\statcell{45.06}{5.8}{} & \cellcolor{Green3!58}\secondstatcell{36.29}{0.2}{} & \cellcolor{Green3!26}\statcell{40.09}{0.4}{} & \cellcolor{Green3!0}\statcell{42.40}{1.0}{} & \cellcolor{Green3!0}\statcell{43.76}{0.7}{} \\
\midrule
\textbf{BELT-B-T} & 19M & \cellcolor{Green3!52}\statcell{37.67}{0.8}{} & \cellcolor{Green3!42}\statcell{44.22}{2.2}{} & \cellcolor{Green3!30}\statcell{48.23}{1.9}{} & \cellcolor{Green3!0}\statcell{35.47}{1.0}{} & \cellcolor{Green3!0}\statcell{39.73}{0.8}{} & \cellcolor{Green3!19}\statcell{42.78}{0.5}{} & \cellcolor{Green3!33}\statcell{44.87}{0.2}{} \\
\textbf{BELT-H-T} & 19M & \cellcolor{Green3!49}\statcell{35.95}{6.5}{} & \cellcolor{Green3!39}\statcell{43.78}{2.3}{} & \cellcolor{Green3!34}\statcell{48.61}{1.7}{} & \cellcolor{Green3!26}\statcell{35.84}{0.5}{} & \cellcolor{Green3!20}\statcell{40.01}{0.6}{} & \cellcolor{Green3!26}\statcell{42.93}{0.3}{} & \cellcolor{Green3!37}\statcell{44.98}{0.6}{} \\
\textbf{BELT-B-R} & 19M & \cellcolor{Green3!58}\secondstatcell{40.63}{1.1}{} & \cellcolor{Green3!41}\statcell{44.03}{0.9}{} & \cellcolor{Green3!49}\statcell{50.24}{1.3}{} & \cellcolor{Green3!44}\statcell{36.10}{0.4}{} & \cellcolor{Green3!46}\statcell{40.37}{0.9}{} & \cellcolor{Green3!44}\statcell{43.31}{0.3}{} & \cellcolor{Green3!49}\statcell{45.41}{0.3}{} \\
\textbf{BELT-H-R} & 19M & \cellcolor{Green3!60}\beststatcell{41.79}{2.4}{} & \cellcolor{Green3!49}\secondstatcell{45.77}{1.6}{} & \cellcolor{Green3!60}\beststatcell{51.41}{0.6}{} & \cellcolor{Green3!53}\statcell{36.22}{0.7}{} & \cellcolor{Green3!60}\beststatcell{40.57}{0.5}{} & \cellcolor{Green3!60}\secondstatcell{43.62}{0.1}{} & \cellcolor{Green3!60}\beststatcell{45.76}{0.5}{} \\
\bottomrule\end{tabular}}
\endgroup
\end{center}
\end{table}

\FloatBarrier

\subsection{ALFWorld results}
\label{sec:alfworld-results}
\ecoparagraph{Main readings.} ALFWorld measures task completion and ScienceWorld the simulator's progress score; neither provides a per-step correctness signal, and an agent episode generates 21--33 times as many tokens as an OLMo3 math trajectory. In ALFWorld, natural completion succeeds on 33.6\% of games at 227k tokens per episode, versus 4.5\% without thinking. At 50\% saving every controller family falls 6--10~pp below natural success, so budgets are read at 20--50\%. Under the same calibration, BELT-B-T saves $12.4\pm13.9$\% of tokens at natural success and $25.4\pm13.0$\% within two points, against 0.0\% and 11.5\% for OSP-1, and retains more success than OSP-1 at a 20\% saving budget. Across 20--50\% saving, BELT-H-T's mean success is within 0.9~pp of OSP-1's (\cref{tab:agentic}; uncertainty values are game-bootstrap SDs).

\begin{figure}[t]\centering
\includegraphics[width=.55\textwidth]{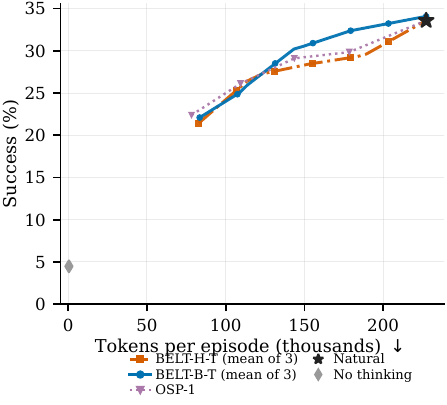}
\caption{ALFWorld test frontiers on the unseen games: success versus generated tokens per episode. Multi-run methods show the mean of the runs' frontiers over their common cost range; single-run methods show the frontier through their members. Star: natural completion; diamond: no thinking.}
\label{fig:alfworld}
\end{figure}

\ecoparagraph{Full grid.} \Cref{tab:alfworld,tab:alfworld-members,tab:alfworld-bootstrap,fig:alfworld} give the full grid, every member on the unseen games, the paired game-bootstrap differences against OSP-1 and the test frontiers. Three members match or exceed natural success with fewer tokens: BELT-B-T s0 at the 20\% and 35\% targets and BELT-H-R s0 at the 20\% target.

\begin{figure}[t]\centering
\includegraphics[width=.55\textwidth]{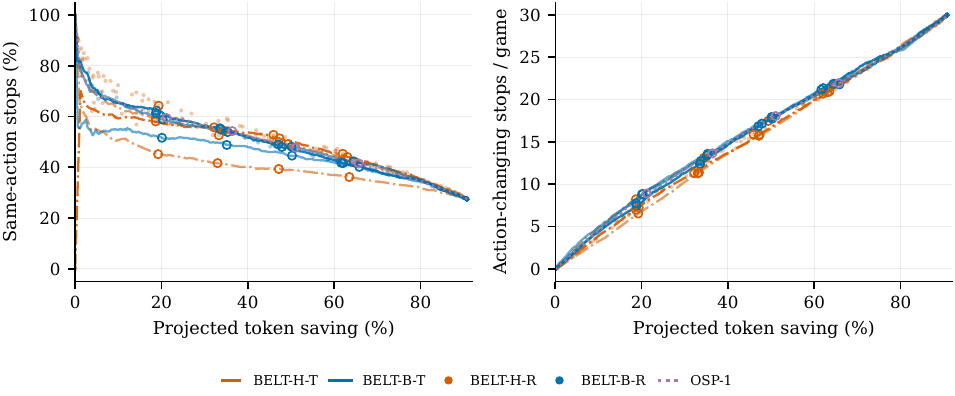}
\caption{Label-free stop diagnostic on the unseen natural trajectories. Left: share of a candidate's projected stops whose forced action equals the natural action; right: projected stops that change the action, per game; both against projected token saving. Lines: threshold grids (one per probe run); points: refit policies; open circles: the calibrated members.}
\label{fig:alfworld-same-stop}
\end{figure}

\ecoparagraph{Where members leave the natural trajectory.} An episode departs from natural completion only at a stop whose forced action differs from the natural action. On the unseen natural trajectories, without outcome labels, every BELT variant and OSP-1 keeps the natural action at 40--42\% of its stops at 65\% projected saving and changes it at 21.2--22.2 stops per game (\cref{fig:alfworld-same-stop}; means over probe runs where there are several).

\begin{table}[!ht]
\caption{ALFWorld per-turn thinking control with OLMo3 (134 unseen games, at most 50 steps, one generation seed). Token saving (\%) relative to natural completion at a success tolerance of 0/2/5~pp, and success (\%) at token-saving budgets. Probes, fitted directions and the OSP-1 checkpoint are transferred from mathematics (\cref{sec:alfworld-protocol}). BELT-T and OSP-1 thresholds are calibrated on the seen split without outcome labels; BELT-R uses the refit policies selected by the same rule. Subscripts: SD of the reading over 5,000 game resamples (methods with several probe training runs are averaged over runs inside each resample); the probe-run spread is in \cref{sec:checks-alfworld}.}\label{tab:alfworld}
\begin{center}
\begingroup\setlength{\tabcolsep}{2.5pt}
\resizebox{\linewidth}{!}{%
\begin{tabular}{@{}l|c|ccc|cccc@{}}\toprule
\multirow{2}{*}{\textbf{Method}} & \multirow{2}{*}{\shortstack{\textbf{Probe}\\\textbf{Params}}} & \multicolumn{3}{c|}{\textbf{Token saving (\%) at tolerance}} & \multicolumn{4}{c}{\textbf{Success (\%) at saving}} \\
 &  & \textbf{0 pp} & \textbf{2 pp} & \textbf{5 pp} & \textbf{50\%} & \textbf{40\%} & \textbf{30\%} & \textbf{20\%} \\
\midrule
Natural & -- & \missingcell& \missingcell& \missingcell& \statcell{33.6}{4.2}{}& \statcell{33.6}{4.2}{}& \statcell{33.6}{4.2}{}& \statcell{33.6}{4.2}{} \\
OSP-1 & 405M & \statcell{0.0}{18.4}{}& \statcell{11.5}{17.3}{}& \secondstatcell{39.5}{15.8}{}& \secondstatcell{26.5}{3.5}{}& \statcell{28.5}{3.6}{}& \statcell{29.4}{3.6}{}& \statcell{30.1}{3.6}{} \\
\midrule
\textbf{BELT-B-T} & 19M & \secondstatcell{12.4}{13.9}{}& \secondstatcell{25.4}{13.0}{}& \statcell{38.0}{11.8}{}& \statcell{25.9}{2.9}{}& \secondstatcell{29.2}{3.0}{}& \secondstatcell{31.1}{3.0}{}& \secondstatcell{32.5}{3.1}{} \\
\textbf{BELT-H-T} & 19M & \statcell{0.0}{12.1}{}& \statcell{13.2}{13.3}{}& \statcell{32.5}{13.8}{}& \statcell{26.4}{2.9}{}& \statcell{27.8}{3.0}{}& \statcell{28.6}{3.0}{}& \statcell{29.2}{3.1}{} \\
\textbf{BELT-B-R} & 19M & \statcell{0.0}{17.5}{}& \statcell{13.4}{16.3}{}& \statcell{34.2}{15.5}{}& \statcell{26.4}{3.7}{}& \statcell{27.8}{3.6}{}& \statcell{29.2}{3.5}{}& \statcell{30.6}{3.6}{} \\
\textbf{BELT-H-R} & 19M & \beststatcell{20.0}{19.4}{}& \beststatcell{30.2}{16.5}{}& \beststatcell{45.4}{14.7}{}& \beststatcell{27.7}{3.8}{}& \beststatcell{29.6}{3.6}{}& \beststatcell{31.6}{3.7}{}& \beststatcell{33.6}{3.8}{} \\
\bottomrule\end{tabular}}
\endgroup
\end{center}
\end{table}

\begin{table}[!ht]
\centering\small
\caption{ALFWorld members on the 134 unseen games: success (\%) / realised token saving (\%), with the projected saving on the seen calibration split in brackets. Thresholded and refit rows: the four members calibrated for 20/35/50/65\% projected saving. Natural completion: 33.6\% success at 227k tokens per episode.}\label{tab:alfworld-members}
\begingroup\setlength{\tabcolsep}{3pt}
\resizebox{\linewidth}{!}{%
\begin{tabular}{@{}llccccc@{}}\toprule
Method & Run & \multicolumn{5}{c}{Success / realised saving [projected saving] (\%)} \\
\midrule
OSP-1 & s0 & 29.9 / 21.5 [20.0] & 29.1 / 36.8 [35.0] & 26.1 / 51.9 [50.0] & 22.4 / 65.6 [65.0] &  \\
BELT-B-T & s0 & 35.1 / 21.5 [20.0] & 33.6 / 37.1 [35.0] & 23.1 / 52.3 [50.0] & 20.9 / 68.3 [65.0] &  \\
BELT-B-T & s1 & 32.1 / 21.3 [20.0] & 29.9 / 35.7 [35.0] & 26.1 / 50.1 [50.0] & 19.4 / 63.3 [65.0] &  \\
BELT-B-T & s2 & 25.4 / 18.4 [20.0] & 27.6 / 34.7 [35.0] & 26.9 / 51.8 [50.0] & 24.6 / 68.2 [65.0] &  \\
BELT-H-T & s0 & 24.6 / 13.9 [20.0] & 24.6 / 29.4 [35.0] & 29.9 / 45.8 [50.0] & 22.4 / 63.5 [65.0] &  \\
BELT-H-T & s1 & 29.9 / 19.7 [20.0] & 29.1 / 33.5 [35.0] & 26.9 / 50.2 [50.0] & 17.9 / 64.3 [65.0] &  \\
BELT-H-T & s2 & 26.1 / 16.9 [20.0] & 18.7 / 34.8 [35.0] & 23.1 / 51.4 [50.0] & 23.1 / 70.5 [65.0] &  \\
BELT-B-R & s0 & 30.6 / 19.9 [19.4] & 28.4 / 35.7 [35.2] & 26.1 / 51.8 [49.1] & 19.4 / 68.1 [64.6] &  \\
BELT-H-R & s0 & 33.6 / 20.0 [20.4] & 26.9 / 31.3 [35.4] & 27.6 / 50.3 [49.2] & 21.6 / 61.8 [64.7] &  \\
\bottomrule\end{tabular}}\endgroup
\end{table}

\begin{table}[!ht]
\centering\small
\caption{ALFWorld paired bootstrap over the 134 unseen games (5,000 resamples; both methods re-read on every resample, multi-run methods averaged inside it): difference in token saving (pp) at each tolerance and in success (pp) at each budget, with the 95\% percentile interval. The resampling covers games, not probe or training seeds.}\label{tab:alfworld-bootstrap}
\begingroup\setlength{\tabcolsep}{2pt}
\resizebox{\linewidth}{!}{%
\begin{tabular}{@{}l|ccc|cccc@{}}\toprule
\multirow{2}{*}{\textbf{Comparison}} & \multicolumn{3}{c|}{\textbf{Saving difference at tolerance}} & \multicolumn{4}{c}{\textbf{Success difference at saving}} \\
 & \textbf{0 pp} & \textbf{2 pp} & \textbf{5 pp} & \textbf{50\%} & \textbf{40\%} & \textbf{30\%} & \textbf{20\%} \\
\midrule
BELT-B-T vs.\ OSP-1 & $+12.4$\,{\scriptsize[$-28.5$, $+37.5$]} & $+13.9$\,{\scriptsize[$-21.7$, $+28.6$]} & $-1.5$\,{\scriptsize[$-19.0$, $+25.5$]} & $-0.6$\,{\scriptsize[$-5.7$, $+5.2$]} & $+0.8$\,{\scriptsize[$-4.5$, $+6.2$]} & $+1.7$\,{\scriptsize[$-3.5$, $+6.3$]} & $+2.4$\,{\scriptsize[$-3.1$, $+6.4$]} \\
BELT-H-T vs.\ OSP-1 & $+0.0$\,{\scriptsize[$-40.8$, $+27.2$]} & $+1.7$\,{\scriptsize[$-31.1$, $+20.1$]} & $-7.0$\,{\scriptsize[$-27.6$, $+20.0$]} & $-0.1$\,{\scriptsize[$-6.1$, $+5.1$]} & $-0.7$\,{\scriptsize[$-5.8$, $+4.3$]} & $-0.8$\,{\scriptsize[$-5.7$, $+3.6$]} & $-0.9$\,{\scriptsize[$-5.7$, $+3.5$]} \\
BELT-H-R vs.\ OSP-1 & $+20.0$\,{\scriptsize[$-39.8$, $+52.7$]} & $+18.7$\,{\scriptsize[$-31.4$, $+40.0$]} & $+5.9$\,{\scriptsize[$-28.6$, $+32.6$]} & $+1.2$\,{\scriptsize[$-7.5$, $+8.5$]} & $+1.2$\,{\scriptsize[$-7.4$, $+7.8$]} & $+2.2$\,{\scriptsize[$-7.2$, $+8.5$]} & $+3.5$\,{\scriptsize[$-5.4$, $+9.8$]} \\
\bottomrule\end{tabular}}\endgroup
\end{table}

\FloatBarrier

\subsection{ScienceWorld results}
\label{sec:sciworld-results}
\Cref{tab:sciworld-full,tab:sciworld-members,tab:sciworld-bootstrap} give the full grid, every member and the paired game-bootstrap differences. Of 63 controller members, 60 score at or above natural completion. Every calibrated family reaches its 95\% member, the first-probe stop, at or above the natural score, so the three tolerance readings coincide at 85\% saving for BELT and OSP-1 alike and only the budgets at 50--80\% saving distinguish them. There BELT-H-R has the highest score at 50, 60 and 70\% and BELT-H-T at 80\%; every BELT family is within 1.2 points of OSP-1 at every budget, and every paired interval against OSP-1 includes zero. Reading the operating points on one task-stratified half of the games and reporting on the other (200 splits) moves the BELT tolerance readings by at most 2.2~pp and the budget readings by at most 2.6 points, with OSP-1 moving by up to 2.7.
\begin{table}[!ht]
\caption{ScienceWorld per-turn thinking control with OLMo3 (90 test games, three per task type, at most 50 steps, one generation seed). Token saving (\%) relative to natural completion at a score tolerance of 0/2/5 points, and mean score (0--100) at token-saving budgets. Conventions follow \cref{tab:alfworld}: probes, fitted directions and the OSP-1 checkpoint are transferred from mathematics; BELT-T and OSP-1 thresholds and the BELT-R refit policies are calibrated on the disjoint development variations without outcome labels, at projected savings of 20--95\% (\cref{sec:sciworld-protocol}). Subscripts: SD of the reading over 5,000 game resamples (methods with several probe training runs are averaged over runs inside each resample); the probe-run spread is in \cref{sec:checks-alfworld}.}\label{tab:sciworld-full}
\begin{center}
\begingroup\setlength{\tabcolsep}{2.5pt}
\resizebox{\linewidth}{!}{%
\begin{tabular}{@{}l|c|ccc|cccc@{}}\toprule
\multirow{2}{*}{\textbf{Method}} & \multirow{2}{*}{\shortstack{\textbf{Probe}\\\textbf{Params}}} & \multicolumn{3}{c|}{\textbf{Token saving (\%) at tolerance}} & \multicolumn{4}{c}{\textbf{Score at saving}} \\
 &  & \textbf{0 pt} & \textbf{2 pt} & \textbf{5 pt} & \textbf{80\%} & \textbf{70\%} & \textbf{60\%} & \textbf{50\%} \\
\midrule
Natural & -- & \missingcell& \missingcell& \missingcell& \statcell{21.4}{3.7}{}& \statcell{21.4}{3.7}{}& \statcell{21.4}{3.7}{}& \statcell{21.4}{3.7}{} \\
OSP-1 & 405M & \secondstatcell{84.9}{5.8}{}& \secondstatcell{84.9}{2.6}{}& \secondstatcell{84.9}{1.6}{}& \statcell{25.2}{3.2}{}& \statcell{25.7}{3.1}{}& \secondstatcell{26.9}{3.2}{}& \statcell{28.0}{3.4}{} \\
\midrule
\textbf{BELT-B-T} & 19M & \beststatcell{85.0}{8.1}{}& \beststatcell{85.0}{3.5}{}& \beststatcell{85.0}{1.7}{}& \secondstatcell{25.4}{3.1}{}& \statcell{25.9}{3.0}{}& \statcell{26.5}{2.9}{}& \statcell{27.0}{2.9}{} \\
\textbf{BELT-H-T} & 19M & \beststatcell{85.0}{7.4}{}& \beststatcell{85.0}{2.9}{}& \beststatcell{85.0}{1.7}{}& \beststatcell{25.6}{3.1}{}& \secondstatcell{26.3}{3.0}{}& \statcell{26.6}{3.0}{}& \statcell{26.8}{3.0}{} \\
\textbf{BELT-B-R} & 19M & \secondstatcell{84.9}{6.6}{}& \secondstatcell{84.9}{3.0}{}& \secondstatcell{84.9}{1.6}{}& \statcell{25.0}{3.1}{}& \statcell{25.4}{3.0}{}& \statcell{26.7}{3.0}{}& \secondstatcell{28.2}{3.1}{} \\
\textbf{BELT-H-R} & 19M & \secondstatcell{84.9}{6.6}{}& \secondstatcell{84.9}{3.1}{}& \secondstatcell{84.9}{1.7}{}& \secondstatcell{25.4}{3.2}{}& \beststatcell{26.5}{3.2}{}& \beststatcell{27.6}{3.4}{}& \beststatcell{28.5}{3.5}{} \\
\bottomrule\end{tabular}}
\endgroup
\end{center}
\end{table}

\begin{table}[!ht]
\centering\small
\caption{ScienceWorld members on the 90 test games: score / realised token saving (\%), with the projected saving on the development calibration split in brackets. Each row lists the seven members calibrated for 20/35/50/65/80/90/95\% projected saving; the 95\% target lies above the 91.5\% ceiling of the label-free rule on the calibration split, so every family's 95\% member is the first-probe stop (the same actions in 87 of 90 games across families). Natural completion: score 21.4 at 144k tokens per episode; no thinking: 8.9.}\label{tab:sciworld-members}
\begingroup\setlength{\tabcolsep}{3pt}
\resizebox{\linewidth}{!}{%
\begin{tabular}{@{}llccccccc@{}}\toprule
Method & Run & \multicolumn{7}{c}{Score / realised saving [projected saving] (\%)} \\
\midrule
OSP-1 & s0 & 20.7 / 10.9 [20.0] & 23.3 / 26.3 [35.0] & 27.3 / 34.9 [50.0] & 28.0 / 50.2 [65.0] & 25.7 / 69.9 [80.0] & 25.0 / 83.0 [90.0] & 24.8 / 84.9 [91.5] \\
BELT-B-T & s0 & 26.5 / 13.9 [20.0] & 28.0 / 28.9 [35.0] & 26.2 / 38.1 [50.0] & 26.0 / 49.1 [65.0] & 26.5 / 65.4 [80.0] & 23.9 / 83.6 [90.0] & 24.8 / 84.9 [91.5] \\
BELT-B-T & s1 & 23.8 / 12.8 [20.0] & 22.1 / 21.4 [35.0] & 18.5 / 34.9 [50.0] & 23.4 / 54.3 [65.0] & 24.4 / 67.6 [80.0] & 24.4 / 81.8 [90.0] & 24.8 / 84.9 [91.5] \\
BELT-B-T & s2 & 24.6 / 10.3 [20.0] & 25.2 / 23.9 [35.0] & 28.3 / 38.2 [50.0] & 29.2 / 48.1 [65.0] & 26.9 / 69.6 [80.0] & 24.8 / 83.0 [90.0] & 25.7 / 85.0 [91.5] \\
BELT-H-T & s0 & 18.1 / 15.6 [20.0] & 27.4 / 27.4 [35.0] & 26.5 / 40.6 [50.0] & 28.2 / 55.5 [65.0] & 22.6 / 74.5 [80.0] & 21.4 / 83.2 [90.0] & 24.8 / 84.9 [91.5] \\
BELT-H-T & s1 & 22.5 / 12.9 [20.0] & 25.6 / 23.8 [35.0] & 24.2 / 39.9 [50.0] & 26.9 / 53.8 [65.0] & 27.4 / 70.0 [80.0] & 26.5 / 83.1 [90.0] & 24.8 / 85.0 [91.5] \\
BELT-H-T & s2 & 23.4 / 16.0 [19.9] & 22.1 / 28.7 [35.0] & 24.5 / 46.8 [50.0] & 24.5 / 54.6 [65.0] & 22.3 / 69.7 [80.0] & 22.4 / 83.5 [90.0] & 24.8 / 84.9 [91.5] \\
BELT-B-R & s0 & 26.8 / 8.9 [20.7] & 28.1 / 28.4 [35.0] & 25.4 / 40.8 [50.4] & 28.2 / 50.2 [65.6] & 25.5 / 67.7 [80.0] & 22.0 / 83.4 [89.9] & 24.8 / 84.9 [91.5] \\
BELT-H-R & s0 & 26.1 / 9.8 [20.8] & 25.3 / 29.1 [35.1] & 28.9 / 31.1 [51.0] & 28.4 / 52.5 [65.3] & 22.8 / 70.3 [80.1] & 21.9 / 83.7 [90.0] & 24.8 / 84.9 [91.5] \\
\bottomrule\end{tabular}}\endgroup
\end{table}

\begin{table}[!ht]
\centering\small
\caption{ScienceWorld paired bootstrap over the 90 test games (5,000 resamples shared across pairs; both methods re-read on every resample, multi-run methods averaged inside it): difference in token saving (pp) at each tolerance and in score (points) at each budget, with the 95\% percentile interval. The resampling covers games, not probe or training runs.}\label{tab:sciworld-bootstrap}
\begingroup\setlength{\tabcolsep}{2pt}
\resizebox{\linewidth}{!}{%
\begin{tabular}{@{}l|ccc|cccc@{}}\toprule
\multirow{2}{*}{\textbf{Comparison}} & \multicolumn{3}{c|}{\textbf{Saving difference at tolerance}} & \multicolumn{4}{c}{\textbf{Score difference at saving}} \\
 & \textbf{0 pt} & \textbf{2 pt} & \textbf{5 pt} & \textbf{80\%} & \textbf{70\%} & \textbf{60\%} & \textbf{50\%} \\
\midrule
BELT-H-T vs.\ OSP-1 & $+0.0$\,{\scriptsize[$-13.0$, $+0.4$]} & $+0.0$\,{\scriptsize[$-0.6$, $+0.2$]} & $+0.0$\,{\scriptsize[$-0.0$, $+0.1$]} & $+0.5$\,{\scriptsize[$-2.0$, $+1.4$]} & $+0.5$\,{\scriptsize[$-3.9$, $+2.2$]} & $-0.2$\,{\scriptsize[$-4.5$, $+2.7$]} & $-1.2$\,{\scriptsize[$-5.4$, $+2.9$]} \\
BELT-B-T vs.\ OSP-1 & $+0.0$\,{\scriptsize[$-20.0$, $+0.5$]} & $+0.0$\,{\scriptsize[$-0.3$, $+0.0$]} & $+0.0$\,{\scriptsize[$-0.0$, $+0.0$]} & $+0.2$\,{\scriptsize[$-2.0$, $+1.4$]} & $+0.2$\,{\scriptsize[$-3.8$, $+2.1$]} & $-0.4$\,{\scriptsize[$-4.9$, $+2.6$]} & $-1.0$\,{\scriptsize[$-6.4$, $+3.2$]} \\
BELT-H-R vs.\ OSP-1 & $+0.0$\,{\scriptsize[$-11.0$, $+2.3$]} & $+0.0$\,{\scriptsize[$-0.3$, $+0.1$]} & $+0.0$\,{\scriptsize[$+0.0$, $+0.0$]} & $+0.2$\,{\scriptsize[$-2.8$, $+0.9$]} & $+0.8$\,{\scriptsize[$-4.8$, $+2.4$]} & $+0.7$\,{\scriptsize[$-5.2$, $+3.9$]} & $+0.5$\,{\scriptsize[$-6.1$, $+5.2$]} \\
BELT-B-R vs.\ OSP-1 & $+0.0$\,{\scriptsize[$-10.3$, $+1.1$]} & $+0.0$\,{\scriptsize[$-0.5$, $+0.0$]} & $+0.0$\,{\scriptsize[$+0.0$, $+0.0$]} & $-0.2$\,{\scriptsize[$-2.5$, $+0.8$]} & $-0.4$\,{\scriptsize[$-4.5$, $+2.4$]} & $-0.2$\,{\scriptsize[$-4.9$, $+3.9$]} & $+0.2$\,{\scriptsize[$-5.8$, $+5.3$]} \\
\bottomrule\end{tabular}}\endgroup
\end{table}

\FloatBarrier

\subsection{Utility of dev-selected policies}
\label{sec:source-transfer}
This protocol fixes the price and selects policies on mathematics development utility before target-population evaluation (\cref{tab:source-transfer}). BELT-B-R has higher mean dev-selected utility than OSP at the three lowest math OOD prices and the two lowest ID prices, and lower utility at the other prices. On OLMo3, it exceeds full OSP at all five prices. Both BELT interfaces exceed OSP-1 at every reported Qwen3 OOD and OLMo3 mathematics price. On the OLMo3 code sets, BELT-H-R exceeds OSP-1 at all five prices on both sets; other code rankings depend on split and price.
\begin{table}[t]
\caption{Utility of dev-selected policies. Reporting convention follows \cref{tab:allbaselines}; all values are micro averages.}\label{tab:source-transfer}
\begin{center}
\begingroup\setlength{\tabcolsep}{4pt}
\begin{tabular}{@{}lccccc@{}}\toprule
\textbf{Method} & \multicolumn{5}{c}{\textbf{Utility by $\lambda$ ($\times10^{-6}$)}} \\
 & \textbf{2} & \textbf{5} & \textbf{10} & \textbf{20} & \textbf{50} \\\midrule
\multicolumn{6}{c}{\textbf{Math ID}} \\
\midrule
OSP & \cellcolor{Green3!24}\statcell{0.5464}{0.0030}{}& \cellcolor{Green3!32}\statcell{0.5340}{0.0026}{}& \cellcolor{Green3!60}\beststatcell{0.5133}{0.0019}{}& \cellcolor{Green3!60}\beststatcell{0.4719}{0.0008}{}& \cellcolor{Green3!60}\beststatcell{0.3698}{0.0010}{} \\
OSP-1 & \cellcolor{Green3!0}\statcell{0.5363}{0.0089}{}& \cellcolor{Green3!0}\statcell{0.5238}{0.0046}{}& \cellcolor{Green3!0}\statcell{0.4989}{0.0045}{}& \cellcolor{Green3!0}\statcell{0.4552}{0.0089}{}& \cellcolor{Green3!0}\statcell{0.3584}{0.0031}{} \\
BELT-B-R & \cellcolor{Green3!60}\beststatcell{0.5619}{0.0011}{}& \cellcolor{Green3!60}\beststatcell{0.5430}{0.0009}{}& \cellcolor{Green3!53}\secondstatcell{0.5116}{0.0007}{}& \cellcolor{Green3!37}\statcell{0.4654}{0.0049}{}& \cellcolor{Green3!16}\secondstatcell{0.3614}{0.0052}{} \\
BELT-H-R & \cellcolor{Green3!35}\secondstatcell{0.5513}{0.0017}{}& \cellcolor{Green3!37}\secondstatcell{0.5356}{0.0014}{}& \cellcolor{Green3!50}\statcell{0.5109}{0.0011}{}& \cellcolor{Green3!42}\secondstatcell{0.4670}{0.0015}{}& \cellcolor{Green3!14}\statcell{0.3610}{0.0021}{} \\
\midrule
\multicolumn{6}{c}{\textbf{Math OOD}} \\
\midrule
OSP & \cellcolor{Green3!16}\statcell{0.7270}{0.0057}{}& \cellcolor{Green3!22}\statcell{0.7177}{0.0049}{}& \cellcolor{Green3!41}\statcell{0.7022}{0.0037}{}& \cellcolor{Green3!60}\beststatcell{0.6713}{0.0013}{}& \cellcolor{Green3!60}\beststatcell{0.5914}{0.0038}{} \\
OSP-1 & \cellcolor{Green3!0}\statcell{0.7205}{0.0037}{}& \cellcolor{Green3!0}\statcell{0.7098}{0.0027}{}& \cellcolor{Green3!0}\statcell{0.6915}{0.0019}{}& \cellcolor{Green3!0}\statcell{0.6563}{0.0015}{}& \cellcolor{Green3!0}\statcell{0.5802}{0.0047}{} \\
BELT-B-R & \cellcolor{Green3!60}\beststatcell{0.7454}{0.0006}{}& \cellcolor{Green3!60}\beststatcell{0.7310}{0.0005}{}& \cellcolor{Green3!60}\beststatcell{0.7072}{0.0006}{}& \cellcolor{Green3!39}\statcell{0.6660}{0.0024}{}& \cellcolor{Green3!26}\statcell{0.5851}{0.0042}{} \\
BELT-H-R & \cellcolor{Green3!35}\secondstatcell{0.7352}{0.0015}{}& \cellcolor{Green3!37}\secondstatcell{0.7227}{0.0012}{}& \cellcolor{Green3!42}\secondstatcell{0.7024}{0.0015}{}& \cellcolor{Green3!50}\secondstatcell{0.6689}{0.0011}{}& \cellcolor{Green3!32}\secondstatcell{0.5861}{0.0020}{} \\
\midrule
\multicolumn{6}{c}{\textbf{LiveCodeBench Pre}} \\
\midrule
OSP & \cellcolor{Green3!19}\statcell{0.6287}{0.0083}{}& \cellcolor{Green3!26}\statcell{0.6146}{0.0082}{}& \cellcolor{Green3!34}\secondstatcell{0.5911}{0.0080}{}& \cellcolor{Green3!60}\beststatcell{0.5440}{0.0084}{}& \cellcolor{Green3!46}\secondstatcell{0.4152}{0.0116}{} \\
OSP-1 & \cellcolor{Green3!0}\statcell{0.6099}{0.0031}{}& \cellcolor{Green3!0}\statcell{0.5878}{0.0042}{}& \cellcolor{Green3!0}\statcell{0.5569}{0.0027}{}& \cellcolor{Green3!0}\statcell{0.4972}{0.0019}{}& \cellcolor{Green3!0}\statcell{0.3970}{0.0138}{} \\
BELT-B-R & \cellcolor{Green3!60}\beststatcell{0.6699}{0.0014}{}& \cellcolor{Green3!60}\beststatcell{0.6502}{0.0015}{}& \cellcolor{Green3!60}\beststatcell{0.6179}{0.0010}{}& \cellcolor{Green3!45}\statcell{0.5325}{0.0134}{}& \cellcolor{Green3!24}\statcell{0.4064}{0.0192}{} \\
BELT-H-R & \cellcolor{Green3!21}\secondstatcell{0.6308}{0.0066}{}& \cellcolor{Green3!26}\secondstatcell{0.6148}{0.0051}{}& \cellcolor{Green3!22}\statcell{0.5795}{0.0188}{}& \cellcolor{Green3!46}\secondstatcell{0.5332}{0.0138}{}& \cellcolor{Green3!60}\beststatcell{0.4206}{0.0125}{} \\
\midrule
\multicolumn{6}{c}{\textbf{LiveCodeBench Post}} \\
\midrule
OSP & \cellcolor{Green3!34}\secondstatcell{0.2901}{0.0050}{}& \cellcolor{Green3!53}\secondstatcell{0.2745}{0.0051}{}& \cellcolor{Green3!60}\beststatcell{0.2485}{0.0060}{}& \cellcolor{Green3!60}\beststatcell{0.1966}{0.0091}{}& \cellcolor{Green3!51}\secondstatcell{0.0970}{0.0280}{} \\
OSP-1 & \cellcolor{Green3!0}\statcell{0.2672}{0.0005}{}& \cellcolor{Green3!0}\statcell{0.2416}{0.0010}{}& \cellcolor{Green3!0}\statcell{0.2002}{0.0038}{}& \cellcolor{Green3!0}\statcell{0.1331}{0.0162}{}& \cellcolor{Green3!24}\statcell{0.0766}{0.0385}{} \\
BELT-B-R & \cellcolor{Green3!60}\beststatcell{0.3073}{0.0007}{}& \cellcolor{Green3!60}\beststatcell{0.2791}{0.0009}{}& \cellcolor{Green3!39}\statcell{0.2316}{0.0030}{}& \cellcolor{Green3!17}\statcell{0.1506}{0.0384}{}& \cellcolor{Green3!0}\statcell{0.0583}{0.0794}{} \\
BELT-H-R & \cellcolor{Green3!28}\statcell{0.2857}{0.0046}{}& \cellcolor{Green3!38}\statcell{0.2656}{0.0019}{}& \cellcolor{Green3!41}\secondstatcell{0.2335}{0.0044}{}& \cellcolor{Green3!54}\secondstatcell{0.1899}{0.0092}{}& \cellcolor{Green3!60}\beststatcell{0.1036}{0.0386}{} \\
\midrule
\multicolumn{6}{c}{\textbf{OLMo3 OOD}} \\
\midrule
OSP & \cellcolor{Green3!53}\statcell{0.8380}{0.0069}{}& \cellcolor{Green3!53}\secondstatcell{0.8251}{0.0059}{}& \cellcolor{Green3!56}\secondstatcell{0.8037}{0.0042}{}& \cellcolor{Green3!57}\statcell{0.7553}{0.0100}{}& \cellcolor{Green3!0}\statcell{0.6454}{0.0062}{} \\
OSP-1 & \cellcolor{Green3!0}\statcell{0.8172}{0.0038}{}& \cellcolor{Green3!0}\statcell{0.8030}{0.0022}{}& \cellcolor{Green3!0}\statcell{0.7781}{0.0047}{}& \cellcolor{Green3!0}\statcell{0.7389}{0.0026}{}& \cellcolor{Green3!7}\statcell{0.6476}{0.0063}{} \\
BELT-B-R & \cellcolor{Green3!60}\secondstatcell{0.8406}{0.0067}{}& \cellcolor{Green3!60}\beststatcell{0.8280}{0.0081}{}& \cellcolor{Green3!60}\beststatcell{0.8053}{0.0054}{}& \cellcolor{Green3!60}\beststatcell{0.7561}{0.0104}{}& \cellcolor{Green3!60}\beststatcell{0.6633}{0.0045}{} \\
BELT-H-R & \cellcolor{Green3!60}\beststatcell{0.8407}{0.0050}{}& \cellcolor{Green3!52}\statcell{0.8246}{0.0040}{}& \cellcolor{Green3!50}\statcell{0.8009}{0.0044}{}& \cellcolor{Green3!59}\secondstatcell{0.7558}{0.0109}{}& \cellcolor{Green3!22}\secondstatcell{0.6520}{0.0036}{} \\
\midrule
\multicolumn{6}{c}{\textbf{OLMo3 LiveCodeBench, before Aug.\ 2024}} \\
\midrule
OSP & \cellcolor{Green3!23}\statcell{0.6798}{0.0136}{}& \cellcolor{Green3!52}\secondstatcell{0.6610}{0.0123}{}& \cellcolor{Green3!60}\beststatcell{0.6297}{0.0103}{}& \cellcolor{Green3!60}\beststatcell{0.5616}{0.0154}{}& \cellcolor{Green3!13}\secondstatcell{0.4012}{0.0060}{} \\
OSP-1 & \cellcolor{Green3!0}\statcell{0.6740}{0.0040}{}& \cellcolor{Green3!0}\statcell{0.6506}{0.0015}{}& \cellcolor{Green3!0}\statcell{0.6127}{0.0141}{}& \cellcolor{Green3!0}\statcell{0.5338}{0.0222}{}& \cellcolor{Green3!0}\statcell{0.3973}{0.0062}{} \\
BELT-B-R & \cellcolor{Green3!36}\secondstatcell{0.6831}{0.0037}{}& \cellcolor{Green3!45}\statcell{0.6595}{0.0026}{}& \cellcolor{Green3!42}\statcell{0.6245}{0.0053}{}& \cellcolor{Green3!19}\statcell{0.5428}{0.0121}{}& \cellcolor{Green3!12}\statcell{0.4011}{0.0031}{} \\
BELT-H-R & \cellcolor{Green3!60}\beststatcell{0.6890}{0.0042}{}& \cellcolor{Green3!60}\beststatcell{0.6626}{0.0059}{}& \cellcolor{Green3!55}\secondstatcell{0.6283}{0.0065}{}& \cellcolor{Green3!51}\secondstatcell{0.5575}{0.0095}{}& \cellcolor{Green3!60}\beststatcell{0.4158}{0.0091}{} \\
\midrule
\multicolumn{6}{c}{\textbf{OLMo3 LiveCodeBench, from Aug.\ 2024}} \\
\midrule
OSP & \cellcolor{Green3!44}\statcell{0.4431}{0.0118}{}& \cellcolor{Green3!60}\beststatcell{0.4238}{0.0099}{}& \cellcolor{Green3!60}\beststatcell{0.3916}{0.0066}{}& \cellcolor{Green3!60}\beststatcell{0.3237}{0.0057}{}& \cellcolor{Green3!60}\beststatcell{0.1915}{0.0022}{} \\
OSP-1 & \cellcolor{Green3!0}\statcell{0.4279}{0.0019}{}& \cellcolor{Green3!0}\statcell{0.3990}{0.0036}{}& \cellcolor{Green3!0}\statcell{0.3673}{0.0029}{}& \cellcolor{Green3!0}\statcell{0.2935}{0.0126}{}& \cellcolor{Green3!18}\statcell{0.1805}{0.0158}{} \\
BELT-B-R & \cellcolor{Green3!48}\secondstatcell{0.4447}{0.0064}{}& \cellcolor{Green3!43}\statcell{0.4169}{0.0033}{}& \cellcolor{Green3!24}\secondstatcell{0.3770}{0.0080}{}& \cellcolor{Green3!28}\statcell{0.3078}{0.0059}{}& \cellcolor{Green3!0}\statcell{0.1758}{0.0146}{} \\
BELT-H-R & \cellcolor{Green3!60}\beststatcell{0.4488}{0.0108}{}& \cellcolor{Green3!51}\secondstatcell{0.4200}{0.0123}{}& \cellcolor{Green3!23}\statcell{0.3765}{0.0184}{}& \cellcolor{Green3!41}\secondstatcell{0.3141}{0.0049}{}& \cellcolor{Green3!50}\secondstatcell{0.1890}{0.0023}{} \\
\bottomrule\end{tabular}\endgroup
\end{center}
\end{table}

\FloatBarrier

\subsection{Inference-time threshold sweeps}
\label{sec:threshold-expansion}
\ecoparagraph{Scope of the main comparison.}
The main tables compare complete training/adaptation recipes, including the single-direction variants, rather than isolating candidate count. On Qwen3, original OSP follows its price-specific stochastic policy recipe; the 13-threshold expansion below is an additional evaluation constructed here, not part of the cited OSP method, and is compared with both parallel BELT adaptation modes (refit and threshold) on the same probe. OLMo3 retains its separately registered thresholded BELT protocol.

Threshold expansion adds candidate policies without retraining weights: BELT-B-R+T/H-R+T threshold multiple refitted directions, and BELT-B-T/H-T one fitted direction. For each of OSP's five checkpoints and each of BELT-B-R/H-R's 17 policies, we take 13 equally spaced source-validation score quantiles and stop at the first threshold crossing. Retaining the original stochastic policies gives 70 OSP and 238 candidates per BELT interface before deduplication, plus natural completion. OSP-1+T adds the single checkpoint's native stochastic policy to its already-frozen 13 thresholds (14 non-natural candidates on Qwen3 and OLMo3); the original OSP-1 row retains its 13 deterministic rules. On OLMo3, OSP+T applies the same 70-candidate construction to the five OLMo3 checkpoints, with thresholds from each checkpoint's own development-split scores; its native members are the OSP row of \cref{tab:olmo-main}. Both members of this comparison use the natural-completion grading and elicited-answer token cost of \cref{sec:natural-elicitation}. Test-frontier constraints and utility selection use micro averages; Qwen3 macro values evaluate the same policies or mixtures.

\ecoparagraph{Price-specific fitting and threshold-only adaptation.}
A single fitted direction with thresholds can match or exceed the family that refits at every price: on Qwen3 OOD, BELT-H-T saves 45.98\% versus refit BELT-H-R's 45.25\% at the one-point tolerance (\cref{tab:allbaselines}); \cref{sec:analysis-single-direction,sec:single-direction-details} analyze these nonnested families and their seed consistency.

\ecoparagraph{Retaining the single checkpoint's native policy.}
Across the five evaluated scopes, OSP-1+T leaves 14 of 15 tolerance saving means unchanged (LiveCodeBench Post at 2~pp improves by 0.12~pp) and raises seven of 20 budget-accuracy means by 0.01--0.09~pp. The gains are small because OSP-1 already spans 13 threshold rules and the extension adds one stochastic member; the saving comparisons retain their ordering.

\ecoparagraph{Competitive trade-offs after threshold expansion.}
Using the best printed mean among BELT-B-R+T/H-R+T/B-T/H-T, BELT exceeds OSP+T in 17 of 28 constrained readings and trails in 11. At fixed saving, BELT wins 9 of 16 accuracy cells, and all seven of its accuracy deficits are below 1~pp (maximum 0.54~pp); the four gaps of 1~pp or more are BELT gains on Pre. Threshold expansion thus preserves competitive accuracy--cost coverage with smaller controller-training and inference requirements (\cref{sec:cost,sec:cost-inference,sec:cost-measured}), while changing some individual rankings; it leaves controller-training costs unchanged.

On Qwen3 OOD at natural accuracy, expansion narrows the refit H-R--OSP saving gap from 13.19 to 1.01~pp; expanded OSP leads expanded refit B-R/H-R in OOD accuracy at the four budgets, but H-T exceeds OSP+T at 40\% saving (\crefrange{tab:threshold-expansion-qwen}{tab:threshold-expansion-qwen-accuracy}). On LiveCodeBench Pre, B-T and B-R+T save more than OSP+T at the one-point tolerance; on Post, OSP+T saves 1.21--2.25~pp more at the three tolerances, while B-R+T leads it at 40\% saving and is within 0.09--0.39~pp of its accuracy at the other three budgets (\crefrange{tab:threshold-expansion-code}{tab:threshold-expansion-code-accuracy}). On OLMo3, BELT-H-R wins five of the seven constrained readings against OSP+T with no loss under the seed-SD rule (\cref{tab:olmo-full-0-1,tab:reb-single-variant-wtl}). Utility rankings are reported separately in \cref{tab:threshold-expansion-utility}.
\begin{table}[!htp]
\caption{Original and threshold-expanded families on Qwen3 (Macro). Reporting convention follows \cref{tab:allbaselines}. NF denotes an infeasible budget in at least one seed. Panel: token saving. OSP-1+T retains the native stochastic policy alongside its existing 13 source-validation thresholds.}\label{tab:threshold-expansion-qwen}
\begin{center}
\begingroup\setlength{\tabcolsep}{4pt}
\begin{tabular}{@{}lccc@{}}\toprule
\textbf{Method} & \textbf{0 pp} & \textbf{1 pp} & \textbf{2 pp} \\\midrule
\multicolumn{4}{c}{\textbf{ID}} \\
OSP & \cellcolor{Green3!55}\statcell{44.77}{0.75}{}& \cellcolor{Green3!57}\statcell{49.72}{1.09}{}& \cellcolor{Green3!58}\statcell{54.15}{0.90}{} \\
OSP+T & \cellcolor{Green3!57}\statcell{45.58}{0.99}{}& \cellcolor{Green3!58}\statcell{49.78}{3.28}{}& \cellcolor{Green3!60}\beststatcell{54.82}{0.24}{} \\
OSP-1+T & \cellcolor{Green3!39}\statcell{39.19}{5.46}{}& \cellcolor{Green3!48}\statcell{46.59}{2.01}{}& \cellcolor{Green3!51}\statcell{51.50}{1.49}{} \\
\textbf{BELT-B-T} & \cellcolor{Green3!0}\statcell{25.21}{1.08}{}& \cellcolor{Green3!0}\statcell{29.57}{1.50}{}& \cellcolor{Green3!0}\statcell{33.54}{1.74}{} \\
\textbf{BELT-H-T} & \cellcolor{Green3!60}\secondstatcell{46.52}{1.51}{}& \cellcolor{Green3!60}\beststatcell{50.64}{0.98}{}& \cellcolor{Green3!59}\statcell{54.29}{0.79}{} \\
BELT-B-R & \cellcolor{Green3!54}\statcell{44.45}{3.00}{}& \cellcolor{Green3!55}\statcell{48.74}{1.90}{}& \cellcolor{Green3!54}\statcell{52.58}{0.60}{} \\
BELT-B-R+T & \cellcolor{Green3!57}\statcell{45.65}{1.25}{}& \cellcolor{Green3!56}\statcell{49.41}{2.25}{}& \cellcolor{Green3!58}\statcell{54.10}{1.06}{} \\
BELT-H-R & \cellcolor{Green3!52}\statcell{43.76}{0.98}{}& \cellcolor{Green3!54}\statcell{48.58}{0.44}{}& \cellcolor{Green3!53}\statcell{52.28}{0.25}{} \\
BELT-H-R+T & \cellcolor{Green3!60}\beststatcell{46.62}{0.82}{}& \cellcolor{Green3!58}\secondstatcell{50.06}{0.68}{}& \cellcolor{Green3!60}\secondstatcell{54.70}{0.64}{} \\
\multicolumn{4}{c}{\textbf{OOD}} \\
OSP & \cellcolor{Green3!32}\statcell{25.92}{22.46}{}& \cellcolor{Green3!54}\statcell{45.19}{1.69}{}& \cellcolor{Green3!55}\statcell{50.68}{2.04}{} \\
OSP+T & \cellcolor{Green3!58}\statcell{40.08}{0.85}{}& \cellcolor{Green3!55}\statcell{45.34}{4.15}{}& \cellcolor{Green3!59}\secondstatcell{51.69}{0.88}{} \\
OSP-1+T & \cellcolor{Green3!0}\statcell{9.07}{15.71}{}& \cellcolor{Green3!23}\statcell{36.89}{0.18}{}& \cellcolor{Green3!31}\statcell{44.37}{0.81}{} \\
\textbf{BELT-B-T} & \cellcolor{Green3!30}\statcell{25.22}{1.01}{}& \cellcolor{Green3!0}\statcell{30.87}{0.92}{}& \cellcolor{Green3!0}\statcell{36.01}{0.76}{} \\
\textbf{BELT-H-T} & \cellcolor{Green3!59}\secondstatcell{40.77}{1.46}{}& \cellcolor{Green3!57}\secondstatcell{45.98}{0.60}{}& \cellcolor{Green3!52}\statcell{49.82}{0.63}{} \\
BELT-B-R & \cellcolor{Green3!53}\statcell{37.14}{1.95}{}& \cellcolor{Green3!49}\statcell{43.76}{1.77}{}& \cellcolor{Green3!48}\statcell{48.74}{0.62}{} \\
BELT-B-R+T & \cellcolor{Green3!51}\statcell{36.23}{2.80}{}& \cellcolor{Green3!52}\statcell{44.76}{1.78}{}& \cellcolor{Green3!50}\statcell{49.23}{1.64}{} \\
BELT-H-R & \cellcolor{Green3!56}\statcell{39.11}{0.63}{}& \cellcolor{Green3!54}\statcell{45.25}{1.12}{}& \cellcolor{Green3!54}\statcell{50.26}{1.09}{} \\
BELT-H-R+T & \cellcolor{Green3!60}\beststatcell{41.09}{1.12}{}& \cellcolor{Green3!60}\beststatcell{46.75}{1.38}{}& \cellcolor{Green3!60}\beststatcell{51.98}{1.80}{} \\
\bottomrule\end{tabular}\endgroup
\end{center}
\end{table}

\begin{table}[!htp]
\caption{Original and threshold-expanded families on Qwen3 (Macro). Reporting convention follows \cref{tab:allbaselines}. NF denotes an infeasible budget in at least one seed. Panel: accuracy.}\label{tab:threshold-expansion-qwen-accuracy}
\begin{center}
\begingroup\setlength{\tabcolsep}{4pt}
\begin{tabular}{@{}lcccc@{}}\toprule
\textbf{Method} & \textbf{70\%} & \textbf{60\%} & \textbf{50\%} & \textbf{40\%} \\\midrule
\multicolumn{5}{c}{\textbf{ID}} \\
OSP & \cellcolor{Green3!55}\statcell{44.83}{0.15}{}& \cellcolor{Green3!56}\statcell{49.24}{0.26}{}& \cellcolor{Green3!59}\secondstatcell{52.44}{0.26}{}& \cellcolor{Green3!58}\statcell{54.41}{0.20}{} \\
OSP+T & \cellcolor{Green3!59}\secondstatcell{45.46}{0.32}{}& \cellcolor{Green3!60}\secondstatcell{49.64}{0.28}{}& \cellcolor{Green3!59}\statcell{52.40}{0.77}{}& \cellcolor{Green3!60}\beststatcell{54.57}{0.27}{} \\
OSP-1+T & \cellcolor{Green3!54}\statcell{44.74}{0.10}{}& \cellcolor{Green3!51}\statcell{48.73}{0.12}{}& \cellcolor{Green3!47}\statcell{51.33}{0.35}{}& \cellcolor{Green3!42}\statcell{53.20}{0.39}{} \\
\textbf{BELT-B-T} & \cellcolor{Green3!0}\statcell{37.46}{2.1}{}& \cellcolor{Green3!0}\statcell{43.11}{1.4}{}& \cellcolor{Green3!0}\statcell{47.14}{0.8}{}& \cellcolor{Green3!0}\statcell{50.09}{0.5}{} \\
\textbf{BELT-H-T} & \cellcolor{Green3!47}\statcell{43.81}{0.3}{}& \cellcolor{Green3!57}\statcell{49.34}{0.4}{}& \cellcolor{Green3!60}\beststatcell{52.49}{0.2}{}& \cellcolor{Green3!58}\secondstatcell{54.42}{0.2}{} \\
BELT-B-R & \cellcolor{Green3!53}\statcell{44.65}{0.11}{}& \cellcolor{Green3!57}\statcell{49.32}{0.12}{}& \cellcolor{Green3!53}\statcell{51.90}{0.17}{}& \cellcolor{Green3!52}\statcell{53.99}{0.43}{} \\
BELT-B-R+T & \cellcolor{Green3!52}\statcell{44.41}{0.27}{}& \cellcolor{Green3!56}\statcell{49.22}{0.34}{}& \cellcolor{Green3!56}\statcell{52.15}{0.33}{}& \cellcolor{Green3!54}\statcell{54.12}{0.15}{} \\
BELT-H-R & \cellcolor{Green3!59}\statcell{45.39}{0.09}{}& \cellcolor{Green3!56}\statcell{49.27}{0.12}{}& \cellcolor{Green3!56}\statcell{52.12}{0.12}{}& \cellcolor{Green3!53}\statcell{54.05}{0.21}{} \\
BELT-H-R+T & \cellcolor{Green3!60}\beststatcell{45.53}{0.23}{}& \cellcolor{Green3!60}\beststatcell{49.69}{0.19}{}& \cellcolor{Green3!59}\statcell{52.41}{0.19}{}& \cellcolor{Green3!56}\statcell{54.26}{0.11}{} \\
\multicolumn{5}{c}{\textbf{OOD}} \\
OSP & \cellcolor{Green3!59}\statcell{58.91}{0.65}{}& \cellcolor{Green3!56}\statcell{62.58}{0.19}{}& \cellcolor{Green3!57}\secondstatcell{65.25}{0.30}{}& \cellcolor{Green3!55}\statcell{66.76}{0.14}{} \\
OSP+T & \cellcolor{Green3!60}\beststatcell{59.10}{0.40}{}& \cellcolor{Green3!60}\beststatcell{63.08}{0.38}{}& \cellcolor{Green3!60}\beststatcell{65.51}{0.31}{}& \cellcolor{Green3!60}\secondstatcell{67.06}{0.11}{} \\
OSP-1+T & \cellcolor{Green3!59}\secondstatcell{58.95}{0.22}{}& \cellcolor{Green3!51}\statcell{62.10}{0.33}{}& \cellcolor{Green3!47}\statcell{64.30}{0.04}{}& \cellcolor{Green3!41}\statcell{65.73}{0.11}{} \\
\textbf{BELT-B-T} & \cellcolor{Green3!0}\statcell{50.01}{2.6}{}& \cellcolor{Green3!0}\statcell{56.20}{1.2}{}& \cellcolor{Green3!0}\statcell{60.04}{0.7}{}& \cellcolor{Green3!0}\statcell{62.80}{0.4}{} \\
\textbf{BELT-H-T} & \cellcolor{Green3!36}\statcell{55.50}{0.7}{}& \cellcolor{Green3!48}\statcell{61.75}{0.3}{}& \cellcolor{Green3!56}\statcell{65.12}{0.0}{}& \cellcolor{Green3!60}\beststatcell{67.09}{0.2}{} \\
BELT-B-R & \cellcolor{Green3!56}\statcell{58.42}{0.23}{}& \cellcolor{Green3!55}\statcell{62.53}{0.29}{}& \cellcolor{Green3!54}\statcell{64.93}{0.12}{}& \cellcolor{Green3!54}\statcell{66.66}{0.38}{} \\
BELT-B-R+T & \cellcolor{Green3!52}\statcell{57.92}{0.72}{}& \cellcolor{Green3!55}\statcell{62.49}{0.40}{}& \cellcolor{Green3!54}\statcell{64.99}{0.39}{}& \cellcolor{Green3!56}\statcell{66.78}{0.24}{} \\
BELT-H-R & \cellcolor{Green3!56}\statcell{58.47}{0.61}{}& \cellcolor{Green3!54}\statcell{62.42}{0.36}{}& \cellcolor{Green3!53}\statcell{64.90}{0.18}{}& \cellcolor{Green3!55}\statcell{66.73}{0.09}{} \\
BELT-H-R+T & \cellcolor{Green3!56}\statcell{58.56}{0.51}{}& \cellcolor{Green3!56}\secondstatcell{62.67}{0.54}{}& \cellcolor{Green3!57}\statcell{65.24}{0.20}{}& \cellcolor{Green3!59}\statcell{67.03}{0.12}{} \\
\bottomrule\end{tabular}\endgroup
\end{center}
\end{table}

\begin{table}[!htp]
\caption{Original and threshold-expanded families on LiveCodeBench. Reporting convention follows \cref{tab:allbaselines}. Panel: token saving. OSP-1+T retains the native stochastic policy alongside its existing 13 source-validation thresholds.}\label{tab:threshold-expansion-code}
\begin{center}
\begingroup\setlength{\tabcolsep}{4pt}
\begin{tabular}{@{}lccc@{}}\toprule
\textbf{Method} & \textbf{0 pp} & \textbf{1 pp} & \textbf{2 pp} \\\midrule
\multicolumn{4}{c}{\textbf{Pre}} \\
OSP & \cellcolor{Green3!49}\statcell{34.76}{3.37}{}& \cellcolor{Green3!43}\statcell{38.25}{2.80}{}& \cellcolor{Green3!38}\statcell{41.74}{2.78}{} \\
OSP+T & \cellcolor{Green3!51}\statcell{36.78}{2.14}{}& \cellcolor{Green3!48}\statcell{41.35}{1.74}{}& \cellcolor{Green3!42}\statcell{43.73}{0.47}{} \\
OSP-1+T & \cellcolor{Green3!0}\statcell{0.03}{0.06}{}& \cellcolor{Green3!0}\statcell{15.25}{7.45}{}& \cellcolor{Green3!0}\statcell{23.71}{7.40}{} \\
\textbf{BELT-B-T} & \cellcolor{Green3!60}\beststatcell{42.96}{0.36}{}& \cellcolor{Green3!60}\beststatcell{47.65}{1.07}{}& \cellcolor{Green3!58}\secondstatcell{51.40}{1.14}{} \\
\textbf{BELT-H-T} & \cellcolor{Green3!47}\statcell{33.40}{6.35}{}& \cellcolor{Green3!43}\statcell{38.25}{6.81}{}& \cellcolor{Green3!39}\statcell{42.22}{6.02}{} \\
BELT-B-R & \cellcolor{Green3!53}\statcell{37.84}{4.79}{}& \cellcolor{Green3!51}\statcell{42.87}{3.28}{}& \cellcolor{Green3!51}\statcell{48.18}{1.92}{} \\
BELT-B-R+T & \cellcolor{Green3!59}\secondstatcell{42.14}{1.78}{}& \cellcolor{Green3!60}\secondstatcell{47.40}{1.24}{}& \cellcolor{Green3!60}\beststatcell{52.31}{0.81}{} \\
BELT-H-R & \cellcolor{Green3!49}\statcell{35.37}{2.51}{}& \cellcolor{Green3!46}\statcell{40.15}{2.32}{}& \cellcolor{Green3!43}\statcell{44.19}{1.56}{} \\
BELT-H-R+T & \cellcolor{Green3!50}\statcell{35.96}{2.65}{}& \cellcolor{Green3!48}\statcell{40.99}{3.04}{}& \cellcolor{Green3!45}\statcell{45.18}{2.44}{} \\
\multicolumn{4}{c}{\textbf{Post}} \\
OSP & \cellcolor{Green3!56}\statcell{54.58}{2.30}{}& \cellcolor{Green3!42}\statcell{62.38}{2.01}{}& \cellcolor{Green3!50}\secondstatcell{70.23}{0.20}{} \\
OSP+T & \cellcolor{Green3!60}\beststatcell{57.69}{2.91}{}& \cellcolor{Green3!60}\beststatcell{66.10}{1.84}{}& \cellcolor{Green3!60}\beststatcell{71.45}{1.75}{} \\
OSP-1+T & \cellcolor{Green3!0}\statcell{14.07}{24.37}{}& \cellcolor{Green3!0}\statcell{53.92}{5.13}{}& \cellcolor{Green3!23}\statcell{66.87}{1.53}{} \\
\textbf{BELT-B-T} & \cellcolor{Green3!53}\statcell{52.77}{0.66}{}& \cellcolor{Green3!25}\statcell{59.08}{1.13}{}& \cellcolor{Green3!0}\statcell{64.04}{1.04}{} \\
\textbf{BELT-H-T} & \cellcolor{Green3!47}\statcell{48.30}{8.74}{}& \cellcolor{Green3!22}\statcell{58.47}{5.73}{}& \cellcolor{Green3!29}\statcell{67.62}{3.12}{} \\
BELT-B-R & \cellcolor{Green3!54}\statcell{53.51}{2.49}{}& \cellcolor{Green3!36}\statcell{61.21}{1.07}{}& \cellcolor{Green3!38}\statcell{68.77}{2.91}{} \\
BELT-B-R+T & \cellcolor{Green3!58}\secondstatcell{56.48}{2.22}{}& \cellcolor{Green3!50}\secondstatcell{64.01}{2.39}{}& \cellcolor{Green3!42}\statcell{69.20}{3.08}{} \\
BELT-H-R & \cellcolor{Green3!50}\statcell{50.48}{5.43}{}& \cellcolor{Green3!24}\statcell{58.76}{3.94}{}& \cellcolor{Green3!14}\statcell{65.80}{2.33}{} \\
BELT-H-R+T & \cellcolor{Green3!50}\statcell{50.72}{3.87}{}& \cellcolor{Green3!36}\statcell{61.30}{5.65}{}& \cellcolor{Green3!25}\statcell{67.11}{3.60}{} \\
\bottomrule\end{tabular}\endgroup
\end{center}
\end{table}

\begin{table}[!htp]
\caption{Original and threshold-expanded families on LiveCodeBench. Reporting convention follows \cref{tab:allbaselines}. Panel: accuracy.}\label{tab:threshold-expansion-code-accuracy}
\begin{center}
\begingroup\setlength{\tabcolsep}{4pt}
\begin{tabular}{@{}lcccc@{}}\toprule
\textbf{Method} & \textbf{70\%} & \textbf{60\%} & \textbf{50\%} & \textbf{40\%} \\\midrule
\multicolumn{5}{c}{\textbf{Pre}} \\
OSP & \infeasiblecell& \cellcolor{Green3!11}\statcell{55.10}{0.74}{}& \cellcolor{Green3!23}\statcell{58.39}{0.81}{}& \cellcolor{Green3!31}\statcell{60.99}{0.70}{} \\
OSP+T & \cellcolor{Green3!45}\statcell{54.04}{0.34}{}& \cellcolor{Green3!41}\statcell{56.88}{0.82}{}& \cellcolor{Green3!31}\statcell{58.96}{0.72}{}& \cellcolor{Green3!44}\statcell{61.95}{0.49}{} \\
OSP-1+T & \cellcolor{Green3!0}\statcell{50.84}{0.30}{}& \cellcolor{Green3!0}\statcell{54.40}{0.41}{}& \cellcolor{Green3!0}\statcell{56.76}{0.24}{}& \cellcolor{Green3!0}\statcell{58.74}{0.90}{} \\
\textbf{BELT-B-T} & \cellcolor{Green3!55}\secondstatcell{54.76}{0.32}{}& \cellcolor{Green3!56}\secondstatcell{57.84}{0.39}{}& \cellcolor{Green3!58}\secondstatcell{60.89}{0.29}{}& \cellcolor{Green3!60}\beststatcell{63.09}{0.06}{} \\
\textbf{BELT-H-T} & \cellcolor{Green3!19}\statcell{52.22}{0.57}{}& \cellcolor{Green3!23}\statcell{55.81}{0.84}{}& \cellcolor{Green3!27}\statcell{58.67}{1.23}{}& \cellcolor{Green3!32}\statcell{61.08}{1.36}{} \\
BELT-B-R & \cellcolor{Green3!29}\statcell{52.89}{0.75}{}& \cellcolor{Green3!42}\statcell{56.99}{1.07}{}& \cellcolor{Green3!47}\statcell{60.06}{0.52}{}& \cellcolor{Green3!49}\statcell{62.29}{1.08}{} \\
BELT-B-R+T & \cellcolor{Green3!60}\beststatcell{55.09}{0.42}{}& \cellcolor{Green3!60}\beststatcell{58.07}{0.44}{}& \cellcolor{Green3!60}\beststatcell{61.00}{0.24}{}& \cellcolor{Green3!58}\secondstatcell{62.91}{0.35}{} \\
BELT-H-R & \cellcolor{Green3!25}\statcell{52.59}{0.15}{}& \cellcolor{Green3!27}\statcell{56.06}{0.48}{}& \cellcolor{Green3!31}\statcell{58.96}{0.31}{}& \cellcolor{Green3!39}\statcell{61.57}{0.48}{} \\
BELT-H-R+T & \cellcolor{Green3!23}\statcell{52.44}{0.75}{}& \cellcolor{Green3!23}\statcell{55.81}{0.79}{}& \cellcolor{Green3!34}\statcell{59.14}{0.38}{}& \cellcolor{Green3!40}\statcell{61.63}{0.57}{} \\
\multicolumn{5}{c}{\textbf{Post}} \\
OSP & \cellcolor{Green3!52}\secondstatcell{26.71}{0.04}{}& \cellcolor{Green3!39}\statcell{27.96}{0.29}{}& \cellcolor{Green3!58}\secondstatcell{29.53}{0.54}{}& \cellcolor{Green3!47}\statcell{30.05}{0.51}{} \\
OSP+T & \cellcolor{Green3!60}\beststatcell{26.92}{0.31}{}& \cellcolor{Green3!60}\beststatcell{28.34}{0.43}{}& \cellcolor{Green3!60}\beststatcell{29.59}{0.41}{}& \cellcolor{Green3!56}\secondstatcell{30.39}{0.53}{} \\
OSP-1+T & \cellcolor{Green3!35}\statcell{26.27}{0.07}{}& \cellcolor{Green3!0}\statcell{27.27}{0.21}{}& \cellcolor{Green3!0}\statcell{27.94}{0.35}{}& \cellcolor{Green3!0}\statcell{28.33}{0.41}{} \\
\textbf{BELT-B-T} & \cellcolor{Green3!0}\statcell{25.39}{0.08}{}& \cellcolor{Green3!13}\statcell{27.50}{0.21}{}& \cellcolor{Green3!41}\statcell{29.08}{0.04}{}& \cellcolor{Green3!54}\statcell{30.30}{0.09}{} \\
\textbf{BELT-H-T} & \cellcolor{Green3!35}\statcell{26.28}{0.51}{}& \cellcolor{Green3!18}\statcell{27.59}{0.53}{}& \cellcolor{Green3!25}\statcell{28.64}{0.76}{}& \cellcolor{Green3!30}\statcell{29.43}{0.76}{} \\
BELT-B-R & \cellcolor{Green3!41}\statcell{26.44}{0.41}{}& \cellcolor{Green3!33}\statcell{27.85}{0.18}{}& \cellcolor{Green3!38}\statcell{28.99}{0.22}{}& \cellcolor{Green3!41}\statcell{29.84}{0.16}{} \\
BELT-B-R+T & \cellcolor{Green3!45}\statcell{26.53}{0.68}{}& \cellcolor{Green3!53}\secondstatcell{28.22}{0.31}{}& \cellcolor{Green3!57}\statcell{29.50}{0.23}{}& \cellcolor{Green3!60}\beststatcell{30.52}{0.17}{} \\
BELT-H-R & \cellcolor{Green3!23}\statcell{25.98}{0.42}{}& \cellcolor{Green3!17}\statcell{27.57}{0.46}{}& \cellcolor{Green3!28}\statcell{28.72}{0.55}{}& \cellcolor{Green3!31}\statcell{29.47}{0.34}{} \\
BELT-H-R+T & \cellcolor{Green3!36}\statcell{26.31}{0.66}{}& \cellcolor{Green3!31}\statcell{27.83}{0.69}{}& \cellcolor{Green3!29}\statcell{28.75}{0.59}{}& \cellcolor{Green3!35}\statcell{29.61}{0.40}{} \\
\bottomrule\end{tabular}\endgroup
\end{center}
\end{table}

\begin{table}[tp]
\caption{Utility of the original and threshold-expanded families. Qwen3 uses Macro averages. Reporting convention follows \cref{tab:allbaselines}.}\label{tab:threshold-expansion-utility}
\begin{center}
\begingroup
\setlength{\tabcolsep}{2.8pt}
\begin{tabular}{@{}lccccc@{}}\toprule
\textbf{Method} & $2\times10^{-6}$ & $5\times10^{-6}$ & $10^{-5}$ & $2\times10^{-5}$ & $5\times10^{-5}$ \\
\midrule
\multicolumn{6}{c}{\textbf{Qwen3 ID (Macro)}} \\
\midrule
OSP & \cellcolor{Green3!0}\statcell{0.5421}{0.0022}{}& \cellcolor{Green3!0}\statcell{0.5300}{0.0018}{}& \cellcolor{Green3!42}\statcell{0.5098}{0.0012}{}& \cellcolor{Green3!51}\secondstatcell{0.4695}{0.0004}{}& \cellcolor{Green3!38}\secondstatcell{0.3714}{0.0009}{} \\
OSP+T & \cellcolor{Green3!30}\secondstatcell{0.5493}{0.0033}{}& \cellcolor{Green3!32}\secondstatcell{0.5342}{0.0030}{}& \cellcolor{Green3!60}\beststatcell{0.5110}{0.0012}{}& \cellcolor{Green3!60}\beststatcell{0.4704}{0.0014}{}& \cellcolor{Green3!60}\beststatcell{0.3746}{0.0020}{} \\
\addlinespace[2pt]
BELT-B-R & \cellcolor{Green3!60}\beststatcell{0.5564}{0.0007}{}& \cellcolor{Green3!60}\beststatcell{0.5379}{0.0006}{}& \cellcolor{Green3!33}\statcell{0.5092}{0.0023}{}& \cellcolor{Green3!24}\statcell{0.4669}{0.0018}{}& \cellcolor{Green3!14}\statcell{0.3678}{0.0010}{} \\
BELT-B-R+T & \cellcolor{Green3!60}\beststatcell{0.5564}{0.0007}{}& \cellcolor{Green3!60}\beststatcell{0.5379}{0.0006}{}& \cellcolor{Green3!56}\secondstatcell{0.5107}{0.0015}{}& \cellcolor{Green3!25}\statcell{0.4670}{0.0018}{}& \cellcolor{Green3!23}\statcell{0.3691}{0.0015}{} \\
\addlinespace[2pt]
BELT-H-R & \cellcolor{Green3!16}\statcell{0.5460}{0.0018}{}& \cellcolor{Green3!6}\statcell{0.5308}{0.0014}{}& \cellcolor{Green3!0}\statcell{0.5070}{0.0013}{}& \cellcolor{Green3!0}\statcell{0.4646}{0.0018}{}& \cellcolor{Green3!0}\statcell{0.3657}{0.0012}{} \\
BELT-H-R+T & \cellcolor{Green3!24}\statcell{0.5479}{0.0012}{}& \cellcolor{Green3!21}\statcell{0.5328}{0.0018}{}& \cellcolor{Green3!44}\statcell{0.5099}{0.0021}{}& \cellcolor{Green3!41}\statcell{0.4686}{0.0017}{}& \cellcolor{Green3!37}\statcell{0.3712}{0.0013}{} \\
\midrule
\multicolumn{6}{c}{\textbf{Qwen3 OOD (Macro)}} \\
\midrule
OSP & \cellcolor{Green3!0}\statcell{0.6653}{0.0047}{}& \cellcolor{Green3!0}\statcell{0.6544}{0.0038}{}& \cellcolor{Green3!0}\statcell{0.6363}{0.0024}{}& \cellcolor{Green3!38}\statcell{0.6001}{0.0017}{}& \cellcolor{Green3!38}\secondstatcell{0.5135}{0.0040}{} \\
OSP+T & \cellcolor{Green3!28}\statcell{0.6748}{0.0031}{}& \cellcolor{Green3!28}\secondstatcell{0.6611}{0.0025}{}& \cellcolor{Green3!26}\statcell{0.6388}{0.0019}{}& \cellcolor{Green3!43}\secondstatcell{0.6006}{0.0023}{}& \cellcolor{Green3!60}\beststatcell{0.5165}{0.0048}{} \\
\addlinespace[2pt]
BELT-B-R & \cellcolor{Green3!60}\beststatcell{0.6855}{0.0005}{}& \cellcolor{Green3!60}\beststatcell{0.6690}{0.0004}{}& \cellcolor{Green3!57}\secondstatcell{0.6417}{0.0002}{}& \cellcolor{Green3!0}\statcell{0.5962}{0.0031}{}& \cellcolor{Green3!4}\statcell{0.5090}{0.0029}{} \\
BELT-B-R+T & \cellcolor{Green3!60}\beststatcell{0.6855}{0.0005}{}& \cellcolor{Green3!60}\beststatcell{0.6690}{0.0004}{}& \cellcolor{Green3!60}\beststatcell{0.6420}{0.0005}{}& \cellcolor{Green3!18}\statcell{0.5980}{0.0016}{}& \cellcolor{Green3!0}\statcell{0.5084}{0.0027}{} \\
\addlinespace[2pt]
BELT-H-R & \cellcolor{Green3!25}\statcell{0.6737}{0.0017}{}& \cellcolor{Green3!21}\statcell{0.6594}{0.0013}{}& \cellcolor{Green3!4}\statcell{0.6367}{0.0015}{}& \cellcolor{Green3!21}\statcell{0.5983}{0.0006}{}& \cellcolor{Green3!30}\statcell{0.5125}{0.0052}{} \\
BELT-H-R+T & \cellcolor{Green3!31}\secondstatcell{0.6756}{0.0027}{}& \cellcolor{Green3!26}\statcell{0.6607}{0.0025}{}& \cellcolor{Green3!26}\statcell{0.6388}{0.0022}{}& \cellcolor{Green3!60}\beststatcell{0.6023}{0.0010}{}& \cellcolor{Green3!60}\beststatcell{0.5165}{0.0014}{} \\
\midrule
\multicolumn{6}{c}{\textbf{LiveCodeBench Pre}} \\
\midrule
OSP & \cellcolor{Green3!0}\statcell{0.6287}{0.0083}{}& \cellcolor{Green3!0}\statcell{0.6146}{0.0082}{}& \cellcolor{Green3!3}\statcell{0.5911}{0.0080}{}& \cellcolor{Green3!13}\statcell{0.5440}{0.0084}{}& \cellcolor{Green3!0}\statcell{0.4241}{0.0054}{} \\
OSP+T & \cellcolor{Green3!17}\secondstatcell{0.6413}{0.0185}{}& \cellcolor{Green3!16}\secondstatcell{0.6247}{0.0164}{}& \cellcolor{Green3!15}\statcell{0.5977}{0.0140}{}& \cellcolor{Green3!20}\statcell{0.5462}{0.0089}{}& \cellcolor{Green3!35}\secondstatcell{0.4424}{0.0030}{} \\
\addlinespace[2pt]
BELT-B-R & \cellcolor{Green3!60}\beststatcell{0.6729}{0.0009}{}& \cellcolor{Green3!60}\beststatcell{0.6533}{0.0012}{}& \cellcolor{Green3!59}\secondstatcell{0.6205}{0.0015}{}& \cellcolor{Green3!54}\secondstatcell{0.5561}{0.0020}{}& \cellcolor{Green3!25}\statcell{0.4370}{0.0089}{} \\
BELT-B-R+T & \cellcolor{Green3!60}\beststatcell{0.6729}{0.0009}{}& \cellcolor{Green3!60}\beststatcell{0.6533}{0.0012}{}& \cellcolor{Green3!60}\beststatcell{0.6208}{0.0013}{}& \cellcolor{Green3!60}\beststatcell{0.5579}{0.0005}{}& \cellcolor{Green3!60}\beststatcell{0.4553}{0.0014}{} \\
\addlinespace[2pt]
BELT-H-R & \cellcolor{Green3!4}\statcell{0.6313}{0.0071}{}& \cellcolor{Green3!2}\statcell{0.6157}{0.0059}{}& \cellcolor{Green3!0}\statcell{0.5898}{0.0042}{}& \cellcolor{Green3!0}\statcell{0.5403}{0.0034}{}& \cellcolor{Green3!11}\statcell{0.4300}{0.0043}{} \\
BELT-H-R+T & \cellcolor{Green3!11}\statcell{0.6371}{0.0058}{}& \cellcolor{Green3!9}\statcell{0.6206}{0.0054}{}& \cellcolor{Green3!7}\statcell{0.5936}{0.0048}{}& \cellcolor{Green3!10}\statcell{0.5432}{0.0052}{}& \cellcolor{Green3!15}\statcell{0.4318}{0.0059}{} \\
\midrule
\multicolumn{6}{c}{\textbf{LiveCodeBench Post}} \\
\midrule
OSP & \cellcolor{Green3!11}\statcell{0.2901}{0.0050}{}& \cellcolor{Green3!26}\statcell{0.2745}{0.0051}{}& \cellcolor{Green3!49}\statcell{0.2485}{0.0060}{}& \cellcolor{Green3!34}\statcell{0.2099}{0.0009}{}& \cellcolor{Green3!0}\statcell{0.1400}{0.0047}{} \\
OSP+T & \cellcolor{Green3!25}\statcell{0.2955}{0.0082}{}& \cellcolor{Green3!33}\statcell{0.2767}{0.0055}{}& \cellcolor{Green3!60}\beststatcell{0.2507}{0.0023}{}& \cellcolor{Green3!60}\beststatcell{0.2152}{0.0021}{}& \cellcolor{Green3!60}\beststatcell{0.1483}{0.0021}{} \\
\addlinespace[2pt]
BELT-B-R & \cellcolor{Green3!58}\secondstatcell{0.3080}{0.0013}{}& \cellcolor{Green3!44}\secondstatcell{0.2798}{0.0012}{}& \cellcolor{Green3!19}\statcell{0.2429}{0.0018}{}& \cellcolor{Green3!24}\statcell{0.2078}{0.0046}{}& \cellcolor{Green3!1}\statcell{0.1401}{0.0046}{} \\
BELT-B-R+T & \cellcolor{Green3!60}\beststatcell{0.3087}{0.0007}{}& \cellcolor{Green3!60}\beststatcell{0.2845}{0.0013}{}& \cellcolor{Green3!57}\secondstatcell{0.2502}{0.0010}{}& \cellcolor{Green3!41}\secondstatcell{0.2113}{0.0032}{}& \cellcolor{Green3!49}\statcell{0.1468}{0.0045}{} \\
\addlinespace[2pt]
BELT-H-R & \cellcolor{Green3!0}\statcell{0.2858}{0.0047}{}& \cellcolor{Green3!0}\statcell{0.2670}{0.0032}{}& \cellcolor{Green3!0}\statcell{0.2392}{0.0053}{}& \cellcolor{Green3!0}\statcell{0.2029}{0.0039}{}& \cellcolor{Green3!22}\statcell{0.1431}{0.0003}{} \\
BELT-H-R+T & \cellcolor{Green3!5}\statcell{0.2876}{0.0056}{}& \cellcolor{Green3!7}\statcell{0.2689}{0.0049}{}& \cellcolor{Green3!16}\statcell{0.2423}{0.0063}{}& \cellcolor{Green3!28}\statcell{0.2086}{0.0054}{}& \cellcolor{Green3!50}\secondstatcell{0.1469}{0.0021}{} \\
\bottomrule\end{tabular}\endgroup
\end{center}
\end{table}

\FloatBarrier

\subsection{Budget control without policy mixtures}
\Crefrange{tab:ceiling}{tab:ceiling-ood} select a single feasible policy without interpolation. A single policy can leave budget unused, whereas a mixture reaches costs between adjacent frontier members; exports retain realized costs for both.
\begin{table}[!htp]
\caption{Macro accuracy (\%) under single-member token budgets. At each target, the most accurate feasible member is selected without interpolation. Entries are mean$_{\pm\mathrm{SD}}$ across seeds; NF denotes an infeasible target. Panel: ID.}\label{tab:ceiling}
\begin{center}
\begingroup\setlength{\tabcolsep}{4pt}
\begin{tabular}{@{}lcccc@{}}\toprule
\textbf{Method} & \textbf{70\%} & \textbf{60\%} & \textbf{50\%} & \textbf{40\%} \\\midrule
OSP & \statcell{43.52}{0.48}{}& \statcell{46.91}{1.52}{}& \statcell{51.43}{0.78}{}& \statcell{51.43}{0.78}{} \\
OSP-1 & \statcell{43.39}{0.86}{}& \statcell{46.94}{1.19}{}& \statcell{50.31}{1.19}{}& \statcell{52.67}{0.64}{} \\
\textbf{BELT-B-T} & \statcell{37.11}{1.98}{}& \statcell{41.98}{1.48}{}& \statcell{45.70}{0.85}{}& \statcell{48.43}{0.65}{} \\
\textbf{BELT-H-T} & \statcell{40.60}{1.73}{}& \statcell{46.88}{1.08}{}& \statcell{51.57}{0.48}{}& \statcell{53.93}{0.34}{} \\
\textbf{BELT-B-R} & \statcell{41.39}{0.82}{}& \statcell{47.33}{1.94}{}& \statcell{50.83}{0.88}{}& \statcell{53.22}{0.69}{} \\
\textbf{BELT-H-R} & \statcell{43.11}{0.93}{}& \statcell{47.24}{1.18}{}& \statcell{51.40}{0.21}{}& \statcell{53.92}{0.29}{} \\
\bottomrule\end{tabular}\endgroup
\end{center}
\end{table}

\begin{table}[!htp]
\caption{Macro accuracy (\%) under single-member token budgets. At each target, the most accurate feasible member is selected without interpolation. Entries are mean$_{\pm\mathrm{SD}}$ across seeds; NF denotes an infeasible target. Panel: OOD.}\label{tab:ceiling-ood}
\begin{center}
\begingroup\setlength{\tabcolsep}{4pt}
\begin{tabular}{@{}lcccc@{}}\toprule
\textbf{Method} & \textbf{70\%} & \textbf{60\%} & \textbf{50\%} & \textbf{40\%} \\\midrule
OSP & \statcell{58.45}{0.98}{}& \statcell{60.32}{1.44}{}& \statcell{64.35}{0.61}{}& \statcell{65.09}{1.49}{} \\
OSP-1 & \statcell{57.78}{0.76}{}& \statcell{61.64}{0.55}{}& \statcell{63.89}{0.39}{}& \statcell{65.51}{0.29}{} \\
\textbf{BELT-B-T} & \statcell{47.14}{3.10}{}& \statcell{53.39}{1.52}{}& \statcell{59.77}{0.73}{}& \statcell{61.92}{0.56}{} \\
\textbf{BELT-H-T} & \statcell{52.72}{2.30}{}& \statcell{59.65}{1.38}{}& \statcell{64.26}{0.37}{}& \statcell{66.14}{0.43}{} \\
\textbf{BELT-B-R} & \statcell{56.56}{0.94}{}& \statcell{61.19}{1.22}{}& \statcell{64.20}{0.63}{}& \statcell{66.31}{0.48}{} \\
\textbf{BELT-H-R} & \statcell{56.61}{0.21}{}& \statcell{61.38}{1.15}{}& \statcell{64.08}{0.14}{}& \statcell{66.60}{0.08}{} \\
\bottomrule\end{tabular}\endgroup
\end{center}
\end{table}

\FloatBarrier

\subsection{In-distribution accuracy--cost trade-offs}
\begin{figure}[!ht]\centering
\includegraphics[width=.7\textwidth]{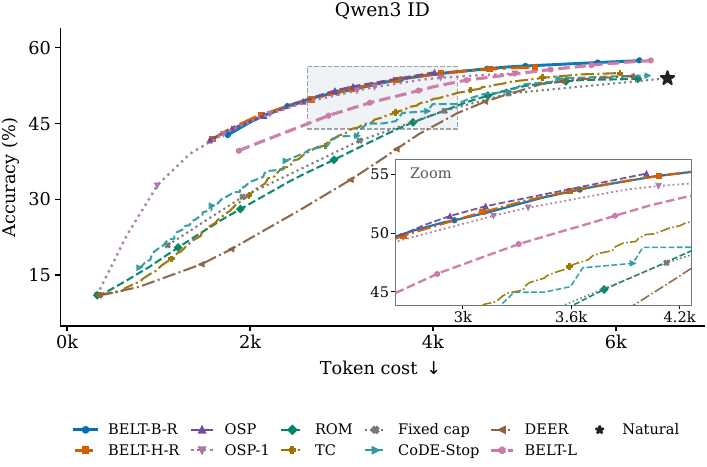}
\caption{Qwen3 ID frontiers; the inset enlarges the central accuracy--cost trade-off. The presentation follows \cref{fig:tradeoff-main}.}
\label{fig:tradeoff-full}
\end{figure}

\clearpage
\subsection{Accuracy--cost trade-offs by dataset}
\label{sec:dataset-tradeoffs}
\begin{figure}[!ht]
\centering
\includegraphics[width=\textwidth]{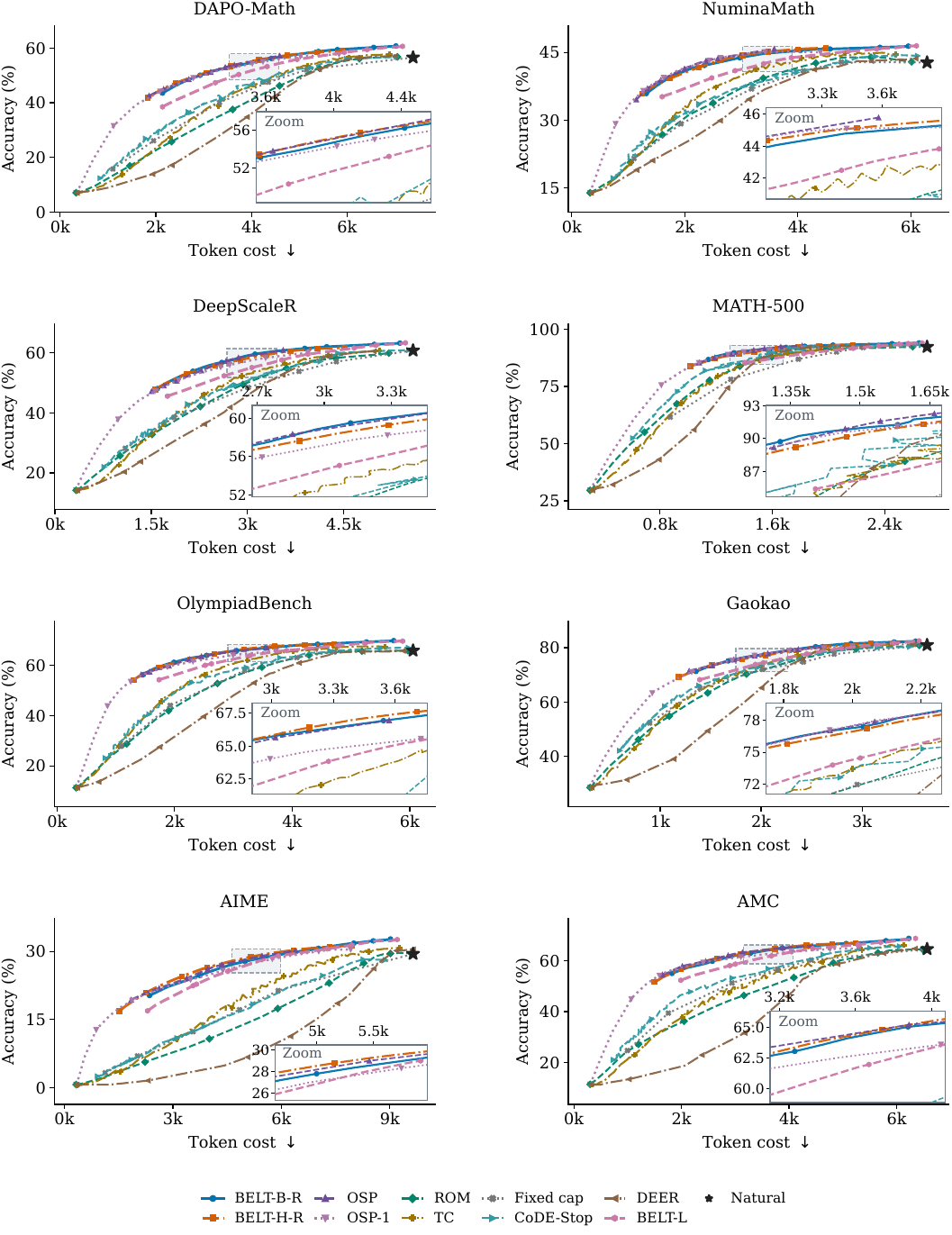}
\caption{Dataset-level frontiers. The presentation follows \cref{fig:tradeoff-main}. Axes differ across panels. Each inset enlarges the same 38--52\% token-saving cost range to distinguish nearby frontiers.}
\label{fig:tradeoff-datasets}
\end{figure}
\clearpage

\subsection{OLMo3 accuracy--token cost trade-offs}
\label{sec:olmo-tradeoffs}
\begin{figure}[!ht]
\centering
\includegraphics[width=\textwidth]{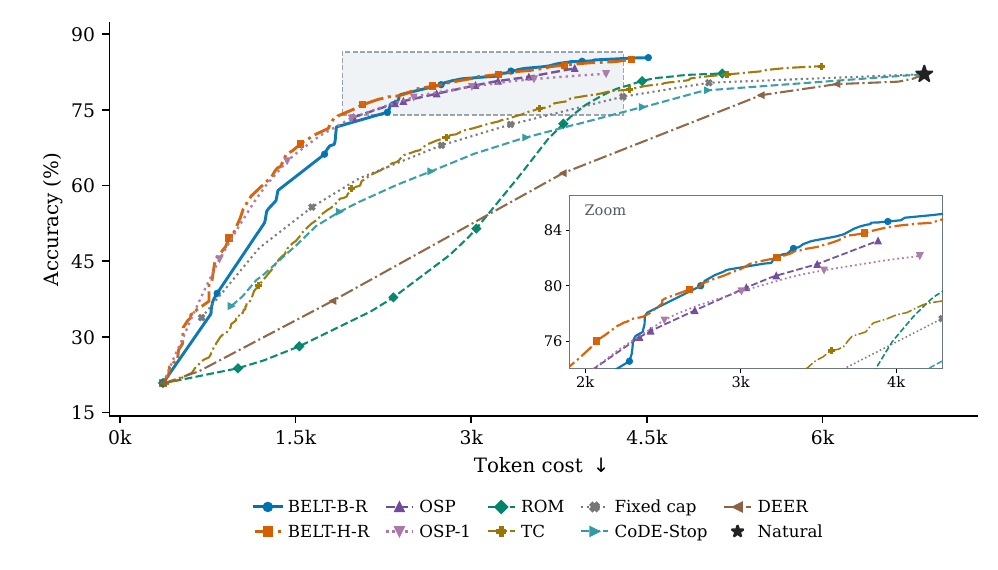}
\caption{OLMo3 OOD frontiers. The presentation follows \cref{fig:tradeoff-main}. Single-direction and OSP-1+T comparisons also appear in Tables~\ref{tab:olmo-full}--\ref{tab:olmo-full-1-1}.}
\label{fig:tradeoff-olmo}
\end{figure}

\Cref{fig:tradeoff-olmo} shows the OLMo3 frontiers and operating points of \cref{tab:olmo-main}, evaluated on the test frontier of each method.
\clearpage

\section{Belief and Policy Analysis}
\label{sec:prediction-diagnostics}
\subsection{Features, policy capacity, and belief ablations}
\label{sec:utility-controls}
\ecoparagraph{Shared evaluation protocol.}
The utility controls in \cref{tab:representation-policy-full,tab:belief-utility,tab:probe-controls,tab:input-subsets} select a policy at each reported token price on the 160-question policy-fitting and development sets of \cref{sec:training-recipe}, then evaluate it on the main 1,025-question OOD roster with strict test grading, total generated-token cost and micro average utility. \Cref{tab:comparison-protocols} lists the grading of each comparison and the source of its SD.

\begin{table}[!ht]
\caption{Grading and variability in utility comparisons. The 320-question bank first selects the BELT probe checkpoint; its disjoint 160-question halves then fit and select policies. The last column gives the source of sample SD.}
\label{tab:comparison-protocols}
\begin{center}
\begingroup\setlength{\tabcolsep}{4pt}
\begin{tabular}{@{}>{\raggedright\arraybackslash}p{.24\textwidth}>{\raggedright\arraybackslash}p{.27\textwidth}>{\raggedright\arraybackslash}p{.17\textwidth}>{\raggedright\arraybackslash}p{.22\textwidth}@{}}\toprule
Comparison & Policy fitting / development selection & Test grading & SD across \\\midrule
Qwen3 BELT utility controls & Numeric-equality labels & Strict math & 3 independent probe training runs \\
Fixed feature controls & Numeric-equality labels & Strict math & 5 policy-fitting starts \\
BELT-L & Numeric-equality labels & Strict math & 3 probe training seeds \\
Matched OSP-1/BELT features & Shared fitter; strict math development utility & Strict math; code unit tests & 3 independent training runs per method \\
OLMo3 BELT & Loose-label utility (refit) & Strict math & 3 independent probe training runs \\\bottomrule
\end{tabular}\endgroup
\end{center}
\end{table}

\ecoparagraph{Features and policy capacity.}
\Cref{tab:representation-policy-full}a reports four fixed-feature controls alongside BELT; their SDs measure optimization variability over policy-fitting starts and are not paired with BELT's training-run SDs. Raw-state and pooling controls change prefix access and architecture; PCA matches the 768-dimensional representation. The pretrained-reference control uses a ReProbe checkpoint with the same architecture and no additional four-belief training; since the main BELT probes start randomly, it varies initialization together with belief supervision. BELT-L maps the 2,048-dimensional layer-16 block-end state linearly to the four targets with the loss in \cref{eq:belief-training} (three seeds, checkpoints selected by minimum validation belief loss), and the \beltb{} policy is fit on its four logits and prefix length; it tests the targets without the sequence probe.

In \cref{tab:representation-policy-full}b, the MLP and linear policies use the same four belief logits, prefix length and three probe training runs. MLP minus linear mean utility ranges from $-0.00099$ to $+0.00345$ across the five prices, so the added capacity gives no uniform gain. The hand rules change the available information, using only their named beliefs.
\begin{table}[t]
\caption{Utility comparing features (a) and controller capacity (b). Mean and SD are reported over five policy-fitting starts for the first four feature controls, and three training seeds for OSP-1, BELT-L and BELT policies.}
\label{tab:representation-policy-full}
\begin{center}
\begingroup\setlength{\tabcolsep}{4pt}
\begin{tabular}{@{}>{\raggedright\arraybackslash}p{112pt}ccccc@{}}\toprule
\textbf{Control} & \multicolumn{5}{c}{\textbf{Utility by $\lambda$ ($\times10^{-6}$)}} \\
 & \textbf{2} & \textbf{5} & \textbf{10} & \textbf{20} & \textbf{50} \\\midrule
\multicolumn{6}{@{}l}{\textit{(a) Features for a linear stopping policy}} \\\midrule
Pretrained ReProbe representation & \cellcolor{Green3!4}\statcell{0.719}{0.004}{}& \cellcolor{Green3!24}\statcell{0.713}{0.001}{}& \cellcolor{Green3!30}\statcell{0.691}{0.001}{}& \cellcolor{Green3!33}\statcell{0.650}{0.000}{}& \cellcolor{Green3!23}\statcell{0.550}{0.000}{} \\
Reasoner state (layer 16) & \cellcolor{Green3!16}\statcell{0.724}{0.000}{}& \cellcolor{Green3!19}\statcell{0.711}{0.001}{}& \cellcolor{Green3!24}\statcell{0.688}{0.001}{}& \cellcolor{Green3!29}\statcell{0.647}{0.001}{}& \cellcolor{Green3!23}\statcell{0.550}{0.002}{} \\
Reasoner state (PCA, 768-d) & \cellcolor{Green3!16}\statcell{0.724}{0.000}{}& \cellcolor{Green3!17}\statcell{0.710}{0.000}{}& \cellcolor{Green3!24}\statcell{0.688}{0.000}{}& \cellcolor{Green3!26}\statcell{0.645}{0.001}{}& \cellcolor{Green3!22}\statcell{0.549}{0.004}{} \\
Mean-pooled reasoner states & \cellcolor{Green3!0}\statcell{0.717}{0.000}{}& \cellcolor{Green3!0}\statcell{0.702}{0.000}{}& \cellcolor{Green3!0}\statcell{0.677}{0.000}{}& \cellcolor{Green3!0}\statcell{0.627}{0.000}{}& \cellcolor{Green3!0}\statcell{0.527}{0.002}{} \\
OSP-1 features & \cellcolor{Green3!24}\statcell{0.728}{0.003}{}& \cellcolor{Green3!30}\statcell{0.716}{0.004}{}& \cellcolor{Green3!41}\statcell{0.696}{0.004}{}& \cellcolor{Green3!46}\secondstatcell{0.659}{0.001}{}& \cellcolor{Green3!57}\secondstatcell{0.583}{0.006}{} \\
OSP-1 features (multi-start) & \cellcolor{Green3!27}\statcell{0.729}{0.002}{}& \cellcolor{Green3!30}\statcell{0.716}{0.004}{}& \cellcolor{Green3!41}\statcell{0.696}{0.004}{}& \cellcolor{Green3!46}\secondstatcell{0.659}{0.001}{}& \cellcolor{Green3!57}\secondstatcell{0.583}{0.006}{} \\
BELT-L (2048$\to$4) & \cellcolor{Green3!60}\beststatcell{0.744}{0.000}{}& \cellcolor{Green3!60}\beststatcell{0.730}{0.000}{}& \cellcolor{Green3!60}\beststatcell{0.705}{0.000}{}& \cellcolor{Green3!44}\statcell{0.658}{0.002}{}& \cellcolor{Green3!29}\statcell{0.556}{0.004}{} \\
BELT probe & \cellcolor{Green3!40}\secondstatcell{0.735}{0.001}{}& \cellcolor{Green3!45}\secondstatcell{0.723}{0.001}{}& \cellcolor{Green3!54}\secondstatcell{0.702}{0.002}{}& \cellcolor{Green3!60}\beststatcell{0.669}{0.001}{}& \cellcolor{Green3!60}\beststatcell{0.586}{0.002}{} \\
\midrule
\multicolumn{6}{@{}l}{\textit{(b) Controllers on a fixed BELT probe}} \\\midrule
Rule: $\pnow\geq\tau$ & \cellcolor{Green3!0}\statcell{0.710}{0.002}{}& \cellcolor{Green3!0}\statcell{0.696}{0.002}{}& \cellcolor{Green3!0}\statcell{0.674}{0.002}{}& \cellcolor{Green3!0}\statcell{0.621}{0.014}{}& \cellcolor{Green3!0}\statcell{0.474}{0.019}{} \\
Rule: $\pnow\geq\pend-m$ & \cellcolor{Green3!36}\secondstatcell{0.731}{0.004}{}& \cellcolor{Green3!36}\statcell{0.717}{0.004}{}& \cellcolor{Green3!36}\statcell{0.694}{0.004}{}& \cellcolor{Green3!38}\statcell{0.650}{0.008}{}& \cellcolor{Green3!55}\statcell{0.579}{0.005}{} \\
Linear & \cellcolor{Green3!60}\beststatcell{0.745}{0.001}{}& \cellcolor{Green3!60}\beststatcell{0.731}{0.000}{}& \cellcolor{Green3!60}\beststatcell{0.707}{0.001}{}& \cellcolor{Green3!59}\secondstatcell{0.666}{0.002}{}& \cellcolor{Green3!58}\secondstatcell{0.585}{0.004}{} \\
MLP & \cellcolor{Green3!60}\beststatcell{0.745}{0.001}{}& \cellcolor{Green3!58}\secondstatcell{0.730}{0.001}{}& \cellcolor{Green3!58}\secondstatcell{0.706}{0.001}{}& \cellcolor{Green3!60}\beststatcell{0.667}{0.002}{}& \cellcolor{Green3!60}\beststatcell{0.588}{0.005}{} \\
\bottomrule\end{tabular}\endgroup
\end{center}
\end{table}

\ecoparagraph{Training targets and policy inputs.}
Target ablations retrain the BELT probe, reselect its checkpoint and fit a policy on the retained beliefs, so they measure supervision, checkpoint selection and the output interface together. Input ablations refit the policy on the fixed four-target probe after removing belief coordinates. Both retain prefix length (\cref{tab:belief-utility}; the full-input row is shared).
\begin{table}[t]
\caption{Utility comparing belief training targets and policy inputs across five token prices. Mean and sample SD cover three BELT probe training seeds. The two panels respectively retrain the BELT probe and refit policies on a fixed four-target BELT probe. Values at $\lambda=10^{-5}$ retain four decimal places.}
\label{tab:belief-utility}
\begin{center}
\begingroup\setlength{\tabcolsep}{4pt}
\begin{tabular}{@{}lccccc@{}}\toprule
\textbf{Control} & \multicolumn{5}{c}{\textbf{Utility by $\lambda$ ($\times10^{-6}$)}} \\
 & \textbf{2} & \textbf{5} & \textbf{10} & \textbf{20} & \textbf{50} \\\midrule
\multicolumn{6}{@{}l}{\textit{BELT probe training targets}} \\\midrule
\emph{Now}, \emph{End} & \cellcolor{Green3!0}\secondstatcell{0.745}{0.001}{}& \cellcolor{Green3!60}\beststatcell{0.731}{0.001}{}& \cellcolor{Green3!17}\statcell{0.7067}{0.0006}{}& \cellcolor{Green3!0}\statcell{0.659}{0.001}{}& \cellcolor{Green3!0}\statcell{0.582}{0.002}{} \\
\emph{Now}, \emph{End}, \emph{Later} & \cellcolor{Green3!0}\secondstatcell{0.745}{0.000}{}& \cellcolor{Green3!0}\secondstatcell{0.730}{0.000}{}& \cellcolor{Green3!0}\statcell{0.7065}{0.0004}{}& \cellcolor{Green3!0}\statcell{0.659}{0.001}{}& \cellcolor{Green3!30}\secondstatcell{0.585}{0.004}{} \\
\emph{Now}, \emph{End}, \emph{Same} & \cellcolor{Green3!60}\beststatcell{0.746}{0.001}{}& \cellcolor{Green3!60}\beststatcell{0.731}{0.001}{}& \cellcolor{Green3!26}\secondstatcell{0.7068}{0.0010}{}& \cellcolor{Green3!51}\secondstatcell{0.665}{0.002}{}& \cellcolor{Green3!60}\beststatcell{0.588}{0.003}{} \\
All four & \cellcolor{Green3!0}\secondstatcell{0.745}{0.001}{}& \cellcolor{Green3!60}\beststatcell{0.731}{0.000}{}& \cellcolor{Green3!60}\beststatcell{0.7072}{0.0006}{}& \cellcolor{Green3!60}\beststatcell{0.666}{0.002}{}& \cellcolor{Green3!30}\secondstatcell{0.585}{0.004}{} \\
\midrule
\multicolumn{6}{@{}l}{\textit{Policy inputs from a fixed four-target BELT probe}} \\\midrule
\emph{Now}, \emph{End} & \cellcolor{Green3!0}\secondstatcell{0.745}{0.001}{}& \cellcolor{Green3!0}\secondstatcell{0.730}{0.001}{}& \cellcolor{Green3!0}\statcell{0.7061}{0.0006}{}& \cellcolor{Green3!9}\statcell{0.660}{0.003}{}& \cellcolor{Green3!20}\statcell{0.583}{0.001}{} \\
\emph{Now}, \emph{End}, \emph{Later} & \cellcolor{Green3!0}\secondstatcell{0.745}{0.001}{}& \cellcolor{Green3!0}\secondstatcell{0.730}{0.001}{}& \cellcolor{Green3!11}\statcell{0.7063}{0.0008}{}& \cellcolor{Green3!0}\statcell{0.659}{0.003}{}& \cellcolor{Green3!0}\statcell{0.582}{0.002}{} \\
\emph{Now}, \emph{End}, \emph{Same} & \cellcolor{Green3!60}\beststatcell{0.746}{0.001}{}& \cellcolor{Green3!60}\beststatcell{0.731}{0.001}{}& \cellcolor{Green3!49}\secondstatcell{0.7070}{0.0010}{}& \cellcolor{Green3!51}\secondstatcell{0.665}{0.003}{}& \cellcolor{Green3!40}\secondstatcell{0.584}{0.003}{} \\
All four & \cellcolor{Green3!0}\secondstatcell{0.745}{0.001}{}& \cellcolor{Green3!60}\beststatcell{0.731}{0.000}{}& \cellcolor{Green3!60}\beststatcell{0.7072}{0.0006}{}& \cellcolor{Green3!60}\beststatcell{0.666}{0.002}{}& \cellcolor{Green3!60}\beststatcell{0.585}{0.004}{} \\
\bottomrule\end{tabular}\endgroup
\end{center}
\end{table}

\begin{figure}[!htp]\centering
\includegraphics[width=\textwidth]{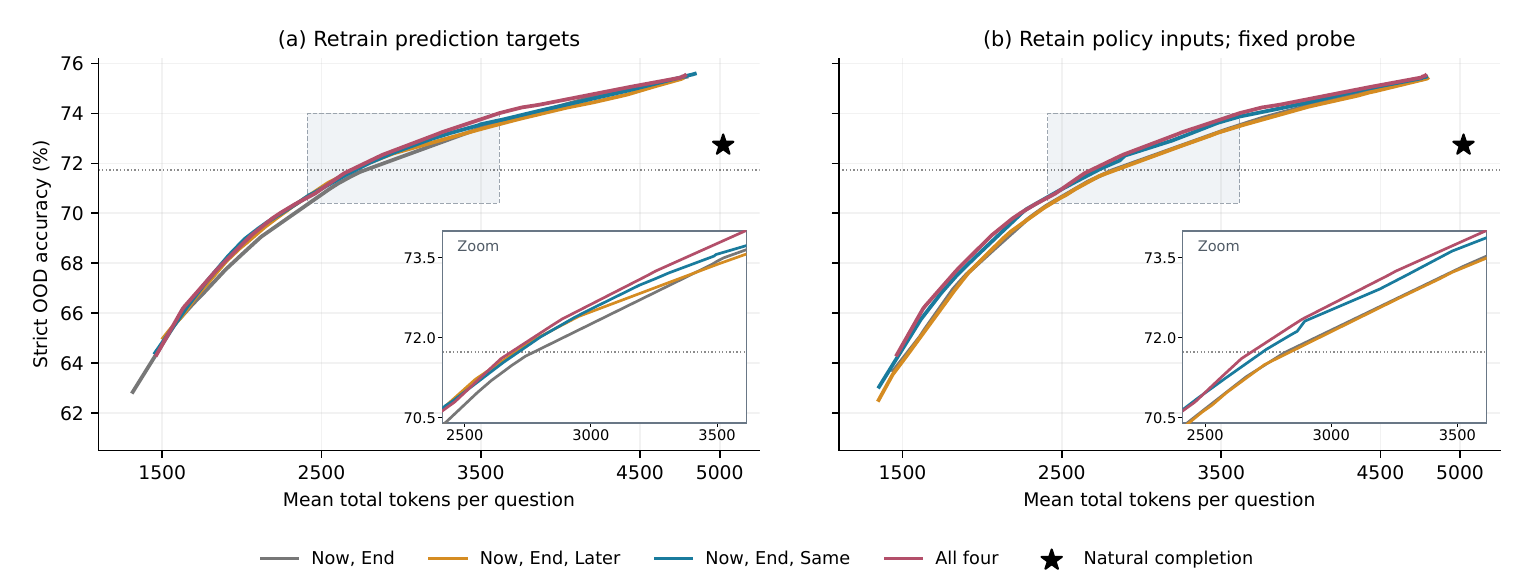}
\caption{Strict OOD accuracy--cost curves for (a) prediction-target retraining and (b) policy-input removal with the four-target probe fixed; prefix length is retained in every arm. Curves show mean interpolated micro accuracy over three probe training runs on each family's common cost support. Insets enlarge the region around the one-point accuracy tolerance (dotted line); stars mark natural completion. \Cref{tab:belief-ablation-frontiers} reports the selected members with macro aggregation.}
\label{fig:belief-ablation-frontiers}
\end{figure}

As a target, adding \emph{Later} to \emph{Now}/\emph{End} decreases mean utility at four of five prices and increases it at the highest. With the probe fixed, removing \emph{Same} lowers mean utility at all five prices, by 0.00052--0.00684, whereas removing \emph{Later} raises it at the two lowest prices, by at most 0.0002, and lowers it at the other three. Same thus makes a consistent mean contribution and Later a price-dependent one.
\ecoparagraph{Belief ablations under accuracy and cost constraints.}
\Cref{tab:belief-ablation-frontiers,fig:belief-ablation-frontiers} recompute both interventions on the same OOD cohort with the main test-family protocol and 17 native policies per seed, without threshold expansion. This descriptive evaluation is distinct from the development-selected policies in \cref{tab:belief-ablation}, and the averaged curves do not define an averaged policy. The full-target and full-input families coincide exactly, and their seven constrained means reproduce the main BELT-B-R OOD row.

Now/End/Same and all-four target training have crossing mean readings at the three tolerances, and their accuracy means differ by at most 0.11~pp at the four budgets. For input ablations, adding Same improves all seven mean constrained readings from both Now/End and Now/End/Later, and adding Later to Now/End/Same improves six of seven. As a prediction target, adding Later to Now/End raises saving by 1.80, 1.98 and 1.67~pp at the 0, 1 and 2~pp tolerances. Its effects are small and depend on the constraint.

\ecoparagraph{Why utility and constrained readings can rank methods differently.}
Utility at the five sampled prices and the constrained readings answer different questions: on the same population and aggregation, maximizing $U_\lambda=A-\lambda C$ over a fixed family selects points where a line of slope $\lambda$ supports its convex hull, and the utility table uses development selection and micro aggregation, whereas constrained readings select test-family members and report macro values. For example, at $\lambda=5\times10^{-5}$ the development-selected Now/End/Same target policies gain 0.00317 utility over all-four with approximately 386 fewer tokens at a 1.61~pp accuracy loss, a favorable exchange at that price (\cref{tab:belief-price-components}).
\begin{table}[!htp]
\centering\small
\caption{OOD accuracy--cost readings for prediction-target retraining and policy-input removal. Both retain prefix length. Each seed contributes 17 native policies plus natural completion, with the main adjacent-member interpolation protocol. Constraints are imposed on micro accuracy/cost; the selected members are then reported as macro token savings or accuracy, as in the main math tables. Entries are mean and sample SD over three probe training runs. NF means that at least one seed cannot reach the budget. All-four is the same BELT-B-R family in both panels and in the main tables. These are exploratory test-family readings, not development-selected deployment guarantees.}
\label{tab:belief-ablation-frontiers}
\begingroup\setlength{\tabcolsep}{3pt}
\begin{tabular}{@{}lrrrrrrr@{}}\toprule
 & \multicolumn{3}{c}{Token saving (\%) at tolerance} & \multicolumn{4}{c}{Accuracy (\%) at saving} \\
Belief set & 0 pp & 1 pp & 2 pp & 70\% & 60\% & 50\% & 40\% \\\midrule
\multicolumn{8}{l}{\textbf{(a) Retrain prediction targets}} \\
Now, End & \statcell{33.26}{1.35}{} & \statcell{41.54}{2.30}{} & \statcell{47.49}{1.30}{} & \statcell{58.46}{0.72}{} & \statcell{62.04}{0.87}{} & \statcell{64.59}{0.37}{} & \statcell{66.19}{0.39}{} \\
Now, End, Later & \statcell{35.06}{3.87}{} & \statcell{43.52}{1.29}{} & \statcell{49.16}{0.67}{} & \infeasiblecell & \statcell{62.38}{0.33}{} & \statcell{64.92}{0.26}{} & \statcell{66.41}{0.29}{} \\
Now, End, Same & \statcell{36.19}{0.58}{} & \statcell{43.28}{0.75}{} & \statcell{49.17}{0.96}{} & \statcell{58.36}{0.82}{} & \statcell{62.61}{0.23}{} & \statcell{64.82}{0.29}{} & \statcell{66.58}{0.11}{} \\
All four & \statcell{37.14}{1.95}{} & \statcell{43.76}{1.77}{} & \statcell{48.74}{0.62}{} & \statcell{58.42}{0.23}{} & \statcell{62.53}{0.29}{} & \statcell{64.93}{0.12}{} & \statcell{66.66}{0.38}{} \\
\midrule
\multicolumn{8}{l}{\textbf{(b) Retain policy inputs; fixed four-target probe}} \\
Now, End & \statcell{32.76}{1.22}{} & \statcell{41.11}{2.39}{} & \statcell{47.01}{1.46}{} & \statcell{57.81}{0.46}{} & \statcell{61.95}{0.25}{} & \statcell{64.59}{0.46}{} & \statcell{66.10}{0.31}{} \\
Now, End, Later & \statcell{32.74}{1.28}{} & \statcell{40.71}{0.70}{} & \statcell{47.28}{1.15}{} & \statcell{57.91}{0.42}{} & \statcell{61.98}{0.14}{} & \statcell{64.54}{0.27}{} & \statcell{66.08}{0.13}{} \\
Now, End, Same & \statcell{36.08}{1.63}{} & \statcell{43.08}{1.27}{} & \statcell{48.84}{1.09}{} & \statcell{58.27}{0.36}{} & \statcell{62.17}{0.39}{} & \statcell{64.90}{0.14}{} & \statcell{66.54}{0.34}{} \\
All four & \statcell{37.14}{1.95}{} & \statcell{43.76}{1.77}{} & \statcell{48.74}{0.62}{} & \statcell{58.42}{0.23}{} & \statcell{62.53}{0.29}{} & \statcell{64.93}{0.12}{} & \statcell{66.66}{0.38}{} \\
\bottomrule\end{tabular}\endgroup
\end{table}
\begin{table}[!htp]
\centering\small
\caption{Accuracy and cost underlying the development-selected utilities for Now/End/Same and all-four target training in \cref{tab:belief-ablation}a. Test accuracy is strict and micro averaged; cost is mean total tokens per question. Entries are mean and sample SD over three separately trained probes. At each price, $U_\lambda=A/100-\lambda C$. These selected points use the archived development selector, rather than test-family constraint selection.}
\label{tab:belief-price-components}
\begingroup\setlength{\tabcolsep}{7pt}
\begin{tabular}{@{}lrrrr@{}}\toprule
 & \multicolumn{2}{c}{Now, End, Same targets} & \multicolumn{2}{c}{All four targets} \\
$\lambda$ ($\times10^{-6}$) & Accuracy (\%) & Total tokens & Accuracy (\%) & Total tokens \\\midrule
2 & \statcell{75.56}{0.12}{} & \statcell{4864.3}{9.9}{} & \statcell{75.50}{0.07}{} & \statcell{4807.3}{42.3}{} \\
5 & \statcell{75.56}{0.12}{} & \statcell{4864.3}{9.9}{} & \statcell{75.50}{0.07}{} & \statcell{4807.3}{42.3}{} \\
10 & \statcell{75.52}{0.06}{} & \statcell{4836.0}{58.5}{} & \statcell{75.49}{0.06}{} & \statcell{4778.3}{33.9}{} \\
20 & \statcell{73.65}{0.63}{} & \statcell{3583.7}{385.2}{} & \statcell{73.67}{0.95}{} & \statcell{3535.2}{561.7}{} \\
50 & \statcell{69.84}{0.57}{} & \statcell{2204.6}{130.3}{} & \statcell{71.46}{1.31}{} & \statcell{2590.2}{325.2}{} \\
\bottomrule\end{tabular}\endgroup
\end{table}

\FloatBarrier

\subsection{Linear belief probe accuracy--cost trade-offs}
\label{sec:linear-probe-frontiers}
BELT-L has the highest mean utility in \cref{tab:representation-policy} at the three lowest prices. \Cref{tab:linear-probe-operating-points} completes its accuracy--cost evaluation on the test family with the same grading, populations and 17-price policy family; OLMo3 was not trained or evaluated. The low-price utility advantage does not carry across all constraints: at a one-point tolerance, BELT-L saves less than H-R on Qwen3 ID and OOD and less than OSP on Post, although on Code Pre its accuracy at 70\% and 60\% saving is above the primary methods'. The sequence probe therefore gives the broader coverage; the linear control's advantage is confined to low-price OOD utility and two code budgets.
\begin{table}[!htp]
\centering\small
\caption{BELT-L operating points under the primary accuracy--cost protocol. Token saving and accuracy are percentages; means and sample SDs cover three probe-training seeds. Math uses macro averages and code uses its single-dataset average. The probe has 8,196 parameters, with a six-parameter policy and 17 fitted prices. NF is an infeasible budget in at least one seed; -- denotes the unevaluated OLMo3 setting. Rows describe different populations and are not ranked against one another.}
\label{tab:linear-probe-operating-points}
\begingroup\setlength{\tabcolsep}{3pt}
\begin{tabular}{@{}lrrrrrrr@{}}\toprule
 & \multicolumn{3}{c}{Token saving (\%) at tolerance} & \multicolumn{4}{c}{Accuracy (\%) at saving} \\
Scope & 0 pp & 1 pp & 2 pp & 70\% & 60\% & 50\% & 40\% \\\midrule
Qwen3 ID & \statcell{30.92}{1.25}{} & \statcell{36.24}{0.27}{} & \statcell{39.94}{0.80}{} & \statcell{40.39}{0.68}{} & \statcell{44.79}{0.56}{} & \statcell{48.65}{0.22}{} & \statcell{51.53}{0.21}{} \\
Qwen3 OOD & \statcell{27.31}{0.59}{} & \statcell{33.72}{0.81}{} & \statcell{38.71}{0.46}{} & \infeasiblecell & \statcell{58.28}{0.36}{} & \statcell{61.78}{0.40}{} & \statcell{64.45}{0.16}{} \\
Code Pre & \statcell{38.90}{2.43}{} & \statcell{43.52}{1.94}{} & \statcell{48.14}{1.45}{} & \statcell{55.06}{0.14}{} & \statcell{57.96}{0.10}{} & \statcell{60.14}{0.28}{} & \statcell{62.32}{0.53}{} \\
Code Post & \statcell{47.94}{2.86}{} & \statcell{55.07}{2.21}{} & \statcell{62.19}{1.58}{} & \statcell{25.57}{0.07}{} & \statcell{26.99}{0.26}{} & \statcell{28.40}{0.39}{} & \statcell{29.81}{0.52}{} \\
OLMo3 OOD & \missingcell & \missingcell & \missingcell & \missingcell & \missingcell & \missingcell & \missingcell \\
\bottomrule\end{tabular}\endgroup
\end{table}

\begin{figure}[!htp]\centering
\includegraphics[width=.95\textwidth]{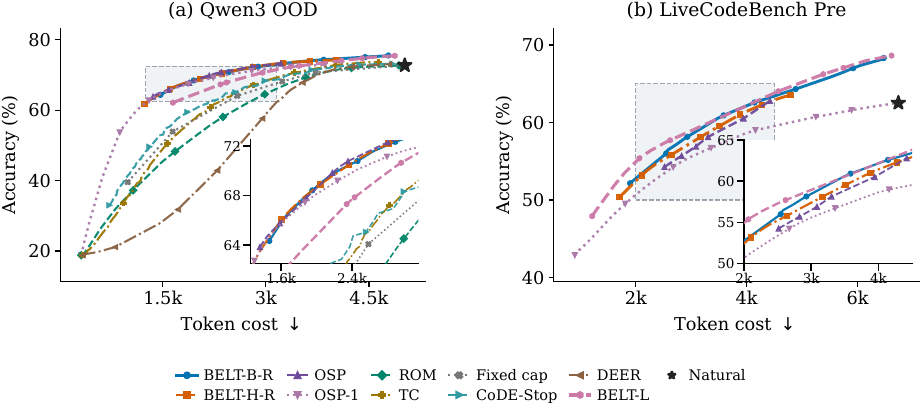}
\caption{Qwen3 OOD and LiveCodeBench Pre frontiers including the BELT-L control. Stars mark natural completion; math here is micro averaged, whereas \cref{tab:linear-probe-operating-points} reports macro averages. BELT-L on ID, individual math datasets and Code Post also appears in \cref{fig:tradeoff-full,fig:tradeoff-datasets,fig:code-post}.}
\label{fig:linear-probe-frontiers}
\end{figure}
\FloatBarrier

\subsection{Matched feature comparison and transfer}
\label{sec:matched-transfer}
\ecoparagraph{Matched policy fitting.}
We freeze three OSP-1 checkpoints trained at $\lambda=10^{-5}$ and extract their 2,048-dimensional boundary states after final normalization. OSP-1 features and BELT-B-R/H-R receive the same 17-price grid, linear fitter, standardization and regularization search, and at each reported price strict-grading micro average development utility selects among the 17 policies and natural completion. The main fit starts from zero weights; a four-start sensitivity adds three random initializations per feature set. This matches fitting opportunities while retaining different architectures and feature dimensions.

BELT-H-R exceeds OSP-1 features at all five OOD prices with the main fit (by 0.0032--0.0094) and also with four fitting starts; BELT-B-R does so with either fitting-start budget. ID rankings depend on the price (\crefrange{tab:matched-features}{tab:matched-features-3}). Applying the same fits and development selections to LiveCodeBench without code-side fitting or selection, BELT-H-R exceeds OSP-1 features at all five Pre prices with either budget; on Post, its main fit exceeds them at the four lower prices and trails at the highest, and four-start Post comparisons vary by price. \Cref{tab:matched-code-pre-seeds,tab:matched-code-post-seeds} report each training run. Question-bootstrap intervals use 5,000 shared resamples with policies and selections fixed, and therefore exclude training-run and selector uncertainty.
\begin{table}[!htp]
\caption{Utility under matched fitting and selection budgets. Reporting convention follows \cref{tab:allbaselines}; all values are micro averages. Panel: Math ID. Token prices $\lambda$ are in $10^{-6}$ units.}\label{tab:matched-features}
\begin{center}
\begingroup\setlength{\tabcolsep}{4pt}
\begin{tabular}{@{}>{\raggedright\arraybackslash}p{73pt}ccccc@{}}\toprule
\textbf{Features / fitting} & \textbf{2} & \textbf{5} & \textbf{10} & \textbf{20} & \textbf{50} \\\midrule
OSP-1 features & \cellcolor{Green3!0}\statcell{0.5409}{0.0011}{}& \cellcolor{Green3!0}\statcell{0.5252}{0.0013}{}& \cellcolor{Green3!0}\statcell{0.4994}{0.0021}{}& \cellcolor{Green3!1}\statcell{0.4582}{0.0018}{}& \cellcolor{Green3!13}\statcell{0.3612}{0.0021}{} \\
OSP-1 features (four starts) & \cellcolor{Green3!4}\statcell{0.5424}{0.0015}{}& \cellcolor{Green3!0}\statcell{0.5252}{0.0013}{}& \cellcolor{Green3!0}\statcell{0.4994}{0.0021}{}& \cellcolor{Green3!0}\statcell{0.4580}{0.0017}{}& \cellcolor{Green3!7}\statcell{0.3611}{0.0021}{} \\
BELT-H-R & \cellcolor{Green3!30}\statcell{0.5513}{0.0017}{}& \cellcolor{Green3!35}\statcell{0.5356}{0.0014}{}& \cellcolor{Green3!56}\statcell{0.5109}{0.0011}{}& \cellcolor{Green3!59}\secondstatcell{0.4670}{0.0015}{}& \cellcolor{Green3!0}\statcell{0.3610}{0.0021}{} \\
BELT-H-R (four starts) & \cellcolor{Green3!29}\statcell{0.5511}{0.0016}{}& \cellcolor{Green3!35}\statcell{0.5356}{0.0014}{}& \cellcolor{Green3!56}\statcell{0.5108}{0.0008}{}& \cellcolor{Green3!60}\beststatcell{0.4671}{0.0015}{}& \cellcolor{Green3!60}\beststatcell{0.3619}{0.0083}{} \\
BELT-B-R & \cellcolor{Green3!60}\secondstatcell{0.5619}{0.0011}{}& \cellcolor{Green3!60}\secondstatcell{0.5430}{0.0009}{}& \cellcolor{Green3!60}\secondstatcell{0.5116}{0.0007}{}& \cellcolor{Green3!49}\statcell{0.4654}{0.0049}{}& \cellcolor{Green3!27}\secondstatcell{0.3614}{0.0052}{} \\
BELT-B-R (four starts) & \cellcolor{Green3!60}\beststatcell{0.5620}{0.0012}{}& \cellcolor{Green3!60}\beststatcell{0.5431}{0.0010}{}& \cellcolor{Green3!60}\beststatcell{0.5117}{0.0008}{}& \cellcolor{Green3!44}\statcell{0.4647}{0.0040}{}& \cellcolor{Green3!7}\statcell{0.3611}{0.0026}{} \\
\bottomrule\end{tabular}\endgroup
\end{center}
\end{table}

\begin{table}[!htp]
\caption{Utility under matched fitting and selection budgets. Reporting convention follows \cref{tab:allbaselines}; all values are micro averages. Panel: Math OOD. Token prices $\lambda$ are in $10^{-6}$ units.}\label{tab:matched-features-1}
\begin{center}
\begingroup\setlength{\tabcolsep}{4pt}
\begin{tabular}{@{}>{\raggedright\arraybackslash}p{73pt}ccccc@{}}\toprule
\textbf{Features / fitting} & \textbf{2} & \textbf{5} & \textbf{10} & \textbf{20} & \textbf{50} \\\midrule
OSP-1 features & \cellcolor{Green3!0}\statcell{0.7284}{0.0033}{}& \cellcolor{Green3!0}\statcell{0.7160}{0.0036}{}& \cellcolor{Green3!0}\statcell{0.6955}{0.0039}{}& \cellcolor{Green3!0}\statcell{0.6594}{0.0012}{}& \cellcolor{Green3!0}\statcell{0.5829}{0.0060}{} \\
OSP-1 features (four starts) & \cellcolor{Green3!4}\statcell{0.7294}{0.0024}{}& \cellcolor{Green3!0}\statcell{0.7160}{0.0036}{}& \cellcolor{Green3!1}\statcell{0.6956}{0.0039}{}& \cellcolor{Green3!0}\statcell{0.6594}{0.0012}{}& \cellcolor{Green3!0}\statcell{0.5829}{0.0059}{} \\
BELT-H-R & \cellcolor{Green3!24}\statcell{0.7352}{0.0015}{}& \cellcolor{Green3!27}\statcell{0.7227}{0.0012}{}& \cellcolor{Green3!35}\statcell{0.7024}{0.0015}{}& \cellcolor{Green3!60}\beststatcell{0.6689}{0.0011}{}& \cellcolor{Green3!44}\secondstatcell{0.5861}{0.0020}{} \\
BELT-H-R (four starts) & \cellcolor{Green3!20}\statcell{0.7340}{0.0011}{}& \cellcolor{Green3!24}\statcell{0.7219}{0.0007}{}& \cellcolor{Green3!34}\statcell{0.7021}{0.0013}{}& \cellcolor{Green3!59}\secondstatcell{0.6688}{0.0011}{}& \cellcolor{Green3!60}\beststatcell{0.5873}{0.0047}{} \\
BELT-B-R & \cellcolor{Green3!60}\beststatcell{0.7454}{0.0006}{}& \cellcolor{Green3!60}\beststatcell{0.7310}{0.0005}{}& \cellcolor{Green3!60}\beststatcell{0.7072}{0.0006}{}& \cellcolor{Green3!42}\statcell{0.6660}{0.0024}{}& \cellcolor{Green3!30}\statcell{0.5851}{0.0042}{} \\
BELT-B-R (four starts) & \cellcolor{Green3!59}\secondstatcell{0.7451}{0.0004}{}& \cellcolor{Green3!59}\secondstatcell{0.7307}{0.0003}{}& \cellcolor{Green3!57}\secondstatcell{0.7066}{0.0001}{}& \cellcolor{Green3!43}\statcell{0.6662}{0.0026}{}& \cellcolor{Green3!19}\statcell{0.5843}{0.0045}{} \\
\bottomrule\end{tabular}\endgroup
\end{center}
\end{table}

\begin{table}[!htp]
\caption{Utility under matched fitting and selection budgets. Reporting convention follows \cref{tab:allbaselines}; all values are micro averages. Panel: LiveCodeBench Pre. Token prices $\lambda$ are in $10^{-6}$ units.}\label{tab:matched-features-2}
\begin{center}
\begingroup\setlength{\tabcolsep}{4pt}
\begin{tabular}{@{}>{\raggedright\arraybackslash}p{73pt}ccccc@{}}\toprule
\textbf{Features / fitting} & \textbf{2} & \textbf{5} & \textbf{10} & \textbf{20} & \textbf{50} \\\midrule
OSP-1 features & \cellcolor{Green3!0}\statcell{0.6170}{0.0129}{}& \cellcolor{Green3!0}\statcell{0.6013}{0.0103}{}& \cellcolor{Green3!0}\statcell{0.5756}{0.0067}{}& \cellcolor{Green3!0}\statcell{0.5240}{0.0060}{}& \cellcolor{Green3!44}\statcell{0.4169}{0.0069}{} \\
OSP-1 features (four starts) & \cellcolor{Green3!2}\statcell{0.6185}{0.0137}{}& \cellcolor{Green3!0}\statcell{0.6015}{0.0104}{}& \cellcolor{Green3!0}\statcell{0.5757}{0.0067}{}& \cellcolor{Green3!0}\statcell{0.5240}{0.0062}{}& \cellcolor{Green3!45}\statcell{0.4170}{0.0070}{} \\
BELT-H-R & \cellcolor{Green3!16}\statcell{0.6308}{0.0066}{}& \cellcolor{Green3!17}\statcell{0.6148}{0.0051}{}& \cellcolor{Green3!6}\statcell{0.5795}{0.0188}{}& \cellcolor{Green3!56}\secondstatcell{0.5332}{0.0138}{}& \cellcolor{Green3!60}\beststatcell{0.4206}{0.0125}{} \\
BELT-H-R (four starts) & \cellcolor{Green3!14}\statcell{0.6290}{0.0112}{}& \cellcolor{Green3!14}\statcell{0.6131}{0.0095}{}& \cellcolor{Green3!7}\statcell{0.5802}{0.0195}{}& \cellcolor{Green3!54}\statcell{0.5329}{0.0136}{}& \cellcolor{Green3!58}\secondstatcell{0.4201}{0.0131}{} \\
BELT-B-R & \cellcolor{Green3!60}\beststatcell{0.6699}{0.0014}{}& \cellcolor{Green3!60}\beststatcell{0.6502}{0.0015}{}& \cellcolor{Green3!60}\beststatcell{0.6179}{0.0010}{}& \cellcolor{Green3!52}\statcell{0.5325}{0.0134}{}& \cellcolor{Green3!0}\statcell{0.4064}{0.0192}{} \\
BELT-B-R (four starts) & \cellcolor{Green3!60}\secondstatcell{0.6698}{0.0020}{}& \cellcolor{Green3!60}\secondstatcell{0.6501}{0.0021}{}& \cellcolor{Green3!59}\secondstatcell{0.6172}{0.0024}{}& \cellcolor{Green3!60}\beststatcell{0.5339}{0.0148}{}& \cellcolor{Green3!2}\statcell{0.4069}{0.0183}{} \\
\bottomrule\end{tabular}\endgroup
\end{center}
\end{table}

\begin{table}[!htp]
\caption{Utility under matched fitting and selection budgets. Reporting convention follows \cref{tab:allbaselines}; all values are micro averages. Panel: LiveCodeBench Post. Token prices $\lambda$ are in $10^{-6}$ units.}\label{tab:matched-features-3}
\begin{center}
\begingroup\setlength{\tabcolsep}{4pt}
\begin{tabular}{@{}>{\raggedright\arraybackslash}p{73pt}ccccc@{}}\toprule
\textbf{Features / fitting} & \textbf{2} & \textbf{5} & \textbf{10} & \textbf{20} & \textbf{50} \\\midrule
OSP-1 features & \cellcolor{Green3!0}\statcell{0.2834}{0.0037}{}& \cellcolor{Green3!0}\statcell{0.2635}{0.0018}{}& \cellcolor{Green3!2}\statcell{0.2312}{0.0083}{}& \cellcolor{Green3!60}\secondstatcell{0.1898}{0.0105}{}& \cellcolor{Green3!60}\secondstatcell{0.1130}{0.0088}{} \\
OSP-1 features (four starts) & \cellcolor{Green3!3}\statcell{0.2844}{0.0043}{}& \cellcolor{Green3!0}\statcell{0.2636}{0.0019}{}& \cellcolor{Green3!0}\statcell{0.2311}{0.0082}{}& \cellcolor{Green3!60}\statcell{0.1896}{0.0106}{}& \cellcolor{Green3!60}\beststatcell{0.1131}{0.0092}{} \\
BELT-H-R & \cellcolor{Green3!6}\statcell{0.2857}{0.0046}{}& \cellcolor{Green3!8}\statcell{0.2656}{0.0019}{}& \cellcolor{Green3!60}\beststatcell{0.2335}{0.0044}{}& \cellcolor{Green3!60}\beststatcell{0.1899}{0.0092}{}& \cellcolor{Green3!51}\statcell{0.1036}{0.0386}{} \\
BELT-H-R (four starts) & \cellcolor{Green3!4}\statcell{0.2848}{0.0066}{}& \cellcolor{Green3!5}\statcell{0.2648}{0.0041}{}& \cellcolor{Green3!50}\secondstatcell{0.2331}{0.0037}{}& \cellcolor{Green3!58}\statcell{0.1885}{0.0115}{}& \cellcolor{Green3!58}\statcell{0.1108}{0.0360}{} \\
BELT-B-R & \cellcolor{Green3!60}\beststatcell{0.3073}{0.0007}{}& \cellcolor{Green3!60}\beststatcell{0.2791}{0.0009}{}& \cellcolor{Green3!12}\statcell{0.2316}{0.0030}{}& \cellcolor{Green3!0}\statcell{0.1506}{0.0384}{}& \cellcolor{Green3!6}\statcell{0.0583}{0.0794}{} \\
BELT-B-R (four starts) & \cellcolor{Green3!59}\secondstatcell{0.3071}{0.0015}{}& \cellcolor{Green3!59}\secondstatcell{0.2789}{0.0017}{}& \cellcolor{Green3!23}\statcell{0.2320}{0.0025}{}& \cellcolor{Green3!2}\statcell{0.1522}{0.0409}{}& \cellcolor{Green3!0}\statcell{0.0526}{0.0718}{} \\
\bottomrule\end{tabular}\endgroup
\end{center}
\end{table}

\begin{table}[t]
\caption{Individual training runs for the matched feature comparison on LiveCodeBench Pre. Seed indices are not paired across methods.}\label{tab:matched-code-pre-seeds}
\begin{center}
\begingroup\setlength{\tabcolsep}{4pt}
\begin{tabular}{@{}lccccc@{}}\toprule
\textbf{Features / fitting / seed} & \multicolumn{5}{c}{\textbf{Utility by $\lambda$ ($\times10^{-6}$)}} \\
 & \textbf{2} & \textbf{5} & \textbf{10} & \textbf{20} & \textbf{50} \\\midrule
OSP-1 features, main, s0 & \cellcolor{Green3!0}0.6027& \cellcolor{Green3!0}0.5897& \cellcolor{Green3!10}0.5679& \cellcolor{Green3!7}0.5202& \cellcolor{Green3!41}0.4172 \\
OSP-1 features, main, s1 & \cellcolor{Green3!22}0.6278& \cellcolor{Green3!19}0.6090& \cellcolor{Green3!21}0.5788& \cellcolor{Green3!8}0.5207& \cellcolor{Green3!32}0.4099 \\
OSP-1 features, main, s2 & \cellcolor{Green3!15}0.6204& \cellcolor{Green3!15}0.6053& \cellcolor{Green3!22}0.5800& \cellcolor{Green3!30}0.5309& \cellcolor{Green3!50}0.4237 \\
OSP-1 features, four starts, s0 & \cellcolor{Green3!0}0.6027& \cellcolor{Green3!0}0.5897& \cellcolor{Green3!10}0.5679& \cellcolor{Green3!7}0.5202& \cellcolor{Green3!42}0.4177 \\
OSP-1 features, four starts, s1 & \cellcolor{Green3!22}0.6278& \cellcolor{Green3!19}0.6090& \cellcolor{Green3!21}0.5788& \cellcolor{Green3!8}0.5207& \cellcolor{Green3!32}0.4097 \\
OSP-1 features, four starts, s2 & \cellcolor{Green3!19}0.6249& \cellcolor{Green3!15}0.6057& \cellcolor{Green3!22}0.5802& \cellcolor{Green3!31}0.5312& \cellcolor{Green3!50}0.4237 \\
\midrule
BELT-H-R, main, s0 & \cellcolor{Green3!18}0.6234& \cellcolor{Green3!19}0.6090& \cellcolor{Green3!0}0.5580& \cellcolor{Green3!1}0.5177& \cellcolor{Green3!52}0.4251 \\
BELT-H-R, main, s1 & \cellcolor{Green3!29}0.6356& \cellcolor{Green3!26}0.6172& \cellcolor{Green3!30}0.5889& \cellcolor{Green3!46}0.5383& \cellcolor{Green3!27}0.4064 \\
BELT-H-R, main, s2 & \cellcolor{Green3!27}0.6335& \cellcolor{Green3!27}0.6183& \cellcolor{Green3!33}0.5918& \cellcolor{Green3!58}\underline{0.5437}& \cellcolor{Green3!58}\underline{0.4302} \\
BELT-H-R, four starts, s0 & \cellcolor{Green3!12}0.6160& \cellcolor{Green3!12}0.6022& \cellcolor{Green3!0}0.5579& \cellcolor{Green3!1}0.5177& \cellcolor{Green3!49}0.4230 \\
BELT-H-R, four starts, s1 & \cellcolor{Green3!29}0.6356& \cellcolor{Green3!26}0.6172& \cellcolor{Green3!30}0.5889& \cellcolor{Green3!44}0.5373& \cellcolor{Green3!26}0.4057 \\
BELT-H-R, four starts, s2 & \cellcolor{Green3!28}0.6353& \cellcolor{Green3!29}0.6197& \cellcolor{Green3!35}0.5939& \cellcolor{Green3!58}\underline{0.5437}& \cellcolor{Green3!60}\textbf{0.4315} \\
\midrule
BELT-B-R, main, s0 & \cellcolor{Green3!60}\underline{0.6714}& \cellcolor{Green3!60}\underline{0.6517}& \cellcolor{Green3!60}\underline{0.6187}& \cellcolor{Green3!50}0.5400& \cellcolor{Green3!49}0.4229 \\
BELT-B-R, main, s1 & \cellcolor{Green3!57}0.6686& \cellcolor{Green3!57}0.6486& \cellcolor{Green3!58}0.6168& \cellcolor{Green3!0}0.5171& \cellcolor{Green3!0}0.3854 \\
BELT-B-R, main, s2 & \cellcolor{Green3!58}0.6697& \cellcolor{Green3!58}0.6504& \cellcolor{Green3!59}0.6181& \cellcolor{Green3!51}0.5405& \cellcolor{Green3!33}0.4110 \\
BELT-B-R, four starts, s0 & \cellcolor{Green3!60}\textbf{0.6719}& \cellcolor{Green3!60}\textbf{0.6521}& \cellcolor{Green3!60}\textbf{0.6190}& \cellcolor{Green3!50}0.5400& \cellcolor{Green3!50}0.4239 \\
BELT-B-R, four starts, s1 & \cellcolor{Green3!56}0.6678& \cellcolor{Green3!56}0.6478& \cellcolor{Green3!56}0.6145& \cellcolor{Green3!0}0.5171& \cellcolor{Green3!3}0.3875 \\
BELT-B-R, four starts, s2 & \cellcolor{Green3!58}0.6697& \cellcolor{Green3!58}0.6504& \cellcolor{Green3!59}0.6181& \cellcolor{Green3!60}\textbf{0.5447}& \cellcolor{Green3!31}0.4093 \\
\midrule
\bottomrule\end{tabular}\endgroup
\end{center}
\end{table}

\begin{table}[t]
\caption{Individual training runs for the matched feature comparison on LiveCodeBench Post. Seed indices are not paired across methods.}\label{tab:matched-code-post-seeds}
\begin{center}
\begingroup\setlength{\tabcolsep}{4pt}
\begin{tabular}{@{}lccccc@{}}\toprule
\textbf{Features / fitting / seed} & \multicolumn{5}{c}{\textbf{Utility by $\lambda$ ($\times10^{-6}$)}} \\
 & \textbf{2} & \textbf{5} & \textbf{10} & \textbf{20} & \textbf{50} \\\midrule
OSP-1 features, main, s0 & \cellcolor{Green3!3}0.2794& \cellcolor{Green3!11}0.2640& \cellcolor{Green3!59}\underline{0.2383}& \cellcolor{Green3!60}\textbf{0.2015}& \cellcolor{Green3!55}0.1223 \\
OSP-1 features, main, s1 & \cellcolor{Green3!17}0.2867& \cellcolor{Green3!4}0.2615& \cellcolor{Green3!0}0.2221& \cellcolor{Green3!46}0.1811& \cellcolor{Green3!51}0.1120 \\
OSP-1 features, main, s2 & \cellcolor{Green3!12}0.2841& \cellcolor{Green3!15}0.2651& \cellcolor{Green3!41}0.2333& \cellcolor{Green3!50}0.1867& \cellcolor{Green3!48}0.1047 \\
OSP-1 features, four starts, s0 & \cellcolor{Green3!3}0.2794& \cellcolor{Green3!11}0.2640& \cellcolor{Green3!59}\underline{0.2383}& \cellcolor{Green3!60}\textbf{0.2015}& \cellcolor{Green3!55}0.1228 \\
OSP-1 features, four starts, s1 & \cellcolor{Green3!17}0.2867& \cellcolor{Green3!4}0.2615& \cellcolor{Green3!0}0.2221& \cellcolor{Green3!46}0.1811& \cellcolor{Green3!51}0.1116 \\
OSP-1 features, four starts, s2 & \cellcolor{Green3!18}0.2872& \cellcolor{Green3!15}0.2652& \cellcolor{Green3!40}0.2330& \cellcolor{Green3!50}0.1862& \cellcolor{Green3!48}0.1047 \\
\midrule
BELT-H-R, main, s0 & \cellcolor{Green3!8}0.2816& \cellcolor{Green3!10}0.2634& \cellcolor{Green3!36}0.2320& \cellcolor{Green3!56}\underline{0.1958}& \cellcolor{Green3!57}\underline{0.1286} \\
BELT-H-R, main, s1 & \cellcolor{Green3!25}0.2906& \cellcolor{Green3!19}0.2664& \cellcolor{Green3!29}0.2301& \cellcolor{Green3!45}0.1793& \cellcolor{Green3!32}0.0591 \\
BELT-H-R, main, s2 & \cellcolor{Green3!14}0.2850& \cellcolor{Green3!20}0.2670& \cellcolor{Green3!60}\textbf{0.2385}& \cellcolor{Green3!55}0.1946& \cellcolor{Green3!55}0.1231 \\
BELT-H-R, four starts, s0 & \cellcolor{Green3!0}0.2777& \cellcolor{Green3!0}0.2602& \cellcolor{Green3!36}0.2319& \cellcolor{Green3!56}0.1957& \cellcolor{Green3!56}0.1263 \\
BELT-H-R, four starts, s1 & \cellcolor{Green3!25}0.2906& \cellcolor{Green3!19}0.2664& \cellcolor{Green3!29}0.2301& \cellcolor{Green3!42}0.1752& \cellcolor{Green3!36}0.0697 \\
BELT-H-R, four starts, s2 & \cellcolor{Green3!16}0.2862& \cellcolor{Green3!23}0.2679& \cellcolor{Green3!56}0.2373& \cellcolor{Green3!55}0.1946& \cellcolor{Green3!60}\textbf{0.1365} \\
\midrule
BELT-B-R, main, s0 & \cellcolor{Green3!59}\underline{0.3081}& \cellcolor{Green3!59}\underline{0.2798}& \cellcolor{Green3!39}0.2327& \cellcolor{Green3!23}0.1472& \cellcolor{Green3!56}0.1252 \\
BELT-B-R, main, s1 & \cellcolor{Green3!56}0.3069& \cellcolor{Green3!53}0.2781& \cellcolor{Green3!22}0.2282& \cellcolor{Green3!0}0.1140& \cellcolor{Green3!0}-0.0294 \\
BELT-B-R, main, s2 & \cellcolor{Green3!56}0.3069& \cellcolor{Green3!58}0.2795& \cellcolor{Green3!43}0.2339& \cellcolor{Green3!53}0.1906& \cellcolor{Green3!39}0.0792 \\
BELT-B-R, four starts, s0 & \cellcolor{Green3!60}\textbf{0.3088}& \cellcolor{Green3!60}\textbf{0.2803}& \cellcolor{Green3!40}0.2330& \cellcolor{Green3!23}0.1472& \cellcolor{Green3!51}0.1117 \\
BELT-B-R, four starts, s1 & \cellcolor{Green3!54}0.3057& \cellcolor{Green3!50}0.2770& \cellcolor{Green3!26}0.2292& \cellcolor{Green3!0}0.1140& \cellcolor{Green3!1}-0.0274 \\
BELT-B-R, four starts, s2 & \cellcolor{Green3!56}0.3069& \cellcolor{Green3!58}0.2795& \cellcolor{Green3!43}0.2339& \cellcolor{Green3!56}0.1954& \cellcolor{Green3!37}0.0736 \\
\midrule
\bottomrule\end{tabular}\endgroup
\end{center}
\end{table}

\FloatBarrier

\subsection{Fixed-direction versus price-specific policies}
\label{sec:single-direction-details}
BELT-T and BELT-R share belief supervision and probe capacity. On Qwen3, the refit variant evaluates $\pi_i=\sigma(w_\lambda^\top x_i+b_\lambda)$ on the shared standardized features $x_i$ at each of 17 fitted prices; T fits $(w_0,b_0)$ once at the intermediate price under stochastic expected utility (\cref{sec:implementation}) and replays $\pi_i=\mathbf{1}\{w_0^\top x_i+b_0\geq t_q\}$ for 13 source-validation score quantiles. Both include natural completion. T thus changes the intercept of a hard stopping rule while preserving its direction; it is not merely a smaller subset of the stochastic price grid. Stochastic evaluation integrates stopping probabilities exactly, so the contrast is not Monte Carlo evaluation noise. An explicit union would guarantee $\max_{\pi\in\mathcal F\cup\mathcal G}U_\lambda(\pi)\geq\max_{\pi\in\mathcal F}U_\lambda(\pi)$, but not for two separate, nonnested families or for development-selected test performance.

\begin{table}[!htp]
\caption{Illustrative comparisons at a 1~pp accuracy tolerance: token savings (\%) and sample SD over three probe training seeds. Differences (pp, computed before rounding) pair the two families by shared probe seed; the last column counts positive within-seed differences. Math uses the main macro reading; code its single-dataset reading. The four rows illustrate both directions of the comparison rather than forming a new evaluation setting.}
\label{tab:single-direction-comparison}
\begin{center}
\begingroup\setlength{\tabcolsep}{4pt}
\begin{tabular}{@{}llcccc@{}}\toprule
Scope & Interface & Refit & Threshold & Threshold$-$Refit & Positive seeds \\\midrule
Math ID & H & \statcell{48.58}{0.44}{} & \statcell{50.64}{0.98}{} & \statcell{+2.07}{0.59}{} & 3/3 \\
Math OOD & H & \statcell{45.25}{1.12}{} & \statcell{45.98}{0.60}{} & \statcell{+0.74}{1.44}{} & 2/3 \\
Code Pre & B & \statcell{42.87}{3.28}{} & \statcell{47.65}{1.07}{} & \statcell{+4.77}{2.22}{} & 3/3 \\
Code Post & B & \statcell{61.21}{1.07}{} & \statcell{59.08}{1.13}{} & \statcell{-2.12}{1.04}{} & 0/3 \\
\bottomrule\end{tabular}\endgroup
\end{center}
\end{table}

\ecoparagraph{When can thresholds reproduce refitting?}
For $w_0,w_\lambda\ne0$, the hard rules
$\mathbf1\{w_\lambda^\top x+b_\lambda\ge0\}$ and
$\mathbf1\{w_0^\top x+b_0\ge t\}$ agree for every
$x\in\mathbb R^d$ if and only if $w_\lambda=cw_0$ for $c>0$,
with $t=b_0-b_\lambda/c$, since identical nonconstant halfspaces
share their boundary hyperplane and occupied side.
Finite observed support can admit agreement without collinearity,
and a finite threshold grid can omit the required threshold.
For sigmoid hazards, scale also matters, so equal hard boundaries
do not imply equal stochastic first-stop masses $S_i\pi_i$.

\ecoparagraph{A linear counterexample.}
Consider one decision with
$x=(a_{\rm stop}-a_{\rm continue},
C_{\rm continue}-C_{\rm stop})$, whose utility advantage is
$x_1+\lambda x_2$.
For $x_A=(-0.05,1)$ and $x_B=(-0.2,20)$,
a direction fitted to this advantage at $\lambda_0=0.001$
ranks A above B (scores $-0.049,-0.18$).
At $\lambda_1=0.02$, stopping is optimal only at B
(scores $-0.03,+0.2$), which no upper threshold on the
$\lambda_0$ score can reproduce.
Both advantage functions are linear; their normals differ.
This illustrative example is not an empirical result.
Sequential continuation values also depend on the future policy
and price, so linear capacity supplies no common-direction theorem.

\ecoparagraph{Measured geometry and its scope.}
The saved BELT-B-R fits in the common standardized coordinates,
with $\kappa=10^{-4}$, rotate by $83.18^\circ$, $81.94^\circ$
and $67.00^\circ$ between the lowest and highest prices across
probe seeds 0--2 (within-seed angles between five-input normals,
excluding the intercept, computed before averaging).
They establish noncollinearity of these B-R fits, but do not measure
H-R geometry, disagreement on observed prefixes, or correspondence
with the T direction fitted at $10^{-5}$.

The Qwen3 fitting grid $\lambda_k=2\times10^{-6}\cdot20^{k/12}$, $k=0,\ldots,16$ (\cref{tab:policy-hyperparameters}), does not contain the single-direction fitting price $10^{-5}$, so even before threshold replay its fitted direction is not guaranteed to occur in the full grid. OLMo3 uses a different full family with 221 thresholded members (\cref{sec:olmo-replication}); the stochastic-versus-deterministic distinction above is specific to Qwen3.

Three effects should be distinguished. First, threshold coverage: adding thresholds to the native H-R policies improves the Qwen3 OOD reading at 1~pp and reverses its mean ordering with H-T, also at the matching micro constraint (46.82\% to 48.44\%, above H-T's 47.83\%). Second, fits optimize micro utility on a finite source bank, whereas frontier points impose an accuracy or cost constraint and sometimes report macro accuracy, so a direction fitted for one utility trade-off need not cover another constraint well. Third, refitted directions can respond to source-bank noise and may transfer differently to ID, OOD, Pre and Post; fixing the direction could reduce this variability, and the present comparison also changes deterministic versus stochastic stopping. The main frontier readings choose candidates on the test population; development selection noise alone cannot explain their ordering. Some small mean gains reverse across probe seeds (\cref{tab:single-direction-comparison}); the paired differences are descriptive.

\ecoparagraph{Refit plus thresholds versus thresholds, paired by seed.}
\ecoparagraph{Counts, random-initialized probes (main line).} Cells where R+T exceeds T by at least 1\,pp on all three seeds, of seven readings: B: Qwen3 ID 7/7, Qwen3 OOD 7/7, LCB Pre 0/7, LCB Post 3/7, OLMo3 1/7; H: Qwen3 ID 1/7, Qwen3 OOD 1/7, LCB Pre 0/7, LCB Post 0/7, OLMo3 1/7. Total 21 of 70; T exceeds R+T by at least 1\,pp on all three seeds in 0.
\ecoparagraph{Counts, ReProbe-initialized probes (analysis family).} Cells where R+T exceeds T by at least 1\,pp on all three seeds, of seven readings: B: Qwen3 ID 7/7, Qwen3 OOD 7/7, LCB Pre 0/7, LCB Post 3/7, OLMo3 1/7; H: Qwen3 ID 0/7, Qwen3 OOD 1/7, LCB Pre 0/7, LCB Post 0/7, OLMo3 1/7. Total 20 of 70; T exceeds R+T by at least 1\,pp on all three seeds in 0.

\ecoparagraph{H interface.} On the H interface, refitting is competitive with the threshold family rather than ahead of it: R+T exceeds T by at least 1\,pp on every seed only at the 30\% token budget on Qwen3 ID, Qwen3 OOD, OLMo3, and at every reading the mean of R+T minus T is at least $-0.59$\,pp.
Here the single fitted direction with thresholds carries the trade-off: on Qwen3 ID it is ahead of the refit family at the 0\,pp tolerance, the 1\,pp tolerance, the 2\,pp tolerance ($+2.76$, $+2.07$, $+2.01$\,pp, ahead on every seed at each; \cref{tab:single-direction-comparison}).
On LiveCodeBench Pre the B refit family is competitive with its threshold family but below it at seven of seven readings (R$-$T at the tolerance readings: $-5.12$, $-4.77$, $-3.22$\,pp).
That threshold family leads full OSP on Pre: it wins 6 of the 6 feasible readings, each robustly (\cref{tab:reb-single-variant-wtl}).

{\fontsize{8}{9}\selectfont\setlength{\tabcolsep}{1pt}
\begin{longtable}{@{}lllccccccc@{}}
\caption{Refit plus thresholds (R+T) and refit (R) minus the threshold family (T), paired by probe seed, random-initialized probes: mean $\pm$ SD over three seed-paired differences (pp); $^\dagger$ at least 1\,pp on all three seeds. OLMo3's registered BELT family is already refit $\times$ threshold, so its R$-$T row is not defined. \missingcell{} = infeasible for one side.}\label{tab:reb-refit-threshold}\\
\toprule Setting & B/H & Contrast & Save 0 & Save 1 & Save 2 & Acc 0.3 & Acc 0.4 & Acc 0.5 & Acc 0.6 \\\midrule\endfirsthead
\toprule Setting & B/H & Contrast & Save 0 & Save 1 & Save 2 & Acc 0.3 & Acc 0.4 & Acc 0.5 & Acc 0.6 \\\midrule\endhead
\bottomrule\endfoot
Qwen3 ID & B & R+T$-$T & \statcell{+20.44}{0.79}{\dagger} & \statcell{+19.85}{0.93}{\dagger} & \statcell{+20.56}{1.11}{\dagger} & \statcell{+6.95}{2.13}{\dagger} & \statcell{+6.11}{1.52}{\dagger} & \statcell{+5.02}{0.50}{\dagger} & \statcell{+4.03}{0.48}{\dagger} \\
 & B & R$-$T & \statcell{+19.24}{1.95}{\dagger} & \statcell{+19.18}{0.60}{\dagger} & \statcell{+19.04}{1.14}{\dagger} & \statcell{+7.20}{2.15}{\dagger} & \statcell{+6.21}{1.40}{\dagger} & \statcell{+4.77}{0.65}{\dagger} & \statcell{+3.90}{0.13}{\dagger} \\
 & H & R+T$-$T & \statcell{+0.09}{0.90}{} & \statcell{-0.58}{1.12}{} & \statcell{+0.41}{0.22}{} & \statcell{+1.72}{0.49}{\dagger} & \statcell{+0.35}{0.34}{} & \statcell{-0.08}{0.28}{} & \statcell{-0.16}{0.19}{} \\
 & H & R$-$T & \statcell{-2.76}{0.89}{} & \statcell{-2.07}{0.59}{} & \statcell{-2.01}{0.54}{} & \statcell{+1.58}{0.18}{\dagger} & \statcell{-0.07}{0.43}{} & \statcell{-0.37}{0.20}{} & \statcell{-0.37}{0.09}{} \\
\midrule
Qwen3 OOD & B & R+T$-$T & \statcell{+11.01}{1.80}{\dagger} & \statcell{+13.88}{0.86}{\dagger} & \statcell{+13.22}{1.72}{\dagger} & \statcell{+7.91}{3.35}{\dagger} & \statcell{+6.29}{1.10}{\dagger} & \statcell{+4.96}{0.66}{\dagger} & \statcell{+3.97}{0.30}{\dagger} \\
 & B & R$-$T & \statcell{+11.92}{0.95}{\dagger} & \statcell{+12.89}{0.85}{\dagger} & \statcell{+12.73}{0.72}{\dagger} & \statcell{+8.41}{2.47}{\dagger} & \statcell{+6.33}{0.95}{\dagger} & \statcell{+4.89}{0.64}{\dagger} & \statcell{+3.86}{0.09}{\dagger} \\
 & H & R+T$-$T & \statcell{+0.32}{1.46}{} & \statcell{+0.76}{1.55}{} & \statcell{+2.16}{1.19}{} & \statcell{+3.06}{1.06}{\dagger} & \statcell{+0.92}{0.55}{} & \statcell{+0.12}{0.21}{} & \statcell{-0.06}{0.19}{} \\
 & H & R$-$T & \statcell{-1.66}{1.33}{} & \statcell{-0.74}{1.44}{} & \statcell{+0.45}{1.04}{} & \statcell{+2.97}{0.96}{\dagger} & \statcell{+0.67}{0.48}{} & \statcell{-0.22}{0.17}{} & \statcell{-0.36}{0.08}{} \\
\midrule
LCB Pre & B & R+T$-$T & \statcell{-0.82}{1.74}{} & \statcell{-0.25}{1.01}{} & \statcell{+0.91}{0.56}{} & \statcell{+0.34}{0.68}{} & \statcell{+0.23}{0.07}{} & \statcell{+0.11}{0.28}{} & \statcell{-0.17}{0.39}{} \\
 & B & R$-$T & \statcell{-5.12}{4.52}{} & \statcell{-4.77}{2.22}{} & \statcell{-3.22}{1.01}{} & \statcell{-1.86}{0.90}{} & \statcell{-0.85}{0.73}{} & \statcell{-0.83}{0.30}{} & \statcell{-0.80}{1.09}{} \\
 & H & R+T$-$T & \statcell{+2.56}{3.81}{} & \statcell{+2.74}{3.96}{} & \statcell{+2.97}{3.67}{} & \statcell{+0.22}{0.37}{} & \statcell{+0.00}{0.80}{} & \statcell{+0.47}{0.87}{} & \statcell{+0.55}{0.81}{} \\
 & H & R$-$T & \statcell{+1.97}{3.87}{} & \statcell{+1.90}{4.49}{} & \statcell{+1.97}{4.53}{} & \statcell{+0.36}{0.62}{} & \statcell{+0.25}{0.38}{} & \statcell{+0.29}{1.00}{} & \statcell{+0.49}{0.87}{} \\
\midrule
LCB Post & B & R+T$-$T & \statcell{+3.70}{1.58}{\dagger} & \statcell{+4.92}{1.37}{\dagger} & \statcell{+5.16}{3.48}{\dagger} & \statcell{+1.14}{0.72}{} & \statcell{+0.72}{0.27}{} & \statcell{+0.42}{0.21}{} & \statcell{+0.22}{0.24}{} \\
 & B & R$-$T & \statcell{+0.73}{1.91}{} & \statcell{+2.12}{1.04}{} & \statcell{+4.72}{3.31}{\dagger} & \statcell{+1.05}{0.45}{} & \statcell{+0.36}{0.18}{} & \statcell{-0.10}{0.19}{} & \statcell{-0.46}{0.21}{} \\
 & H & R+T$-$T & \statcell{+2.42}{5.04}{} & \statcell{+2.82}{2.37}{} & \statcell{-0.51}{0.72}{} & \statcell{+0.03}{0.19}{} & \statcell{+0.24}{0.27}{} & \statcell{+0.11}{0.18}{} & \statcell{+0.18}{0.36}{} \\
 & H & R$-$T & \statcell{+2.18}{3.47}{} & \statcell{+0.29}{2.12}{} & \statcell{-1.82}{1.04}{} & \statcell{-0.30}{0.21}{} & \statcell{-0.02}{0.17}{} & \statcell{+0.08}{0.26}{} & \statcell{+0.04}{0.46}{} \\
\midrule
OLMo3 & B & R+T$-$T & \statcell{+0.46}{0.69}{} & \statcell{+1.21}{3.83}{} & \statcell{+2.89}{2.23}{} & \statcell{+3.52}{2.22}{\dagger} & \statcell{+1.30}{1.19}{} & \statcell{+0.45}{0.48}{} & \statcell{+0.53}{0.49}{} \\
 & H & R+T$-$T & \statcell{+0.38}{1.07}{} & \statcell{+1.12}{2.41}{} & \statcell{+2.01}{3.79}{} & \statcell{+5.37}{3.43}{\dagger} & \statcell{+1.15}{1.83}{} & \statcell{+0.15}{0.56}{} & \statcell{+0.01}{0.15}{} \\
\end{longtable}}

\subsection{BELT probe capacity and input subsets}
\label{sec:probe-controls}
\ecoparagraph{BELT probe capacity.}
\Cref{tab:probe-controls} compares the compact probe with a top-two-layer reasoner branch under \emph{Now}/\emph{End} and \emph{Now}/\emph{End}/\emph{Later} supervision. At $\lambda=10^{-5}$ they are close under both target sets; differences vary by price and target set. These belief-input comparisons jointly vary architecture, readout and parameter count, and include a hardware deviation from the archived branch experiment: the branch was retrained and scored on RTX 4090, the compact random-init probes on GH200.
\begin{table}[!ht]
\caption{Utility comparing BELT probe architectures with belief inputs. Reporting convention follows \cref{tab:belief-utility}.}
\label{tab:probe-controls}
\begin{center}
\begingroup\setlength{\tabcolsep}{3pt}
\begin{tabular}{@{}>{\raggedright\arraybackslash}p{83pt}ccccc@{}}\toprule
Architecture / prediction targets & \multicolumn{5}{c}{Utility by $\lambda$ ($\times10^{-6}$)} \\
 & 2 & 5 & 10 & 20 & 50 \\\midrule
BELT probe / \emph{Now}, \emph{End} & \cellcolor{Green3!60}\beststatcell{0.7451}{0.0005}{}& \cellcolor{Green3!60}\beststatcell{0.7307}{0.0005}{}& \cellcolor{Green3!60}\beststatcell{0.7067}{0.0006}{}& \cellcolor{Green3!40}\secondstatcell{0.6586}{0.0008}{}& \cellcolor{Green3!0}\statcell{0.5820}{0.0021}{} \\
Top-two branch / \emph{Now}, \emph{End} & \cellcolor{Green3!0}\statcell{0.7447}{0.0007}{}& \cellcolor{Green3!0}\statcell{0.7301}{0.0004}{}& \cellcolor{Green3!0}\statcell{0.7061}{0.0003}{}& \cellcolor{Green3!60}\beststatcell{0.6589}{0.0014}{}& \cellcolor{Green3!56}\secondstatcell{0.5851}{0.0027}{} \\
BELT probe / \emph{Now}, \emph{End}, \emph{Later} & \cellcolor{Green3!15}\secondstatcell{0.7448}{0.0002}{}& \cellcolor{Green3!30}\secondstatcell{0.7304}{0.0003}{}& \cellcolor{Green3!40}\secondstatcell{0.7065}{0.0004}{}& \cellcolor{Green3!33}\statcell{0.6585}{0.0007}{}& \cellcolor{Green3!60}\beststatcell{0.5853}{0.0043}{} \\
Top-two branch / \emph{Now}, \emph{End}, \emph{Later} & \cellcolor{Green3!15}\secondstatcell{0.7448}{0.0008}{}& \cellcolor{Green3!20}\statcell{0.7303}{0.0008}{}& \cellcolor{Green3!10}\statcell{0.7062}{0.0008}{}& \cellcolor{Green3!0}\statcell{0.6580}{0.0009}{}& \cellcolor{Green3!25}\statcell{0.5834}{0.0030}{} \\
\bottomrule\end{tabular}\endgroup
\end{center}
\end{table}

\ecoparagraph{Combining beliefs with observed length.}
With the four-target probe fixed and the linear policy refit (\cref{tab:input-subsets}), adding prefix length to \emph{Now} raises mean utility at $\lambda=10^{-5}$ from 0.6766 to 0.7050, and further beliefs raise it to 0.7072 for the complete input set. The additional benefit of the beliefs depends on the token price and the input combination.
\begin{table}[!ht]
\caption{Utility comparing policy input subsets. Reporting convention follows \cref{tab:belief-utility}. The last row is the complete BELT-B-R input set.}
\label{tab:input-subsets}
\begin{center}
\begingroup\setlength{\tabcolsep}{3pt}
\fittable{%
\begin{tabular}{@{}lccccc@{}}\toprule
Policy inputs & \multicolumn{5}{c}{Utility by $\lambda$ ($\times10^{-6}$)} \\
 & 2 & 5 & 10 & 20 & 50 \\\midrule
\emph{Now} only & \cellcolor{Green3!0}\statcell{0.7163}{0.0010}{}& \cellcolor{Green3!0}\statcell{0.7001}{0.0018}{}& \cellcolor{Green3!0}\statcell{0.6766}{0.0004}{}& \cellcolor{Green3!0}\statcell{0.6241}{0.0025}{}& \cellcolor{Green3!0}\statcell{0.4927}{0.0048}{} \\
\emph{Now} + length & \cellcolor{Green3!57}\statcell{0.7440}{0.0008}{}& \cellcolor{Green3!57}\statcell{0.7295}{0.0007}{}& \cellcolor{Green3!56}\statcell{0.7050}{0.0004}{}& \cellcolor{Green3!47}\statcell{0.6569}{0.0007}{}& \cellcolor{Green3!39}\statcell{0.5535}{0.0041}{} \\
\emph{End} + length & \cellcolor{Green3!56}\statcell{0.7437}{0.0014}{}& \cellcolor{Green3!57}\statcell{0.7293}{0.0011}{}& \cellcolor{Green3!56}\statcell{0.7052}{0.0005}{}& \cellcolor{Green3!48}\statcell{0.6573}{0.0002}{}& \cellcolor{Green3!45}\statcell{0.5616}{0.0010}{} \\
\emph{Now}, \emph{Same} + length & \cellcolor{Green3!59}\secondstatcell{0.7451}{0.0002}{}& \cellcolor{Green3!59}\secondstatcell{0.7306}{0.0002}{}& \cellcolor{Green3!59}\secondstatcell{0.7065}{0.0005}{}& \cellcolor{Green3!60}\secondstatcell{0.6657}{0.0021}{}& \cellcolor{Green3!57}\secondstatcell{0.5807}{0.0075}{} \\
\emph{Same} + length & \cellcolor{Green3!59}\statcell{0.7448}{0.0000}{}& \cellcolor{Green3!58}\statcell{0.7302}{0.0000}{}& \cellcolor{Green3!57}\statcell{0.7058}{0.0001}{}& \cellcolor{Green3!54}\statcell{0.6620}{0.0031}{}& \cellcolor{Green3!45}\statcell{0.5627}{0.0027}{} \\
\emph{End}, \emph{Later} + length & \cellcolor{Green3!56}\statcell{0.7435}{0.0006}{}& \cellcolor{Green3!56}\statcell{0.7290}{0.0006}{}& \cellcolor{Green3!56}\statcell{0.7050}{0.0006}{}& \cellcolor{Green3!47}\statcell{0.6569}{0.0006}{}& \cellcolor{Green3!45}\statcell{0.5617}{0.0017}{} \\
\emph{Now}, \emph{End}, \emph{Later}, \emph{Same} + length & \cellcolor{Green3!60}\beststatcell{0.7454}{0.0006}{}& \cellcolor{Green3!60}\beststatcell{0.7310}{0.0005}{}& \cellcolor{Green3!60}\beststatcell{0.7072}{0.0006}{}& \cellcolor{Green3!60}\beststatcell{0.6660}{0.0024}{}& \cellcolor{Green3!60}\beststatcell{0.5851}{0.0042}{} \\
\bottomrule\end{tabular}}\endgroup
\end{center}
\end{table}

BELT exceeds a denser fixed cap, which takes the forced answer at every probe boundary, by $+24.63$\,pp of Qwen3 OOD saving at 1\,pp and $+0.0291$ utility at $\lambda=2\times10^{-5}$, on every seed (\cref{tab:reb-dense-cap}).

\ecoparagraph{Parsimonious controls across settings.}
\Cref{tab:reb-parsimonious-controls} places the \emph{Now} + length control and the dense cap next to BELT and OSP in every setting. Against the \emph{Now} + length control, BELT saves 13.29~pp more at a one-point tolerance on Qwen3 OOD, 17.49~pp more on ID and 9.21~pp more on OLMo3, on all three seeds. Its development-selected utility at $\lambda=2\times10^{-5}$ is higher by 0.0120 (OOD) and 0.0187 (ID).
\begin{table}[!ht]
\centering\small
\caption{Parsimonious input controls next to BELT and OSP (random-initialized probes; OLMo3 has one probe family). Save 1 = token saving (\%) at a 1\,pp accuracy tolerance on the test frontier (Qwen3 and OLMo3 macro, LiveCodeBench single dataset); $U$ = development-selected utility at price $\lambda$ (micro), the protocol of \cref{tab:input-subsets,tab:source-transfer}. Mean over three seeds; the dense cap is deterministic. All four + length is BELT-B-R (on OLMo3 without the length input); the OLMo3 dense cap row is the union of the dense block caps and the main token caps (\cref{tab:reb-dense-cap}).}
\label{tab:reb-parsimonious-controls}
\begingroup\setlength{\tabcolsep}{3pt}
\begin{tabular}{@{}llcccccc@{}}\toprule
Setting & Policy & Save 1 & $U$ 2e-6 & $U$ 5e-6 & $U$ 1e-5 & $U$ 2e-5 & $U$ 5e-5 \\\midrule
Qwen3 OOD & \emph{Now} + length & \statcell{31.96}{}{} & \statcell{0.7440}{}{} & \statcell{0.7295}{}{} & \statcell{0.7050}{}{} & \statcell{0.6569}{}{} & \statcell{0.5535}{}{} \\
 & \emph{Now}, \emph{Same} + length & \statcell{41.01}{}{} & \statcell{0.7451}{}{} & \statcell{0.7306}{}{} & \statcell{0.7065}{}{} & \statcell{0.6657}{}{} & \statcell{0.5807}{}{} \\
 & All four + length (B-R) & \statcell{43.76}{}{} & \statcell{0.7454}{}{} & \statcell{0.7310}{}{} & \statcell{0.7072}{}{} & \statcell{0.6660}{}{} & \statcell{0.5851}{}{} \\
 & Dense forced cap & \statcell{20.62}{}{} & \statcell{0.7260}{}{} & \statcell{0.7118}{}{} & \statcell{0.6880}{}{} & \statcell{0.6398}{}{} & \statcell{0.5138}{}{} \\
 & BELT-H-R & \statcell{45.25}{}{} & \statcell{0.7352}{}{} & \statcell{0.7227}{}{} & \statcell{0.7024}{}{} & \statcell{0.6689}{}{} & \statcell{0.5861}{}{} \\
 & OSP & \statcell{45.19}{}{} & \statcell{0.7270}{}{} & \statcell{0.7177}{}{} & \statcell{0.7022}{}{} & \statcell{0.6713}{}{} & \statcell{0.5914}{}{} \\
 & OSP-1 & \statcell{36.89}{}{} & \statcell{0.7205}{}{} & \statcell{0.7098}{}{} & \statcell{0.6915}{}{} & \statcell{0.6563}{}{} & \statcell{0.5802}{}{} \\
\midrule
Qwen3 ID & \emph{Now} + length & \statcell{31.08}{}{} & \statcell{0.5597}{}{} & \statcell{0.5408}{}{} & \statcell{0.5092}{}{} & \statcell{0.4483}{}{} & \statcell{0.3112}{}{} \\
 & \emph{Now}, \emph{Same} + length & \statcell{44.15}{}{} & \statcell{0.5616}{}{} & \statcell{0.5427}{}{} & \statcell{0.5113}{}{} & \statcell{0.4635}{}{} & \statcell{0.3524}{}{} \\
 & All four + length (B-R) & \statcell{48.74}{}{} & \statcell{0.5619}{}{} & \statcell{0.5430}{}{} & \statcell{0.5116}{}{} & \statcell{0.4654}{}{} & \statcell{0.3614}{}{} \\
 & Dense forced cap & \statcell{16.57}{}{} & \statcell{0.5340}{}{} & \statcell{0.5153}{}{} & \statcell{0.4841}{}{} & \statcell{0.4241}{}{} & \statcell{0.2587}{}{} \\
 & BELT-H-R & \statcell{48.58}{}{} & \statcell{0.5513}{}{} & \statcell{0.5356}{}{} & \statcell{0.5109}{}{} & \statcell{0.4670}{}{} & \statcell{0.3610}{}{} \\
 & OSP & \statcell{49.72}{}{} & \statcell{0.5464}{}{} & \statcell{0.5340}{}{} & \statcell{0.5133}{}{} & \statcell{0.4719}{}{} & \statcell{0.3698}{}{} \\
 & OSP-1 & \statcell{46.59}{}{} & \statcell{0.5363}{}{} & \statcell{0.5238}{}{} & \statcell{0.4989}{}{} & \statcell{0.4552}{}{} & \statcell{0.3584}{}{} \\
\midrule
OLMo3 & \emph{Now} + length & \statcell{46.00}{}{} & \statcell{0.8464}{}{} & \statcell{0.8271}{}{} & \statcell{0.7929}{}{} & \statcell{0.7457}{}{} & \statcell{0.6210}{}{} \\
 & \emph{Now}, \emph{Same} + length & \statcell{54.28}{}{} & \statcell{0.8408}{}{} & \statcell{0.8228}{}{} & \statcell{0.8016}{}{} & \statcell{0.7581}{}{} & \statcell{0.6597}{}{} \\
 & All four + length (B-R) & \statcell{54.30}{}{} & \statcell{0.8406}{}{} & \statcell{0.8280}{}{} & \statcell{0.8053}{}{} & \statcell{0.7561}{}{} & \statcell{0.6633}{}{} \\
 & Dense forced cap & \statcell{18.37}{}{} & \statcell{0.8225}{}{} & \statcell{0.8032}{}{} & \statcell{0.7710}{}{} & \statcell{0.7033}{}{} & \statcell{0.5616}{}{} \\
 & BELT-H-R & \statcell{55.22}{}{} & \statcell{0.8407}{}{} & \statcell{0.8246}{}{} & \statcell{0.8009}{}{} & \statcell{0.7558}{}{} & \statcell{0.6520}{}{} \\
 & OSP & \statcell{49.86}{}{} & \statcell{0.8380}{}{} & \statcell{0.8251}{}{} & \statcell{0.8037}{}{} & \statcell{0.7553}{}{} & \statcell{0.6454}{}{} \\
 & OSP-1 & \statcell{47.54}{}{} & \statcell{0.8172}{}{} & \statcell{0.8030}{}{} & \statcell{0.7781}{}{} & \statcell{0.7389}{}{} & \statcell{0.6476}{}{} \\
\midrule
LiveCodeBench Pre & \emph{Now} + length & \statcell{46.96}{}{} & \statcell{0.6714}{}{} & \statcell{0.6516}{}{} & \statcell{0.6193}{}{} & \statcell{0.5455}{}{} & \statcell{0.4048}{}{} \\
 & \emph{Now}, \emph{Same} + length & \statcell{46.15}{}{} & \statcell{0.6711}{}{} & \statcell{0.6512}{}{} & \statcell{0.6179}{}{} & \statcell{0.5408}{}{} & \statcell{0.4183}{}{} \\
 & All four + length (B-R) & \statcell{42.87}{}{} & \statcell{0.6699}{}{} & \statcell{0.6502}{}{} & \statcell{0.6179}{}{} & \statcell{0.5325}{}{} & \statcell{0.4064}{}{} \\
 & Dense forced cap & \statcell{46.77}{}{} & \statcell{0.6488}{}{} & \statcell{0.6311}{}{} & \statcell{0.6017}{}{} & \statcell{0.5464}{}{} & \statcell{0.4134}{}{} \\
 & BELT-H-R & \statcell{40.15}{}{} & \statcell{0.6308}{}{} & \statcell{0.6148}{}{} & \statcell{0.5795}{}{} & \statcell{0.5332}{}{} & \statcell{0.4206}{}{} \\
 & OSP & \statcell{38.25}{}{} & \statcell{0.6287}{}{} & \statcell{0.6146}{}{} & \statcell{0.5911}{}{} & \statcell{0.5440}{}{} & \statcell{0.4152}{}{} \\
 & OSP-1 & \statcell{15.25}{}{} & \statcell{0.6099}{}{} & \statcell{0.5878}{}{} & \statcell{0.5569}{}{} & \statcell{0.4972}{}{} & \statcell{0.3970}{}{} \\
\midrule
LiveCodeBench Post & \emph{Now} + length & \statcell{54.66}{}{} & \statcell{0.3074}{}{} & \statcell{0.2789}{}{} & \statcell{0.2325}{}{} & \statcell{0.1308}{}{} & \statcell{-0.0286}{}{} \\
 & \emph{Now}, \emph{Same} + length & \statcell{62.35}{}{} & \statcell{0.3088}{}{} & \statcell{0.2800}{}{} & \statcell{0.2319}{}{} & \statcell{0.1431}{}{} & \statcell{0.0285}{}{} \\
 & All four + length (B-R) & \statcell{61.21}{}{} & \statcell{0.3073}{}{} & \statcell{0.2791}{}{} & \statcell{0.2316}{}{} & \statcell{0.1506}{}{} & \statcell{0.0583}{}{} \\
 & Dense forced cap & \statcell{53.90}{}{} & \statcell{0.3001}{}{} & \statcell{0.2748}{}{} & \statcell{0.2325}{}{} & \statcell{0.1507}{}{} & \statcell{-0.0242}{}{} \\
 & BELT-H-R & \statcell{58.76}{}{} & \statcell{0.2857}{}{} & \statcell{0.2656}{}{} & \statcell{0.2335}{}{} & \statcell{0.1899}{}{} & \statcell{0.1036}{}{} \\
 & OSP & \statcell{62.38}{}{} & \statcell{0.2901}{}{} & \statcell{0.2745}{}{} & \statcell{0.2485}{}{} & \statcell{0.1966}{}{} & \statcell{0.0970}{}{} \\
 & OSP-1 & \statcell{53.92}{}{} & \statcell{0.2672}{}{} & \statcell{0.2416}{}{} & \statcell{0.2002}{}{} & \statcell{0.1331}{}{} & \statcell{0.0766}{}{} \\
\bottomrule\end{tabular}\endgroup
\end{table}

\ecoparagraph{Utility across token prices.} On Qwen3 OOD, BELT's development-selected utility is above that of \emph{Now} + length at all five prices ($+0.0014$ to $+0.0316$), with the same sign on every seed at $\lambda=5\times10^{-6}$, $\lambda=1\times10^{-5}$, $\lambda=2\times10^{-5}$, $\lambda=5\times10^{-5}$.

\ecoparagraph{LiveCodeBench Pre.} Transferred from math, BELT is competitive with \emph{Now} + length rather than ahead of it: at 1\,pp its saving is $-6.81$\,pp relative to \emph{Now} + length, lower on every seed (ReProbe-initialized family $-3.46$\,pp).
Among the learned controllers BELT leads on Pre: it is above OSP at 6 of 6 jointly feasible constrained readings and above OSP-1 at 12 of 12 readings, and it saves 9.39\,pp more than full OSP at 1\,pp, ahead on three of three seed pairs (\cref{tab:reb-bootstrap}).

\ecoparagraph{LiveCodeBench Post.} At 1\,pp BELT saves 4.10\,pp more than \emph{Now} + length (ahead on two of three seeds), and at $\lambda=2\times10^{-5}$ its development-selected utility (0.1899) is above both \emph{Now} + length and the dense forced cap (0.1507, the stronger of the two at this price) on every seed (mean margin over the per-seed stronger control $+0.0392$).

\ecoparagraph{OLMo3 with two beliefs.} On OLMo3, BELT is competitive with the belief control \emph{Now}, \emph{Same} + length: $+0.94$\,pp at 1\,pp (positive on two of three seeds) and $-0.0023$ in utility at $\lambda=2\times10^{-5}$.
BELT leads it on every seed at the 30\% token budget ($+6.51$\,pp), and leads the parsimonious \emph{Now} + length by $+9.21$\,pp at 1\,pp on every seed (utility margin at the same price $+0.0101$, positive on two of three seeds).

\begin{table}[!ht]
\centering\small
\caption{Fixed caps with the forced answer at the cap. Main = the five block caps of the main tables (OLMo3: eight thinking-token caps); dense = a cap at every probe boundary (block $4k$), so the main Qwen3 and LiveCodeBench caps (16, 32, 64, 96, 128 blocks) are members of the dense grid; OLMo3 dense $\cup$ main adds the token caps. Save $k$ = token saving (\%) at a $k$\,pp tolerance (test frontier); $U$ = development-selected utility. Deterministic.}
\label{tab:reb-dense-cap}
\begingroup\setlength{\tabcolsep}{3pt}
\begin{tabular}{@{}llcccccc@{}}\toprule
Setting & Cap family & Save 0 & Save 1 & Save 2 & $U$ 1e-5 & $U$ 2e-5 & $U$ 5e-5 \\\midrule
Qwen3 ID & Main & \statcell{0.00}{}{} & \statcell{8.41}{}{} & \statcell{16.82}{}{} & \statcell{0.4735}{}{} & \statcell{0.4079}{}{} & \statcell{0.2684}{}{} \\
 & Dense & \statcell{11.73}{}{} & \statcell{16.57}{}{} & \statcell{21.26}{}{} & \statcell{0.4841}{}{} & \statcell{0.4241}{}{} & \statcell{0.2587}{}{} \\
\midrule
Qwen3 OOD & Main & \statcell{0.00}{}{} & \statcell{13.71}{}{} & \statcell{27.23}{}{} & \statcell{0.6771}{}{} & \statcell{0.6269}{}{} & \statcell{0.5196}{}{} \\
 & Dense & \statcell{15.14}{}{} & \statcell{20.62}{}{} & \statcell{27.25}{}{} & \statcell{0.6880}{}{} & \statcell{0.6398}{}{} & \statcell{0.5138}{}{} \\
\midrule
OLMo3 & Main & \statcell{0.00}{}{} & \statcell{15.10}{}{} & \statcell{26.90}{}{} & \statcell{0.7536}{}{} & \statcell{0.7033}{}{} & \statcell{0.5616}{}{} \\
 & Dense & \statcell{15.14}{}{} & \statcell{18.37}{}{} & \statcell{21.77}{}{} & \statcell{0.7710}{}{} & \statcell{0.7066}{}{} & \statcell{0.5284}{}{} \\
 & Dense $\cup$ main & \statcell{15.14}{}{} & \statcell{18.37}{}{} & \statcell{26.11}{}{} & \statcell{0.7710}{}{} & \statcell{0.7033}{}{} & \statcell{0.5616}{}{} \\
\midrule
LiveCodeBench Pre & Main & \statcell{41.00}{}{} & \statcell{46.76}{}{} & \statcell{50.76}{}{} & \statcell{0.5579}{}{} & \statcell{0.4905}{}{} & \statcell{0.4359}{}{} \\
 & Dense & \statcell{44.10}{}{} & \statcell{46.77}{}{} & \statcell{53.18}{}{} & \statcell{0.6017}{}{} & \statcell{0.5464}{}{} & \statcell{0.4134}{}{} \\
\midrule
LiveCodeBench Post & Main & \statcell{49.02}{}{} & \statcell{53.33}{}{} & \statcell{57.63}{}{} & \statcell{0.1903}{}{} & \statcell{0.0940}{}{} & \statcell{0.0411}{}{} \\
 & Dense & \statcell{49.02}{}{} & \statcell{53.90}{}{} & \statcell{59.97}{}{} & \statcell{0.2325}{}{} & \statcell{0.1507}{}{} & \statcell{-0.0242}{}{} \\
\bottomrule\end{tabular}\endgroup
\end{table}
On Qwen3 OOD the dense grid raises the cap's 1\,pp saving from 13.71\% to 20.62\% and its 0\,pp saving from 0.00\% to 15.14\%; BELT exceeds the dense cap at every constrained reading and price on every seed (1\,pp: $+24.63$\,pp).

\FloatBarrier

\subsection{Information provided to the controller by the four beliefs}
\label{sec:belief-details}
\ecoparagraph{Weights and stopping regions.}
In \cref{fig:belief-roles}a, normalized coefficients from probe outputs and refitted policies change with price: length dominates at the lowest price, and positive \emph{Same} and negative \emph{End} gain relative weight at higher prices. Across three probe runs, the mean directional threshold falls from 2.16 to $-0.33$; direction and threshold change simultaneously. Panels b and c show stop probabilities from one representative run on the OOD reporting bank.

\ecoparagraph{Coefficient interpretation.}
The coefficient study uses the 160-question fitting and development splits with $\kappa=10^{-4}$ fixed across probe training runs and prices. Inputs follow \cref{eq:policy}: belief logits and prefix length are standardized on fitting boundaries, with one transformation across prices within each training run, so $x_j=(\operatorname{logit}(\operatorname{clip}(p_j))-\mu_j)/\sigma_j$ for belief $j$, and a unit increase in $x_j$ changes stopping log-odds by $w_j$ with other inputs fixed. Correlations are 0.82--0.98 for $\pnow,\psame$, 0.95--0.97 for $\pend,\plater$, and $-0.59$ to $-0.52$ for $\pend$ and prefix length, so coefficients describe conditional roles rather than individual input contributions.

For $\lVert w_\lambda\rVert_2>0$, we separate the direction, threshold, and scale of each fitted policy by writing
\begin{equation}
\begin{aligned}
\pi_i&=\sigma\!\left(a_\lambda[\hat w_\lambda^\top x_i-t_\lambda]\right),\\
\hat w_\lambda&=w_\lambda/\lVert w_\lambda\rVert_2,\qquad
 t_\lambda=-b_\lambda/\lVert w_\lambda\rVert_2,\qquad
 a_\lambda=\lVert w_\lambda\rVert_2.
\end{aligned}
\label{eq:policy-geometry}
\end{equation}
Here $\hat w_\lambda$ is the unit direction over all five input weights (excluding the intercept), $t_\lambda$ locates the 0.5 stop-probability boundary along it, and $a_\lambda$ controls the sharpness of the transition.

\begin{table}[t]

\caption{Policy geometry at three token prices; $\lambda$ is in $10^{-6}$ units. Mean$_{\pm\mathrm{SD}}$ over three BELT probe seeds; probe size 16M.}
\label{tab:policy-geometry}
\begin{center}
\setlength{\tabcolsep}{2pt}
\begin{tabular}{@{}lccc@{}}\toprule
$\lambda$ & 2.00 & 18.91 & 108.58 \\\midrule
Length $\hat w_d$ & \statcell{0.931}{0.021}{}& \statcell{0.155}{0.158}{}& \statcell{-0.116}{0.167}{} \\
Threshold $t_\lambda$ & \statcell{2.164}{0.047}{}& \statcell{0.993}{0.288}{}& \statcell{-0.326}{0.118}{} \\
Scale $a_\lambda$ & \statcell{6.233}{0.100}{}& \statcell{6.131}{0.940}{}& \statcell{7.774}{0.492}{} \\
\bottomrule\end{tabular}
\end{center}
\end{table}

\Cref{tab:policy-geometry} reports mean and sample SD after transforming each saved fit separately; these summaries do not define an averaged policy. The mean normalized length coefficient changes from $+0.931$ at the lowest price to $-0.116$ at the highest. Length is an observed covariate: cost already incurred is shared by stopping and continuing, while utility compares subsequent cost and answer outcomes; its coefficient describes the conditional role of observed length in the fitted policy.

\ecoparagraph{Heatmap construction.}
The heatmaps average full-policy stop probabilities at $\lambda=2\times10^{-6}$ and $1.085767\times10^{-4}$ over the 212,101 forced-answer boundaries of the OOD reporting bank (probe seed 0), binned on a $60\times60$ \emph{Now}--\emph{End} grid of standardized logits, with each boundary keeping its own $\plater,\psame,d$ values; gray contours interpolate $\log_{10}$ counts at levels 1, 2, and 3. Counts and colors are not weighted by question or policy survival; counts show observed support, not independent questions or policy visitation, and equal axis coordinates need not denote equal correctness probabilities. The histogram is diagnostic; operating-point results come from the separate test-frontier evaluation.

\Needspace{20\baselineskip}
\FloatBarrier

\subsection{Prediction quality across questions and positions}
\label{sec:prediction-quality}
\begin{wrapfigure}[16]{r}{0.47\textwidth}
\centering
\includegraphics[width=\linewidth]{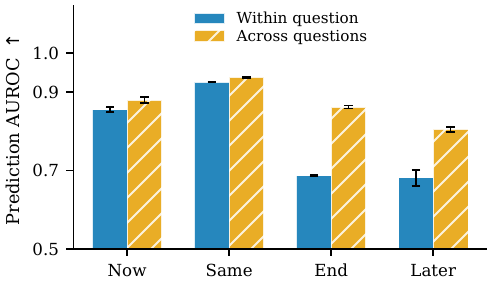}
\caption{Within- and across-question belief AUROC, shown as mean and sample SD over three independent BELT probe training runs.}
\label{fig:belief-prediction}
\end{wrapfigure}

\emph{Now} and \emph{Same} have within-question AUROCs of 0.855 and 0.926 on the source-selection diagnostic bank. \emph{End} and \emph{Later} distinguish boundaries from different questions more strongly than boundaries within the same question.

The bank has 1,065 questions (10,648 trajectories), distinct from the 320-question policy bank, with 385,277 labelled positions for $\pnow,\pend,\plater$ and 203,454 for $\psame$, which requires a parseable natural-completion answer after elicitation; an unparseable forced answer has label zero. Target-set comparisons share labelled boundaries within each belief. Within-question AUROC equally averages questions with both label classes; across-question AUROC compares positive/negative boundaries from different questions, not question means (\cref{fig:belief-prediction}). These diagnostics characterize outcome prediction; the policy-input ablations in \cref{sec:analysis-information} quantify contributions to stopping utility.
\FloatBarrier

\subsection{ReProbe initialization}
\label{sec:init-push-further}
Main-line BELT probes are randomly initialized. \Cref{tab:init-push-further} compares ReProbe initialization for the refit variants only, paired by probe seed over four settings and twelve readings. It raises H-R's OOD one-point saving from 45.25\% to 48.43\% ($+3.19$~pp, positive on all three seed pairs); across the 48 cells, H-R improves in 22 on all three paired seeds, and no cell worsens on all three. The three probe seeds share one ReProbe pretraining run (2.36~h on one four-GPU GH200 node), which is additional to the main configuration's cost.
\begin{table}[t]
\centering\small
\caption{Initialization pushes further. Seed-paired change from random to ReProbe initialization (ReProbe $-$ random, mean $\pm$ SD over three seed pairs; $^\dagger$ all three pairs share a sign). Saving in pp at a $k$~pp tolerance; accuracy in pp at a token budget; utility $\times10^{-4}$, development-selected. OLMo-3 is excluded: its ReProbe-initialized probes exist only at layer 24.}
\label{tab:init-push-further}
\begingroup\setlength{\tabcolsep}{3pt}
\begin{tabular}{@{}lcccccccc@{}}\toprule
 & \multicolumn{2}{c}{Qwen ID} & \multicolumn{2}{c}{Qwen OOD} & \multicolumn{2}{c}{LCB Pre} & \multicolumn{2}{c}{LCB Post} \\
Reading & H-R & B-R & H-R & B-R & H-R & B-R & H-R & B-R \\\midrule
Save 0 & $+1.6^\dagger$ & $+0.6$ & $+1.5^\dagger$ & $+0.4$ & $+2.8$ & $-0.2$ & $+2.2$ & $-2.3$ \\
Save 1 & $+1.6^\dagger$ & $+1.9^\dagger$ & $+3.2^\dagger$ & $+1.1^\dagger$ & $+3.2^\dagger$ & $+0.4$ & $+2.8$ & $-1.7$ \\
Save 2 & $+2.2^\dagger$ & $+1.8^\dagger$ & $+3.7^\dagger$ & $+1.0$ & $+4.2^\dagger$ & $-0.1$ & $+3.6$ & $-2.1$ \\
Acc 0.3 & $+0.4^\dagger$ & $+0.7^\dagger$ & $+0.7$ & $+0.7^\dagger$ & $+1.5^\dagger$ & $+1.1^\dagger$ & $+0.5$ & $-0.5$ \\
Acc 0.4 & $+0.8^\dagger$ & $+0.2^\dagger$ & $+0.6^\dagger$ & $+0.2$ & $+1.3^\dagger$ & $+0.4$ & $+0.3$ & $-0.2$ \\
Acc 0.5 & $+0.5^\dagger$ & $+0.4^\dagger$ & $+0.3^\dagger$ & $+0.3^\dagger$ & $+1.3^\dagger$ & $+0.3$ & $+0.3$ & $-0.2$ \\
Acc 0.6 & $+0.2$ & $+0.1$ & $-0.0$ & $+0.1$ & $+0.6$ & $-0.2$ & $+0.3$ & $+0.1$ \\
$U$ 2e-06 & $+6.0$ & $-1.2$ & $-16.1$ & $-0.9$ & $+53.4$ & $-2.0$ & $+29.6^\dagger$ & $+0.2$ \\
$U$ 5e-06 & $+14.1^\dagger$ & $-1.0$ & $-16.2$ & $-0.7$ & $+39.1$ & $-2.7$ & $+5.0$ & $-2.2$ \\
$U$ 1e-05 & $+10.6^\dagger$ & $-0.8$ & $-18.7$ & $-2.8$ & $+110.5$ & $-9.1$ & $-26.6$ & $+0.4$ \\
$U$ 2e-05 & $+12.5$ & $-18.5$ & $-6.3$ & $-7.0$ & $+98.2$ & $-64.6$ & $+55.6$ & $-126.9$ \\
$U$ 5e-05 & $+74.1^\dagger$ & $+38.7$ & $+82.2^\dagger$ & $+21.7^\dagger$ & $+158.1^\dagger$ & $+74.9^\dagger$ & $-8.8$ & $-55.8$ \\
\bottomrule\end{tabular}\endgroup
\end{table}

\section{Controller Parameters and Computational Cost} \label{sec:cost-analysis}
FLOP estimates count a multiply--accumulate as two operations and omit normalization, activations, and embedding lookups.

\subsection{Inference arithmetic} \label{sec:cost-inference}
\ecoparagraph{Reasoner and controller inference.}
Let $L$ be the number of reasoner blocks, $d_q$ the total query-head width, $d_{\mathrm{kv}}$ the total key/value-head width, and $P_{\mathrm{blk}}$ the parameters per block. For Qwen3, $L=28$, $d_q=2{,}048$, $d_{\mathrm{kv}}=1{,}024$, and $P_{\mathrm{blk}}\approx50.3$M. With $P_{\mathrm{LLM}}=LP_{\mathrm{blk}}$ and an output embedding of approximately $0.31$B parameters, a decode step at context length $t$ costs approximately
\begin{equation}
  \mathcal F_{\mathrm{LLM}}(t)
  \simeq2(P_{\mathrm{LLM}}+P_{\mathrm{emb}})+4Ld_qt
  \simeq3.44\ \mathrm{GFLOPs}+2.29\times10^5t\ \mathrm{FLOPs}.
  \label{eq:cost-llm}
\end{equation}
The key/value cache has $2Ld_{\mathrm{kv}}=57{,}344$ entries per token; bytes depend on precision.

The BELT probe uses cached layer-16 states, width $d_{\mathrm p}=768$, and $L_{\mathrm p}=2$ causal Transformer layers. Of its $P_{\mathrm p}=15{,}957{,}765$ parameters, $P_{\mathrm{mat}}=15{,}335{,}424$ enter token-wise matrix multiplications; boundary readouts and the unused inherited classifier are excluded. With incremental attention over cached history, the leading per-token cost is
\begin{equation}
  \mathcal F_{\mathrm{BELT}}(t)\simeq2P_{\mathrm{mat}}+4L_{\mathrm p}d_{\mathrm p}t
  \simeq3.07\times10^7+6.14\times10^3t\ \mathrm{FLOPs}.
  \label{eq:cost-belt}
\end{equation}
Its weight term, attention term and $2L_{\mathrm p}d_{\mathrm p}=3{,}072$ cache entries per token are about $0.89\%$, $2.68\%$ and $5.36\%$ of the reasoner's. Four belief readouts add about $6{,}144$ FLOPs per boundary, and the BELT-B-R/H-R policy a 5/768-dimensional dot product and an intercept.

OSP adds two reasoner blocks and a stopping head; the original blocks still serve generation. Its approximate per-token cost is $2\cdot2P_{\mathrm{blk}}+4\cdot2d_qt\simeq2.01\times10^8+1.64\times10^4t$ FLOPs, and both its attention term and its $4d_{\mathrm{kv}}=4{,}096$ cache entries are $2/28=7.14\%$ of the reasoner's. A dense-attention counter, which counts full-sequence attention rather than the causal token-by-token arithmetic of \cref{eq:cost-llm,eq:cost-belt}, gives 85.5M FLOPs per banked token for the BELT probe and 347.6M for OSP, a ratio of approximately 4.1.

\subsection{Reference-schedule training estimates} \label{sec:cost-measured}
The archived reference schedules on one RTX 4090 estimate controller training at 10.0 GPU-hours for three BELT epochs and 50.9 GPU-hours for one seven-epoch OSP run, so BELT needs approximately $5.1\times$ fewer GPU-hours than one OSP checkpoint; five OSP runs, one per token price, give 254.5 GPU-hours (\cref{tab:cost-summary}). The BELT estimate extrapolates its first epoch to three epochs; the five-checkpoint OSP estimate assumes each run costs as much as the recorded $\lambda=10^{-5}$ run. The estimates cover 31,248 training and 2,554 validation trajectories, with batch sizes 8 for BELT and 3 for OSP. Frozen reasoner states are recomputed each epoch, reused only within a bounded store; permanent caching would change the comparison by removing this repeated frozen-prefix computation after initial extraction.

On OLMo3 GH200 GPUs, the three BELT probe runs took 8.5--10.4 GPU-hours each (mean 9.8) for three epochs on one GPU. OSP trains at world size 3; its measured GH200 rate of 3.41~s per global step (median of nine single-node segments) and 474~s of validation per epoch give 5.1~h per epoch, or approximately 107 GPU-hours for seven epochs and 534 for five checkpoints, one per token price. Later OSP epochs ran on other GPU types, so this estimate applies the GH200 rate to all seven epochs. BELT thus uses approximately $11\times$ fewer training GPU-hours than one OSP checkpoint and $55\times$ fewer than five. All estimates here include validation and exclude shared trajectory generation, feature export, and CPU policy fitting.

\subsection{Preference adaptation and parameter storage} \label{sec:cost-adaptation}
Both BELT variants use one belief training run and feature export; refit adds one CPU policy fit per token price, and threshold one fitted policy with scalar cutoffs. On Qwen3, $P_{\mathrm{policy}}=6$ or 769 for BELT-B-R/H-R. For $K$ deployed policies, refit stores $P_{\mathrm p}+K P_{\mathrm{policy}}$, threshold $P_{\mathrm p}+P_{\mathrm{policy}}$ plus $K$ scalar cutoffs, and full OSP $K P_{\mathrm{OSP}}$, excluding the shared reasoner and standardization statistics. Each request uses only its selected policy or checkpoint; OSP-1 uses one checkpoint with varied thresholds.

\begin{table}[!ht]
  \caption{Trainable controller parameters, excluding the frozen reasoner. OSP counts one two-block branch and its stopping head.} \label{tab:controller-parameters}
  \begin{center}
    \begin{tabular}{@{}lrr@{}}
      \toprule
      Backbone & BELT & OSP \\
      \midrule
      Qwen3 & 15,957,765 & 100,676,098 \\
      OLMo3 & 19,103,494 & 404,791,298 \\
      \bottomrule
    \end{tabular}
  \end{center}
\end{table}

\section{Additional Comparisons and Protocol Checks}
\label{sec:checks}
This appendix documents operating-point sensitivity, baseline implementations and evaluation-set checks for \cref{sec:main-results,sec:analysis}.

\subsection{Stability of test-frontier readings}
\label{sec:checks-reading}
\begin{table}[t]\centering
\caption{Split-half sensitivity of test-frontier readings. $n$: members per seed; med.: median shrinkage; max$|\cdot|$: largest absolute shrinkage; SD: median cross-reading SD, pooled over splits and seeds for LiveCodeBench. Tolerance and budget entries use percentage points; price entries use utility units. Summaries pool these protocols and are descriptive.}
\label{tab:cross-reading}
\begingroup\scriptsize\setlength{\tabcolsep}{2.5pt}\renewcommand{\arraystretch}{1.1}
\resizebox{\textwidth}{!}{\begin{tabular}{@{}l|ccccc|ccccc|ccccc@{}}\toprule
\multirow{2}{*}{\textbf{Method}} & \multicolumn{5}{c|}{\textbf{Qwen3 ID/OOD (2,719)}} & \multicolumn{5}{c|}{\textbf{OLMo3 OOD (1,025)}} & \multicolumn{5}{c}{\textbf{LiveCodeBench (549 + 416)}} \\
 & \shortstack{\textbf{Probe}\\\textbf{Params}} & $n$ & med. & max$|\cdot|$ & SD & \shortstack{\textbf{Probe}\\\textbf{Params}} & $n$ & med. & max$|\cdot|$ & SD & \shortstack{\textbf{Probe}\\\textbf{Params}} & $n$ & med. & max$|\cdot|$ & SD \\\midrule
Natural & -- & 1 & +0.000& 0.00& 0.01& -- & -- & -- & -- & -- & -- & -- & -- & -- & -- \\
Fixed cap & 0 & 5 & +0.000& 1.13& 1.07& 0 & 8 & +0.002& 1.20& 0.66& -- & -- & -- & -- & -- \\
CoDE-Stop & 0 & 225 & +0.001& 1.52& 1.07& 0 & 75 & +0.001& 0.79& 0.65& -- & -- & -- & -- & -- \\
DEER & 0 & 54 & +0.001& 1.31& 1.05& 0 & 18 & +0.000& 0.99& 0.61& -- & -- & -- & -- & -- \\
TC & 514 & 300 & +0.002& 0.86& 1.00& 514 & 150 & +0.002& 0.93& 1.14& -- & -- & -- & -- & -- \\
ROM & 5M & 21 & +0.002& 1.43& 1.03& 8M & 21 & +0.002& 2.54& 1.27& -- & -- & -- & -- & -- \\
OSP & 101M $\times 5$ & 5 & +0.000& 5.66& 1.03& 405M $\times 5$ & 5 & -0.000& 0.28& 1.00& 101M $\times 5$ & 5 & +0.011& 1.25& 2.28 \\
OSP-1 & 101M & 13 & +0.002& 7.53& 1.07& 405M & 13 & +0.002& 7.15& 1.02& 101M & 13 & +0.000& 4.02& 2.24 \\
BELT-B-T & -- & -- & -- & -- & -- & 19M & 13 & +0.001& 0.34& 0.98& -- & -- & -- & -- & -- \\
BELT-H-T & -- & -- & -- & -- & -- & 19M & 13 & +0.001& 0.22& 0.98& -- & -- & -- & -- & -- \\
\textbf{BELT-B-R} & 16M & 17 & +0.000& 0.36& 1.04& 19M & 221 & +0.002& 2.77& 0.96& 16M & 17 & +0.006& 0.38& 2.34 \\
\textbf{BELT-H-R} & 16M & 17 & +0.000& 0.20& 1.02& 19M & 221 & +0.003& 1.21& 0.96& 16M & 17 & +0.001& 0.55& 2.25 \\
\bottomrule\end{tabular}}\endgroup
\end{table}

Test-frontier operating points select members and mixture weights on test questions (\cref{sec:testcurve-reading}). We re-read them on 200 stratified half splits of each test population, selecting on one half and evaluating on the other in both directions; shrinkage is the whole-population reading minus the mean cross-reading (\cref{tab:cross-reading}). On Qwen3, maximum absolute shrinkage is 0.36~pp for BELT-B-R and 0.20~pp for BELT-H-R, whereas full OSP and OSP-1 reach 5.66 and 7.53~pp: BELT's operating points are stable under re-reading, whereas individual OSP and OSP-1 points shift by several points even where median shrinkage is small. The mean 0\,pp shifts of \cref{sec:cross-fitted} (\cref{tab:xfit-shift}) differ from these maxima by construction.

\subsection{Single-variant comparisons and paired bootstrap}
\label{sec:reb-single-variant}
\begin{table}[!ht]
\centering\small
\caption{BELT against full OSP and OSP-1 at a 1\,pp accuracy tolerance (token saving, \%; random-initialized probes; mean over three seeds). BELT is the recommended configuration, BELT-H-T (\cref{sec:main-results}); the other three variants are in \cref{tab:reb-single-variant-wtl,tab:reb-single-variant-margins}. Verdict: W/L when the difference of seed means exceeds the larger seed SD, else T. Interval: pooled question $\times$ seed 95\% bootstrap interval, with whether $|\Delta|>2$\,SE; shown for the pre-registered comparisons (\cref{tab:reb-bootstrap}).}
\label{tab:reb-belt-vs-baselines}
\begingroup\setlength{\tabcolsep}{3pt}
\begin{tabular}{@{}llccccl@{}}\toprule
Setting & Comparator & BELT & Comparator & $\Delta$ & Verdict & Pooled interval; 2\,SE \\\midrule
Qwen3 ID & full OSP & 50.64 & 49.72 & $+0.92$ & T & -- \\
Qwen3 ID & OSP-1 & 50.64 & 46.59 & $+4.05$ & W & $[-0.05,+9.67]$; no \\
Qwen3 OOD & full OSP & 45.98 & 45.19 & $+0.79$ & T & -- \\
Qwen3 OOD & OSP-1 & 45.98 & 36.89 & $+9.09$ & W & $[+3.30,+16.22]$; yes \\
LiveCodeBench Pre & full OSP & 38.25 & 38.25 & $-0.00$ & T & -- \\
LiveCodeBench Pre & OSP-1 & 38.25 & 15.25 & $+23.00$ & W & $[+10.23,+36.37]$; yes \\
LiveCodeBench Post & full OSP & 58.47 & 62.38 & $-3.91$ & T & -- \\
LiveCodeBench Post & OSP-1 & 58.47 & 53.92 & $+4.55$ & T & $[-9.78,+27.37]$; no \\
OLMo3 & full OSP & 54.09 & 49.86 & $+4.23$ & W & -- \\
OLMo3 & OSP-1 & 54.09 & 47.54 & $+6.55$ & W & $[+1.17,+15.57]$; no \\
\bottomrule\end{tabular}\endgroup
\end{table}

\ecoparagraph{Decision rule.} Each BELT variant is compared on its own with each comparator, and every count below concerns a single variant. A reading is a win (W) or loss (L) when the difference of seed means exceeds the larger of the two seed standard deviations, and a tie (T) otherwise. A W or L is robust (superscript r) when the verdict is unchanged after leaving out any single comparator seed or any single BELT seed in turn. NF = the reading is infeasible for one side (excluded). Readings: token saving at 0/1/2\,pp tolerance and accuracy at 30/40/50/60\% token budgets (token savings of 70/60/50/40\%). OSP+T adds 13 source-validation thresholds to every full-OSP checkpoint while the BELT variants in that column carry no threshold expansion (BELT refit plus thresholds is in \cref{tab:reb-refit-threshold}); on OLMo3 the refit variants are the registered 221-member family, which already includes per-direction thresholds (\cref{sec:olmo-replication}). Per-reading margins are in \cref{tab:reb-single-variant-margins}; the ReProbe-initialized family is in \cref{tab:reb-single-variant-wtl-fb}.

\begin{table}[!ht]
\centering\small
\caption{Single-variant win/tie/loss counts over the seven constrained readings per setting, random-initialized probes (main line). Cells: W/T/L with robust W/L counts as superscript. One row per variant; no count combines variants. OSP+T is not evaluated on the OLMo3 LiveCodeBench sets (--); its Total covers the other five settings.}
\label{tab:reb-single-variant-wtl}
\begingroup\setlength{\tabcolsep}{3pt}
\resizebox{\linewidth}{!}{\begin{tabular}{@{}llcccccccc@{}}\toprule
Variant & Comparator & Qwen3 ID & Qwen3 OOD & LCB Pre & LCB Post & OLMo3 & \shortstack{OLMo3 LCB\\before 8/2024} & \shortstack{OLMo3 LCB\\from 8/2024} & Total \\\midrule
H-R & full OSP & 1/1/5$^{1/2}$ & 0/6/1$^{0/0}$ & 1/5/0$^{0/0}$\,(NF 1) & 0/3/4$^{0/2}$ & 6/1/0$^{6/0}$ & 5/2/0$^{1/0}$ & 0/6/1$^{0/0}$ & 13/24/11$^{8/4}$ \\
 & OSP-1 & 4/3/0$^{4/0}$ & 5/2/0$^{5/0}$ & 7/0/0$^{7/0}$ & 3/4/0$^{1/0}$ & 7/0/0$^{6/0}$ & 7/0/0$^{5/0}$ & 5/2/0$^{3/0}$ & 38/11/0$^{31/0}$ \\
 & OSP+T & 0/3/4$^{0/4}$ & 0/1/6$^{0/3}$ & 0/5/2$^{0/1}$ & 0/0/7$^{0/7}$ & 5/2/0$^{4/0}$ & \missingcell & \missingcell & 5/11/19$^{4/15}$ \\
\midrule
H-T & full OSP & 1/5/1$^{0/1}$ & 1/4/2$^{1/2}$ & 0/6/0$^{0/0}$\,(NF 1) & 0/6/1$^{0/0}$ & 2/4/1$^{1/0}$ & 2/5/0$^{0/0}$ & 0/4/3$^{0/1}$ & 6/34/8$^{2/4}$ \\
 & OSP-1 & 6/0/1$^{6/1}$ & 5/0/2$^{5/1}$ & 7/0/0$^{7/0}$ & 2/5/0$^{0/0}$ & 5/1/1$^{3/0}$ & 7/0/0$^{2/0}$ & 3/4/0$^{2/0}$ & 35/10/4$^{25/2}$ \\
 & OSP+T & 0/6/1$^{0/1}$ & 0/3/4$^{0/3}$ & 0/5/2$^{0/1}$ & 0/0/7$^{0/1}$ & 1/5/1$^{0/0}$ & \missingcell & \missingcell & 1/19/15$^{0/6}$ \\
\midrule
B-R & full OSP & 0/4/3$^{0/2}$ & 0/6/1$^{0/0}$ & 5/1/0$^{3/0}$\,(NF 1) & 0/6/1$^{0/0}$ & 6/0/1$^{5/0}$ & 0/7/0$^{0/0}$ & 0/6/1$^{0/1}$ & 11/30/7$^{8/3}$ \\
 & OSP-1 & 4/3/0$^{3/0}$ & 6/0/1$^{5/1}$ & 7/0/0$^{7/0}$ & 5/2/0$^{5/0}$ & 6/1/0$^{6/0}$ & 4/3/0$^{1/0}$ & 3/4/0$^{2/0}$ & 35/13/1$^{29/1}$ \\
 & OSP+T & 0/3/4$^{0/3}$ & 0/1/6$^{0/4}$ & 2/4/1$^{2/1}$ & 0/1/6$^{0/2}$ & 5/1/1$^{4/0}$ & \missingcell & \missingcell & 7/10/18$^{6/10}$ \\
\midrule
B-T & full OSP & 0/0/7$^{0/7}$ & 0/1/6$^{0/6}$ & 6/0/0$^{6/0}$\,(NF 1) & 0/3/4$^{0/4}$ & 4/2/1$^{3/1}$ & 0/7/0$^{0/0}$ & 0/3/4$^{0/1}$ & 10/16/22$^{9/19}$ \\
 & OSP-1 & 0/0/7$^{0/7}$ & 1/0/6$^{0/6}$ & 7/0/0$^{7/0}$ & 5/0/2$^{3/2}$ & 5/1/1$^{5/1}$ & 2/5/0$^{0/0}$ & 3/4/0$^{2/0}$ & 23/10/16$^{17/16}$ \\
 & OSP+T & 0/0/7$^{0/7}$ & 0/0/7$^{0/7}$ & 7/0/0$^{6/0}$ & 0/1/6$^{0/6}$ & 2/4/1$^{0/1}$ & \missingcell & \missingcell & 9/5/21$^{6/21}$ \\
\bottomrule\end{tabular}}\endgroup
\end{table}

{\fontsize{8}{9}\selectfont\setlength{\tabcolsep}{1.5pt}
\begin{longtable}{@{}lllccccccc@{}}
\caption{Single-variant margins (BELT minus comparator, difference of seed means; pp) and verdicts, random-initialized probes. $^{\mathrm{r}}$ robust, $^{\mathrm{e}}$ threshold edge (verdict changes when one seed is left out).}\label{tab:reb-single-variant-margins}\\
\toprule Setting & Variant & Comparator & Save 0 & Save 1 & Save 2 & Acc 0.3 & Acc 0.4 & Acc 0.5 & Acc 0.6 \\\midrule\endfirsthead
\toprule Setting & Variant & Comparator & Save 0 & Save 1 & Save 2 & Acc 0.3 & Acc 0.4 & Acc 0.5 & Acc 0.6 \\\midrule\endhead
\bottomrule\endfoot
Qwen3 ID & H-R & full OSP & $-1.01$\,L$^{\mathrm{e}}$ & $-1.14$\,L$^{\mathrm{e}}$ & $-1.86$\,L$^{\mathrm{r}}$ & $+0.56$\,W$^{\mathrm{r}}$ & $+0.03$\,T & $-0.32$\,L$^{\mathrm{e}}$ & $-0.35$\,L$^{\mathrm{r}}$ \\
 & H-R & OSP-1 & $+4.58$\,T & $+1.98$\,T & $+0.79$\,T & $+0.66$\,W$^{\mathrm{r}}$ & $+0.55$\,W$^{\mathrm{r}}$ & $+0.79$\,W$^{\mathrm{r}}$ & $+0.86$\,W$^{\mathrm{r}}$ \\
 & H-R & OSP+T & $-1.82$\,L$^{\mathrm{r}}$ & $-1.20$\,T & $-2.54$\,L$^{\mathrm{r}}$ & $-0.07$\,T & $-0.37$\,L$^{\mathrm{r}}$ & $-0.28$\,T & $-0.52$\,L$^{\mathrm{r}}$ \\
 & H-T & full OSP & $+1.75$\,W$^{\mathrm{e}}$ & $+0.92$\,T & $+0.15$\,T & $-1.02$\,L$^{\mathrm{r}}$ & $+0.10$\,T & $+0.05$\,T & $+0.02$\,T \\
 & H-T & OSP-1 & $+7.34$\,W$^{\mathrm{r}}$ & $+4.05$\,W$^{\mathrm{r}}$ & $+2.80$\,W$^{\mathrm{r}}$ & $-0.92$\,L$^{\mathrm{r}}$ & $+0.62$\,W$^{\mathrm{r}}$ & $+1.16$\,W$^{\mathrm{r}}$ & $+1.23$\,W$^{\mathrm{r}}$ \\
 & H-T & OSP+T & $+0.95$\,T & $+0.87$\,T & $-0.53$\,T & $-1.65$\,L$^{\mathrm{r}}$ & $-0.30$\,T & $+0.09$\,T & $-0.15$\,T \\
 & B-R & full OSP & $-0.32$\,T & $-0.98$\,T & $-1.57$\,L$^{\mathrm{r}}$ & $-0.18$\,L$^{\mathrm{e}}$ & $+0.08$\,T & $-0.54$\,L$^{\mathrm{r}}$ & $-0.42$\,T \\
 & B-R & OSP-1 & $+5.27$\,T & $+2.15$\,W$^{\mathrm{e}}$ & $+1.08$\,T & $-0.08$\,T & $+0.60$\,W$^{\mathrm{r}}$ & $+0.57$\,W$^{\mathrm{r}}$ & $+0.79$\,W$^{\mathrm{r}}$ \\
 & B-R & OSP+T & $-1.13$\,T & $-1.04$\,T & $-2.25$\,L$^{\mathrm{r}}$ & $-0.81$\,L$^{\mathrm{r}}$ & $-0.32$\,L$^{\mathrm{e}}$ & $-0.49$\,T & $-0.59$\,L$^{\mathrm{r}}$ \\
 & B-T & full OSP & $-19.56$\,L$^{\mathrm{r}}$ & $-20.15$\,L$^{\mathrm{r}}$ & $-20.61$\,L$^{\mathrm{r}}$ & $-7.37$\,L$^{\mathrm{r}}$ & $-6.14$\,L$^{\mathrm{r}}$ & $-5.30$\,L$^{\mathrm{r}}$ & $-4.31$\,L$^{\mathrm{r}}$ \\
 & B-T & OSP-1 & $-13.97$\,L$^{\mathrm{r}}$ & $-17.03$\,L$^{\mathrm{r}}$ & $-17.96$\,L$^{\mathrm{r}}$ & $-7.27$\,L$^{\mathrm{r}}$ & $-5.61$\,L$^{\mathrm{r}}$ & $-4.19$\,L$^{\mathrm{r}}$ & $-3.11$\,L$^{\mathrm{r}}$ \\
 & B-T & OSP+T & $-20.36$\,L$^{\mathrm{r}}$ & $-20.21$\,L$^{\mathrm{r}}$ & $-21.28$\,L$^{\mathrm{r}}$ & $-8.00$\,L$^{\mathrm{r}}$ & $-6.53$\,L$^{\mathrm{r}}$ & $-5.26$\,L$^{\mathrm{r}}$ & $-4.48$\,L$^{\mathrm{r}}$ \\
\midrule
Qwen3 OOD & H-R & full OSP & $+13.20$\,T & $+0.05$\,T & $-0.41$\,T & $-0.44$\,T & $-0.17$\,T & $-0.34$\,L$^{\mathrm{e}}$ & $-0.03$\,T \\
 & H-R & OSP-1 & $+30.04$\,W$^{\mathrm{r}}$ & $+8.35$\,W$^{\mathrm{r}}$ & $+5.89$\,W$^{\mathrm{r}}$ & $-0.45$\,T & $+0.32$\,T & $+0.60$\,W$^{\mathrm{r}}$ & $+1.00$\,W$^{\mathrm{r}}$ \\
 & H-R & OSP+T & $-0.96$\,L$^{\mathrm{e}}$ & $-0.09$\,T & $-1.42$\,L$^{\mathrm{e}}$ & $-0.63$\,L$^{\mathrm{e}}$ & $-0.66$\,L$^{\mathrm{r}}$ & $-0.61$\,L$^{\mathrm{r}}$ & $-0.33$\,L$^{\mathrm{r}}$ \\
 & H-T & full OSP & $+14.86$\,T & $+0.79$\,T & $-0.86$\,T & $-3.41$\,L$^{\mathrm{r}}$ & $-0.84$\,L$^{\mathrm{r}}$ & $-0.13$\,T & $+0.32$\,W$^{\mathrm{r}}$ \\
 & H-T & OSP-1 & $+31.70$\,W$^{\mathrm{r}}$ & $+9.09$\,W$^{\mathrm{r}}$ & $+5.45$\,W$^{\mathrm{r}}$ & $-3.42$\,L$^{\mathrm{r}}$ & $-0.35$\,L$^{\mathrm{e}}$ & $+0.82$\,W$^{\mathrm{r}}$ & $+1.36$\,W$^{\mathrm{r}}$ \\
 & H-T & OSP+T & $+0.69$\,T & $+0.64$\,T & $-1.87$\,L$^{\mathrm{r}}$ & $-3.60$\,L$^{\mathrm{r}}$ & $-1.34$\,L$^{\mathrm{r}}$ & $-0.39$\,L$^{\mathrm{e}}$ & $+0.03$\,T \\
 & B-R & full OSP & $+11.22$\,T & $-1.43$\,T & $-1.93$\,T & $-0.49$\,T & $-0.06$\,T & $-0.32$\,L$^{\mathrm{e}}$ & $-0.11$\,T \\
 & B-R & OSP-1 & $+28.07$\,W$^{\mathrm{r}}$ & $+6.87$\,W$^{\mathrm{r}}$ & $+4.37$\,W$^{\mathrm{r}}$ & $-0.50$\,L$^{\mathrm{r}}$ & $+0.43$\,W$^{\mathrm{e}}$ & $+0.63$\,W$^{\mathrm{r}}$ & $+0.93$\,W$^{\mathrm{r}}$ \\
 & B-R & OSP+T & $-2.94$\,L$^{\mathrm{r}}$ & $-1.58$\,T & $-2.94$\,L$^{\mathrm{r}}$ & $-0.68$\,L$^{\mathrm{r}}$ & $-0.55$\,L$^{\mathrm{e}}$ & $-0.58$\,L$^{\mathrm{r}}$ & $-0.40$\,L$^{\mathrm{e}}$ \\
 & B-T & full OSP & $-0.70$\,T & $-14.32$\,L$^{\mathrm{r}}$ & $-14.66$\,L$^{\mathrm{r}}$ & $-8.90$\,L$^{\mathrm{r}}$ & $-6.39$\,L$^{\mathrm{r}}$ & $-5.21$\,L$^{\mathrm{r}}$ & $-3.96$\,L$^{\mathrm{r}}$ \\
 & B-T & OSP-1 & $+16.15$\,W$^{\mathrm{e}}$ & $-6.02$\,L$^{\mathrm{r}}$ & $-8.36$\,L$^{\mathrm{r}}$ & $-8.92$\,L$^{\mathrm{r}}$ & $-5.90$\,L$^{\mathrm{r}}$ & $-4.26$\,L$^{\mathrm{r}}$ & $-2.93$\,L$^{\mathrm{r}}$ \\
 & B-T & OSP+T & $-14.86$\,L$^{\mathrm{r}}$ & $-14.47$\,L$^{\mathrm{r}}$ & $-15.68$\,L$^{\mathrm{r}}$ & $-9.09$\,L$^{\mathrm{r}}$ & $-6.88$\,L$^{\mathrm{r}}$ & $-5.47$\,L$^{\mathrm{r}}$ & $-4.26$\,L$^{\mathrm{r}}$ \\
\midrule
LCB Pre & H-R & full OSP & $+0.61$\,T & $+1.90$\,T & $+2.45$\,T & \infeasiblecell & $+0.96$\,W$^{\mathrm{e}}$ & $+0.58$\,T & $+0.57$\,T \\
 & H-R & OSP-1 & $+35.34$\,W$^{\mathrm{r}}$ & $+24.90$\,W$^{\mathrm{r}}$ & $+20.48$\,W$^{\mathrm{r}}$ & $+1.78$\,W$^{\mathrm{r}}$ & $+1.72$\,W$^{\mathrm{r}}$ & $+2.21$\,W$^{\mathrm{r}}$ & $+2.83$\,W$^{\mathrm{r}}$ \\
 & H-R & OSP+T & $-1.41$\,T & $-1.20$\,T & $+0.46$\,T & $-1.46$\,L$^{\mathrm{r}}$ & $-0.82$\,L$^{\mathrm{e}}$ & $+0.00$\,T & $-0.39$\,T \\
 & H-T & full OSP & $-1.36$\,T & $-0.00$\,T & $+0.47$\,T & \infeasiblecell & $+0.71$\,T & $+0.28$\,T & $+0.09$\,T \\
 & H-T & OSP-1 & $+33.37$\,W$^{\mathrm{r}}$ & $+23.00$\,W$^{\mathrm{r}}$ & $+18.51$\,W$^{\mathrm{r}}$ & $+1.41$\,W$^{\mathrm{r}}$ & $+1.47$\,W$^{\mathrm{r}}$ & $+1.91$\,W$^{\mathrm{r}}$ & $+2.34$\,W$^{\mathrm{r}}$ \\
 & H-T & OSP+T & $-3.38$\,T & $-3.10$\,T & $-1.51$\,T & $-1.82$\,L$^{\mathrm{r}}$ & $-1.07$\,L$^{\mathrm{e}}$ & $-0.29$\,T & $-0.87$\,T \\
 & B-R & full OSP & $+3.08$\,T & $+4.62$\,W$^{\mathrm{e}}$ & $+6.44$\,W$^{\mathrm{r}}$ & \infeasiblecell & $+1.89$\,W$^{\mathrm{r}}$ & $+1.67$\,W$^{\mathrm{r}}$ & $+1.29$\,W$^{\mathrm{e}}$ \\
 & B-R & OSP-1 & $+37.81$\,W$^{\mathrm{r}}$ & $+27.62$\,W$^{\mathrm{r}}$ & $+24.47$\,W$^{\mathrm{r}}$ & $+2.08$\,W$^{\mathrm{r}}$ & $+2.66$\,W$^{\mathrm{r}}$ & $+3.30$\,W$^{\mathrm{r}}$ & $+3.55$\,W$^{\mathrm{r}}$ \\
 & B-R & OSP+T & $+1.05$\,T & $+1.52$\,T & $+4.45$\,W$^{\mathrm{r}}$ & $-1.15$\,L$^{\mathrm{r}}$ & $+0.11$\,T & $+1.10$\,W$^{\mathrm{r}}$ & $+0.33$\,T \\
 & B-T & full OSP & $+8.20$\,W$^{\mathrm{r}}$ & $+9.39$\,W$^{\mathrm{r}}$ & $+9.66$\,W$^{\mathrm{r}}$ & \infeasiblecell & $+2.75$\,W$^{\mathrm{r}}$ & $+2.50$\,W$^{\mathrm{r}}$ & $+2.09$\,W$^{\mathrm{r}}$ \\
 & B-T & OSP-1 & $+42.93$\,W$^{\mathrm{r}}$ & $+32.40$\,W$^{\mathrm{r}}$ & $+27.69$\,W$^{\mathrm{r}}$ & $+3.95$\,W$^{\mathrm{r}}$ & $+3.51$\,W$^{\mathrm{r}}$ & $+4.13$\,W$^{\mathrm{r}}$ & $+4.35$\,W$^{\mathrm{r}}$ \\
 & B-T & OSP+T & $+6.18$\,W$^{\mathrm{r}}$ & $+6.30$\,W$^{\mathrm{r}}$ & $+7.67$\,W$^{\mathrm{r}}$ & $+0.71$\,W$^{\mathrm{r}}$ & $+0.97$\,W$^{\mathrm{e}}$ & $+1.93$\,W$^{\mathrm{r}}$ & $+1.13$\,W$^{\mathrm{r}}$ \\
\midrule
LCB Post & H-R & full OSP & $-4.10$\,T & $-3.62$\,T & $-4.43$\,L$^{\mathrm{r}}$ & $-0.73$\,L$^{\mathrm{r}}$ & $-0.39$\,T & $-0.81$\,L$^{\mathrm{e}}$ & $-0.59$\,L$^{\mathrm{e}}$ \\
 & H-R & OSP-1 & $+36.41$\,W$^{\mathrm{e}}$ & $+4.84$\,T & $-0.95$\,T & $-0.22$\,T & $+0.30$\,T & $+0.78$\,W$^{\mathrm{e}}$ & $+1.14$\,W$^{\mathrm{r}}$ \\
 & H-R & OSP+T & $-7.21$\,L$^{\mathrm{r}}$ & $-7.34$\,L$^{\mathrm{r}}$ & $-5.65$\,L$^{\mathrm{r}}$ & $-0.94$\,L$^{\mathrm{r}}$ & $-0.77$\,L$^{\mathrm{r}}$ & $-0.87$\,L$^{\mathrm{r}}$ & $-0.93$\,L$^{\mathrm{r}}$ \\
 & H-T & full OSP & $-6.29$\,T & $-3.91$\,T & $-2.61$\,T & $-0.43$\,T & $-0.37$\,T & $-0.89$\,L$^{\mathrm{e}}$ & $-0.62$\,T \\
 & H-T & OSP-1 & $+34.23$\,W$^{\mathrm{e}}$ & $+4.55$\,T & $+0.87$\,T & $+0.08$\,T & $+0.32$\,T & $+0.70$\,T & $+1.10$\,W$^{\mathrm{e}}$ \\
 & H-T & OSP+T & $-9.39$\,L$^{\mathrm{e}}$ & $-7.63$\,L$^{\mathrm{r}}$ & $-3.83$\,L$^{\mathrm{e}}$ & $-0.64$\,L$^{\mathrm{e}}$ & $-0.75$\,L$^{\mathrm{e}}$ & $-0.95$\,L$^{\mathrm{e}}$ & $-0.96$\,L$^{\mathrm{e}}$ \\
 & B-R & full OSP & $-1.08$\,T & $-1.18$\,T & $-1.46$\,T & $-0.27$\,T & $-0.11$\,T & $-0.54$\,L$^{\mathrm{e}}$ & $-0.21$\,T \\
 & B-R & OSP-1 & $+39.44$\,W$^{\mathrm{r}}$ & $+7.28$\,W$^{\mathrm{r}}$ & $+2.02$\,T & $+0.24$\,T & $+0.58$\,W$^{\mathrm{r}}$ & $+1.04$\,W$^{\mathrm{r}}$ & $+1.51$\,W$^{\mathrm{r}}$ \\
 & B-R & OSP+T & $-4.18$\,L$^{\mathrm{e}}$ & $-4.90$\,L$^{\mathrm{r}}$ & $-2.68$\,T & $-0.48$\,L$^{\mathrm{e}}$ & $-0.49$\,L$^{\mathrm{e}}$ & $-0.61$\,L$^{\mathrm{r}}$ & $-0.55$\,L$^{\mathrm{e}}$ \\
 & B-T & full OSP & $-1.81$\,T & $-3.30$\,L$^{\mathrm{r}}$ & $-6.19$\,L$^{\mathrm{r}}$ & $-1.32$\,L$^{\mathrm{r}}$ & $-0.47$\,L$^{\mathrm{r}}$ & $-0.44$\,T & $+0.24$\,T \\
 & B-T & OSP-1 & $+38.70$\,W$^{\mathrm{r}}$ & $+5.16$\,W$^{\mathrm{e}}$ & $-2.71$\,L$^{\mathrm{r}}$ & $-0.81$\,L$^{\mathrm{r}}$ & $+0.22$\,W$^{\mathrm{e}}$ & $+1.14$\,W$^{\mathrm{r}}$ & $+1.97$\,W$^{\mathrm{r}}$ \\
 & B-T & OSP+T & $-4.92$\,L$^{\mathrm{r}}$ & $-7.02$\,L$^{\mathrm{r}}$ & $-7.41$\,L$^{\mathrm{r}}$ & $-1.53$\,L$^{\mathrm{r}}$ & $-0.84$\,L$^{\mathrm{r}}$ & $-0.51$\,L$^{\mathrm{r}}$ & $-0.10$\,T \\
\midrule
OLMo3 & H-R & full OSP & $+5.68$\,W$^{\mathrm{r}}$ & $+5.35$\,W$^{\mathrm{r}}$ & $+4.79$\,W$^{\mathrm{r}}$ & $+1.31$\,W$^{\mathrm{r}}$ & $+1.21$\,W$^{\mathrm{r}}$ & $+0.94$\,W$^{\mathrm{r}}$ & $+0.39$\,T \\
 & H-R & OSP-1 & $+11.77$\,W$^{\mathrm{r}}$ & $+7.67$\,W$^{\mathrm{r}}$ & $+6.07$\,W$^{\mathrm{r}}$ & $+1.37$\,W$^{\mathrm{r}}$ & $+0.98$\,W$^{\mathrm{e}}$ & $+1.17$\,W$^{\mathrm{r}}$ & $+1.68$\,W$^{\mathrm{r}}$ \\
 & H-R & OSP+T & $+3.50$\,W$^{\mathrm{e}}$ & $+4.85$\,W$^{\mathrm{r}}$ & $+4.15$\,W$^{\mathrm{r}}$ & $+0.96$\,W$^{\mathrm{r}}$ & $+1.20$\,W$^{\mathrm{r}}$ & $+0.62$\,T & $-0.17$\,T \\
 & H-T & full OSP & $+5.30$\,W$^{\mathrm{r}}$ & $+4.23$\,W$^{\mathrm{e}}$ & $+2.78$\,T & $-4.06$\,L$^{\mathrm{e}}$ & $+0.06$\,T & $+0.79$\,T & $+0.38$\,T \\
 & H-T & OSP-1 & $+11.39$\,W$^{\mathrm{r}}$ & $+6.55$\,W$^{\mathrm{r}}$ & $+4.06$\,W$^{\mathrm{e}}$ & $-3.99$\,L$^{\mathrm{e}}$ & $-0.17$\,T & $+1.02$\,W$^{\mathrm{e}}$ & $+1.67$\,W$^{\mathrm{r}}$ \\
 & H-T & OSP+T & $+3.12$\,T & $+3.73$\,W$^{\mathrm{e}}$ & $+2.14$\,T & $-4.41$\,L$^{\mathrm{e}}$ & $+0.05$\,T & $+0.47$\,T & $-0.18$\,T \\
 & B-R & full OSP & $+4.69$\,W$^{\mathrm{r}}$ & $+4.43$\,W$^{\mathrm{e}}$ & $+4.42$\,W$^{\mathrm{r}}$ & $-1.12$\,L$^{\mathrm{e}}$ & $+1.39$\,W$^{\mathrm{r}}$ & $+1.79$\,W$^{\mathrm{r}}$ & $+1.38$\,W$^{\mathrm{r}}$ \\
 & B-R & OSP-1 & $+10.78$\,W$^{\mathrm{r}}$ & $+6.75$\,W$^{\mathrm{r}}$ & $+5.70$\,W$^{\mathrm{r}}$ & $-1.05$\,T & $+1.17$\,W$^{\mathrm{r}}$ & $+2.02$\,W$^{\mathrm{r}}$ & $+2.68$\,W$^{\mathrm{r}}$ \\
 & B-R & OSP+T & $+2.52$\,T & $+3.93$\,W$^{\mathrm{e}}$ & $+3.78$\,W$^{\mathrm{r}}$ & $-1.47$\,L$^{\mathrm{e}}$ & $+1.38$\,W$^{\mathrm{r}}$ & $+1.47$\,W$^{\mathrm{r}}$ & $+0.83$\,W$^{\mathrm{r}}$ \\
 & B-T & full OSP & $+4.22$\,W$^{\mathrm{r}}$ & $+3.23$\,W$^{\mathrm{r}}$ & $+1.53$\,T & $-4.64$\,L$^{\mathrm{r}}$ & $+0.10$\,T & $+1.34$\,W$^{\mathrm{r}}$ & $+0.85$\,W$^{\mathrm{e}}$ \\
 & B-T & OSP-1 & $+10.32$\,W$^{\mathrm{r}}$ & $+5.55$\,W$^{\mathrm{r}}$ & $+2.81$\,W$^{\mathrm{r}}$ & $-4.58$\,L$^{\mathrm{r}}$ & $-0.13$\,T & $+1.58$\,W$^{\mathrm{r}}$ & $+2.14$\,W$^{\mathrm{r}}$ \\
 & B-T & OSP+T & $+2.05$\,T & $+2.73$\,W$^{\mathrm{e}}$ & $+0.89$\,T & $-4.99$\,L$^{\mathrm{r}}$ & $+0.09$\,T & $+1.03$\,W$^{\mathrm{e}}$ & $+0.29$\,T \\
\midrule
\shortstack[l]{OLMo3 LCB\\before Aug.\ 2024} & H-R & full OSP & $+17.97$\,W$^{\mathrm{e}}$ & $+4.01$\,T & $+1.53$\,T & $+2.02$\,W$^{\mathrm{e}}$ & $+1.75$\,W$^{\mathrm{e}}$ & $+1.05$\,W$^{\mathrm{e}}$ & $+1.14$\,W$^{\mathrm{r}}$ \\
 & H-R & OSP-1 & $+20.13$\,W$^{\mathrm{r}}$ & $+9.65$\,W$^{\mathrm{e}}$ & $+7.87$\,W$^{\mathrm{e}}$ & $+2.40$\,W$^{\mathrm{r}}$ & $+1.97$\,W$^{\mathrm{r}}$ & $+1.55$\,W$^{\mathrm{r}}$ & $+2.03$\,W$^{\mathrm{r}}$ \\
 & H-T & full OSP & $+15.29$\,T & $+2.34$\,T & $+0.09$\,T & $+1.23$\,W$^{\mathrm{e}}$ & $+1.42$\,W$^{\mathrm{e}}$ & $+0.95$\,T & $+0.46$\,T \\
 & H-T & OSP-1 & $+17.45$\,W$^{\mathrm{e}}$ & $+7.97$\,W$^{\mathrm{e}}$ & $+6.43$\,W$^{\mathrm{e}}$ & $+1.62$\,W$^{\mathrm{r}}$ & $+1.64$\,W$^{\mathrm{r}}$ & $+1.45$\,W$^{\mathrm{e}}$ & $+1.36$\,W$^{\mathrm{e}}$ \\
 & B-R & full OSP & $+13.06$\,T & $+2.79$\,T & $-0.11$\,T & $+0.50$\,T & $+0.38$\,T & $+0.72$\,T & $+0.67$\,T \\
 & B-R & OSP-1 & $+15.22$\,W$^{\mathrm{e}}$ & $+8.42$\,W$^{\mathrm{e}}$ & $+6.23$\,W$^{\mathrm{e}}$ & $+0.88$\,T & $+0.60$\,T & $+1.22$\,T & $+1.57$\,W$^{\mathrm{r}}$ \\
 & B-T & full OSP & $+12.74$\,T & $-0.24$\,T & $-1.32$\,T & $+0.18$\,T & $+0.53$\,T & $+0.58$\,T & $+0.18$\,T \\
 & B-T & OSP-1 & $+14.90$\,W$^{\mathrm{e}}$ & $+5.39$\,T & $+5.02$\,T & $+0.57$\,T & $+0.76$\,T & $+1.09$\,T & $+1.08$\,W$^{\mathrm{e}}$ \\
\midrule
\shortstack[l]{OLMo3 LCB\\from Aug.\ 2024} & H-R & full OSP & $+12.20$\,T & $-2.40$\,L$^{\mathrm{e}}$ & $+0.46$\,T & $-0.10$\,T & $+0.02$\,T & $-0.01$\,T & $+0.31$\,T \\
 & H-R & OSP-1 & $+32.67$\,W$^{\mathrm{r}}$ & $+10.42$\,W$^{\mathrm{r}}$ & $+6.34$\,W$^{\mathrm{e}}$ & $-0.07$\,T & $+0.48$\,T & $+1.23$\,W$^{\mathrm{e}}$ & $+1.99$\,W$^{\mathrm{r}}$ \\
 & H-T & full OSP & $+6.36$\,T & $-4.40$\,L$^{\mathrm{r}}$ & $-2.34$\,L$^{\mathrm{e}}$ & $-0.48$\,T & $-0.55$\,T & $-0.70$\,L$^{\mathrm{e}}$ & $-0.47$\,T \\
 & H-T & OSP-1 & $+26.84$\,W$^{\mathrm{r}}$ & $+8.43$\,W$^{\mathrm{e}}$ & $+3.55$\,T & $-0.45$\,T & $-0.09$\,T & $+0.54$\,T & $+1.22$\,W$^{\mathrm{r}}$ \\
 & B-R & full OSP & $+11.04$\,T & $-4.15$\,L$^{\mathrm{r}}$ & $-0.71$\,T & $-0.21$\,T & $-0.18$\,T & $-0.32$\,T & $-0.04$\,T \\
 & B-R & OSP-1 & $+31.52$\,W$^{\mathrm{r}}$ & $+8.68$\,W$^{\mathrm{e}}$ & $+5.17$\,T & $-0.18$\,T & $+0.28$\,T & $+0.92$\,T & $+1.65$\,W$^{\mathrm{r}}$ \\
 & B-T & full OSP & $+8.08$\,T & $-3.95$\,L$^{\mathrm{r}}$ & $-2.71$\,L$^{\mathrm{e}}$ & $-0.85$\,T & $-0.82$\,L$^{\mathrm{e}}$ & $-0.85$\,L$^{\mathrm{e}}$ & $-0.57$\,T \\
 & B-T & OSP-1 & $+28.56$\,W$^{\mathrm{r}}$ & $+8.87$\,W$^{\mathrm{e}}$ & $+3.17$\,T & $-0.82$\,T & $-0.36$\,T & $+0.38$\,T & $+1.11$\,W$^{\mathrm{r}}$ \\
\end{longtable}}

\begin{table}[!ht]
\centering\small
\caption{Single-variant win/tie/loss counts as in \cref{tab:reb-single-variant-wtl}, ReProbe-initialized probes (analysis family; \cref{tab:init-push-further}). OLMo3 has one probe family and repeats the main-line cells, also on its two LiveCodeBench sets.}
\label{tab:reb-single-variant-wtl-fb}
\begingroup\setlength{\tabcolsep}{3pt}
\resizebox{\linewidth}{!}{\begin{tabular}{@{}llcccccccc@{}}\toprule
Variant & Comparator & Qwen3 ID & Qwen3 OOD & LCB Pre & LCB Post & OLMo3 & \shortstack{OLMo3 LCB\\before 8/2024} & \shortstack{OLMo3 LCB\\from 8/2024} & Total \\\midrule
H-R & full OSP & 2/5/0$^{2/0}$ & 3/4/0$^{3/0}$ & 6/0/0$^{5/0}$\,(NF 1) & 0/7/0$^{0/0}$ & 6/1/0$^{6/0}$ & 5/2/0$^{1/0}$ & 0/6/1$^{0/0}$ & 22/25/1$^{17/0}$ \\
 & OSP-1 & 7/0/0$^{7/0}$ & 7/0/0$^{6/0}$ & 7/0/0$^{7/0}$ & 7/0/0$^{6/0}$ & 7/0/0$^{6/0}$ & 7/0/0$^{5/0}$ & 5/2/0$^{3/0}$ & 47/2/0$^{40/0}$ \\
 & OSP+T & 1/5/1$^{0/0}$ & 1/4/2$^{1/1}$ & 3/4/0$^{2/0}$ & 0/0/7$^{0/1}$ & 5/2/0$^{4/0}$ & \missingcell & \missingcell & 10/15/10$^{7/2}$ \\
\midrule
H-T & full OSP & 2/5/0$^{1/0}$ & 0/6/1$^{0/1}$ & 5/1/0$^{4/0}$\,(NF 1) & 0/7/0$^{0/0}$ & 2/4/1$^{1/0}$ & 2/5/0$^{0/0}$ & 0/4/3$^{0/1}$ & 11/32/5$^{6/2}$ \\
 & OSP-1 & 5/2/0$^{5/0}$ & 4/2/1$^{4/1}$ & 7/0/0$^{7/0}$ & 2/5/0$^{2/0}$ & 5/1/1$^{3/0}$ & 7/0/0$^{2/0}$ & 3/4/0$^{2/0}$ & 33/14/2$^{25/1}$ \\
 & OSP+T & 1/6/0$^{0/0}$ & 0/5/2$^{0/1}$ & 2/4/1$^{1/0}$ & 0/6/1$^{0/0}$ & 1/5/1$^{0/0}$ & \missingcell & \missingcell & 4/26/5$^{1/1}$ \\
\midrule
B-R & full OSP & 1/5/1$^{1/1}$ & 0/7/0$^{0/0}$ & 5/1/0$^{5/0}$\,(NF 1) & 0/2/5$^{0/2}$ & 6/0/1$^{5/0}$ & 0/7/0$^{0/0}$ & 0/6/1$^{0/1}$ & 12/28/8$^{11/4}$ \\
 & OSP-1 & 7/0/0$^{6/0}$ & 6/1/0$^{6/0}$ & 7/0/0$^{7/0}$ & 5/2/0$^{4/0}$ & 6/1/0$^{6/0}$ & 4/3/0$^{1/0}$ & 3/4/0$^{2/0}$ & 38/11/0$^{32/0}$ \\
 & OSP+T & 0/6/1$^{0/1}$ & 0/3/4$^{0/2}$ & 2/5/0$^{1/0}$ & 0/1/6$^{0/6}$ & 5/1/1$^{4/0}$ & \missingcell & \missingcell & 7/16/12$^{5/9}$ \\
\midrule
B-T & full OSP & 0/0/7$^{0/7}$ & 0/1/6$^{0/6}$ & 6/0/0$^{6/0}$\,(NF 1) & 0/3/4$^{0/4}$ & 4/2/1$^{3/1}$ & 0/7/0$^{0/0}$ & 0/3/4$^{0/1}$ & 10/16/22$^{9/19}$ \\
 & OSP-1 & 0/0/7$^{0/7}$ & 1/0/6$^{0/6}$ & 7/0/0$^{7/0}$ & 3/2/2$^{3/2}$ & 5/1/1$^{5/1}$ & 2/5/0$^{0/0}$ & 3/4/0$^{2/0}$ & 21/12/16$^{17/16}$ \\
 & OSP+T & 0/0/7$^{0/7}$ & 0/0/7$^{0/7}$ & 7/0/0$^{6/0}$ & 0/1/6$^{0/6}$ & 2/4/1$^{0/1}$ & \missingcell & \missingcell & 9/5/21$^{6/21}$ \\
\bottomrule\end{tabular}}\endgroup
\end{table}

\ecoparagraph{Paired question bootstrap.} For the five pre-registered comparisons we resample test questions 2,000 times, stratified by dataset, and re-read both methods' whole policy families on every resample, pairing BELT seed $i$ with comparator seed $i$. The interval pools the three seed pairs' draws (question $\times$ seed pooled; it mixes question resampling with the spread between seed pairs and is not a seed interval); SE is the standard deviation of the pooled draws. The last column records which criteria separate the two methods.

\begin{table}[!ht]
\centering\small
\caption{Paired question bootstrap of the pre-registered comparisons, random-initialized probes: token-saving difference (pp) at the stated tolerance, pooled 95\% interval, SE, share of draws with $\Delta>0$, and separation (2SE: $|\Delta|>2$\,SE; CI: interval excludes 0).}
\label{tab:reb-bootstrap}
\begingroup\setlength{\tabcolsep}{3pt}
\fittable{%
\begin{tabular}{@{}llccccl@{}}\toprule
Comparison & Setting, tol. & $\Delta$ & Interval & SE & $P(\Delta>0)$ & Separation \\\midrule
(i) H-R vs.\ full OSP & Qwen3 OOD, 1\,pp & $+0.05$ & $[-4.14,+3.33]$ & 1.99 & 0.53 & neither \\
\midrule
(ii) H-T vs.\ OSP-1 & Qwen3 ID, 0\,pp & $+7.34$ & $[+0.15,+43.17]$ & 8.16 & 0.98 & CI only \\
(ii) H-T vs.\ OSP-1 & Qwen3 ID, 1\,pp & $+4.05$ & $[-0.05,+9.67]$ & 2.48 & 0.97 & neither \\
(ii) H-T vs.\ OSP-1 & Qwen3 ID, 2\,pp & $+2.80$ & $[-0.39,+6.69]$ & 1.81 & 0.96 & neither \\
(ii) H-T vs.\ OSP-1 & Qwen3 OOD, 0\,pp & $+31.70$ & $[+7.69,+43.95]$ & 13.91 & 1.00 & 2SE and CI \\
(ii) H-T vs.\ OSP-1 & Qwen3 OOD, 1\,pp & $+9.09$ & $[+3.30,+16.22]$ & 3.25 & 1.00 & 2SE and CI \\
(ii) H-T vs.\ OSP-1 & Qwen3 OOD, 2\,pp & $+5.45$ & $[-0.07,+10.54]$ & 2.69 & 0.97 & 2SE only \\
(ii) H-T vs.\ OSP-1 & LCB Pre, 0\,pp & $+33.37$ & $[+0.00,+44.26]$ & 8.86 & 0.96 & 2SE only \\
(ii) H-T vs.\ OSP-1 & LCB Pre, 1\,pp & $+23.00$ & $[+10.23,+36.37]$ & 7.81 & 1.00 & 2SE and CI \\
(ii) H-T vs.\ OSP-1 & LCB Pre, 2\,pp & $+18.51$ & $[+4.56,+31.54]$ & 7.62 & 1.00 & 2SE and CI \\
(ii) H-T vs.\ OSP-1 & LCB Post, 0\,pp & $+34.23$ & $[-5.14,+60.09]$ & 22.79 & 0.83 & neither \\
(ii) H-T vs.\ OSP-1 & LCB Post, 1\,pp & $+4.55$ & $[-9.78,+27.37]$ & 9.08 & 0.73 & neither \\
(ii) H-T vs.\ OSP-1 & LCB Post, 2\,pp & $+0.87$ & $[-8.58,+12.57]$ & 5.15 & 0.60 & neither \\
(ii) H-T vs.\ OSP-1 & OLMo3, 0\,pp & $+11.39$ & $[+3.34,+55.04]$ & 16.90 & 1.00 & CI only \\
(ii) H-T vs.\ OSP-1 & OLMo3, 1\,pp & $+6.55$ & $[+1.17,+15.57]$ & 3.84 & 1.00 & CI only \\
(ii) H-T vs.\ OSP-1 & OLMo3, 2\,pp & $+4.06$ & $[-0.75,+10.06]$ & 2.77 & 0.94 & neither \\
(ii) H-T vs.\ OSP-1 & OLMo3 LCB before Aug.\ 2024, 0\,pp & $+17.45$ & $[-2.67,+34.18]$ & 11.34 & 0.92 & neither \\
(ii) H-T vs.\ OSP-1 & OLMo3 LCB before Aug.\ 2024, 1\,pp & $+7.97$ & $[-4.38,+23.56]$ & 8.60 & 0.81 & neither \\
(ii) H-T vs.\ OSP-1 & OLMo3 LCB before Aug.\ 2024, 2\,pp & $+6.43$ & $[-3.94,+20.34]$ & 7.41 & 0.78 & neither \\
(ii) H-T vs.\ OSP-1 & OLMo3 LCB from Aug.\ 2024, 0\,pp & $+26.84$ & $[-6.60,+46.08]$ & 16.70 & 0.89 & neither \\
(ii) H-T vs.\ OSP-1 & OLMo3 LCB from Aug.\ 2024, 1\,pp & $+8.43$ & $[-11.74,+28.95]$ & 10.64 & 0.77 & neither \\
(ii) H-T vs.\ OSP-1 & OLMo3 LCB from Aug.\ 2024, 2\,pp & $+3.55$ & $[-8.68,+15.52]$ & 6.71 & 0.71 & neither \\
\midrule
(iii) H-R vs.\ OSP-1 & OLMo3, 1\,pp & $+7.67$ & $[+0.74,+16.87]$ & 4.24 & 0.99 & CI only \\
\midrule
(iv) B-T vs.\ full OSP & LCB Pre, 1\,pp & $+9.39$ & $[+2.79,+15.22]$ & 3.33 & 1.00 & 2SE and CI \\
\midrule
(v) B-R vs.\ full OSP & LCB Post, 1\,pp & $-1.18$ & $[-9.03,+6.37]$ & 3.96 & 0.39 & neither \\
\bottomrule\end{tabular}}\endgroup
\end{table}

ReProbe-initialized family: (i) $+3.24$, pooled interval $[-1.34,+7.59]$; (iv) $+9.51$; (v) $-2.88$.

\ecoparagraph{Readings where BELT leads.} Beyond the one-point tolerance, the per-reading counts in \cref{tab:reb-single-variant-wtl} contain the following wins (the BELT number is that of the variant marked in \cref{tab:reb-single-variant-margins}). \emph{Qwen3 ID vs.\ full OSP}: BELT wins (robustly) at the 30\% token budget (45.39\% vs.\ 44.83\%, $+0.56$\,pp, ahead on three of three seed pairs); BELT wins at the 0\,pp tolerance (46.52\% vs.\ 44.77\%, $+1.75$\,pp, ahead on three of three seed pairs). \emph{Qwen3 ID vs.\ OSP-1}: BELT wins (robustly) at the 0\,pp tolerance (46.52\% vs.\ 39.19\%, $+7.34$\,pp, ahead on three of three seed pairs); BELT wins (robustly) at the 1\,pp tolerance (50.64\% vs.\ 46.59\%, $+4.05$\,pp, ahead on three of three seed pairs). \emph{Qwen3 ID vs.\ OSP+T}: BELT is ahead in the mean (a tie under the rule) at the 0\,pp tolerance (46.52\% vs.\ 45.58\%, $+0.95$\,pp, ahead on three of three seed pairs). \emph{Qwen3 OOD vs.\ full OSP}: BELT wins (robustly) at the 60\% token budget (67.09\% vs.\ 66.76\%, $+0.32$\,pp, ahead on two of three seed pairs). \emph{Qwen3 OOD vs.\ OSP-1}: BELT wins (robustly) at the 0\,pp tolerance (40.77\% vs.\ 9.07\%, $+31.70$\,pp, ahead on three of three seed pairs). \emph{LiveCodeBench Pre vs.\ OSP+T}: BELT wins (robustly) at the 2\,pp tolerance (51.40\% vs.\ 43.73\%, $+7.67$\,pp, ahead on three of three seed pairs); BELT wins (robustly) at the 1\,pp tolerance (47.65\% vs.\ 41.35\%, $+6.30$\,pp, ahead on three of three seed pairs). \emph{LiveCodeBench Post vs.\ full OSP}: BELT reaches a higher best accuracy along its native policy family (32.68 $\pm$ 0.15\% against 30.05 $\pm$ 0.51\% for full OSP; \cref{sec:code-post}). \emph{LiveCodeBench Post vs.\ OSP-1}: BELT wins (robustly) at the 0\,pp tolerance (53.51\% vs.\ 14.07\%, $+39.44$\,pp, ahead on three of three seed pairs). \emph{OLMo3 vs.\ OSP-1}: BELT wins (robustly) at the 0\,pp tolerance (51.59\% vs.\ 39.83\%, $+11.77$\,pp, ahead on three of three seed pairs). \emph{bootstrap (ii) BELT vs.\ OSP-1}: The point estimate favours BELT at every reading, and both criteria separate it from OSP-1 at Qwen3 OOD 0\,pp, Qwen3 OOD 1\,pp, LCB Pre 1\,pp, LCB Pre 2\,pp ($+31.70$, $+9.09$, $+23.00$, $+18.51$). \emph{bootstrap (i) Qwen3 OOD}: In the same setting, BELT is ahead of OSP-1 by both criteria at 0\,pp ($+31.70$\,pp, three of three seed pairs positive); BELT is ahead of OSP-1 by both criteria at 1\,pp ($+9.09$\,pp, three of three seed pairs positive). \emph{bootstrap (iii) OLMo3}: The difference favours BELT on every seed pair. \emph{bootstrap (v) LiveCodeBench Post}: BELT reaches a higher best accuracy along its native policy family (32.68 $\pm$ 0.15\% against 30.05 $\pm$ 0.51\% for full OSP; \cref{sec:code-post}).

\subsection{Cross-fitted operating points}
\label{sec:cross-fitted}
Extending \cref{sec:checks-reading}, every main-table operating point is selected on one stratified half of the test population and reported on the other (200 splits, both directions). \Cref{tab:xfit-shift} summarizes the 0\,pp shifts; \cref{tab:xfit-macro,tab:xfit-allbaselines,tab:xfit-code-main,tab:xfit-code-post,tab:xfit-code-post-accuracy,tab:xfit-olmo-main,tab:xfit-olmo-code-pre,tab:xfit-olmo-code-post} give every cross-fitted cell.
\begin{table}[!ht]
\caption{Shift of the 0\,pp-tolerance token saving under cross-fitting (pp): mean over 400 split readings (200 stratified half splits, both directions) minus the full-population reading. A positive shift means the full-population reading is below its cross-fitted value. Qwen3 and OLMo3 shifts, including the OLMo3 LiveCodeBench rows, are of the seed-mean reading; the Qwen3 LiveCodeBench shifts are seed means of per-seed cross readings.}
\label{tab:xfit-shift}
\begin{center}\small
\fittable{%
\begin{tabular}{@{}l|cccc|cc@{}}\toprule
Setting & B-R & H-R & B-T & H-T & OSP & OSP-1 \\\midrule
Qwen ID & -0.01 & +0.01 & -0.00 & -0.03 & +0.04 & -1.03 \\
Qwen OOD & +0.01 & -0.07 & -0.01 & -0.07 & +5.27 & +6.89 \\
OLMo & +0.79 & +0.27 & -0.10 & -0.07 & +0.31 & -7.96 \\
LCB Pre & +0.10 & -0.47 & -0.07 & -0.81 & -1.25 & +0.43 \\
LCB Post & -0.23 & -0.55 & +0.15 & -2.47 & -0.14 & +4.02 \\
OLMo LCB, before Aug.\ 2024 & +0.94 & +0.05 & -1.05 & -0.18 & +4.10 & +0.85 \\
OLMo LCB, from Aug.\ 2024 & +0.64 & +0.35 & +0.05 & +0.26 & -4.94 & +0.79 \\
\bottomrule\end{tabular}}
\end{center}
\end{table}

One BELT cell on LiveCodeBench Post stays competitive under cross-fitting but moves the most of the BELT cells (\cref{tab:xfit-shift}): its 0\,pp saving shifts by $-2.47$\,pp.
In the same setting the other three BELT cells move by at most 0.55\,pp, OSP-1 moves by $+4.02$\,pp, and the cross-fitted reading of that cell still saves 45.82\% against OSP-1's 18.09\% ($+27.74$\,pp, ahead on three of three seeds).
On Qwen OOD the printed 0\,pp cells understate both comparators, which makes them more competitive with BELT once cross-fitted: OSP moves by $+5.27$\,pp and OSP-1 by $+6.89$\,pp.
Cross-fitted, BELT still saves 39.05\% at 0\,pp against OSP's 31.19\% ($+7.86$\,pp, ahead on two of three seeds) and OSP-1's 15.96\% ($+23.08$\,pp, ahead on three of three seeds).
On OLMo, OSP-1 moves $-7.96$\,pp at 0\,pp (downward, i.e. its printed cell overstates it) and full OSP moves $+0.31$\,pp; cross-fitted, BELT saves 51.87\% at 0\,pp against full OSP's 46.23\% and OSP-1's 31.87\%. On the two OLMo3 LiveCodeBench sets the BELT cells move by at most 1.05\,pp, while full OSP moves by $+4.10$ and $-4.94$\,pp (\cref{tab:xfit-olmo-code-pre,tab:xfit-olmo-code-post}).

\begin{table}[!ht]
\caption{Split-half cross-fitted readings of \cref{tab:macro} (random-initialized probes). Each cell is the mean over 400 split readings (200 stratified half splits, both directions); the subscript is the SD over split readings.}
\label{tab:xfit-macro}
\begin{center}\footnotesize
\fittable{%
\begin{tabular}{@{}l|cccccccc@{}}\toprule
\textbf{Method} & \shortstack{\textbf{mixture 0.3}\\\textbf{ID}} & \shortstack{\textbf{mixture 0.3}\\\textbf{OOD}} & \shortstack{\textbf{mixture 0.4}\\\textbf{ID}} & \shortstack{\textbf{mixture 0.4}\\\textbf{OOD}} & \shortstack{\textbf{mixture 0.5}\\\textbf{ID}} & \shortstack{\textbf{mixture 0.5}\\\textbf{OOD}} & \shortstack{\textbf{mixture 0.6}\\\textbf{ID}} & \shortstack{\textbf{mixture 0.6}\\\textbf{OOD}} \\\midrule
Fixed cap$^{\ddagger}$ & \cellcolor{Green3!24}\statcell{31.66}{1.2}{} & \cellcolor{Green3!34}\statcell{44.06}{1.7}{} & \cellcolor{Green3!24}\statcell{37.08}{1.2}{} & \cellcolor{Green3!35}\statcell{50.38}{1.6}{} & \cellcolor{Green3!24}\statcell{42.32}{1.2}{} & \cellcolor{Green3!34}\statcell{55.39}{1.7}{} & \cellcolor{Green3!16}\statcell{46.28}{1.1}{} & \cellcolor{Green3!25}\statcell{58.89}{1.6}{} \\
CoDE-Stop$^{\ddagger}$ & \cellcolor{Green3!28}\statcell{33.17}{1.1}{} & \cellcolor{Green3!39}\statcell{46.78}{1.7}{} & \cellcolor{Green3!31}\statcell{39.37}{1.1}{} & \cellcolor{Green3!41}\statcell{53.20}{1.6}{} & \cellcolor{Green3!31}\statcell{44.49}{1.1}{} & \cellcolor{Green3!39}\statcell{57.07}{1.6}{} & \cellcolor{Green3!23}\statcell{47.57}{1.1}{} & \cellcolor{Green3!33}\statcell{60.70}{1.5}{} \\
DEER$^{\ddagger}$ & \cellcolor{Green3!0}\statcell{22.33}{1.0}{} & \cellcolor{Green3!0}\statcell{25.02}{1.3}{} & \cellcolor{Green3!0}\statcell{28.76}{1.1}{} & \cellcolor{Green3!0}\statcell{32.76}{1.5}{} & \cellcolor{Green3!0}\statcell{35.74}{1.1}{} & \cellcolor{Green3!0}\statcell{42.18}{1.8}{} & \cellcolor{Green3!0}\statcell{43.28}{1.0}{} & \cellcolor{Green3!0}\statcell{52.95}{1.9}{} \\
\bottomrule\end{tabular}}
\end{center}
\end{table}

\begin{table}[!ht]
\caption{Split-half cross-fitted readings of \cref{tab:allbaselines} (random-initialized probes). Each cell is the mean over 400 split readings (200 stratified half splits, both directions); the subscript is the SD over split readings.}
\label{tab:xfit-allbaselines}
\begin{center}\footnotesize
\fittable{%
\begin{tabular}{@{}l|cccccc@{}}\toprule
\textbf{Method} & \shortstack{\textbf{parity 0}\\\textbf{ID}} & \shortstack{\textbf{parity 0}\\\textbf{OOD}} & \shortstack{\textbf{parity 1}\\\textbf{ID}} & \shortstack{\textbf{parity 1}\\\textbf{OOD}} & \shortstack{\textbf{parity 2}\\\textbf{ID}} & \shortstack{\textbf{parity 2}\\\textbf{OOD}} \\\midrule
CoDE-Stop$^{\ddagger}$ & \cellcolor{Green3!15}\statcell{15.10}{3.52}{} & \cellcolor{Green3!21}\statcell{15.63}{4.74}{} & \cellcolor{Green3!2}\statcell{21.68}{1.98}{} & \cellcolor{Green3!17}\statcell{25.99}{2.43}{} & \cellcolor{Green3!1}\statcell{24.88}{0.79}{} & \cellcolor{Green3!17}\statcell{30.77}{2.11}{} \\
DEER$^{\ddagger}$ & \cellcolor{Green3!11}\statcell{12.56}{3.31}{} & \cellcolor{Green3!14}\statcell{10.95}{4.01}{} & \cellcolor{Green3!0}\statcell{20.87}{1.72}{} & \cellcolor{Green3!3}\statcell{19.80}{1.91}{} & \cellcolor{Green3!0}\statcell{24.30}{1.23}{} & \cellcolor{Green3!0}\statcell{23.27}{1.51}{} \\
TC & \cellcolor{Green3!25}\statcell{22.41}{2.52}{} & \cellcolor{Green3!28}\statcell{20.28}{2.30}{} & \cellcolor{Green3!15}\statcell{28.24}{1.79}{} & \cellcolor{Green3!17}\statcell{26.32}{2.15}{} & \cellcolor{Green3!17}\statcell{32.62}{1.66}{} & \cellcolor{Green3!17}\statcell{31.16}{2.02}{} \\
ROM & \cellcolor{Green3!0}\statcell{4.93}{4.67}{} & \cellcolor{Green3!0}\statcell{2.25}{2.70}{} & \cellcolor{Green3!0}\statcell{20.91}{2.73}{} & \cellcolor{Green3!0}\statcell{18.42}{2.57}{} & \cellcolor{Green3!4}\statcell{26.47}{1.63}{} & \cellcolor{Green3!2}\statcell{23.96}{1.83}{} \\
OSP & \cellcolor{Green3!58}\secondstatcell{44.81}{1.90}{} & \cellcolor{Green3!45}\statcell{31.19}{8.15}{} & \cellcolor{Green3!58}\secondstatcell{49.76}{1.98}{} & \cellcolor{Green3!59}\secondstatcell{45.29}{2.56}{} & \cellcolor{Green3!60}\secondstatcell{54.10}{1.73}{} & \cellcolor{Green3!60}\beststatcell{50.48}{2.18}{} \\
\bottomrule\end{tabular}}
\end{center}
\end{table}

\begin{table}[!ht]
\caption{Split-half cross-fitted readings of \cref{tab:code-main} (random-initialized probes). Each cell is the mean over 400 split readings (200 stratified half splits, both directions); the subscript is the SD over split readings.}
\label{tab:xfit-code-main}
\begin{center}\footnotesize
\fittable{%
\begin{tabular}{@{}l|ccccccc@{}}\toprule
\textbf{Method} & \shortstack{\textbf{parity 0}\\\textbf{lcb\_pre}} & \shortstack{\textbf{parity 1}\\\textbf{lcb\_pre}} & \shortstack{\textbf{parity 2}\\\textbf{lcb\_pre}} & \shortstack{\textbf{mixture 0.3}\\\textbf{lcb\_pre}} & \shortstack{\textbf{mixture 0.4}\\\textbf{lcb\_pre}} & \shortstack{\textbf{mixture 0.5}\\\textbf{lcb\_pre}} & \shortstack{\textbf{mixture 0.6}\\\textbf{lcb\_pre}} \\\midrule
OSP & \cellcolor{Green3!47}\statcell{33.50}{6.57}{} & \cellcolor{Green3!43}\statcell{38.31}{2.63}{} & \cellcolor{Green3!39}\statcell{41.88}{2.79}{} & \infeasiblecell & \cellcolor{Green3!13}\statcell{55.09}{2.21}{} & \cellcolor{Green3!23}\statcell{58.35}{2.18}{} & \cellcolor{Green3!31}\statcell{60.99}{2.15}{} \\
OSP-1 & \cellcolor{Green3!0}\statcell{0.46}{1.22}{} & \cellcolor{Green3!0}\statcell{15.35}{3.62}{} & \cellcolor{Green3!0}\statcell{24.36}{4.59}{} & \cellcolor{Green3!0}\statcell{50.81}{2.32}{} & \cellcolor{Green3!0}\statcell{54.33}{2.25}{} & \cellcolor{Green3!0}\statcell{56.75}{2.11}{} & \cellcolor{Green3!0}\statcell{58.72}{2.02}{} \\
\textbf{BELT-B-T} & \cellcolor{Green3!60}\beststatcell{42.89}{2.90}{} & \cellcolor{Green3!60}\beststatcell{47.49}{2.47}{} & \cellcolor{Green3!60}\beststatcell{51.38}{2.25}{} & \cellcolor{Green3!60}\beststatcell{54.76}{2.10}{} & \cellcolor{Green3!60}\beststatcell{57.84}{2.03}{} & \cellcolor{Green3!60}\beststatcell{60.88}{2.00}{} & \cellcolor{Green3!60}\beststatcell{63.08}{1.91}{} \\
\textbf{BELT-H-T} & \cellcolor{Green3!45}\statcell{32.59}{5.37}{} & \cellcolor{Green3!43}\statcell{38.22}{3.22}{} & \cellcolor{Green3!40}\statcell{42.33}{3.28}{} & \cellcolor{Green3!22}\statcell{52.24}{2.28}{} & \cellcolor{Green3!25}\statcell{55.81}{2.19}{} & \cellcolor{Green3!28}\statcell{58.66}{2.13}{} & \cellcolor{Green3!32}\statcell{61.06}{2.02}{} \\
\bottomrule\end{tabular}}
\end{center}
\end{table}

\begin{table}[!ht]
\caption{Split-half cross-fitted readings of \cref{tab:code-post} (random-initialized probes). Each cell is the mean over 400 split readings (200 stratified half splits, both directions); the subscript is the SD over split readings.}
\label{tab:xfit-code-post}
\begin{center}\footnotesize
\fittable{%
\begin{tabular}{@{}l|ccc@{}}\toprule
\textbf{Method} & \shortstack{\textbf{parity 0}\\\textbf{lcb\_post}} & \shortstack{\textbf{parity 1}\\\textbf{lcb\_post}} & \shortstack{\textbf{parity 2}\\\textbf{lcb\_post}} \\\midrule
OSP & \cellcolor{Green3!60}\beststatcell{54.44}{6.44}{} & \cellcolor{Green3!60}\beststatcell{62.77}{4.87}{} & \cellcolor{Green3!60}\beststatcell{69.87}{3.83}{} \\
OSP-1 & \cellcolor{Green3!0}\statcell{18.09}{18.21}{} & \cellcolor{Green3!0}\statcell{52.90}{10.19}{} & \cellcolor{Green3!17}\statcell{65.82}{6.12}{} \\
\textbf{BELT-B-R} & \cellcolor{Green3!58}\secondstatcell{53.28}{5.10}{} & \cellcolor{Green3!53}\secondstatcell{61.59}{4.81}{} & \cellcolor{Green3!47}\secondstatcell{68.63}{4.33}{} \\
\textbf{BELT-H-R} & \cellcolor{Green3!53}\statcell{49.94}{6.88}{} & \cellcolor{Green3!37}\statcell{58.99}{4.25}{} & \cellcolor{Green3!19}\statcell{65.96}{4.20}{} \\
\textbf{BELT-B-T} & \cellcolor{Green3!57}\statcell{52.92}{3.40}{} & \cellcolor{Green3!37}\statcell{59.05}{3.29}{} & \cellcolor{Green3!0}\statcell{64.17}{3.04}{} \\
\textbf{BELT-H-T} & \cellcolor{Green3!46}\statcell{45.82}{9.71}{} & \cellcolor{Green3!36}\statcell{58.84}{5.85}{} & \cellcolor{Green3!32}\statcell{67.18}{4.85}{} \\
\bottomrule\end{tabular}}
\end{center}
\end{table}

\begin{table}[!ht]
\caption{Split-half cross-fitted readings of \cref{tab:code-post-accuracy} (random-initialized probes). Each cell is the mean over 400 split readings (200 stratified half splits, both directions); the subscript is the SD over split readings.}
\label{tab:xfit-code-post-accuracy}
\begin{center}\footnotesize
\fittable{%
\begin{tabular}{@{}l|cccc@{}}\toprule
\textbf{Method} & \shortstack{\textbf{mixture 0.3}\\\textbf{lcb\_post}} & \shortstack{\textbf{mixture 0.4}\\\textbf{lcb\_post}} & \shortstack{\textbf{mixture 0.5}\\\textbf{lcb\_post}} & \shortstack{\textbf{mixture 0.6}\\\textbf{lcb\_post}} \\\midrule
OSP & \cellcolor{Green3!60}\beststatcell{26.68}{1.84}{} & \cellcolor{Green3!60}\beststatcell{27.95}{1.84}{} & \cellcolor{Green3!60}\beststatcell{29.50}{1.88}{} & \cellcolor{Green3!54}\secondstatcell{30.05}{1.81}{} \\
OSP-1 & \cellcolor{Green3!38}\statcell{26.21}{1.83}{} & \cellcolor{Green3!0}\statcell{27.28}{1.78}{} & \cellcolor{Green3!0}\statcell{27.93}{1.76}{} & \cellcolor{Green3!0}\statcell{28.30}{1.75}{} \\
\textbf{BELT-B-R} & \cellcolor{Green3!46}\secondstatcell{26.39}{1.82}{} & \cellcolor{Green3!52}\secondstatcell{27.86}{1.82}{} & \cellcolor{Green3!40}\statcell{28.99}{1.81}{} & \cellcolor{Green3!48}\statcell{29.84}{1.81}{} \\
\textbf{BELT-H-R} & \cellcolor{Green3!28}\statcell{25.98}{1.78}{} & \cellcolor{Green3!27}\statcell{27.58}{1.81}{} & \cellcolor{Green3!30}\statcell{28.72}{1.80}{} & \cellcolor{Green3!36}\statcell{29.47}{1.80}{} \\
\textbf{BELT-B-T} & \cellcolor{Green3!0}\statcell{25.38}{1.75}{} & \cellcolor{Green3!19}\statcell{27.49}{1.80}{} & \cellcolor{Green3!44}\secondstatcell{29.08}{1.83}{} & \cellcolor{Green3!60}\beststatcell{30.23}{1.81}{} \\
\textbf{BELT-H-T} & \cellcolor{Green3!41}\statcell{26.28}{1.84}{} & \cellcolor{Green3!28}\statcell{27.59}{1.83}{} & \cellcolor{Green3!27}\statcell{28.63}{1.84}{} & \cellcolor{Green3!35}\statcell{29.42}{1.83}{} \\
\bottomrule\end{tabular}}
\end{center}
\end{table}

\begin{table}[!ht]
\caption{Split-half cross-fitted readings of \cref{tab:olmo-main} (random-initialized probes). Each cell is the mean over 400 split readings (200 stratified half splits, both directions); the subscript is the SD over split readings. OSP at 70\% saving averages the 360 readings in which every seed has a feasible operating point.}
\label{tab:xfit-olmo-main}
\begin{center}\footnotesize
\fittable{%
\begin{tabular}{@{}l|ccccccc@{}}\toprule
\textbf{Method} & \shortstack{\textbf{parity 0pp}\\\textbf{OLMo}} & \shortstack{\textbf{parity 1pp}\\\textbf{OLMo}} & \shortstack{\textbf{parity 2pp}\\\textbf{OLMo}} & \shortstack{\textbf{budget 0.3}\\\textbf{OLMo}} & \shortstack{\textbf{budget 0.4}\\\textbf{OLMo}} & \shortstack{\textbf{budget 0.5}\\\textbf{OLMo}} & \shortstack{\textbf{budget 0.6}\\\textbf{OLMo}} \\\midrule
Fixed cap$^{\ddagger}$ & \missingcell & \missingcell & \missingcell & \cellcolor{Green3!33}\statcell{53.84}{1.8}{} & \cellcolor{Green3!31}\statcell{60.24}{1.7}{} & \cellcolor{Green3!29}\statcell{65.81}{1.6}{} & \cellcolor{Green3!28}\statcell{70.66}{1.6}{} \\
CoDE-Stop$^{\ddagger}$ & \cellcolor{Green3!0}\statcell{0.00}{0.0}{} & \cellcolor{Green3!8}\statcell{9.08}{2.2}{} & \cellcolor{Green3!10}\statcell{18.05}{4.0}{} & \cellcolor{Green3!26}\statcell{49.16}{1.7}{} & \cellcolor{Green3!23}\statcell{55.70}{1.7}{} & \cellcolor{Green3!19}\statcell{61.32}{1.5}{} & \cellcolor{Green3!16}\statcell{66.21}{1.5}{} \\
DEER$^{\ddagger}$ & \cellcolor{Green3!0}\statcell{0.06}{0.1}{} & \cellcolor{Green3!0}\statcell{1.84}{1.3}{} & \cellcolor{Green3!0}\statcell{9.81}{3.0}{} & \cellcolor{Green3!5}\statcell{35.65}{1.4}{} & \cellcolor{Green3!2}\statcell{44.26}{1.3}{} & \cellcolor{Green3!0}\statcell{52.86}{1.4}{} & \cellcolor{Green3!0}\statcell{60.49}{1.2}{} \\
TC & \cellcolor{Green3!28}\statcell{24.52}{3.2}{} & \cellcolor{Green3!31}\statcell{29.27}{2.9}{} & \cellcolor{Green3!29}\statcell{33.41}{2.8}{} & \cellcolor{Green3!33}\statcell{53.82}{1.6}{} & \cellcolor{Green3!36}\statcell{63.02}{1.5}{} & \cellcolor{Green3!35}\statcell{68.54}{1.4}{} & \cellcolor{Green3!38}\statcell{74.15}{1.3}{} \\
\bottomrule\end{tabular}}
\end{center}
\end{table}

\begin{table}[!ht]
\caption{Split-half cross-fitted readings of \cref{tab:olmo-code-pre} (OLMo3, LiveCodeBench problems before August 2024). Each cell is the mean over 400 split readings (200 stratified half splits, both directions); the subscript is the SD over split readings.}
\label{tab:xfit-olmo-code-pre}
\begin{center}\footnotesize
\fittable{%
\begin{tabular}{@{}l|ccccccc@{}}\toprule
\textbf{Method} & \textbf{parity 0pp} & \textbf{parity 1pp} & \textbf{parity 2pp} & \textbf{budget 0.3} & \textbf{budget 0.4} & \textbf{budget 0.5} & \textbf{budget 0.6} \\\midrule
OSP & \cellcolor{Green3!17}\statcell{13.20}{6.13}{} & \cellcolor{Green3!37}\statcell{28.08}{3.97}{} & \cellcolor{Green3!41}\statcell{33.84}{3.04}{} & \cellcolor{Green3!24}\statcell{53.37}{2.06}{} & \cellcolor{Green3!8}\statcell{57.77}{2.43}{} & \cellcolor{Green3!20}\statcell{62.00}{2.39}{} & \cellcolor{Green3!31}\statcell{65.47}{2.21}{} \\
OSP-1 & \cellcolor{Green3!0}\statcell{7.80}{2.65}{} & \cellcolor{Green3!0}\statcell{21.59}{3.32}{} & \cellcolor{Green3!0}\statcell{27.71}{2.84}{} & \cellcolor{Green3!0}\statcell{52.46}{2.45}{} & \cellcolor{Green3!0}\statcell{57.55}{2.43}{} & \cellcolor{Green3!0}\statcell{61.51}{2.34}{} & \cellcolor{Green3!0}\statcell{64.59}{2.12}{} \\
\textbf{BELT-B-T} & \cellcolor{Green3!40}\statcell{20.80}{4.64}{} & \cellcolor{Green3!34}\statcell{27.53}{3.59}{} & \cellcolor{Green3!34}\statcell{32.85}{3.75}{} & \cellcolor{Green3!15}\statcell{53.04}{2.17}{} & \cellcolor{Green3!28}\statcell{58.32}{2.31}{} & \cellcolor{Green3!44}\statcell{62.59}{2.21}{} & \cellcolor{Green3!38}\statcell{65.66}{2.06}{} \\
\textbf{BELT-H-T} & \cellcolor{Green3!51}\secondstatcell{24.21}{3.55}{} & \cellcolor{Green3!47}\secondstatcell{29.77}{2.91}{} & \cellcolor{Green3!44}\statcell{34.24}{2.98}{} & \cellcolor{Green3!43}\secondstatcell{54.08}{2.05}{} & \cellcolor{Green3!59}\secondstatcell{59.18}{2.07}{} & \cellcolor{Green3!60}\beststatcell{62.97}{2.07}{} & \cellcolor{Green3!48}\statcell{65.96}{1.95}{} \\
\textbf{BELT-B-R} & \cellcolor{Green3!48}\statcell{23.10}{4.49}{} & \cellcolor{Green3!45}\statcell{29.52}{3.87}{} & \cellcolor{Green3!46}\secondstatcell{34.56}{3.70}{} & \cellcolor{Green3!21}\statcell{53.24}{2.23}{} & \cellcolor{Green3!27}\statcell{58.31}{2.43}{} & \cellcolor{Green3!44}\statcell{62.59}{2.29}{} & \cellcolor{Green3!49}\secondstatcell{65.99}{2.14}{} \\
\bottomrule\end{tabular}}
\end{center}
\end{table}

\begin{table}[!ht]
\caption{Split-half cross-fitted readings of \cref{tab:olmo-code-post} (OLMo3, LiveCodeBench problems from August 2024). Each cell is the mean over 400 split readings (200 stratified half splits, both directions); the subscript is the SD over split readings.}
\label{tab:xfit-olmo-code-post}
\begin{center}\footnotesize
\fittable{%
\begin{tabular}{@{}l|ccccccc@{}}\toprule
\textbf{Method} & \textbf{parity 0pp} & \textbf{parity 1pp} & \textbf{parity 2pp} & \textbf{budget 0.3} & \textbf{budget 0.4} & \textbf{budget 0.5} & \textbf{budget 0.6} \\\midrule
OSP & \cellcolor{Green3!27}\statcell{24.65}{11.45}{} & \cellcolor{Green3!58}\secondstatcell{46.92}{4.46}{} & \cellcolor{Green3!58}\secondstatcell{51.41}{2.58}{} & \cellcolor{Green3!60}\beststatcell{36.32}{2.74}{} & \cellcolor{Green3!60}\beststatcell{40.51}{2.60}{} & \cellcolor{Green3!60}\beststatcell{43.61}{2.67}{} & \cellcolor{Green3!58}\secondstatcell{45.42}{2.44}{} \\
OSP-1 & \cellcolor{Green3!0}\statcell{9.90}{8.46}{} & \cellcolor{Green3!0}\statcell{34.68}{6.52}{} & \cellcolor{Green3!0}\statcell{44.41}{4.04}{} & \cellcolor{Green3!59}\secondstatcell{36.30}{2.65}{} & \cellcolor{Green3!26}\statcell{40.06}{2.59}{} & \cellcolor{Green3!0}\statcell{42.39}{2.54}{} & \cellcolor{Green3!0}\statcell{43.76}{2.41}{} \\
\textbf{BELT-B-T} & \cellcolor{Green3!52}\statcell{37.72}{4.35}{} & \cellcolor{Green3!43}\statcell{43.80}{3.64}{} & \cellcolor{Green3!33}\statcell{48.36}{3.21}{} & \cellcolor{Green3!0}\statcell{35.51}{2.48}{} & \cellcolor{Green3!0}\statcell{39.72}{2.45}{} & \cellcolor{Green3!19}\statcell{42.77}{2.43}{} & \cellcolor{Green3!39}\statcell{44.86}{2.37}{} \\
\textbf{BELT-H-T} & \cellcolor{Green3!49}\statcell{36.21}{4.60}{} & \cellcolor{Green3!42}\statcell{43.43}{3.72}{} & \cellcolor{Green3!34}\statcell{48.56}{3.14}{} & \cellcolor{Green3!25}\statcell{35.85}{2.37}{} & \cellcolor{Green3!22}\statcell{40.01}{2.45}{} & \cellcolor{Green3!26}\statcell{42.92}{2.45}{} & \cellcolor{Green3!42}\statcell{44.95}{2.40}{} \\
\textbf{BELT-B-R} & \cellcolor{Green3!58}\secondstatcell{41.26}{3.57}{} & \cellcolor{Green3!54}\statcell{46.05}{3.54}{} & \cellcolor{Green3!50}\statcell{50.50}{3.33}{} & \cellcolor{Green3!32}\statcell{35.94}{2.53}{} & \cellcolor{Green3!26}\statcell{40.06}{2.57}{} & \cellcolor{Green3!34}\statcell{43.09}{2.44}{} & \cellcolor{Green3!50}\statcell{45.19}{2.42}{} \\
\bottomrule\end{tabular}}
\end{center}
\end{table}

\subsection{Development-split selection at a fixed tolerance}
\label{sec:dev-selected}
Instead of reading the operating point on the test frontier, each method selects, on the development split, the policy with the largest token saving whose development accuracy drop is at most 1~pp; \cref{tab:dev-selected-tolerance} reports the realized drop and saving on the test population.
\begin{table}[!ht]
\caption{Operating points selected on the development split at a 1\,pp accuracy tolerance, realized on the test population. Cell: realized accuracy drop (pp) / token saving (\%), mean over three seeds; (k/3) = seeds whose realized drop exceeds 1\,pp. Development split: math dev-160 for Qwen3 and LiveCodeBench (source-side selection), OLMo3 dev-160 for OLMo3. TC and ROM have no development-split rows and are not reported.}
\label{tab:dev-selected-tolerance}
\begin{center}\small
\fittable{%
\begin{tabular}{@{}l|cccccc@{}}\toprule
Setting & B-R & H-R & B-T & H-T & OSP & OSP-1 \\\midrule
Qwen ID & -0.35 / 42.8 (0/3) & -0.44 / 41.8 (0/3) & -0.24 / 24.5 (0/3) & -0.54 / 43.8 (0/3) & -0.42 / 43.4 (0/3) & -0.40 / 34.6 (0/3) \\
Qwen OOD & 0.02 / 37.8 (0/3) & 0.17 / 39.9 (0/3) & 0.35 / 23.9 (0/3) & 0.03 / 41.2 (0/3) & 0.40 / 41.5 (1/3) & 0.55 / 33.5 (0/3) \\
OLMo & 2.68 / 58.4 (3/3) & 3.50 / 60.5 (3/3) & 2.75 / 57.1 (3/3) & 3.20 / 58.2 (3/3) & 2.08 / 54.8 (3/3) & 1.81 / 52.6 (2/3) \\
LCB Pre & -1.25 / 27.1 (0/3) & 3.12 / 45.8 (2/3) & -2.97 / 24.5 (0/3) & 2.20 / 42.1 (1/3) & 0.55 / 37.4 (1/3) & 1.34 / 16.1 (2/3) \\
LCB Post & -1.01 / 40.1 (0/3) & 1.31 / 59.3 (1/3) & -2.53 / 31.0 (0/3) & 0.78 / 56.4 (1/3) & -0.49 / 52.8 (0/3) & 0.31 / 24.9 (0/3) \\
\bottomrule\end{tabular}}
\end{center}
\end{table}

On OLMo, BELT is competitive with full OSP and OSP-1 under development-split selection, but every method's seed-mean realized drop exceeds the 1\,pp tolerance on the test population (BELT 3.50\,pp, full OSP 2.08\,pp, OSP-1 1.81\,pp).
At this operating point BELT saves 60.49\% of tokens against full OSP's 54.77\% ($+5.72$\,pp, more on three of three seeds) and OSP-1's 52.58\% ($+7.90$\,pp, more on three of three seeds), with a realized drop 1.42\,pp larger than full OSP's and 1.69\,pp larger than OSP-1's.
On LiveCodeBench Pre, BELT is competitive, but the saving of one of its policy families varies across seeds under source-side selection (saving SD 20.1\,pp; realized drop above the tolerance on two of three seeds).
In the same setting a second BELT policy family (both are columns of \cref{tab:dev-selected-tolerance}) stays within the tolerance on three of three seeds and saves 27.12\% against OSP-1's 16.07\% ($+11.06$\,pp, more on three of three seeds), with a lower realized drop on three of three seeds.

\subsection{Direct utility training of the BELT probe}
\label{sec:checks-direct}
\begin{table}[t]\centering
\begingroup\footnotesize\setlength{\tabcolsep}{4pt}\renewcommand{\arraystretch}{1.1}
\resizebox{\linewidth}{!}{%
\begin{tabular}{@{}llcrrrr@{}}\toprule
\textbf{Seed} & \textbf{Recipe} & \shortstack{\textbf{Near-never-stop}\\\textbf{checkpoints}} & \shortstack{\textbf{Saving (\%)} $\uparrow$\\\textbf{at 1\,pp}} & \shortstack{\textbf{Acc. (\%)} $\uparrow$\\\textbf{at 0.5 cost}} & \shortstack{\textbf{Acc. (\%)} $\uparrow$\\\textbf{at 0.3 cost}} & \textbf{Utility} $\uparrow$ \\
\midrule
0 & Direct & 2/4 & \cellcolor{Green3!7}6.47& \cellcolor{Green3!47}62.98& \cellcolor{Green3!41}52.36& \cellcolor{Green3!0}\underline{0.627} \\
0 & Belief warm-up $\to$ direct & 0/4 & \cellcolor{Green3!6}5.89& \cellcolor{Green3!45}61.83& \cellcolor{Green3!41}52.46& \cellcolor{Green3!0}\underline{0.627} \\
\addlinespace[2pt]
1 & Direct & 0/2 & \cellcolor{Green3!5}4.75& \cellcolor{Green3!45}62.12& \cellcolor{Green3!43}53.72& \cellcolor{Green3!0}\underline{0.627} \\
1 & Belief warm-up $\to$ direct & 0/2 & \cellcolor{Green3!2}2.94& \cellcolor{Green3!32}54.05& \cellcolor{Green3!24}41.32& \cellcolor{Green3!0}\underline{0.627} \\
\addlinespace[2pt]
2 & Direct & 0/1 & \cellcolor{Green3!8}7.35& \cellcolor{Green3!51}65.48& \cellcolor{Green3!47}56.45& \cellcolor{Green3!0}\underline{0.627} \\
2 & Belief warm-up $\to$ direct & 1/1 & \cellcolor{Green3!0}1.21& \cellcolor{Green3!0}34.39& \cellcolor{Green3!0}25.25& \cellcolor{Green3!0}\underline{0.627} \\
\midrule
0 & Warm-up $-$ direct & -- & \cellcolor{Green3!60}\textbf{-0.58}& \cellcolor{Green3!60}\textbf{-1.15}& \cellcolor{Green3!60}\textbf{+0.09}& \cellcolor{Green3!0}\textbf{0.000} \\
1 & Warm-up $-$ direct & -- & \cellcolor{Green3!47}\underline{-1.81}& \cellcolor{Green3!46}\underline{-8.07}& \cellcolor{Green3!36}\underline{-12.40}& \cellcolor{Green3!0}\textbf{0.000} \\
2 & Warm-up $-$ direct & -- & \cellcolor{Green3!0}-6.14& \cellcolor{Green3!0}-31.09& \cellcolor{Green3!0}-31.20& \cellcolor{Green3!0}\textbf{0.000} \\
\midrule
-- & Natural completion & -- & 0.00 & -- & -- & 0.627 \\
-- & \beltb (three probe seeds) & -- & \cellcolor{Green3!59}\underline{46.30} $\pm$ 1.76& \cellcolor{Green3!60}\underline{71.02} $\pm$ 0.08& \cellcolor{Green3!60}\underline{64.81} $\pm$ 0.22& \cellcolor{Green3!60}\textbf{0.669} $\pm$ 0.002 \\
-- & \belth (three probe seeds) & -- & \cellcolor{Green3!60}\textbf{46.82} $\pm$ 1.08& \cellcolor{Green3!60}\textbf{71.13} $\pm$ 0.25& \cellcolor{Green3!60}\textbf{65.08} $\pm$ 0.43& \cellcolor{Green3!60}\textbf{0.669} $\pm$ 0.001 \\
\bottomrule\end{tabular}}\endgroup
\caption{Belief warm-up and direct utility training on the OOD test frontier. Each family combines retained checkpoints, source-validation thresholds, and native stochastic policies. Near-never-stop counts use native token saving below 0.5\%. Paired differences are within training seed; frozen \beltb/\belth\ references show mean $\pm$ SD over three probe seeds. Utility uses $\lambda=2\times10^{-5}$. Natural completion achieves 72.73\% accuracy at 5,022 total tokens.}
\label{tab:direct-testcurve}
\end{table}

\Cref{tab:reb-direct-training} lists, per seed and initialization, the checkpoint selected by the source-validation criterion and its OOD utility next to the like-for-like BELT reference.
\begin{table}[!ht]
\centering\small
\caption{Direct utility training of the BELT probe, checkpoint chosen by the recorded source-validation criterion (best chain-weighted source-validation utility over epochs 4--7). Native = that checkpoint's own stochastic policy on the OOD roster (question micro), nothing chosen on OOD; utility at $\lambda=2\times10^{-5}$. 1\,pp = token saving of \{native, natural\} at a 1\,pp tolerance. n.c.\ = the selected epoch is outside the retained roster and was never scored on OOD (its source-validation utility equals that of epoch 7); $^{*}$ = selected epoch outside the retained roster (ReProbe-initialized runs: scored in the registered epoch 4--7 window).}
\label{tab:reb-direct-training}
\begingroup\setlength{\tabcolsep}{3pt}
\fittable{%
\begin{tabular}{@{}llcccccccc@{}}\toprule
Init. & Recipe & Seed & Epoch & Src.-val.\ util. & Acc.\ (\%) & Tokens & Native util. & 1\,pp save & Epoch-7 util. \\\midrule
Random & Direct & 0 & e6 & 0.4379 & 72.73 & 5,022 & 0.6269 & 0.00 & 0.6269 \\
 & Direct & 1 & e4 & 0.0941 & 18.77 & 313 & 0.1814 & 1.74 & 0.1814 \\
 & Direct & 2 & e4$^{*}$ & 0.0941 & n.c. & n.c. & n.c. & n.c. & 0.1814 \\
 & Warm-up $\to$ direct & 0 & e4 & 0.0941 & 18.77 & 313 & 0.1814 & 1.74 & 0.1814 \\
 & Warm-up $\to$ direct & 1 & e4$^{*}$ & 0.0941 & n.c. & n.c. & n.c. & n.c. & 0.1814 \\
 & Warm-up $\to$ direct & 2 & e7 & 0.4379 & 72.73 & 5,022 & 0.6269 & 0.00 & 0.6269 \\
\midrule
ReProbe & Direct & 0 & e7 & 0.4379 & 72.73 & 5,022 & 0.6269 & 0.00 & 0.6269 \\
 & Direct & 1 & e5$^{*}$ & 0.4856 & 72.78 & 3,032 & 0.6672 & 39.63 & 0.6102 \\
 & Direct & 2 & e5$^{*}$ & 0.4450 & 66.88 & 1,617 & 0.6365 & 11.59 & 0.4268 \\
 & Warm-up $\to$ direct & 0 & e4 & 0.4379 & 72.73 & 5,022 & 0.6269 & 0.00 & 0.6269 \\
 & Warm-up $\to$ direct & 1 & e7 & 0.4379 & 72.73 & 5,022 & 0.6269 & 0.00 & 0.6269 \\
 & Warm-up $\to$ direct & 2 & e7 & 0.4379 & 72.73 & 5,022 & 0.6269 & 0.00 & 0.6269 \\
\midrule
\multicolumn{7}{@{}l}{BELT-H-R, development-selected (random init)} & \statcell{0.6689}{0.0011}{} & & \\
\multicolumn{7}{@{}l}{BELT-B-R, development-selected (random init)} & \statcell{0.6660}{0.0024}{} & & \\
\multicolumn{7}{@{}l}{OSP (full), development-selected (random init)} & \statcell{0.6713}{0.0013}{} & & \\
\multicolumn{7}{@{}l}{BELT-H-R, development-selected (ReProbe init)} & \statcell{0.6682}{0.0010}{} & & \\
\multicolumn{7}{@{}l}{BELT-B-R, development-selected (ReProbe init)} & \statcell{0.6653}{0.0064}{} & & \\
\multicolumn{7}{@{}l}{Natural completion} & 0.6269 & 0.00 & \\
\bottomrule\end{tabular}}\endgroup
\end{table}

From random initialization, direct utility training of the shared probe under the recorded recipe ends at never-stop or always-stop checkpoints (0.6269 / 0.1814 / n.c. for direct training, 0.1814 / n.c. / 0.6269 with the \emph{Now}/\emph{End} warm-up, by seed; natural completion 0.6269), against 0.6689 for BELT selected on the same development data.
From ReProbe initialization the same recipe reaches 0.6269 / 0.6672 / 0.6365: seed 1's selected checkpoint (0.6672) is within 0.0010 of BELT with the same initialization (0.6682), so from a belief-trained start direct training reaches BELT-level utility in one of three seeds.
The direct-training seeds that learn to stop lose utility with further training (seed 1: 0.6672 at epoch 5 to 0.6102 at epoch 7; seed 2: 0.6365 at epoch 5 to 0.4268 at epoch 7)
while belief-trained BELT, selected by the same development rule, holds 0.6682 $\pm$ 0.0010 across its three seeds with the same initialization (0.6689 $\pm$ 0.0011 from random initialization).
The recipe was not tuned. Training cards per direct seed (random init): seed 0 NVIDIA A100 80GB PCIe, seed 1 NVIDIA GeForce RTX 4090, seed 2 NVIDIA GeForce RTX 4090; the warm-up arms share one seed-0 \emph{Now}/\emph{End} trunk.

To separate belief training from architecture, we keep the reasoner frozen and train the BELT probe with a linear stopping head directly on the utility in \cref{eq:stopping-background}, following OS-Pruner (three seeds; price annealed to $\lambda=2\times10^{-5}$ by zero-based epoch 4; last epoch 7). A second variant first trains the probe on \emph{Now} and \emph{End} labels. \Cref{tab:direct-testcurve} reads every retained checkpoint, with its native stochastic policy and 13 source-validation logit-quantile thresholds, on the OOD roster under the test-frontier protocol of \cref{sec:testcurve-reading} (micro average).

We keep a fixed per-arm/seed checkpoint roster (four, two and one checkpoints for seeds 0, 1 and 2 in each recipe), selected with labels without answer elicitation, and do not reselect epochs. With random initialization, direct training includes both near-never-stop and always-stop checkpoints; seed 0 has two of four near-never-stop checkpoints, and seeds 1 and 2 have none. Nevertheless, the test-family maximum utility at $\lambda=2\times10^{-5}$ rounds to 0.627 in all three seeds, within $4\times10^{-5}$ of natural completion, against 0.669 for both BELT-B-R and BELT-H-R. The \emph{Now}/\emph{End} warm-up, whose three arms share one seed-0 trunk and vary actor initialization and optimizer randomness by seed, gives no material utility gain at the reported precision in this random-init roster and does not remove saturation. Threshold-expanded frontiers can show nontrivial saving even for saturated native checkpoints, so near-never-stop counts and frontier readings describe different aspects of these runs. This diagnostic covers the documented recipe with three seeds.

\ecoparagraph{Gradient attenuation in direct training.}
For one trajectory's unregularized return $R_\lambda=\sum_{i=1}^{T}\rho_i u_i$ with $u_i=y_i-\lambda C_i$, let $z_k$ be the stop logit ($\pi_k=\sigma(z_k)$) at a candidate boundary $k<T$ and $W_{k+1}$ the expected utility conditional on continuing past $k$. With the other logits fixed, only $S_k\{\pi_k u_k+(1-\pi_k)W_{k+1}\}$ depends on $z_k$ ($S_k$ and $W_{k+1}$ are fixed), so
\begin{equation}
\frac{\partial R_\lambda}{\partial z_k}=S_k\,\pi_k(1-\pi_k)\,(u_k-W_{k+1}).
\label{eq:stop-gradient}
\end{equation}
The factor $\pi_k(1-\pi_k)$ vanishes at either saturation extreme, and early stopping also reduces later reach $S_k$; if utilities span an interval of width $G$, $|\partial R_\lambda/\partial z_k|\le G S_k e^{-|z_k|}$. Parameter gradients also depend on logit Jacobians and cross-position aggregation; attenuation transfers when these Jacobians remain bounded. For a chain-averaged binary belief loss with weight $\omega$ and $n$ labelled positions,
\begin{equation}
\frac{\partial\mathcal L_{\rm BCE}}{\partial b_k}=\frac{\omega}{n}\left(\sigma(b_k)-y_k\right).
\label{eq:belief-coordinate-gradient}
\end{equation}
This residual has neither policy reach nor an additional sigmoid-derivative factor; the comparison concerns gradient structure, not matched parameter-gradient magnitudes. We freeze the belief probe and fit the policy from a bounded start (\cref{tab:policy-hyperparameters}); the attenuation is one mechanism consistent with the observed saturation, and generalization also affects utility.

\ecoparagraph{A conditional recovery-time comparison.}
With all other logits fixed, positive $S_k\Delta_k$ ($\Delta_k=u_k-W_{k+1}$) and unpreconditioned scalar gradient flow, moving a wrong saturated logit from $-M$ to zero, $M>0$, by direct ascent $\dot z=\eta S_k\Delta_k\sigma(z)(1-\sigma(z))$ takes
\[
T_{\rm direct}=\frac{e^M-e^{-M}+2M}{\eta S_k\Delta_k},
\]
whereas BCE descent $\dot b=\eta c(1-\sigma(b))$ for a binary label $y=1$, $c=\omega/n>0$, takes
\[
T_{\rm BCE}=\frac{M+1-e^{-M}}{\eta c}.
\]
With positive signal scales fixed, recovery grows exponentially versus linearly in $M$; this holds for fixed continuation values and scalar gradient flow, not for Adam on a shared network. Direct negative utility also has coordinate curvature $-S_k\Delta_k\pi_k(1-\pi_k)(1-2\pi_k)$, which changes sign when $S_k\Delta_k\ne0$, whereas binary cross-entropy is convex in a fixed linear belief predictor.

\subsection{Direct-training sensitivity to initialization}
\label{sec:checks-direct-sensitivity}
This separate diagnostic uses ReProbe-initialized direct probes,
not the random-init family in \cref{tab:direct-testcurve}, with the
diagnostic receipts' training bank (labels as in \cref{sec:natural-elicitation}), strict grading
and total-cost accounting. Seeds 1 and 2 use RTX4090 and seed 0
A100. Price is fixed at $2\times10^{-5}$ from zero-based epoch 4,
so the epoch-5-to-7 contrasts lie within one price stage. The immutable source-validation criterion
selects epoch 5 for seeds 1 and 2; the OOD comparison uses native
stochastic policies without OOD-selected thresholds or checkpoints.

\begin{table}[t]
\centering\small
\caption{Separately scoped ReProbe-initialization sensitivity.
Source-validation utility is chain weighted; OOD utility is
question-micro on the 1,025-question roster.
Intervals use 2,000 paired, dataset-stratified question resamples
and are conditional on the trained policies, not training-seed intervals.
An earlier inconsistent validation-score export was excluded; this diagnostic uses verified repeats with unchanged bank and trainer controls.}
\label{tab:direct-init-sensitivity}
\begin{tabular}{@{}lccc@{}}\toprule
Run & Source-val $e5\to e7$ & Native OOD $e5\to e7$ & OOD $\Delta$ [95\% CI] \\\midrule
Seed 1 & $.4856\to.4186$ & $.6672\to.6102$ & $-.0570$ [$-.0701,-.0453$] \\
Seed 2 & $.4450\to.2099$ & $.6365\to.4268$ & $-.2097$ [$-.2310,-.1893$] \\
\bottomrule\end{tabular}
\end{table}

Across the same 12,219 source-validation probes at 9k--12k tokens,
seed 2's mean absolute utility advantage changes from 0.0443 to
0.0379, while mean absolute logit credit changes from
$3.81\times10^{-6}$ to $1.80\times10^{-28}$.
Seed 0 shows the other attenuation path: reach stays near one
while the sigmoid derivative becomes tiny; its native OOD utility
is approximately natural completion, so it supplies no comparable
late utility-drop claim.
These probe-weighted coordinate signals come from accepted native
logits, with stop and continue probabilities evaluated separately;
they are not fit-bank optimizer parameter gradients or causal interventions.

Overfitting remains a possible contributor: seed 1's epoch-average online training loss improves while held-out utility deteriorates, although online epoch loss averages different parameter states and is not frozen-checkpoint fit utility. Belief training is exposed to the same risk and is selected by minimum validation loss: the nine Now/End, Now/End/Later and Now/End/Same ablations select epoch 1 and have higher validation loss at epoch 3 (these are not full four-target BELT training trajectories).

\subsection{Agent probe runs}
\label{sec:checks-alfworld}
In ALFWorld, BELT-H-T's three probe runs differ at the tolerance readings: saving within two points of natural success ranges from 4.5\% to 24.6\% (\cref{tab:alfworld-members}). The lowest run (s2) also keeps the natural action least often at its calibrated stops, 39--45\% at the 20--50\% targets against 49--59\% for the other two (\cref{fig:alfworld-same-stop}). With four members per run, a tolerance reading is set by the members nearest natural success and moves in steps; the budget readings interpolate between members and vary less, with a standard deviation across runs of at most 3.0~pp. The game bootstrap (\cref{tab:alfworld-bootstrap}) resamples games, not probe runs, so the main text compares run means. In ScienceWorld, the three probe runs of BELT-H-T and BELT-B-T agree at the tolerance readings and differ by at most 1.7 and 2.1 points at the budgets (\cref{tab:sciworld-members}).

\subsection{OS-Pruner reproduction}
\label{sec:checks-osp}
With no public code available at the time, we independently implemented OS-Pruner from its published architecture and objective, adapted to the shared trajectory, grading and total-token protocol (\cref{sec:implementation}); the Qwen3 checkpoints are trained on the outcomes of \cref{sec:natural-elicitation} and selected by validation utility after price annealing, and OSP also receives the threshold expansion of \cref{sec:threshold-expansion}. It is stronger than TC, ROM and the training-free rules at the one-point Qwen3 tolerances (\cref{tab:allbaselines}), and full OSP generally improves budget accuracy over OSP-1, though not at every reported operating point (\cref{tab:macro}).

\subsection{Comparators on OLMo3}
\label{sec:checks-olmo}
Full OSP on OLMo3 follows the Qwen3 recipe on the same outcomes: one approximately 405M-parameter checkpoint per training price, five per run and fifteen for three runs, each selected by validation utility. OSP-1 is a single-checkpoint reproduction trained on labels without answer elicitation (\cref{sec:training-recipe}) and evaluated under the shared grading. As on Qwen3, full OSP improves on OSP-1 at most operating points (\cref{tab:olmo-main,tab:olmo-full}). OLMo3 BELT-B-R omits prefix length (\cref{sec:olmo-replication}); \cref{tab:olmo-main} reports both BELT interfaces.

\subsection{Evaluation sets}
\label{sec:checks-ood}
OOD identifies benchmarks outside the three controller-training sources, not greater difficulty: natural accuracy is 66.7\% on OOD and 53.4\% on ID (\cref{tab:macro}), whose sets draw from competition-level subsets of NuminaMath, DAPO-Math and DeepScaleR. Training/evaluation overlap was screened by normalized-text deduplication and by exact and 13-word overlap against recorded external references; the screen removed nine AIME repetitions, and a versioned question registry fixes the retained roster (\cref{sec:dataset-recipe}). For semantic screening, we embedded all 4,241 training-bank and 1,025 OOD questions with \texttt{bge-base-en-v1.5} (cross-checked with \texttt{all-MiniLM-L6-v2}) and inspected the 30 highest-cosine training neighbors by hand. Two OOD questions (one Gaokao, one AMC 2023) repeat a DAPO-Math training problem with different LaTeX formatting (cosine 0.97 and 0.96), and one MATH-500 question shares a template with a DeepScaleR question but has a different polynomial and answer. Every other pair down to cosine 0.88 contains different problems sharing a topic or definition; no OOD question reaches cosine 0.98, and no AIME question has a training neighbor above 0.90. The two identified overlaps are retained in the fixed integer-answer roster used by every method for all reported comparisons.

\begin{samepage}
\section{Future Work}
\label{sec:future-work}
The reusable-probe framework suggests three directions for further study. First, a full serving-stack evaluation with one forced-answer generation after stopping would quantify end-to-end latency and live decoding behavior beyond the banked mathematics and coding trajectories. An initial micro-benchmark in a standard HF generate loop (Qwen3-1.7B, one RTX 4090, batch 1, 49 OOD questions) adds 0.95\% per-token latency, agrees with offline re-prefill stopping decisions at 99.7\% of query points, and reduces wall-clock time by 21.3\% at equal accuracy. Second, the training approach could incorporate a belief about remaining computation cost, as in ZIP-RC \citep{ICLR2026_b22c5b6a}, alongside the four answer-outcome beliefs. Third, symbolic-answer mathematics and further open-ended tasks would broaden coverage beyond the current integer-answer, coding and text-game evaluations.
\end{samepage}

\end{document}